%% file: thesis.tex
\documentclass[PhD,binding=0cm,noexaminfo,oneside]{sapthesis}
\usepackage{microtype} 
\usepackage[utf8]{inputenx}
\usepackage{hyperref}
\usepackage[numbers,square]{natbib}
\usepackage{algorithm2e}
\usepackage{wrapfig}

\usepackage{booktabs}
\usepackage{multirow}
\usepackage{hhline}
\usepackage{xspace}

\usepackage{epsfig}
\usepackage{graphicx}
\usepackage{overpic}
\usepackage{float}
\floatstyle{plaintop}
\restylefloat{table}
\usepackage[export]{adjustbox}
\usepackage{multirow}
\usepackage{subfig}
\usepackage{amsmath}
\usepackage[titletoc]{appendix}
\usepackage{hyperref}

\title{\input{title}}
\author{Antonio D'Innocente}
\IDnumber{1792918}
\course[Computer Science]{Ingegneria Informatica}
\courseorganizer{Department of Computer, Control and Management Engineering ”AntonioRuberti“, Sapienza - University of Rome}
\cycle{XXXIII}
\submitdate{January 2021}
\copyyear{2020}
\advisor{Prof. Barbara Caputo}
\authoremail{ant.dinnocente@gmail.com}

\newcommand{\noauthorea}{}
\input{header}

\begin{document}

\frontmatter

\maketitle

\input{acknowledgments}

\begin{abstract}
\input{abstract}
\end{abstract}

\tableofcontents

\mainmatter

\input{introduction}
\input{related}

\chapter{Domain Generalization-specific approaches}
\label{ch:DG}
\newpage
\input{DSAM}
\label{sec:dsam}
\newpage
\input{Rethinking}
\label{sec:rethinking}

\chapter{Self-Supervised approaches for Domain Generalization and Domain Adaptation}
\label{ch:ss}
\newpage
\input{Jigen}
\label{sec:jigen}
\newpage
\input{iciap}
\label{sec:iciap}
\newpage
\input{oshot}
\label{sec:oshot}

\chapter{Conclusions and Future Work}
\label{ch:conclusions}
\input{conclusions}

\backmatter

\bibliographystyle{abbrvnat}
\bibliography{thesis.bib}

\end{document}

%% file: title.tex
Exploring Data Aggregation and Transformations to Generalize across Visual Domains

%% file: header.tex
\usepackage{color}
\usepackage{latexsym}
\usepackage{listings}
\usepackage{enumitem}
\usepackage{amsfonts,amsmath,amssymb}
\usepackage{tabularx}
\usepackage{cprotect}
\usepackage{multirow}
\usepackage{adjustbox}

\let\oldnl\nl
\newcommand{\nonl}{\renewcommand{\nl}{\let\nl\oldnl}}

\ifdefined\noauthorea
\usepackage{cleveref} 
\AtBeginDocument{\renewcommand{\ref}[1]{\Cref{#1}}}
\AtBeginDocument{\renewcommand{\eqref}[1]{\Cref{#1}}}
\fi

\ifdefined\noauthorea

\else

\fi

\ifdefined\noauthorea

\else

\fi

\newcolumntype{C}[1]{>{\centering\arraybackslash}m{#1}}
\newcolumntype{R}[1]{>{\raggedleft\arraybackslash}m{#1}}

%% file: acknowledgments.tex
\section*{Acknowledgments}

Ringrazio la mia advisor, la professoressa Barbara Caputo, per avermi accolto nel suo laboratorio ed avermi fatto lavorare in prima linea nei problemi della Computer Vision. La professoressa Caputo ha una grande attitudine nel trasmettere la dedizione, mi ha insegnato cos’è e come si fa la ricerca e mi ha sempre incoraggiato a lavorare su quegli argomenti che di più mi appassionavano. Ringrazio la professoressa Tatiana Tommasi per la perfetta supervisione e per aver sempre ascoltato e valorizzato le mie idee. La professoressa Tommasi ha dato di gran lunga i più grandi contributi nei nostri lavori, in ogni aspetto, senza di lei quasi tutte le pubblicazioni descritte in questa tesi non sarebbero state possibili.

La fortuna di lavorare in diverse realtà mi ha portato a conoscere molti colleghi ed amici con cui ho condiviso un fantastico percorso:  Arjan, Chiara, Fabio M. C., Fabio C., Massimiliano, Mirco, Nizar, Paolo, Silvia, Valentina, vi ricordo tutti con affetto. Una speciale menzione va al gruppo di Milano: Valentina, Chiara e Arjan, siete grandi ragazzi, è impossibile pensarvi e non ricordare qualcosa di esilerante, spero che ci rivedremo presto. A tutti gli altri amici e colleghi di Torino che si sono nel tempo aggiunti al lab, sono contentissimo di avervi incontrati e mi dispiace che non abbiamo avuto più tempo per conoscerci meglio, state formando un gran gruppo, continuate così.

%% file: abstract.tex
Computer vision has flourished in recent years thanks to Deep Learning advancements, fast and scalable hardware solutions and large availability of structured image data. Convolutional Neural Networks trained on supervised tasks with backpropagation learn to extract meaningful representations from raw pixels automatically, and surpass shallow methods in image understanding. Though convenient, data-driven feature learning is prone to dataset bias: a network learns its parameters from training signals alone, and will usually perform poorly if train and test distribution differ. To alleviate this problem, research on Domain Generalization (DG), Domain Adaptation (DA) and their variations is increasing.

This thesis contributes to these research topics by presenting novel and effective ways to solve the dataset bias problem in its various settings. We propose new frameworks for Domain Generalization and Domain Adaptation which make use of feature aggregation strategies and visual transformations via data-augmentation and multi-task integration of self-supervision. We also design an algorithm that adapts an object detection model to any out of distribution sample at test time.

With through experimentation, we show how our proposed solutions outperform competitive state-of-the-art approaches in established DG and DA benchmarks.

%% file: introduction.tex
\chapter{Introduction}

The release of AlexNet \cite{NIPS2012alexnet} in 2012 ended a thirty year long AI winter and changed computer vision history. Thanks to fast, parallel implementation on graphic processing units and clever design, training a deep learning model with millions of parameters on a large scale database became a reality and, as as a result, a Convolutional Neural Network was winning the ILSVRC \cite{imagenet} and reducing error rates by 10 points over previous state of the art approaches using Fisher Vectors. Since then, Deep Learning research dominated the computer vision field with several new ground breaking approaches. Generative adversarial networks \cite{Goodfellow:GAN:NIPS2014} for content generation, Deep Residual Learning \cite{he2016deep} for training deeper models, Faster-RCNN \cite{ren2015faster} for real-time object detection to name a few, all these methods built on top of Deep Learning, Convolutions, new hardware advancements and the increasing availability of image data. 

The rise of Deep Learning marked a turning point in industry as well. Thanks to the performance reached by these new models, companies started to integrate computer vision solutions in their products. Face recognition for surveillance tasks, detection for robots and navigation, semantic segmentation and counting for product inspection, ranked retrieval for visual search and indexing, anomaly detection for medical image analysis. Industrial interest in Deep Learning shifted the focus of research: from ideas to applications, from theoretical investigation to practical problem solving.

While proven to be the best approach for many Computer Vision applications, Deep Learning is far from perfect. At the time of writing this thesis, dataset bias is one of the most difficult challenges. Consider a model trained for detecting objects and people in an urban scenario (i.e. camera mounted on a moving car). If the training data is biased, as in the case of containing all images taken during a sunny day, then this model will fail when test data is acquired on a different setting, such as night images or bad weather. Backpropagation learns all the cues for minimizing errors on the training set, ignoring which of these cues are specific to the training set. This leads to poor performances when conditions change. An oracle solution for this problem consists of collecting training data from the same distribution where test data comes from. However, this approach might be unfeasible: Convolutional Neural Networks require large amounts of labeled data to be trained effectively, and collecting these annotations for all possible inference settings is often too expensive. Furthermore, we might not be able to anticipate how test conditions will shift after deployment.

To alleviate this issue we employ Domain Generalization and Domain Adaptation. \emph{Domain Generalization} (DG) is a challenging setting, used when we want train a model to predict effectively on any unseen target distribution. It often assumes multiple distributions (domains) available at training time. We use \emph{Domain Adaptation} (DA) when we have some, often unlabeled, test distribution data at training time, and need to adapt the model to that specific distribution.

Many DA and DG solution use feature based approaches for separating domain-specific from domain-agnostic features (DG) or aligning features from train and test data with adversarial learning (DA). In this thesis, we take a different route, and explore alternative approaches for learning domain invariant representations that leverage data aggregations and transformations. In particular, we propose to use the self-supervised learning paradigm for solving generalization and adaptation problem. We show how the joint training of an auxiliary self-supervised task on unlabeled data can be used to bridge the gap between train and test, learn domain invariant representation, and provide auxiliary targets for adapting deep learning models on one sample at test time. Furthermore, we introduce a novel Domain Generalization framework, and study the effect of refined data augmentation on state-of-the-art DG methods.

\newpage
\section{Contributions}

The contribution of this thesis is to propose novel Domain Adaptation and Domain Generalization solutions by exporing data aggregation and visual transformation strategies. In three of our works, we show how a novel self-supervised-based approach can achieve solid generalization and adaptation results without explicit loss functions. We also present a DG framework which models an agnostic backbone within its architecture, and show how most state-of-the-art DG solutions become uneffective when we apply strong data augmentation to our sources. We present: 

\begin{itemize}
    \item  \textbf{A novel model-based approach for Domain Generalization} \cite{Antonio_GCPR18}. This algorithm leverages a multi-branch architecture and feature aggregation strategy to achieve a separation between domain-specific and agnostic information. We demonstrate how principled feature extraction from this model has led us to achieve state-of-the-art results in two Domain Generalization benchmarks.
    \item \textbf{A novel self-supervised based approach for Domain Generalization and Domain Adaptation} \cite{jigsawCVPR19}. We design a multi-task algorithm integrating supervised and self-supervised signals from the training samples. We show for the first time how an auxiliary self-supervised objective broadens the semantic understanding across domains, gaining state-of-the-art results in Domain Generalization. Furthermore, it is shown the use of self-supervision alone can prime adaptation on the unlabeled target in the unsupervised Domain Adaptation setting.
    \item \textbf{A self-supervised based algorithm for Partial Domain Adaptation} \cite{tackling_iciap19}. By expanding on the method proposed in \cite{jigsawCVPR19}, we integrate our self-supervised based approach into Partial Domain Adaptation algorithm, and improveme over the state-of-the-art in the difficult PDA setting.
    \item \textbf{A novel cross-domain object detection algorithm able to perform adaptation without a target} \cite{oshot}. All existing cross-domain approaches need a sizable amount of unlabeled target data during training. Here, we introduce the one-shot cross-domain setting and present the first approach specifically designed for adapting a model without the need to anticipate and collect the target set. Our OSHOT algorithm performs adaptation on the fly to each test sample by exploiting self-supervised signals from the sample itself. We compare our algorithm with top performing approaches for cross-domain detection and the most recent one-shot style-transfer technique, achieving state-of-the-art results in our new setting.
    \item \textbf{An extensive study on the real effectiveness of state-of-the-art Domain Generalization methods when strong regularization is used on the sources} \cite{borlino2021rethinking}. All existing DG algorithms are tested using weakly-augmented labeled data. By applying powerful data-augmentation techniques, we are able to reach state-of-the-art results with the naive Deep All algorithm. Furthermore, we show how improvements of top performing methods are negated when strong data augmentation is applied on the sources. In light of this, we have to rethink Domain Generalization benchmarks in order to move towards algorithms that are orthogonal to regularized data. 
\end{itemize}

\newpage
\section{Outline}

\ref{ch:literature} overviews related works. \ref{sec:dg} presents algorithms developed for Domain Generalization, \ref{sec:da} focuses on approaches for the Unsupervised Domain Adaptation setting and its variations. The two following sections provide background context for our self-supervised based solutions: \ref{sec:ss} covers self-supervision and its significance for training on unlabeled data, while \ref{sec:mt} briefly describes multi-task learning, a paradigm we use to join self-supervision with adaptation and generalization tasks.

\ref{ch:DG} presents two Domain Generalization specific approaches: D-SAM (\ref{sec:dsam}), a multi-branch architecture separating domain-specific and domain-agnostic modules explicitly, and a study on the effect of data augmentation for DG, showing how refined image pre-processing enables a model trained with simple ERM to reach state-of-the-art performances (\ref{sec:rethinking}).

\ref{ch:ss} focuses on our proposed self-supervised solutions for both Domain Generalization and Domain Adaptation. The first two sections present Jigen, a multi-task self-supervised approach for DG and DA (\ref{sec:jigen}), and its extension to Partial Domain Adaptation (\ref{sec:iciap}). \ref{sec:oshot} describes OSHOT, an approach for adapting model on one image in object detection tasks.

Finally, \ref{ch:conclusions} wraps up our findings and outlines proposals for future work.

\newpage
\section{Publications}

This is a list of the author's publications to Computer Vision conferences

\begin{itemize}
    \item D’Innocente A, Carlucci FM, Colosi M, Caputo B. Bridging between computer and robot vision through data augmentation: a case study on object recognition. InInternational Conference on Computer Vision Systems 2017 Jul 10 (pp. 384-393). Springer, Cham.
    \item D’Innocente A, Caputo B. Domain generalization with domain-specific aggregation modules. InGerman Conference on Pattern Recognition 2018 Oct 9 (pp. 187-198). Springer, Cham.
    \item Carlucci FM, D'Innocente A, Bucci S, Caputo B, Tommasi T. Domain generalization by solving jigsaw puzzles. InProceedings of the IEEE Conference on Computer Vision and Pattern Recognition 2019 (pp. 2229-2238).
    \item Bucci S, D’Innocente A, Tommasi T. Tackling partial domain adaptation with self-supervision. InInternational Conference on Image Analysis and Processing 2019 Sep 9 (pp. 70-81). Springer, Cham.
    \item D'Innocente A, Borlino FC, Bucci S, Caputo B, Tommasi T. One-Shot Unsupervised Cross-Domain Detection. arXiv preprint arXiv:2005.11610. 2020 May 23.
\end{itemize}

%% file: related.tex
\chapter{Related Works}
\label{ch:literature}

The dataset bias problem is well known in the computer vision community \cite{ponce2007toward,TorralbaEfros_bias}. Convolutional Neural Networks rely on substantial amount of labeled data for learning meaningful representations \cite{imagenet}, yet even large scale image datasets cannot capture all of the variety and complexity of the visual world \cite{ponce2007toward,TorralbaEfros_bias} and, as a result, algorithms struggle if the distribution shifts after deployment. 

Several algorithms have been developed to cope with domain shift, mainly in two different settings: Domain Generalization (DG) and Domain Adaptation (DA).

\newpage
\section{Domain Generalization}
\label{sec:dg}

True risk can be approximated by empirical risk, provided that we have enough data and that all samples are drawn from the same distribution (domain). However, when the test domain has not been seen at training time, performances are bound to drop substantially.

To overcome these limitations computer vision has studied the Domain Generalization (DG) problem \cite{ECCV12_Khosla}, where, given multiple training sources, the algorithm has to learn how to generalize to unseen distributions. More formally, we assume to have $n$ source domains $(D_{1}...D_{n})$, where $D_{i}$ is the $i^{th}$ source domain containing $N_{i}$ data-label pairs $(x_{i}^{j},y_{i}^{j})_{j=1}^{N_{i}}$. Our goal is to learn a function $f: x \rightarrow y$ that generalizes well to any novel testing domain $D_{T}$ not seen during training. The problem is studied under the assumption that all domains share the same feature space and label space.

\vspace{4mm}\textbf{Single-Source Domain Generalization}, also known as Out-Of-Distribution Generalization, is an extension of DG in which only $n=1$ source distribution is available during training. Since it can be difficult to separate style from semantics given only one observed distribution, it can be argued that the problem is ill-posed. 

\subsection{Literature Overview}

Several strategies has been studied as solutions for the Domain Generalization problem DG. 

Feature-level strategies focus on learning domain invariant data representations mainly by minimizing 
different domain shift measures. The metric learning approach of \cite{doretto2017} exploits a Siamese architecture to learn an embedding subspace in which the semantics of the mapped domains are aligned while their visual style is maximally separated. Autoencoder based approaches have been used: MTAE \cite{DGautoencoders} applies data-augmentation induced by deep autoencoders to transform the original image in the style of different related visual domains, \cite{Li_2018_CVPR} extends the traditional autoencoder framework with a Maximum-Mean-Discrepancy loss to align the distribution of different visual domains.

\begin{figure}[!t]
    \centering
\includegraphics[width=\textwidth]{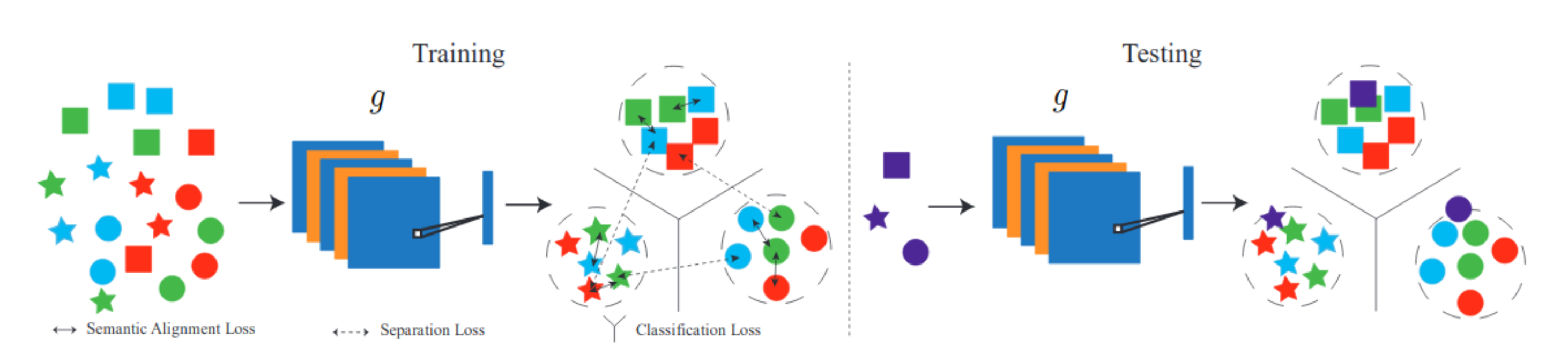}   
    \caption{Feature-level domain generalization approach of \cite{doretto2017}. At training time, the semantic alignment loss minimizes the distance between samples from different domains but the same class label and the separation loss maximizes the distance between samples from different domains and class labels. At test time, the embedding function embeds samples from unseen distributions to the domain invariant space and the prediction function classifies them (right). In this figure, different colors represent different domain distributions and different shapes represent different classes. \cite{doretto2017}}
    \label{fig:doretto}\vspace{-2mm}
\end{figure}

Model-level strategies, by leveraging the multi-source nature of DG, focus on architectural choices and/or optimization techniques to simulate out of distribution testing scenarios during training. Many studies leverage the meta-learning framework, MLDG \cite{MLDG_AAA18} defines meta-training and meta-testing domains at each iteration, and increases the performance of the model on the meta-testing set with a second order optimization inspired by \cite{MAML}, MetaReg \cite{NIPS2018_metareg} defines a new regularization function which is learned through a learning-to-learn approach. \cite{episodic_hospedales} uses a multi-branch network and adopts an episodic training scheme to train the final feature extractor and classifier to be robust to unusual signals. \cite{mancini2018best} trains multiple source-specifc predictors and, at test time, recombines these models based on the sample similarity to each source.  Other works use low-rank network parameter decomposition \cite{hospedalesPACS,Ding2017DeepDG} with the goal of identifying and neglecting domain-specific signatures. \cite{ECCV12_Khosla} aims at neglecting domain specific signatures through a shallow method that exploits multi-task learning.

Finally, data-level employ data-augmentation techniques to synthesize new images. Several works in this direction employ variants of the Generative Adversarial Networks (GANs, \cite{Goodfellow:GAN:NIPS2014}), \cite{DG_ICLR18,Volpi_2018_NIPS} learn how to properly perturb the source samples, even in the challenging case of DG from a single source. CROSSGRAD \cite{CrossGrad} adopts a Bayesian Network and a sampling step in which inputs are augmented through a domain-guided perturbation. \cite{Volpi_2018_NIPS} address the single-source DG problem through an adversarial data-augmentation scheme that systematically mimics a novel distribution by generating hard samples. In DDAIG \cite{zhou2020deep}, a domain transformation network is trained to map source data to unseen domains by optimizing an objective that preserves the label prediction but confuses the domain classifier. The combination of data and feature-level strategies has also shown further  improvements in results \cite{ADAGE}.

\vspace{4mm}\textbf{The Deep-All algorithm} is a standard ERM classifier that naively trains a deep model on the available source labeled data with backpropagation. Despite the existence of several solutions designed specifically for DG, it is not trivial to perform better than Deep-All \cite{hospedalesPACS,episodic_hospedales}. Recent work claims that, when properly regularized, the ERM algorithm is able to beat even the most performing DG algorithms \cite{gulrajani2020search}.

\newpage
\section{Unsupervised Domain Adaptation}
\label{sec:da}

In the Unsupervised Domain Adaptation (UDA) setting, we are given $N_{s}$ annotated samples from a source domain $D_{s}=(x_{s}^{j},y_{s}^{j})_{j=1}^{N_{s}}$, drawn from the distribution $S$, and $N_{T}$ unlabeled samples of the target domain $D_{T}=(x_{T}^{j})_{j=1}^{N_{T}}$ drawn from a different distribution $T$. The aim of UDA algorithms is to learn a function $f: x \rightarrow y$ that performs well on samples drawn from $T$. The setting can be either transductive (we want to deploy the model on $D_{T}$) or non-transductive (the model is going to be deployed on different data drawn from the distribution $T$).

UDA problems assume that all of the data shares the same feature and label space.

\vspace{4mm}\textbf{Multi-Source Domain Adaptation} is a variation of the Unsupervised Domain Adaptation setting in which source data is collected from multiple source domains, so that we have our training set consisting of $n$ source domains $(D_{1}...D_{n})$, where $D_{i}$ is the $i^{th}$ source domain containing $N_{i}$ data-label pairs $(x_{i}^{j},y_{i}^{j})_{j=1}^{N_{i}}$. Solutions for multi-source UDA problems may leverage domain annotation \cite{domainnet} compared to the single-source setting, although in some instances the domain label of the sources may be unknown \cite{mancini2018boosting, hoffman_eccv12,carlucci2017auto}.

\vspace{4mm}\textbf{Partial Domain Adaptation} (in this thesis the acronym PDA is used) is a recently introduced UDA scenario in which the label space of the target domain is strictly contained in that of the source domain $Y_{T} \subset Y_{S}$. Thus, besides dealing with the marginal shift in standard unsupervised domain adaptation, it is necessary to take care of the difference in the label space which makes the problem even more challenging. If this information is neglected and the matching between the whole source and target data is forced, any adaptive method may incur in a degenerate case producing worse performance than its plain non-adaptive version. Still the objective remains that of learning both class discriminative and domain invariant feature models.

\vspace{4mm}\textbf{Predictive Domain Adaptation} is an intermediate setting between Domain Adaptation and Domain Generalization. In the predictive Domain Adaptation setting, we are given $N_{s}$ annotated samples from a source domain $D_{s}=(x_{s}^{j},y_{s}^{j})_{j=1}^{N_{s}}$, drawn from the distribution $S$, and $a$ auxiliary domains $(D_{1}...D_{a})$ where the $i^{th}$ auxiliary domain contains $N_{i}$ unlabeled samples $(x_{i}^{j})_{j=1}^{N_{i}}$. The objective is to train a model on the source domain, while at the same time leveraging the meta-information in the auxiliary domains, in order to be able to generalize to any target distribution $T$ not seen at training time.

\subsection{Literature Overview}

\vspace{4mm}\noindent\textbf{Single-Source and Multi-Source Unsupervised Domain Adaptation}

\vspace{4mm}Feature-level strategies have been long studied in Domain Adaptation. Compared to DG for which the target is unknown, given the availability of unlabeled target data during training in DA, it is possible to employ distance metrics and reduce the gap between the source's and target's representations. \cite{hdivergence} introduced the H-divergence metric between unlabeled data from two different distributions: given a domain $\mathcal{X}$ with $\mathcal{D}$ and $\mathcal{D'}$ probability distributions over $\mathcal{X}$,let $\mathcal{H}$ be a hypothesis class on $\mathcal{X}$ and denote by $I(h)$ the set for which $h \in \mathcal{H}$ is the characteristic function; that is, $x \in I(h) \iff h(x)=1$. The H-divergence between $\mathcal{D}$ and $\mathcal{D'}$ is

\begin{equation}
    d_{\mathcal{H}}(\mathcal{D},\mathcal{D'}) = 2 \sup_{h \in \mathcal{H}} | Pr_{\mathcal{D}}[I(h)] - Pr_{\mathcal{D'}}[I(h)]|
\end{equation}

It is furthermore proven in \cite{hdivergence} that the H-divergence measure can, under some assumptions, bound the difference in error between the two distributions. Several works build upon the thoretical derivations of \cite{hdivergence}. DAN \cite{Long:2015} proposes a deep architecture with hidden representations embedded in a reproducing kernel Hilbert space where the means of different distributions are explicitly matched. DANN \cite{Ganin:DANN:JMLR16} uses a multi-task model with a domain discriminator that, through adversarial training, trains the backbone to extract similar representations for both the source and the target domain. This adversarial framework has become popular in the DA community and several works build on top of it with different flavors: ADDA \cite{Hoffman:Adda:CVPR17} adds a GAN loss to DANN, \cite{saito2017maximum} and \cite{Li_2018_ECCV} align the distributions of source and target by utilizing also the task-specific decision boundaries.  Deep CORAL \cite{dcoral} learns a non-linear transformation to align correlations of layer activations in a deep neural network. Metric learning has also been successfully used in \cite{doretto2017} by employing point-wise surrogates of distribution distances. Finally, \cite{Bousmalis:DSN:NIPS16} builds an autoencoder architecture with private and shared components, the model performs the main task in the source domain and the partitioned representation is also used to reconstruct images from both source and target.

Model-level strategies for UDA include the solution of \cite{carlucci2017auto}, which introduces domain alignment layers in standard learning networks that, through batch-normalization and an entropy-loss regularization, align the statistics of source and target data. Those layers can also be used in multi-source DA to evaluate the relation between the sources and target and then perform source model weighting \cite{MassiRAL,cocktail_CVPR18}.

\begin{figure}[!t]
    \centering
\includegraphics[width=\textwidth]{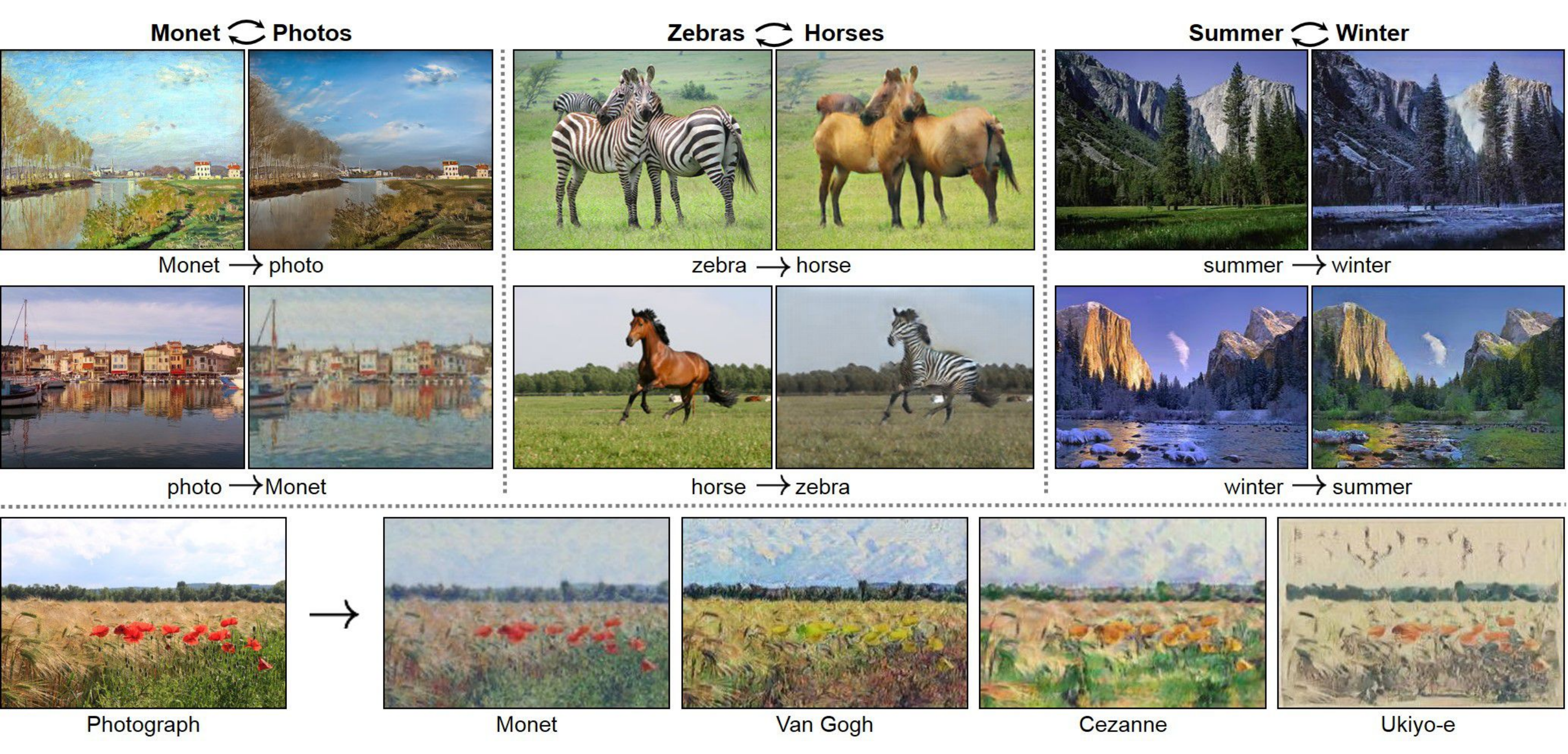}   
    \caption{CycleGAN \cite{CycleGAN2017} performs unsupervised image-to-image translation with Generative networks and a cycle consistency loss. It is a style transfer technique, and it has been successfully used in Unsupervised Domain Adaptation to create target-like labeled images from source data due to the availability of the target set during training.}
    \label{fig:cyclegan}\vspace{-2mm}
\end{figure}

Data-level techniques use data-augmentation to synthesize new images, and, as in the DG counterpart, many techniques use variants of Generative Adversarial Networks framework \cite{Goodfellow:GAN:NIPS2014}. Particularly effective are the methods based style transfer algorithms such as CycleGAN \cite{CycleGAN2017}. Given two unlabeled distributions $A$ and $B$, CycleGAN learns two mapping functions $F: x_{A} \rightarrow x_{B}$ and $G: x_{B} \rightarrow x_{A}$ by minimizing a cycle-consistency loss. SBADAGAN \cite{russo17sbadagan} and CyCADA \cite{cycada} both integrate the CycleGAN approach in UDA settings by adding a semantic contraint through labels (for the source) and pseudo-labels (for the target) in order to generate target-like labeled images using source data. Furthermore, \cite{ADAGE,sankaranarayanan2017generate} combine data-level and feature-level strategies for improved results.

\vspace{4mm}\noindent\textbf{Partial Domain Adaptation}

\vspace{4mm}The first work which considered this setting focused on localizing domain specific and generic image regions \cite{LOAD_ICRA}. The attention maps produced by this initial procedure are less sensitive to the difference in class set with respect to the standard domain classification procedure and
allow to guide the training of a robust source classification model.
Although suitable for robotics applications, this solution is insufficient when each domain has spatially diffused characteristics. 
In those cases the more commonly used PDA technique consists in adding a re-weight source sample strategy to a standard domain adaptation learning process. Both the Selective Adversarial Network (SAN, \cite{SAN}) and the Partial Adversarial Domain Adaptation (PADA, \cite{PADA_eccv18}) approaches build over the domain-adversarial neural network architecture \cite{Ganin:DANN:JMLR16} and exploit the source classification 
model predictions on the target samples to evaluate a statistics on the class distribution. The estimated contribution of each source class  either weights the class-specific domain classifiers \cite{SAN}, or re-scales the respective classification loss and a single overall domain classifier \cite{PADA_eccv18}. A different solution is proposed in \cite{IWAN}, where each domain has its own feature
extractor and the source sample weight is obtained from the domain recognition model rather than from the source classifier.
An alternative view on the PDA problem is presented in two recent preprints \cite{TWIN_PDA,featurenorm_PDA}. The first work uses two separate deep classifiers to reduce the domain shift by enforcing a minimal inconsistency between their predictions on the target. Moreover the class-importance weight is formulated analogously to PADA, but averaging over the output of both the source classifiers. The second work does not attempt to aligning the whole domain distributions and focuses instead on matching the feature norm of source and target. This choice makes the proposed approach robust to negative transfer with good results in the PDA setting without any heuristic weighting mechanism.

\vspace{4mm}\noindent\textbf{Cross-Domain Object Detection}

\begin{figure}[!t]
    \centering
\includegraphics[width=0.7\textwidth]{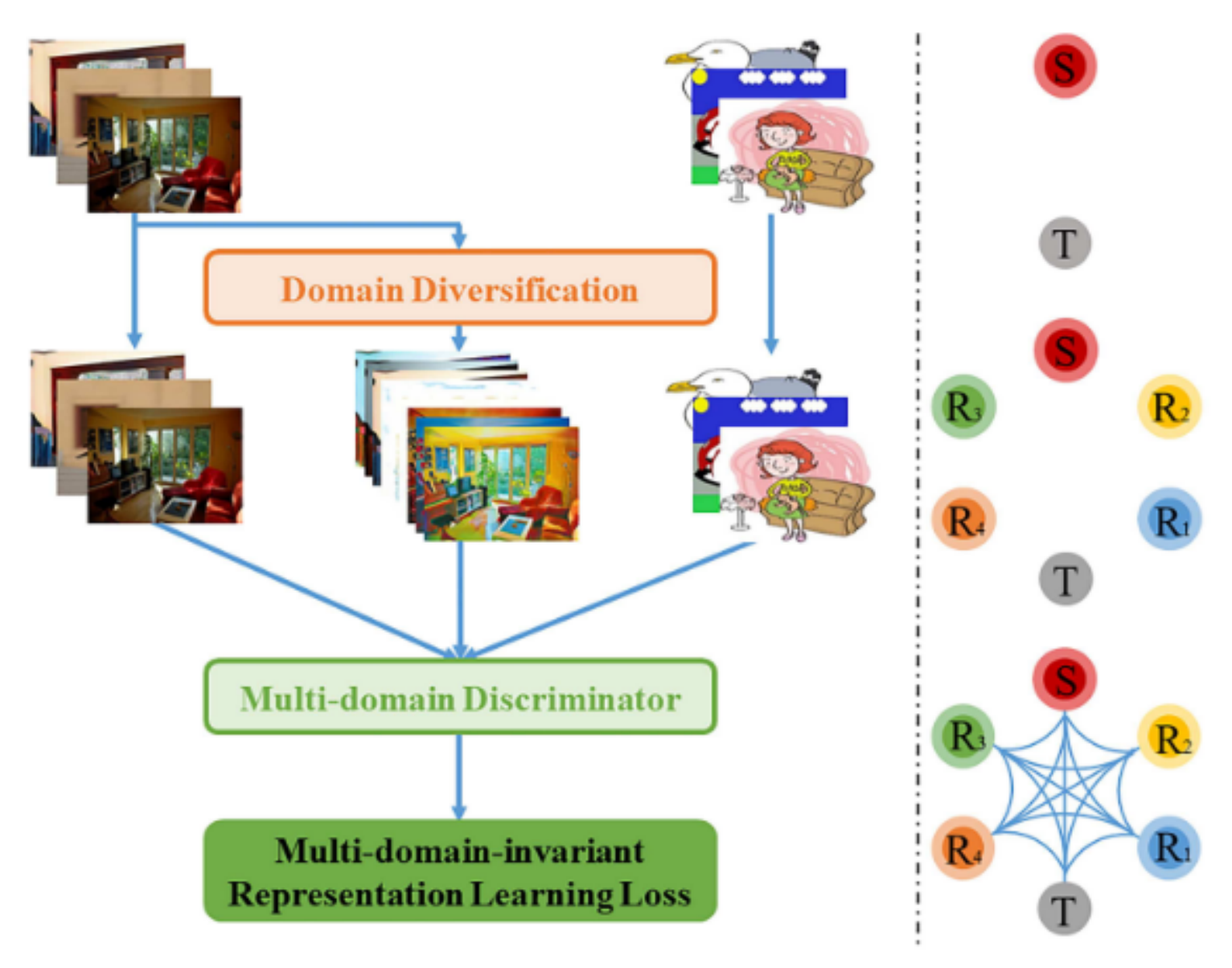}   
    \caption{Cross-Domain paradigm proposed in \cite{diversify_match_Kim_2019_CVPR}. In the diversification stage, multiple versions of the source domain are created through domain randomization via CycleGAN \cite{CycleGAN2017}. In the marching stage, representations distance of the source, target, and randomized domains is then minimized via adversarial learning. The framework exhibits competitive performance in cross-domain detection scenarios.}
    \label{fig:divmatch}
\end{figure}

\vspace{4mm}Many successful object detection approaches have been developed during the past several years, starting from the original sliding window methods based on handcrafted features, till the most recent deep-learning empowered solutions. Regardless of the specific implementation, the detector robustness across visual domain remains a major issue.

Most of the Unsupervised Domain Adaptation literature has focused on object classification, while the task of cross-domain object detection has received relatively less attention. Only in the last two years adaptive detection methods have been developed considering three main components: (i) including multiple and increasingly more accurate feature alignment modules at different internal stages, (ii) adding a preliminary pixel-level adaptation and (iii) pseudo-labeling. The last one is also known as self-training and consists in using the output of the source model detector as coarse annotation on the target. 
The importance of considering both global and local domain adaptation,
together with a consistency regularizer to bridge the two, was first highlighted in  \cite{Chen_2018_CVPR}. The Strong-Weak (SW) method of \cite{Saito_2019_CVPR} improves over the previous one pointing out the need of a better balanced alignment with strong global and weak local adaptation and is further extended by \cite{Xie_2019_ICCV_Workshops} where the adaptive steps are multiplied at different depth in the network.
By generating new source images that look like those of the target, the Domain-Transfer (DT, \cite{inoue2018cross}) method was the first to adopt pixel adaptation for object detection and combine it with pseudo-labeling. More recently the Div-Match approach \cite{diversify_match_Kim_2019_CVPR} re-elaborated the idea of domain randomization \cite{Tobin2017DomainRF}: multiple CycleGAN \cite{CycleGAN2017} applications with different constraints produce three extra source variants with which the target can be aligned at different extent through an adversarial multi-domain discriminator.
A weak self-training procedure (WST) to reduce false negatives is combined with adversarial background score regularization (BSR) in \cite{kim2019selftraining}. 
Finally, \cite{robust_Khodabandeh_2019_ICCV} followed the pseudo-labeling strategy including an approach to deal with noisy annotations.

\vspace{4mm}\noindent\textbf{Adaptive Learning on a Budget}

\vspace{4mm}There is a wide literature on learning from a limited amount of data, both for classification and detection. However, in case of domain shift, learning on a target budget becomes extremely challenging. Indeed, the standard assumption for adaptive learning is that a large amount of unsupervised target samples are available at training time so that a model can capture the domain style from them and close the gap with respect to the source.
Only few attempts have been done to reduce the target cardinality. In \cite{fewshotNIPS17} the considered setting is that of few-shot supervised domain adaptation: only a few target samples are available but they are fully labeled. In \cite{oneshotNIPS2018,Cohen_2019_ICCV} the focus is on one-shot unsupervised style transfer with a large source dataset and a single unsupervised target image. These works propose time-costly autoencoder-based methods to generate a version of the target image that maintains its content but visually resembles the source in its global appearance. Thus the goal is image generation with no discriminative purpose. 
A related setting is that of online domain adaptation where unsupervised target samples are initially scarce but accumulate in time \cite{Hoffman_CVPR2014,Wulfmeier2017IncrementalAD,mancini2018kitting}. In this case target samples belong to a continuous data stream with smooth domain changing, so the coherence among subsequent samples can be exploited for adaptation.

\newpage
\section{Self-Supervised Learning}
\label{sec:ss}

The success of deep learning approaches for computer vision relies on the availability of large quantities of annotated data. Although data acquisition might not be a problem, data annotation could be costly. Self-supervision, a subset of unsupervised learning, is used to avoid paying the price of human annotation and learning features from images using only unlabeled data. Many self-supervised learning methods exists, but the vast majority of them consists on using a pretext task that exploits inherent data attributes to automatically generate surrogate labels: part of the existing knowledge about the images is manually removed (e.g. the color, the orientation, the patch order) and the task consists in recovering it. It has been shown that the first layers of a network trained in this way capture useful semantic knowledge \cite{asano20a-critical}. The second step of the learning process consists in transferring the self-supervised learned model of those initial layers to a supervised downstream task (e.g. classification, detection), while the ending part of the network is newly trained. 
As the number of annotated samples of the downstream task gets lower, the advantage provided by the transferred model generally gets more evident  \cite{S4L_iccv19,asano20a-critical}. A literature survey on the most notable self-supervised learning methods for computer vision follows.

\begin{figure}[!b]
    \centering
\includegraphics[width=\textwidth]{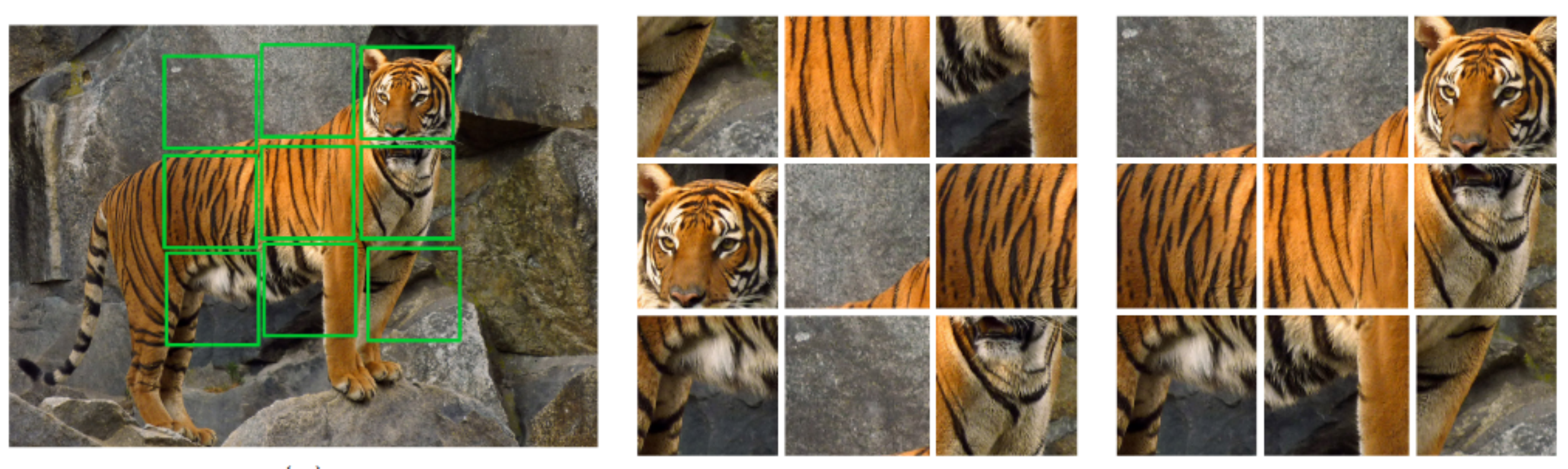}   
    \caption{The self-supervised jigsaw puzzle solving task proposed in \cite{noroozi2016unsupervised}. A subsection of the image is split in 9 tile from a 3x3 square grid. The tiles are then randomly re-arranged and the task is to retrieve the correct order. For simplicity, the problem is cast as classification by selecting only 1000 out of all the 9! possible permutations. Some data-augmentation is applied to the tiles so that the network cannot solve the task using low-level features, such as lines and color aberrations.}
    \label{fig:puzzle}
\end{figure}

\subsection{Literature Overview}

The possible pretext tasks can be organized in three main groups. 
One group relies only on original visual cues and involves either the whole image with
geometric transformations (e.g. translation, scaling, rotation \cite{gidaris2018unsupervised,NIPS2014_geometric}), 
clustering  \cite{caron2018deep}, inpainting \cite{pathakCVPR16context} and colorization \cite{zhang2016colorful},
or considers image patches focusing on their equivariance (learning to count \cite{learningtocount})
and relative position (solving jigsaw puzzles \cite{NorooziF16,Noroozi_2018_CVPR}). 
A second group uses external sensory information either real or synthetic: this solution
is often applied for multi-cue (visual-to-audio \cite{audiovisual}, RGB-to-depth \cite{ren-cvpr2018}) 
and robotic data \cite{grasp2vec, visiontouch}.
Finally, the third group relies on video and on the regularities introduced by the temporal dimension 
\cite{videosiccv15,SSLvideo}.
The most recent SSL research trends are mainly two. On one side there is the proposal of novel 
pretext tasks, compared on the basis of their ability to initialize a downstream task with 
respect to using supervised models as in standard transfer learning  
\cite{gidaris2020learning,jenni2020steering}. 
On the other side there are new approaches to combine several pretext tasks together in multi-task settings \cite{multitaskSSL,ren-cvpr2018}.

\newpage
\section{Multi-Task Learning}
\label{sec:mt}

Multi-task learning (MTL) is a learning paradigm in machine learning and its aim is to leverage useful information in multiple related tasks to help improve the generalization of all the tasks \cite{DBLP:journals/corr/ZhangY17aa}. In deep learning, the idea of multi-task learning is that one task can teach the network to extract features which are useful, yet orthogonal, to different tasks. MTL improves generalization by leveraging the domain-specific information contained in the training signals of related tasks, this is referred to as the inductive bias \cite{Caruana:1997}. Most works presented in this thesis exploit the multi-task learning paradigm by combining a main task (e.g. classification, object detection) with an auxiliary self-supervised task used to induce Domain Adaptation or Generalization for the primary objective.

\subsection{Literature Overview}

\vspace{4mm}\noindent\textbf{Hard and Soft parameter sharing}

\begin{figure}[!b]
    \centering
\includegraphics[width=\textwidth]{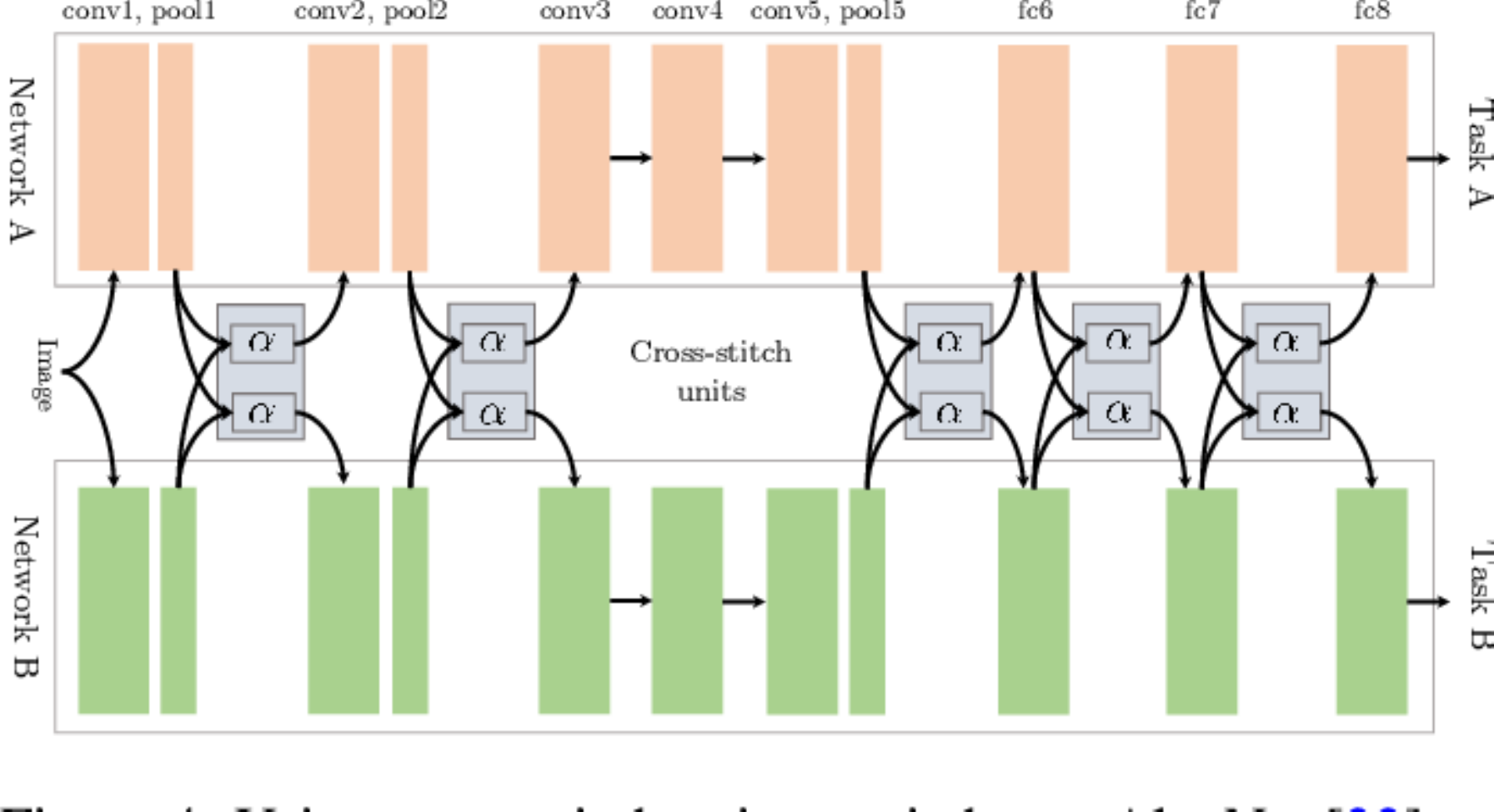}   
    \caption{Cross-Stich Networks \cite{misra2016cross} with two tasks. The two models soft-share their parameters}
    \label{fig:csnetworks}
\end{figure}

\vspace{4mm}In deep learning, a common approach for MTL problems is hard parameter sharing, i.e. the sharing of a part of the model between different tasks \cite{Caruana93multitasklearning:,ubernet,mordanrevisiting,doersch2017multi}. The first approach using hard parameter sharing \cite{Caruana93multitasklearning:} consists on having one shared backbone and multiple task-specific heads. Even though it has been shown this type of approach reduces overfitting \cite{Baxter1997}, the amount of sharing is in essence an hyperparameter, and different levels of sharing work better for different tasks \cite{misra2016cross}. For these reasons, research has also explored soft parameter sharing, in which each task has his own dedicated model whose parameters are interconnected \cite{misra2016cross,duong2015low,yang2016trace}.

\vspace{4mm}\noindent\textbf{Multi-Task Networks}

\vspace{4mm}In Ubernet \cite{ubernet}, an algorithm is presented for containing the computational costs of training on 7 different tasks through skip connections, while also facing the realistic scenario of having unlabeled samples. \cite{mordanrevisiting} is the first work to formally introduce the primary-auxiliary multi-task paradigm, in which auxiliary task are used only to increase performances on the main task. It leverages residual blocks to collect auxiliary knowledge with a privileged information approach. Cross-Stich Networks \cite{misra2016cross} use soft parameter sharing: multiple networks are instantiated and each learns a different task. The parameters are then shared between layers through Cross-Stich units: linear combinations of layer outputs that learn how to better combine cross-task information. \cite{ruder122017sluice} improves over these Cross-Stich units by learning an optimal feature subspace for merging knowledge.  \cite{doersch2017multi} learn jointly four self-supervised tasks in a multi-task framework, and shows how this approach leads to better unsupervised feature learning, which is, in some instances, competitive with full supervision. Routing Networks \cite{rosenbaum2017routing} proposes a Reinfocement Learning approach for finding the optimal layer path for the feature extractor in a multi-task setting.

\vspace{4mm}\noindent\textbf{Task weighting in Multi-Task Learning}

\vspace{4mm}The Multi-Task Learning paradigm suffers from disruptive gradients: without optimal task weighting, the learned representations will not benefit from the inductive bias and the effects may be negative for all tasks involved \cite{kendall2017multi}. This leads to resource consuming grid searches to find the optimal hyperparameters for the algorithm. \cite{kendall2017multi} propose a novel loss for minimizing the omoschedastic uncertanty, thus obtaining an automatic task balancing effect which also adjusts the weights dynamically during training. Finally, \cite{guo2018dynamic} proposes a reinforcement learning strategy for prioritizing the most difficult tasks during the learning process by defining dynamic difficulty measures.

\newpage
\section{Datasets}

The following section describes all the databases that were used to benchmark the algorithms presented in this thesis. All of these datasets are established testbeds widely used for Domain Generalization and Domain Adaptation research, with a focus on evaluating performances under different domain shifts scenarios.

\subsection{Image Classification Datasets}

\vspace{4mm}\noindent\textbf{PACS}

\begin{figure}[!b]
    \centering
\includegraphics[width=0.7\textwidth]{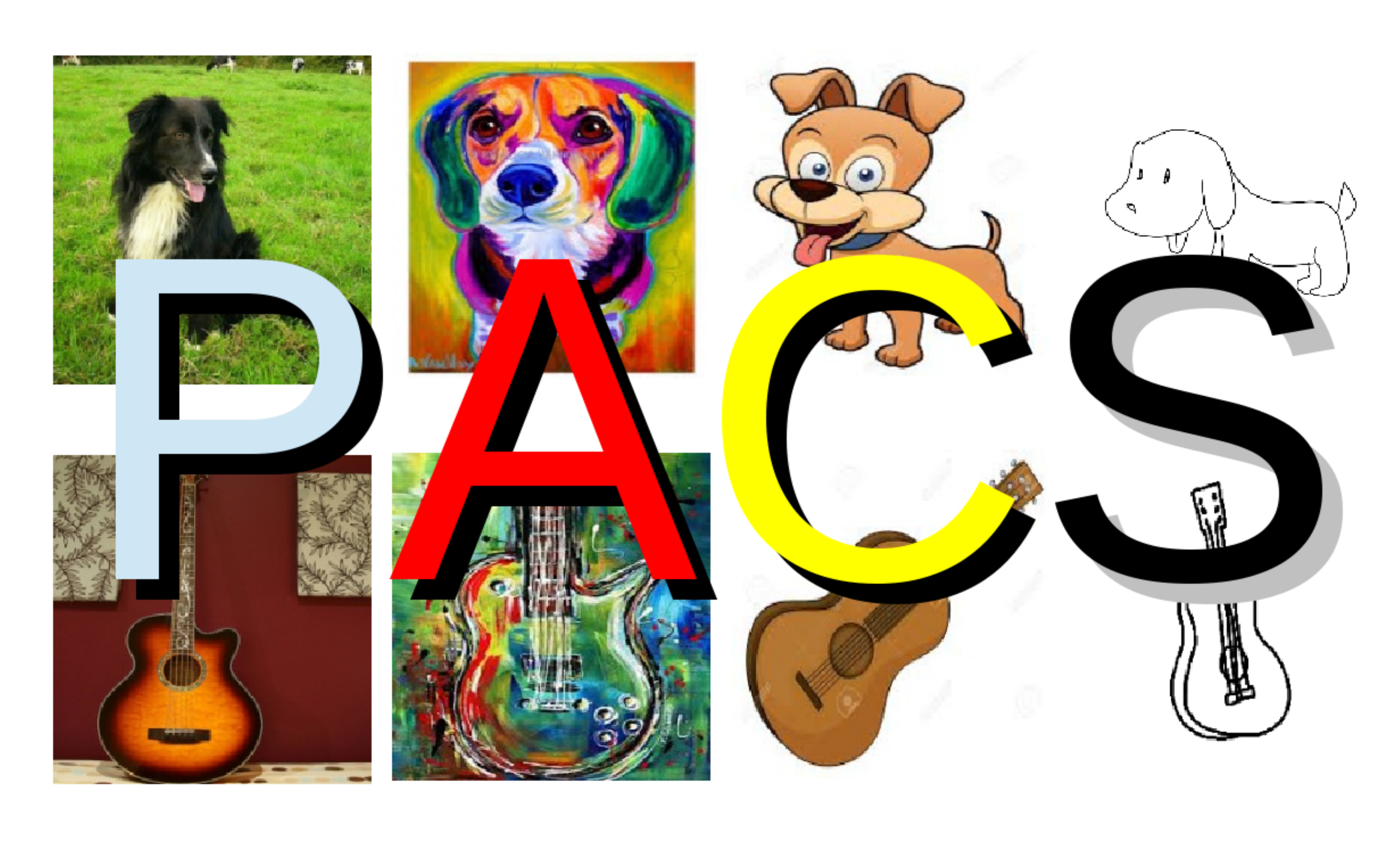}   
    \caption{The PACS dataset consists of images captured from different visual domains: Photos, Art painings, Cartoons and Sketches.}
    \label{fig:pacs}
\end{figure}

\vspace{4mm}The PACS database has been introduced in \cite{hospedalesPACS} as a testbed for Domain Generalization, and has quickly became a standard for evaluating DG algorithms in image classification. It consists of 9.991 images, of resolution 227×227, taken from four different visual domains (Photo,  Art  paintings,  Cartoon  and  Sketches),  depicting  seven  categories. The established evaluation protocol consists of using three domains as source and the remaining domain as a target. PACS is a challenging benchmark, consisting of domains with very large distribution shifts. It has also been used for testing Domain Adaptation methods.

\vspace{4mm}\noindent\textbf{Office-Home}

\vspace{4mm}Much like PACS, Office-Home tests performances under large distribution shifts between source and target. It has been first used in \cite{venkateswara2017Deep} for evaluating Domain Adaptation algorithms. It provides images from four different domains: Artistic images, Clip art, Product images and Real-world images. Each domain depicts 65 object categories that can be found typically in office and home settings. High accuracy on the Office-Home dataset is difficult to obtain, besides the large distribution shifts, it deals with changes in scale, resolution and orientation of objects. The original experimental protocol involved a Domain Adaptation setting with one domain as source and a different domain as target. In our works we repurposed Office-Home as a Domain Generalization benchmark by using three domains as source and the remaining domain as target, and it has subsequently been adopted as a standard benchmark from the DG community.

\vspace{4mm}\noindent\textbf{VLCS}

\vspace{4mm}VLCS \cite{TorralbaEfros_bias} is  built upon 4 different datasets: PASCAL VOC  2007, Labelme, Caltech and SUN and contains 5 object categories. Differently from other popular Domain Generalization testbeds, all the domains are composed of real world photos with the shift mainly due to camera type, illumination conditions, point of view, etc. Moreover, while Caltech is composed by object-centered images, the other three domains contain scene images containing objects at different scales. We use this dataset for Domain Generalization experiments, and follow the standard protocol of \cite{DGautoencoders}, using three domains as source and one as target, and  dividing  each  domain  into  a  training  set (70\%) and a test set (30\%) by random selection from the overall dataset.

\vspace{4mm}\noindent\textbf{Office-31}

\vspace{4mm}Historically, Office-31 \cite{Saenko:2010} has been widely used in Domain Adaptation. It contains 4652 images of 31 object categories common in office environment. Samples are drawn from three annotated distributions: Amazon (A), Webcam (W) and DSLR (D). The established Unsupervised Domain Adaptation setting considers one domain as a source and a different domain as a target. In our works, we use the Office-31 datasets for benchmarking Partial Domain Adaptation algorithms.

\vspace{4mm}\noindent\textbf{VisDA2017}

\begin{figure}[!t]
    \centering
\includegraphics[width=0.7\textwidth]{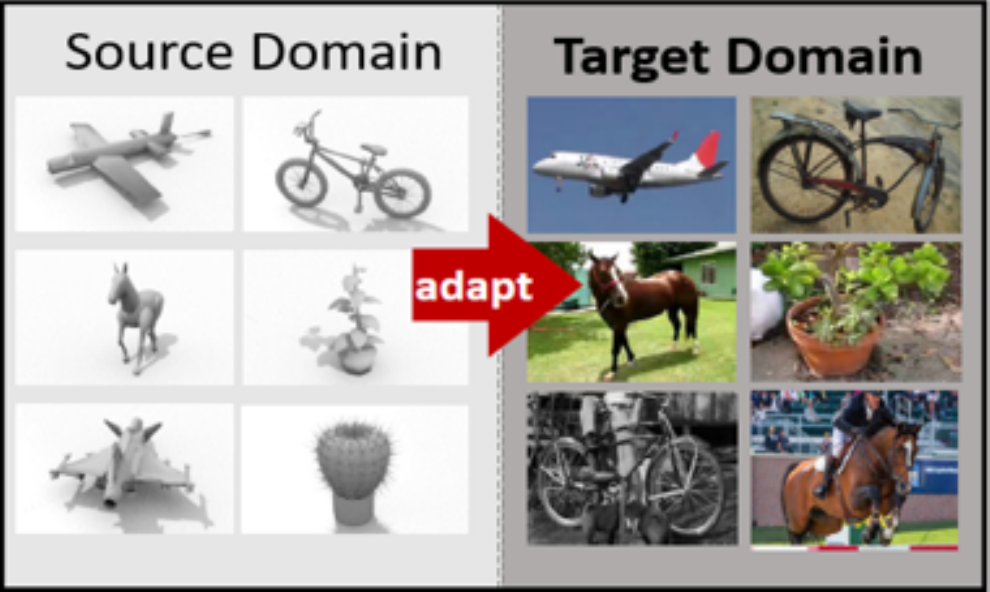}   
    \caption{The 2017 Visual Domain Adaptation challenge dataset. It consists two distribution: a synthetic database and real images. Algorithms are trained on the synthetic to real transfer.}
    \label{fig:visda}
\end{figure}

\vspace{4mm}VisDA2017 is the dataset used in the Visual Domain Adaptation challenge (classification track). It has two domains, synthetic 2D object renderings and real images with a total of 208000 images organized in 12 categories. It is often used in the synthetic to real Domain Adaptation shift, in our experiments we used it for benchmarking Partial Domain Adaptation under the same distribution shift. With respect to most DA benchmarks, VisDA allows to investigate approaches on a very large scale sample size scenario.

\vspace{4mm}\noindent\textbf{CompCars}

\vspace{4mm}The Comprehensive Cars (CompCars) \cite{Yang_2015_CVPR} is a large scale dataset. We used this dataset for Predictive Domain Adaptation experiments, by following the experimental protocol described in \cite{adagraph}, we selected a subset of 24151 images with 4 categories (MPV, SUV, sedan and hatchback) which are type of cars produced between 2009 and 2014 and taken under 5 different viewpoints (front, front-side, side, rear, rear-side). Each view point and each manufacturing year define a separate domain, leading of a total of 30 domains. In the PDA setting, we select one domain as source and a different domain as target, and the images in the remaining 28 domains are used as auxiliary unlabeled data. Considering all possible pairs, we have a total of 870 experiments.

\vspace{4mm}\noindent\textbf{Digits Datasets}

\begin{figure}[!t]
    \centering
\includegraphics[width=0.8\textwidth]{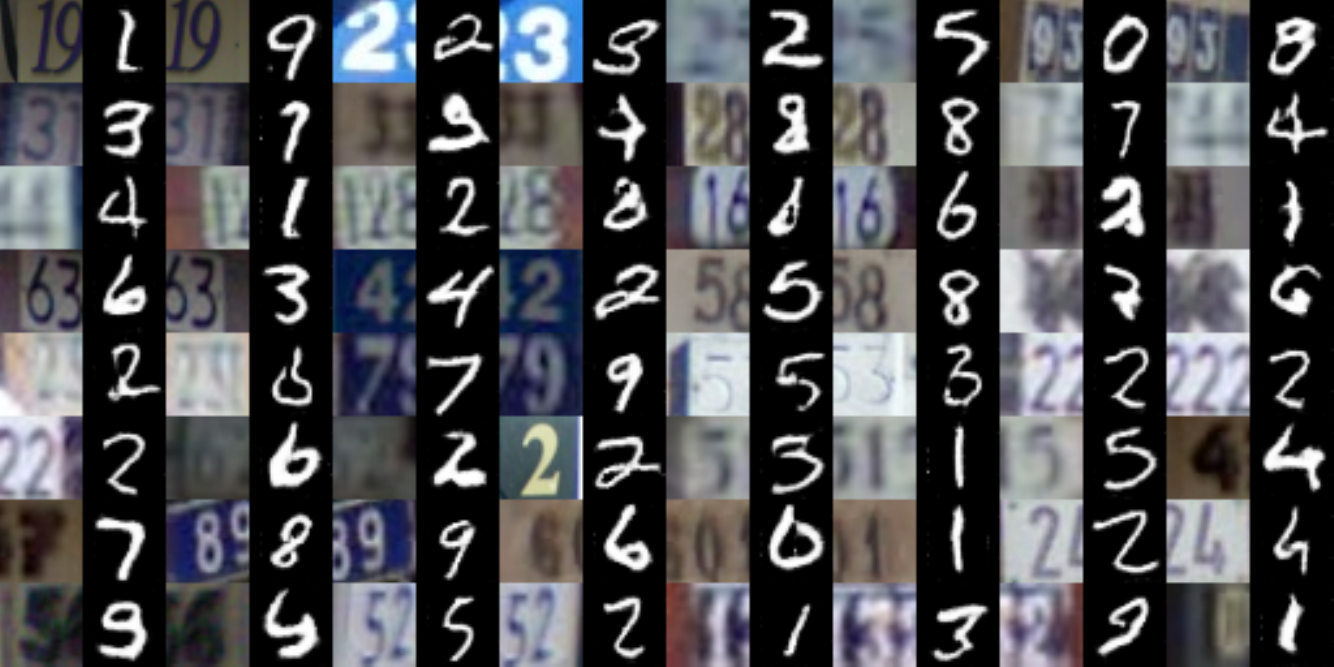}   
    \caption{Side-by-side comparison of MNIST samples (even columns) and SVHN samples (odd columns). The domain shift is large, and these two datasets are often used jointly as benchmark for Domain Adaptation and Domain Generalization methods.}
    \label{fig:mnist}
\end{figure}

\vspace{4mm}Digits dataset are all fairly similar, consisting of ten classes corresponding to the digits and low intra-domain variability. Digits recognition is considered a solved problem in computer vision, (the record accuracy of MNIST \cite{lecun1998gradient} is 99.82\%), however they are still used for evaluating cross-domain methods due to domain shifts occurring under different acquisitions. In this thesis, we use three of these datasets for evaluating single-source Domain Generalization performances.

MNIST \cite{lecun1998gradient} contains 60000 samples for training and 10000 samples for testing. The digits are handwritten and all images are 28x28 one-channel black-or-white centered crops.

MNIST-M \cite{Ganin:DANN:JMLR16} is obtained from MNIST by substituting the black background with a random patch from RGB photos.

SVHN \cite{netzer2011reading} stands for Street View House Numbers and it is a MNIST-like digits dataset obtained by house numbers from Google Street View images. It has 32x32 RGB samples, 73257 are used for training and 26032. Images are centered around the significant digit, as many of them contain different numbers at the sides.

\subsection{Object Detection Datasets}

\vspace{4mm}\noindent\textbf{Pascal-VOC}

\vspace{4mm}Pascal-VOC \cite{everingham2010pascal} is the standard real-world image dataset for object detection benchmarks. VOC2007 and VOC2012 both contain bounding boxes annotations of 20 common categories. VOC2007 has 5011 images in the train-val split and 4952 images in the test split, while VOC2012 contains 11540 images in the train-val split. In a Domain Adaptation setting, Pascal-VOC is often used as a source domain.

\vspace{4mm}\noindent\textbf{Artistic Media Datasets}

\vspace{4mm}Clipart1k, Comic2k and Watercolor2k \cite{inoue2018cross} are referred to as the Artistic Media Datasets (AMD). These are three object detection databases designed for benchmarking Domain Adaptation methods when the source domain is Pascal-VOC. Clipart1k shares its 20 categories with Pascal-VOC: it has 500 images in the training set and 500 images in the test set. Comic2k and Watercolor2k both have the same 6 classes (a subset of the 20 classes of Pascal-VOC), and 1000-1000 images in the training-test splits each.

\vspace{4mm}\noindent\textbf{Cityscapes}

\begin{figure}[!t]
    \centering
\includegraphics[width=0.9\textwidth]{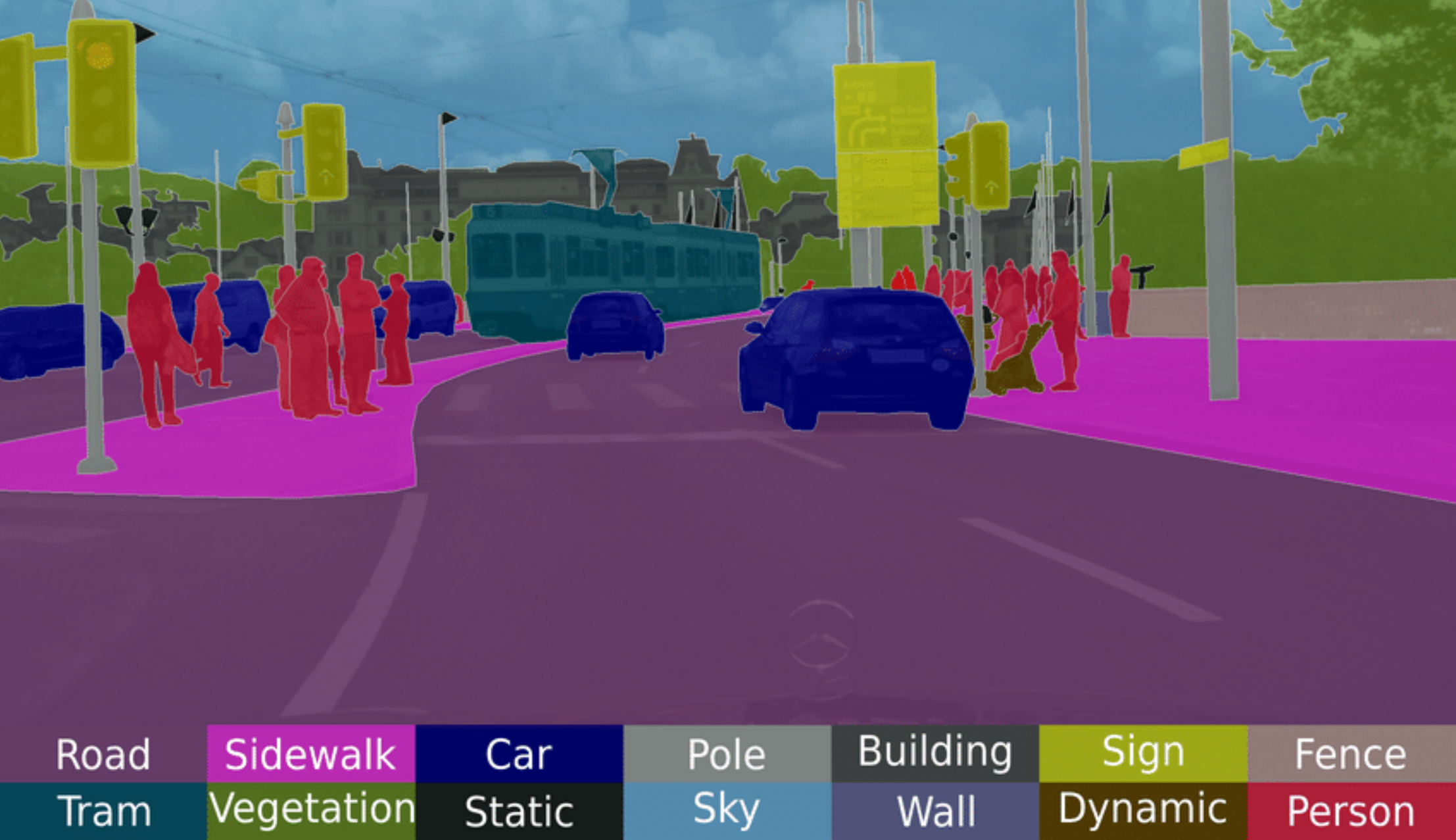}   
    \caption{The cityscapes dataset contains pixel level annotations of urban driving scenes. This dataset is often repurposed for object detection by computing bounding boxes of objects using semantic tags and instance masks \cite{Chen_2018_CVPR}.}
    \label{fig:cityscapes}
\end{figure}

\vspace{4mm}{Cityscapes} \cite{cordts2016cityscapes} is an urban street scene dataset with pixel level annotations of 8 categories. It has 2975 and 500 images respectively in the training and validation splits. For usage in object detection, we use the pixel annotations along with instance masks to generate bounding boxes of objects, as in \cite{Chen_2018_CVPR}. The Cityscapes dataset is used for evaluating cross-domain detection methods as a source or target dataset.

\vspace{4mm}\noindent\textbf{Foggy Cityscapes}

\vspace{4mm}Foggy Cityscapes \cite{sakaridis2018semantic} is obtained by adding different levels of synthetic fog to Cityscapes images. It is used as a target dataset for Domain Adaptation with Cityscapes as its source. We follow established protocols and only consider images with the highest amount of artificial fog, thus training-validation splits have 2975-500 images respectively.

\vspace{4mm}\noindent\textbf{KITTI}

\vspace{4mm}KITTI \cite{kitti} is a dataset of images  depicting  several  driving  urban scenarios. It is used in Domain Adaptation in conjunction with Cityscapes as a source or target domain and in this setting only the boxes with "car" label are considered. By following \cite{Chen_2018_CVPR}, we use the full 7481 images for both training (when used as source) and evaluation (when used as target).

%% file: DSAM.tex
\section{DSAM: a model based approach to Domain Generalization}

\textit{Visual recognition systems are meant to work in the real world. For this to happen, they must work robustly in any visual domain, and not only on the data used during training.
To this end, research on domain adaptation has proposed over the last years several solutions to reduce the covariate shift among the source and target data distributions. 
Within this context, a very realistic scenario deals with domain generalization, i.e. the ability to build visual recognition algorithms able to work robustly several visual domains, without having access to any information about target data statistic. This paper contributes to this research thread, proposing a deep architecture that maintains separated the information about the available source domains data while at the same time leveraging over generic perceptual information. We achieve this by introducing domain-specific aggregation modules that through an aggregation layer strategy are able to merge generic and specific information in an effective manner. Experiments on two different benchmark databases show the power of our approach, reaching the new state of the art in domain generalization.}

\vspace{4mm}As artificial intelligence, fueled by machine and deep learning, is entering more and more into our everyday lives, there is a growing need for visual recognition algorithms able to leave the controlled lab settings and work robustly in the wild. This problem has long been investigated in the community under the name of Domain Adaptation (DA): considering the underlying statistics generating the data used during training (source domain), and those expected at test time (target domain), DA assumes that the robustness issues are due to a covariate shift among the source and target distributions, and it attempts to align such distributions so to increase the recognition performances on the target domain. Since its definition \cite{Saenko:2010}, the vast majority of works has focused on the scenario where one single source is available at training time, and one specific target source is taken into consideration at test time, with or without any labeled data. Although useful, this setup is somewhat limited: given the large abundance of visual data  produced daily worldwide and uploaded on the Web, it is very reasonable to assume that several source domains might be available at training time. Moreover, the assumption to have access to data representative of the underlying statistic of the target domain, regardless of annotation, is not always realistic. Rather than equipping a seeing machine with a DA algorithm able to solve the domain gap for a specific single target, one would hope to have methods able to solve the problem for any target domain. This last scenario, much closer to realistic settings, goes under the name of Domain Generalization (DG, \cite{hospedalesPACS}), and is the focus of our work. 

Current approaches to DG tend to follow two alternative routes: the first tries to use all source data together in order to learn a joint, general representation for the categories of interest strong enough to work on any target domain \cite{MLDG_AAA18}. The second instead opts for keeping separated the information coming from each source domain, trying to estimate at test time the similarity between the target domain represented by the incoming data and the known sources, and use only the classifier branch trained on that specific source for classification \cite{mancini2018best}. Our approach sits across these two philosophies, attempting to get the best of both worlds. Starting from a generic convnet, pre-trained  on a general knowledge database like ImageNet \cite{imagenet}, we build a new multi-branch architecture with as many branches as the source domains available at training time. Each branch leverages over the general knowledge contained into the pre-trained convnet through a deep layer aggregation strategy inspired by \cite{shelhamerdeep}, that we call Domain-Specific Aggregation Modules (D-SAM). The resulting architecture is trained so that all three branches contribute to the classification stage through an aggregation strategy. The resulting convnet can be used in an end-to-end fashion, or its learned representations can be used as features in a linear SVM. 
We tested both options on two different architectures and two different domain generalization databases, benchmarking against all recent approaches to the problem. Results show that our D-SAM architecture, in all cases, consistently achieve the state of the art.

\subsection{Domain Specific Aggregation Modules}
	\label{dsam:sec:methodology}
    
In this section we describe our aggregation strategy for DG.    
We will assume to have $S$ source domains and $T$ target domains,  denoting with $N_{i}$ the cardinality of the $i_{th}$ source domain, for which we have $\{x_{j}^{i},y_{j}^{i}\}_{j=1}^{N_{i}}$ labeled samples. Source and target domains share the same classification task; however, unlike DA, the target distribution is unknown and the algorithm is expected to generalize  to new domains without ever having access to target data, and hence without any possibility to estimate the underlying statistic for the target domain.
    
The most basic approach, Deep All, consists of ignoring the domain membership of the  images available from all training sources, and training a generic algorithm on the combined source samples. Despite its simplicity, Deep All outperforms many engineered methods in domain generalization, as shown in \cite{hospedalesPACS}. 
    The domain specific aggregation modules we propose can be seen as a way to augment the generalization abilities of given CNN architectures by maintaining a generic core, while at the same time explicitly modeling the single domain specific features separately, in a whole coherent structure.
    
    Our architecture consists of a main branch $\Theta$ and a collection of domain specific aggregation modules $\Lambda = \{\lambda_{1}...\lambda_{n}\}$, each specialized on a single source domain. The main branch $\Theta$ is the backbone of our model, and it can be in principle any pre-trained off-the shelf convnet. 
 Aggregation modules, which we design inspired by an iterative aggregation protocol described in \cite{shelhamerdeep}, 
 receive inputs from $\Theta$ and learn to combine features at different levels to produce classification outputs. At training time, each domain-specific aggregation module learns to specialize on a single source domain. In the validation phase, we use a variation of a leave-one-domain-out strategy: we average predictions of each module but, for each ${i_{th}}$ source domain, we exclude the corresponding domain-specific module $\lambda_{i}$ from the evaluation. We test the model in both an end-to-end fashion and by running a linear classifier on the extracted features. In the rest of the section we describe into detail the various components of our approach (section \ref{dsam:sec:aggregation}-\ref{dsam:sec:architecture}) and the training protocol (section \ref{dsam:sec:learning}). 
 
    \subsubsection{Aggregation Module}
    \label{dsam:sec:aggregation}
    
    \begin{figure}[t]
	\centering
		\includegraphics[width=\textwidth]{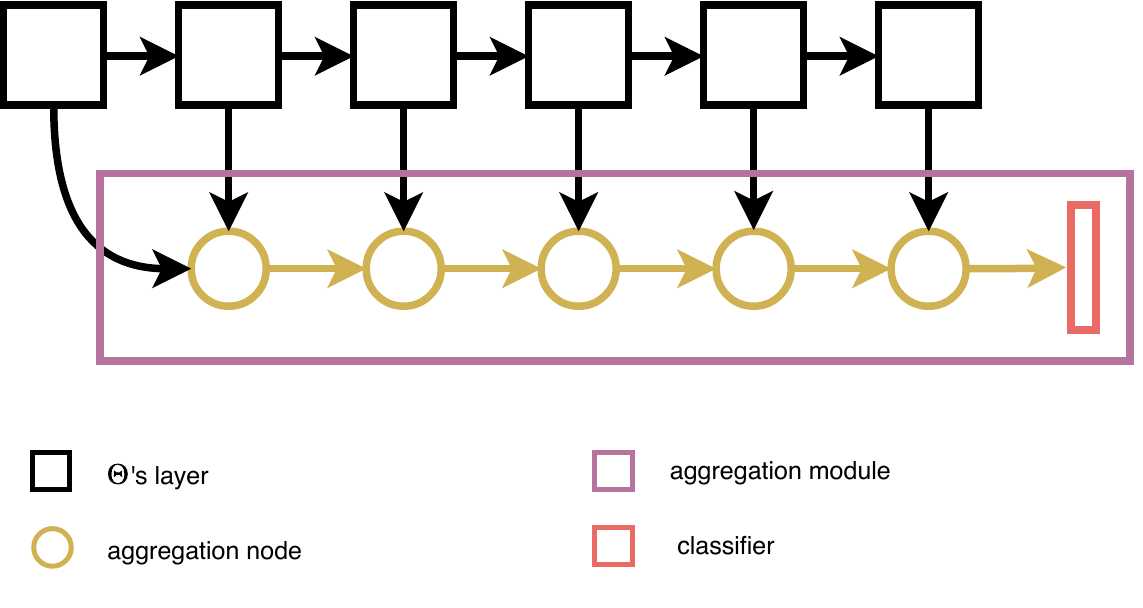}
		\caption{Architecture of an aggregation module (purple) augmenting a CNN model. Aggregation nodes (yellow) iteratively process input from $\Theta$'s layers and propagate them to the classifier.}
		\label{dsam:fig:aggr}
	\end{figure}
    
    Deep Layer Aggregation \cite{shelhamerdeep} is a feature fusion strategy designed to augment a fully convolutional architecture with a parallel, layered structure whose task is to better process and propagate features from the original network to the classifier. Aggregation nodes, the main building block of the augmenting structure, learn to combine convolutional outputs from multiple layers with a compression technique, which in \cite{shelhamerdeep} is implemented with 1x1 convolutions followed by batch normalization and nonlinearity. The arrangement of connections between aggregation nodes and the augmented network's original layers yields an architecture more capable of extracting the full spectrum of spatial and semantical informations from the original model \cite{shelhamerdeep}.
    
    Inspired by the aggregations of \cite{shelhamerdeep}, we implement aggregation modules as parallel feature processing branches pluggable in any CNN architecture. Our aggregation consists of a stacked sequence of aggregation nodes $N$, with each node iteratively combining outputs from $\Theta$ and from the previous node, as shown in Figure \ref{dsam:fig:aggr}. The nodes we use are implemented as 1x1 convolutions followed by nonlinearity. Our aggregation module visually resembles the Iterative Deep Aggregation (IDA) strategy described in \cite{shelhamerdeep}, but the two are different. IDA is an aggregation pattern for merging different scales, and is implemented on top of a hierarchical structure. Our aggregation module is a pluggable augmentation which merges features from various layers sequentially. Compared to \cite{shelhamerdeep}, our structure can be merged with any existing pre-trained model without disrupting the original features' propagation. We also extend its usage to non-fully convolutional models by viewing 2-dimensional outputs of fully connected nodes as 4-dimensional (N x C x H x W) tensors whose H and W dimension are collapsed. As we designed these modules having in mind the DG problem and their usage for domain specific learning, we call them Domain-Specific Aggregation Modules (D-SAM).

    \subsubsection{D-SAM Architecture for Domain Generalization}
    \label{dsam:sec:architecture}
    
    The modular nature of our D-SAMs allows the stacking of multiple augmentations on the same backbone network. Given a DG setting in which we have $S$ source domains, we choose a pre-trained model $\Theta$ and  augment it with $S$ aggregation modules, each of which implements its own classifier while learning to specialize on an individual domain. The overall architecture is shown in Figure \ref{dsam:fig:train}. 
    
    Our intention is to model the domain specific part and the domain generic part within the architecture. While aggregation modules are domain specific, we may see $\Theta$ as the domain generic part that, via backpropagation, learns to yield general features which aggregation modules specialize upon. Although not explicitly trained to do so, our feature evaluations suggest that thanks to our training procedure, the backbone $\Theta$ implicitly learns more domain generic representations compared to the corresponding backbone model trained without aggregations.
    
    \subsubsection{Training and Testing}
    \label{dsam:sec:learning}
    
    \begin{figure}[t]
	\centering
		\includegraphics[width=\linewidth]{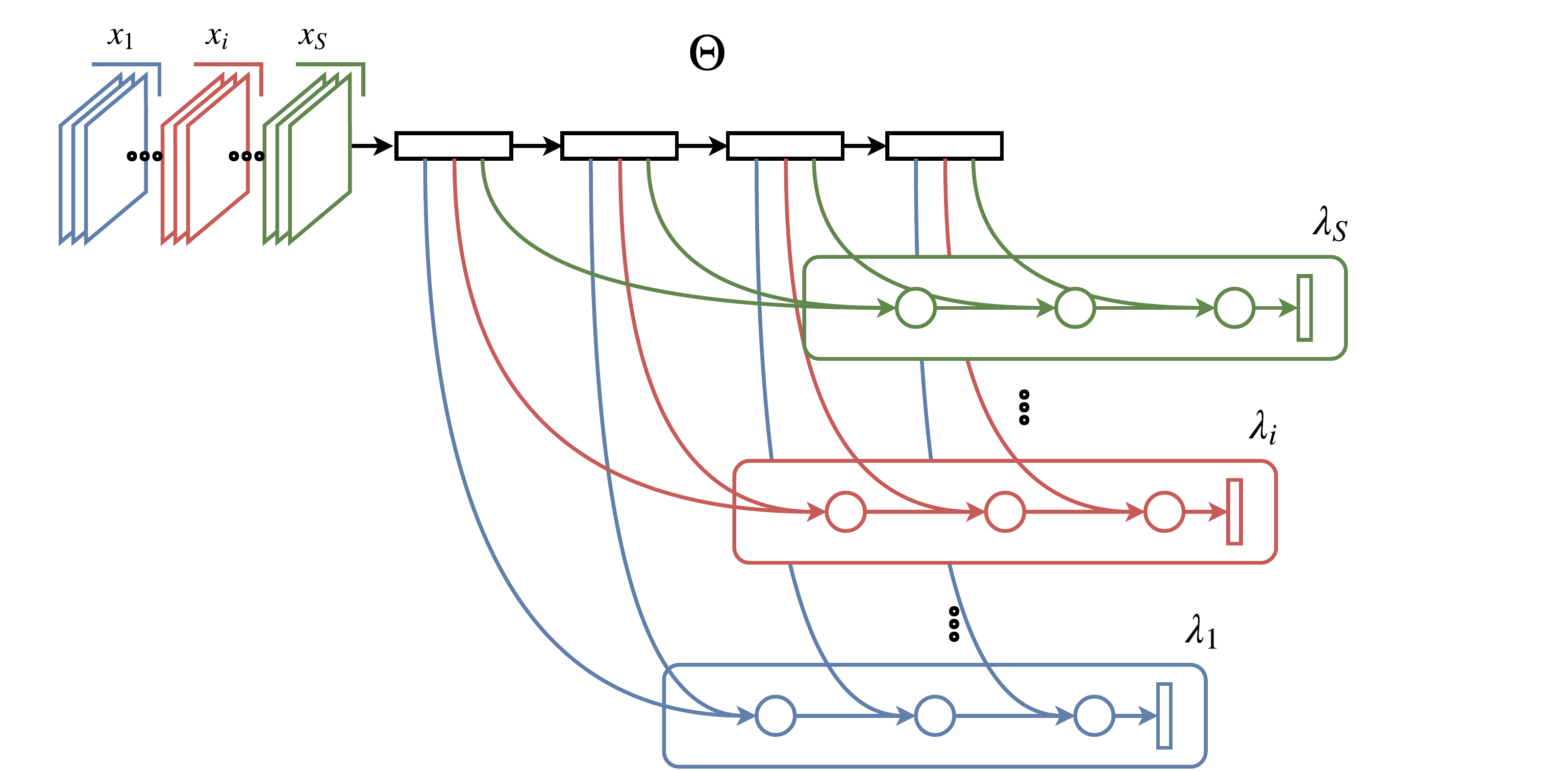}
		\caption{Simplified architecture with 3 aggregation nodes per aggregation module. The main branch $\Theta$ shares features with $S$ specialized modules. At training time, the $i_{th}$ aggregation module only processes outputs relative to the $i_{th}$ domain.}
		\label{dsam:fig:train}
	\end{figure}
    
    We train our model so that the backbone $\Theta$ processes all the input images, while each aggregation module learns to specialize on a single domain. To accomplish this, at each iteration we feed to the network $S$ equal sized mini-batches grouped by domain. Given an input mini-batch $x_{i}$ from the $i_{th}$ source domain, the corresponding output of our function, as also shown graphically in Figure \ref{dsam:fig:train}, is:
    
    \begin{equation}
    f(x_{i}) = \lambda_{i}(\Theta(x_{i})).
    \end{equation}
    
    We optimize our model by minimizing the cross entropy loss function $L_{C} = \displaystyle\sum_{c}y_{x_{j}^{i},c}\log(p_{x_{j}^{i},c})$, which for a training iteration we formalize as:
    
    \begin{equation}
    L(\Theta, \Lambda) = \displaystyle\sum_{i=1}^{S}L_{C}((\lambda_{i}\circ\Theta)(x_{i})).
    \end{equation}
    
    We validate our model by combining probabilities of the outputs of aggregation modules. One problem of the DG setting is that performance on the validation set is not very informative, since accuracy on source domains doesn't give much indication of the generalization ability. We partially mitigate this problem in our algorithm by calculating probabilities for validation as:
    
    \begin{equation}
    p_{x_{j}^v} = \sigma(\displaystyle\sum_{i=1, i\neq v}^{S}\lambda_{i}(\Theta(x_{j}^v))),
    \end{equation}
    
    where $\sigma$ is the softmax function. Given an input image belonging to the $k$ source domain, all aggregation modules besides $\lambda_{k}$ participate in the evaluation. With our validation we keep the model whose aggregation modules are general enough to distinguish between unseen distributions, while still training the main branch on all input data. 
    
    We test our model both in an end to end fashion and as a feature extractor. For end-to-end classification we calculate probabilities for the label as:
    
    \begin{equation}
    p_{x_{j}^t} = \sigma(\displaystyle\sum_{i=1}^{S}\lambda_{i}(\Theta(x_{j}^t))),
    \end{equation}
    
    When testing our algorithm as a feature extractor, we evaluate $\Theta$'s and $\Lambda$'s features by running an SVM Linear Classifier on the DG task.
    
\subsection{Experiments}
\label{dsam:sec:expers}

In this section we report experiments assessing the effectiveness of our DSAM-based architecture in the DG scenario, using two different backbone architectures (a ResNet-18 \cite{he2016deep} and an AlexNet \cite{NIPS2012alexnet}), on two different databases. We first report the model setup (section \ref{dsam:sec:model}), and then we proceed to report the training protocol adopted (section \ref{dsam:sec:training}). Section \ref{dsam:sec:results} reports and comments upon the experimental results obtained.

\subsubsection{Model setup}
	\label{dsam:sec:model}

	\vspace{4mm}\noindent\textbf{Aggregation Nodes} 
	
	\vspace{4mm}We implemented the aggregation nodes as 1x1 convolutional filters followed by nonlinearity. Compared to \cite{shelhamerdeep}, we did not use batch normalization in the aggregations, since we empirically found it detrimental for our difficult DG targets. Whenever the inputs of a node have different scales, we downsampled with the same strategy used in the backbone model. For ResNet-18 experiments, we further regularized the convolutional inputs of our aggregations with dropout.
    
    \vspace{4mm}\noindent\textbf{Aggregation of Fully Connected Layers} 
    
    \vspace{4mm}We observe that a fully connected layer's output can be seen as a 4-dimensional (N, C, H, W) tensor with collapsed height and width dimensions, as each unit's output is a function of the entire input image. A 1x1 convolutional layer whose input is such a tensor coincides with a fully connected layer whose input is a 2-dimensional (N, C) tensor, so for simplicity we implemented those aggregations with fully connected layers instead of convolutions.
    
    \vspace{4mm}\noindent\textbf{Model Initialization}
    
    \vspace{4mm}We experimented with two different backbone models: AlexNet and ResNet-18, both of which are pre-trained on the ImageNet 1000 object categories \cite{imagenet}. We initialized our aggregation modules  $\Lambda$ with random uniform initialization. We connected the aggregation nodes with the output of the AlexNet's layers when using AlexNet as backbone, or with the exit of each residual block when using ResNet-18. 
    
    \subsubsection{Training setup}
    \label{dsam:sec:training}
    We finetuned our models on $S=3$ source domains and tested on the remaining target. We splitted our training sets in 90\% train and 10\% validation, and used the best performing model on the validation set for the final test, following the validation strategy described in Section \ref{dsam:sec:methodology}. For preprocessing, we used random zooming with rescaling, horizontal flipping, brightness/contrast/saturation/hue perturbations and normalization using ImageNet's statistics. We used a batch size of 96 (32 images per source domain) and trained using SGD with momentum set at 0.9 and initial learning rate at 0.01 and 0.007 for ResNet's and AlexNet's experiments respectively. We considered an epoch as the minimum number of steps necessary to iterate over the largest source domain and we trained our models for 30 epochs, scaling the learning rate by a factor of 0.2 every 10 epochs. We used the same setup to train our ResNet-18 Deep All baselines. We repeated each experiment 5 times, averaging the results.
    
    \subsubsection{Results}
    \label{dsam:sec:results}
    
    We run a first set of experiments with the D-SAMs using an AlexNet as backbone, to compare our results with those reported in the literature by previous works, as AlexNet has been so far the convnet of choice in DG. Results are reported in table \ref{dsam:table:pacs}. We see that our approach outperforms previous work by a sizable margin, showing the value of our architecture. Particularly, we underline that D-SAMs obtain remarkable performances on the challenging setting where the 'Sketch' domain acts as target.
    
    We then run a second set of experiments, using both the PACS and Office-Home dataset, using as backbone architecture a ResNet-18. The goal of this set of experiments is on one side to showcase how our approach can be easily used with different $\Theta$ networks, on the other side to perform an ablation study with respect to the possibility to use D-SAMs not only in an end-to-end classification framework, but also to learn feature representations, suitable for domain generalization. To this end, we report results on both databases using the end-to-end approach tested in the AlexNet experiments, plus results obtained using the feature representations learned by $\Theta$, $\Lambda$ and the combination of the two. Specifically, we extract and $l_2$ normalize features from 
the last pooling layer of each component. We integrate features of $\Lambda$’s modules 
with concatenation, and train the SVM classifier leaving the hyperparameter $C$ 
at the default value. Our results in table \ref{dsam:table:resnetpacs} and \ref{dsam:table:resnetoh} show that the SVM classifier 
trained on the $l_2$ normalized features always outperforms the corresponding end-
 to-end models, and that $\Theta$’s and $\Lambda$’s features have similar performance, with $\Theta$’s 
 features outperforming the corresponding Deep All features while requiring no 
 computational overhead for inference.

    \begin{table}
\centering
\caption{PACS end-to-end results using D-SAMs coupled with the AlexNet architecture.}
\label{dsam:table:pacs}
\begin{tabular}{p{2cm}p{2cm}p{2cm}p{2cm}p{2cm}p{2cm}}
\hline
\textbf{}    & Deep All \cite{hospedalesPACS} & TF \cite{hospedalesPACS} & MLDG \cite{MLDG_AAA18}           & SSN \cite{mancini2018best}         & D-SAMs           \\
\hline
art painting & 64.91    & 62.86        & \textbf{66.23} & 64.10          & 63.87          \\
cartoon      & 64.28    & 66.97        & 66.88          & 66.80          & \textbf{70.70} \\
photo        & 86.67    & 89.50        & 88.00          & \textbf{90.20} & 85.55          \\
sketch       & 53.08    & 57.51        & 58.96          & 60.10          & \textbf{64.66} \\
\hline
avg          & 67.24    & 69.21        & 70.01          & 70.30          & \textbf{71.20}
\end{tabular}
\end{table}

\begin{table}
\centering
\caption{PACS results with ResNet-18 using features (top-rows) and end-to-end accuracy (bottom rows).}
\label{dsam:table:resnetpacs}
\begin{tabular}{p{2.5cm}p{1.9cm}p{1.9cm}p{1.9cm}p{1.9cm}|p{1.9cm}}
\hline
\textbf{}   & art painting & cartoon & sketch & photo & Avg   \\
\hline
Deep All (feat.) & 77.06        & \textbf{77.81}   & 74.09  & 93.28 & 80.56 \\
$\Theta$ (feat.)    & \textbf{79.57}        & 76.94   & 75.47  & 94.16 & 81.54 \\
$\Lambda$ (feat.)    & 79.48        & 77.13   & 75.30  & 94.30 & \textbf{81.55} \\
$\Theta + \Lambda$ (feat.) & 79.44 & 77.22 & 75.33 & 94.19 & 81.54
\\
\hline
Deep All    & 77.84        & 75.89   & 69.27  & 95.19  & 79.55 \\
D-SAMs       & 77.33        & 72.43   & \textbf{77.83}  &  \textbf{95.30} & 80.72
\end{tabular}
\end{table}

\begin{table}
\caption{OfficeHome results with ResNet-18 using features (top rows) and end-to-end accuracy (bottom rows).}
\label{dsam:table:resnetoh}
\begin{tabular}
{p{2.5cm}p{1.9cm}p{1.9cm}p{1.9cm}p{1.9cm}|p{1.9cm}}
\hline
\textbf{}      & Art   & Clipart & Product & Real-World & Avg   \\
\hline
Deep All (feat.) & 52.66 & 48.35   & 71.37   & 71.47      & 60.96 \\
$\Theta$ (feat.) & 54.55 & \textbf{49.37}   & 71.38   & \textbf{72.17}      & \textbf{61.87} \\
$\Lambda$ (feat.)     & 54.53    & 49.04      & 71.57      & 71.90         & 61.76    \\
$\Theta + \Lambda$ (feat.) & 54.54 & 49.05 & \textbf{71.58} & 72.03 & 61.80 \\
\hline
Deep All       & 55.59 & 42.42   & 70.34   & 70.86      & 59.81 \\
D-SAMs          & \textbf{58.03} & 44.37   & 69.22   & 71.45      & 60.77
\end{tabular}
\end{table}

\subsection{Conclusions}

We presented a Domain Generalization architecture inspired by recent work on deep layer aggregation. We developed a convnet that, starting from a pre-trained model carrying generic perceptual knowledge, aggregates layers iteratively for as many branches as the available source domains data at training time. The model can be used in an end-to-end fashion, or its convolutional layers can be used as features in a linear SVM. Both approaches, tested with two popular pre-trained architectures on two benchmark databases, achieved state of the art.
Future work in this direcrion should further study deep layer aggregation strategies within the context of domain generalization, as well as scalability with respect to the number of sources.

%% file: Rethinking.tex
\section{Rethinking Domain Generalization Baselines}

\textit{Despite being very powerful in standard learning settings, deep learning models can be extremely brittle when deployed in scenarios different from those on which they were trained. Domain generalization methods investigate this problem and data augmentation strategies have shown to be helpful tools to increase data variability, supporting model robustness across domains. 
In our work we focus on style transfer data augmentation and we present how it can be implemented with a simple and inexpensive strategy to improve generalization. Moreover, we analyze the behavior of current state of the art domain generalization methods when integrated with this augmentation solution: our thorough experimental evaluation shows that their original effect almost always disappears with respect to the augmented baseline. This issue open new scenarios for domain generalization research, highlighting the need of novel methods properly able to take advantage of the introduced data variability to further push domain generalization research.}

\vspace{4mm}The real world offers such a large diversity that the standard machine learning assumption of collecting train and test data under the same conditions, thus from the same domain/distribution, is broadly violated. Domain adaptation and domain generalization methods tackle this problem under different points of view. In the first case, unlabeled test data are considered available at training time, allowing the learning model to peek into the characteristics of the target set and adapt to it \cite{csurka_book}. Domain generalization is a more challenging task because target data are fed to the system only during deployment \cite{NIPS2011_blanchard,shallowDG}. In this last setting it is crucial to train robust model, possibly exploiting multiple available sources. Towards this goal, most of the existing domain generalization strategies try to incorporate the observed data invariances, capturing them at feature \cite{Li_2018_CVPR} or model (meta-learning \cite{episodic_hospedales} and self-supervision \cite{lopez_rotation}) level, in the hypothesis that analogous invariances hold for future test domains. An alternative solution consists in extending the source domains by synthesizing new images. This is usually done by learning generative models with the specific constraint of preserving the object content but varying the global image appearance, with the aim of better spanning the data space and include a larger variability in the training set. Thanks to the developments in generative learning, it is becoming more and more evident that their integration into domain generalization approaches is effective \cite{zhou2020deep}. However their performance tends to grow together with the complexity of the learning model which may involve one or multiple generator modules that are notoriously difficult to train. We also noticed a particular trend in the most recent domain generalization research. 
Several papers discuss the merit of the proposed data augmentation solutions in comparison with feature and model-based generalization techniques \cite{zhou2020deep,zhang2020learning}. 
However, newly introduced feature and model-based approaches avoid benchmarks against data augmentation strategies considering them unfair competitors due to the extended training set \cite{wang2019learning,huang2020selfchallenging}. We believe that the field needs some clarification and we dedicate our work on this topic. Specifically our main contributions are:\\

\begin{figure}
\centering
\begin{tabular}{|c|cc|}
\hline
 & \multicolumn{2}{c|}{content images}\\\hline
& \includegraphics[width=0.20\linewidth]{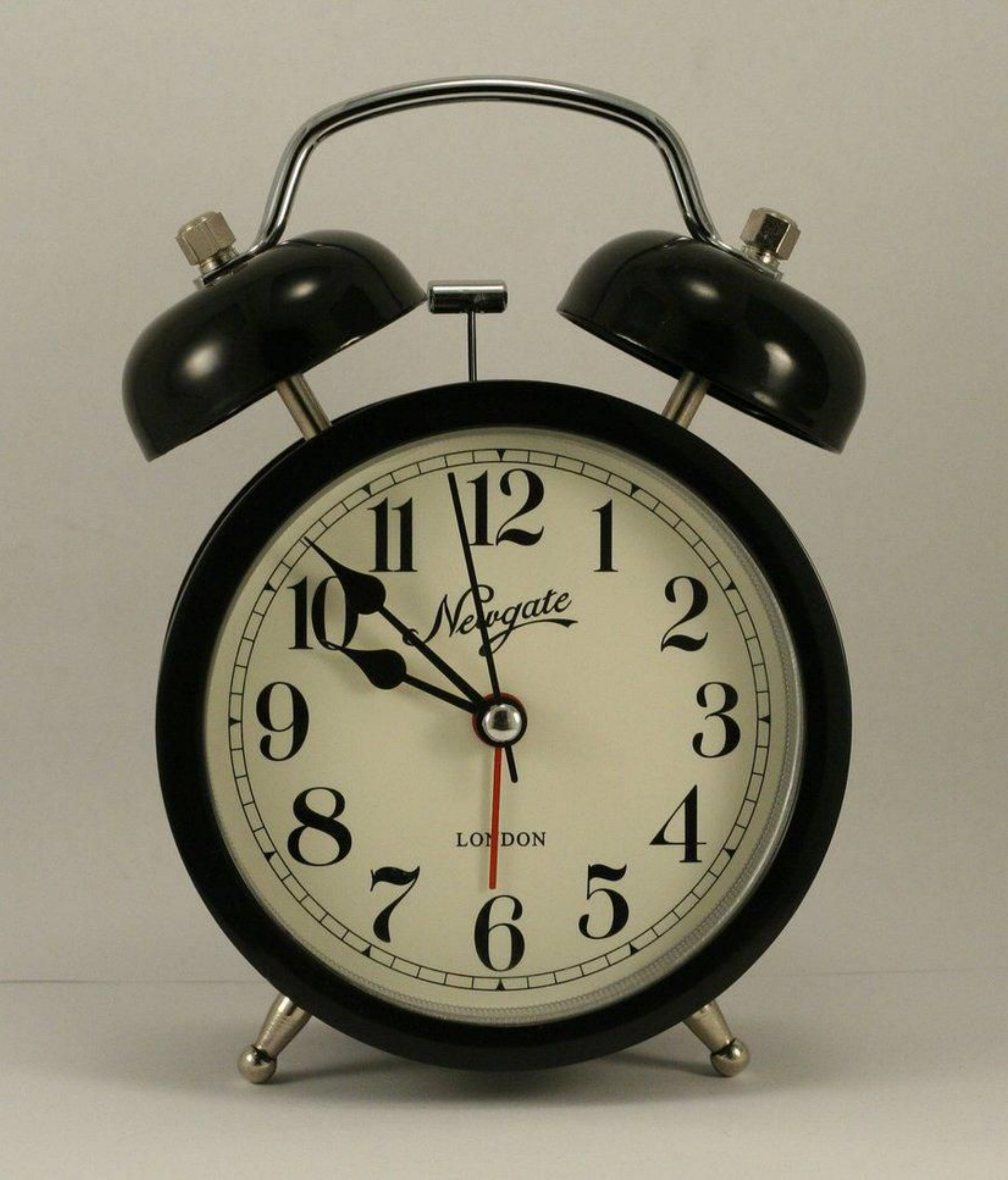} &
\includegraphics[width=0.20\linewidth]{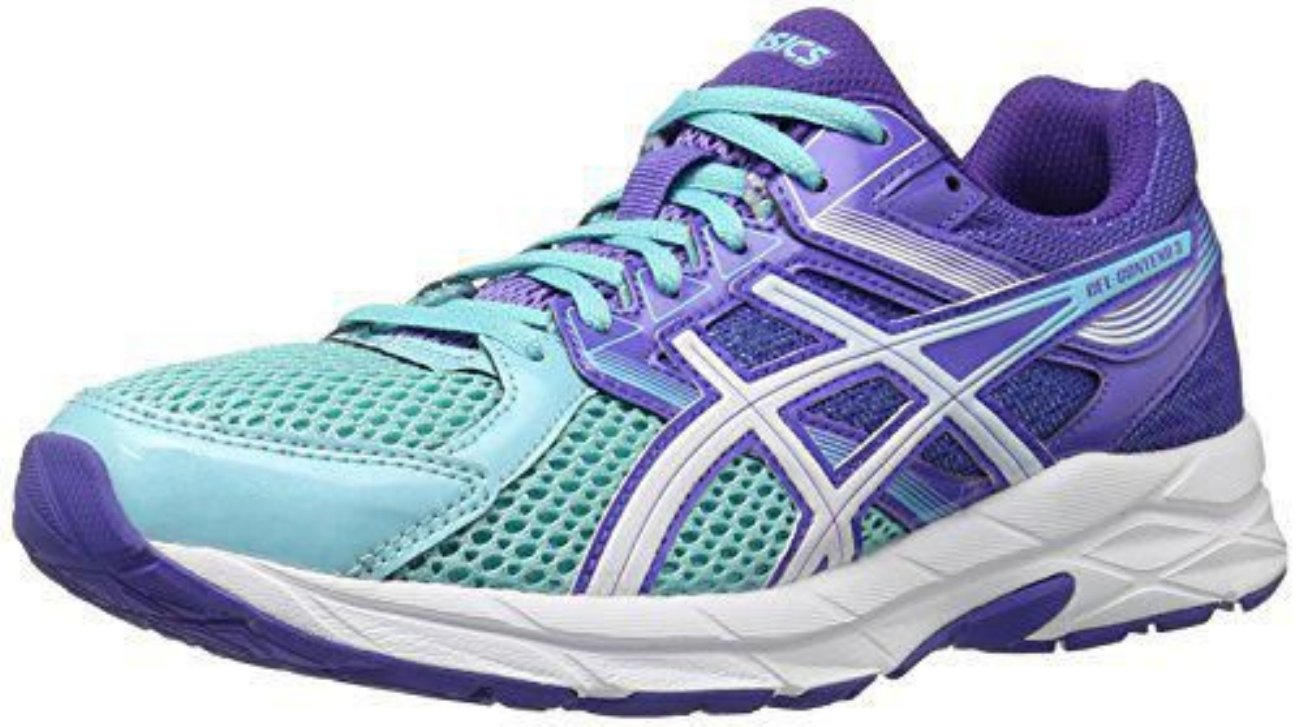}\\
\hline
style image & \multicolumn{2}{c|}{stylized images}\\ \hline
\includegraphics[width=0.20\linewidth]{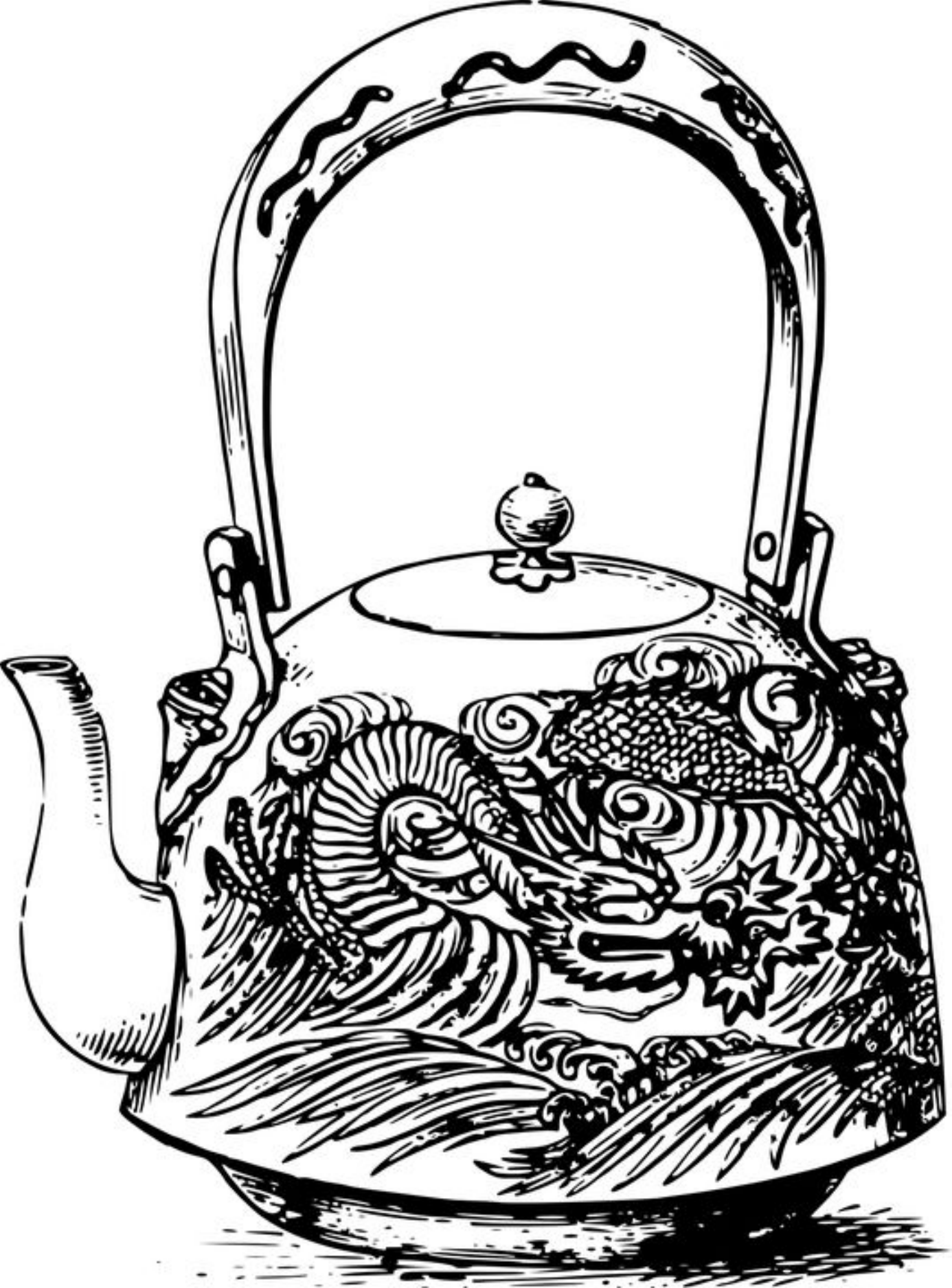} &
\includegraphics[width=0.20\linewidth]{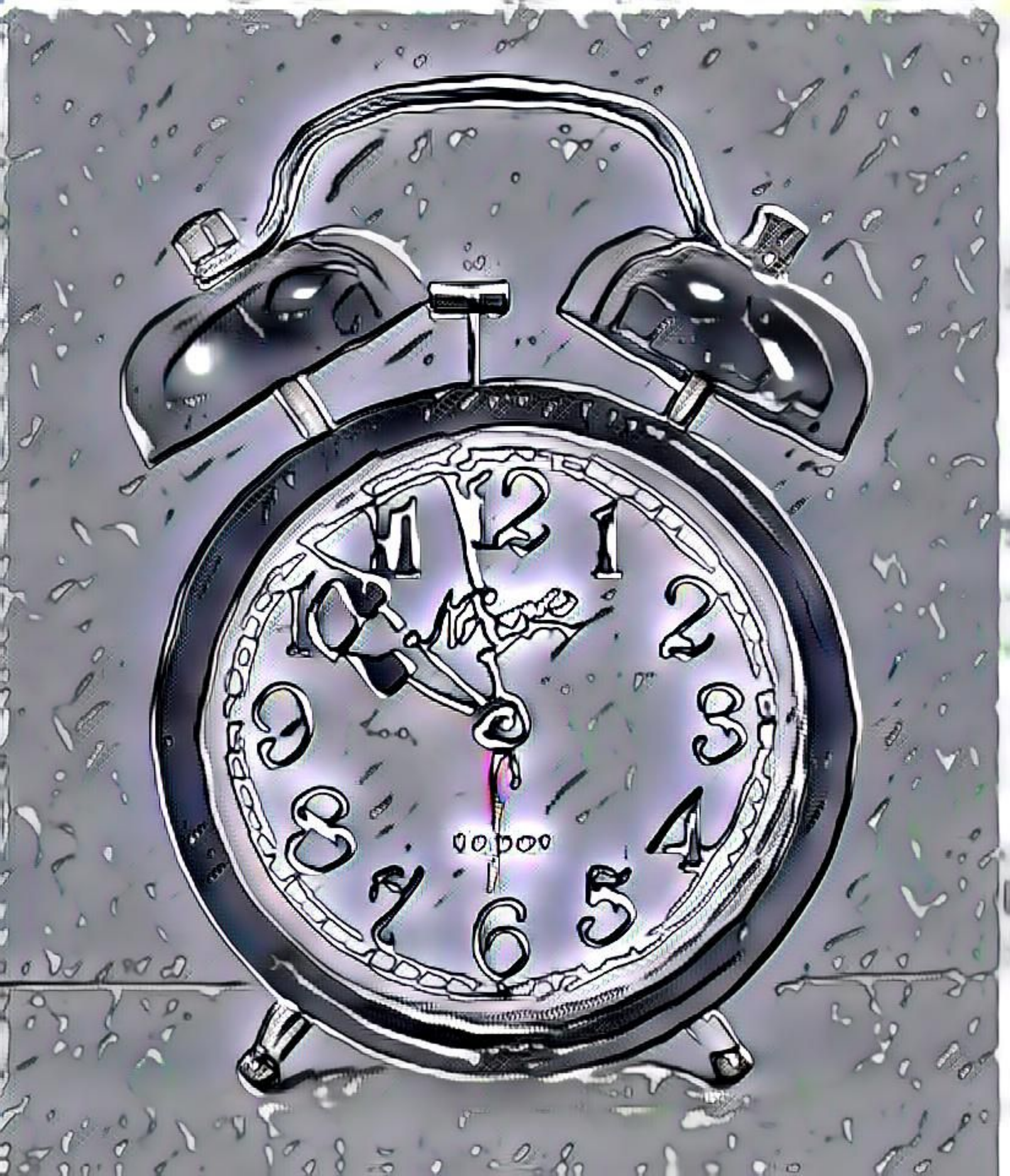} &
\includegraphics[width=0.20\linewidth]{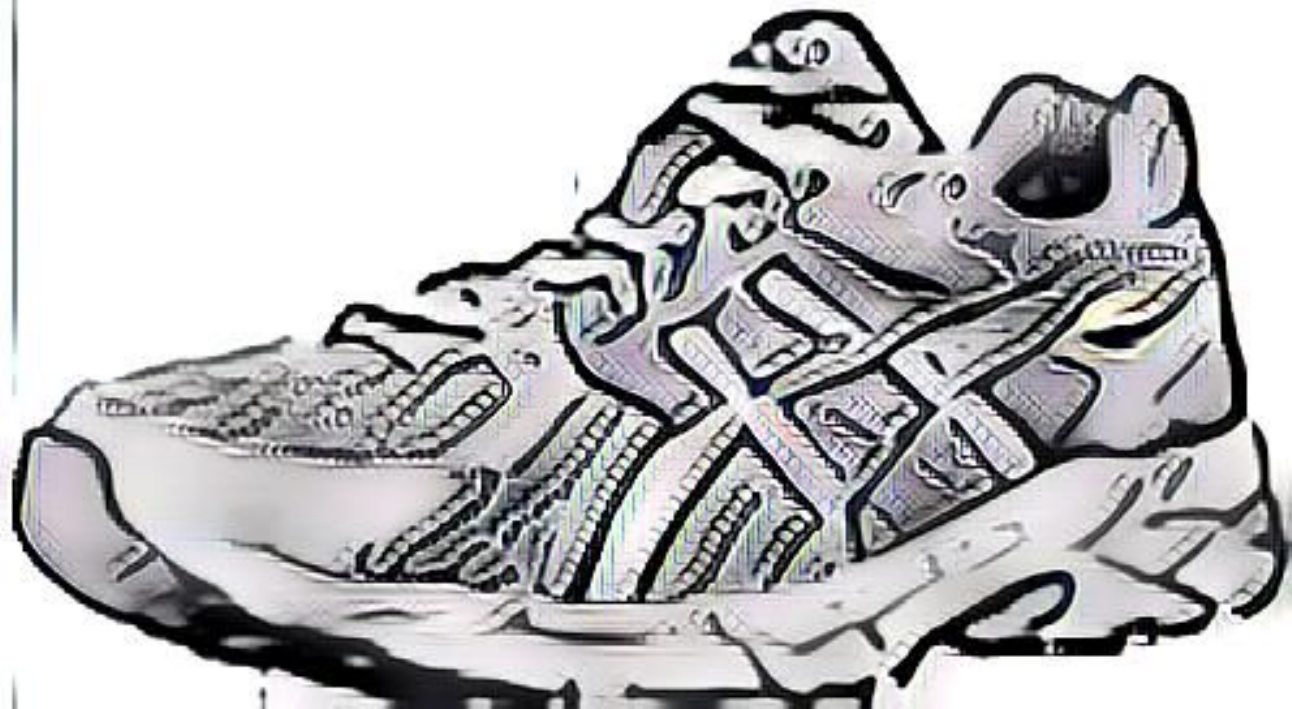} \\
\hline
\includegraphics[width=0.20\linewidth]{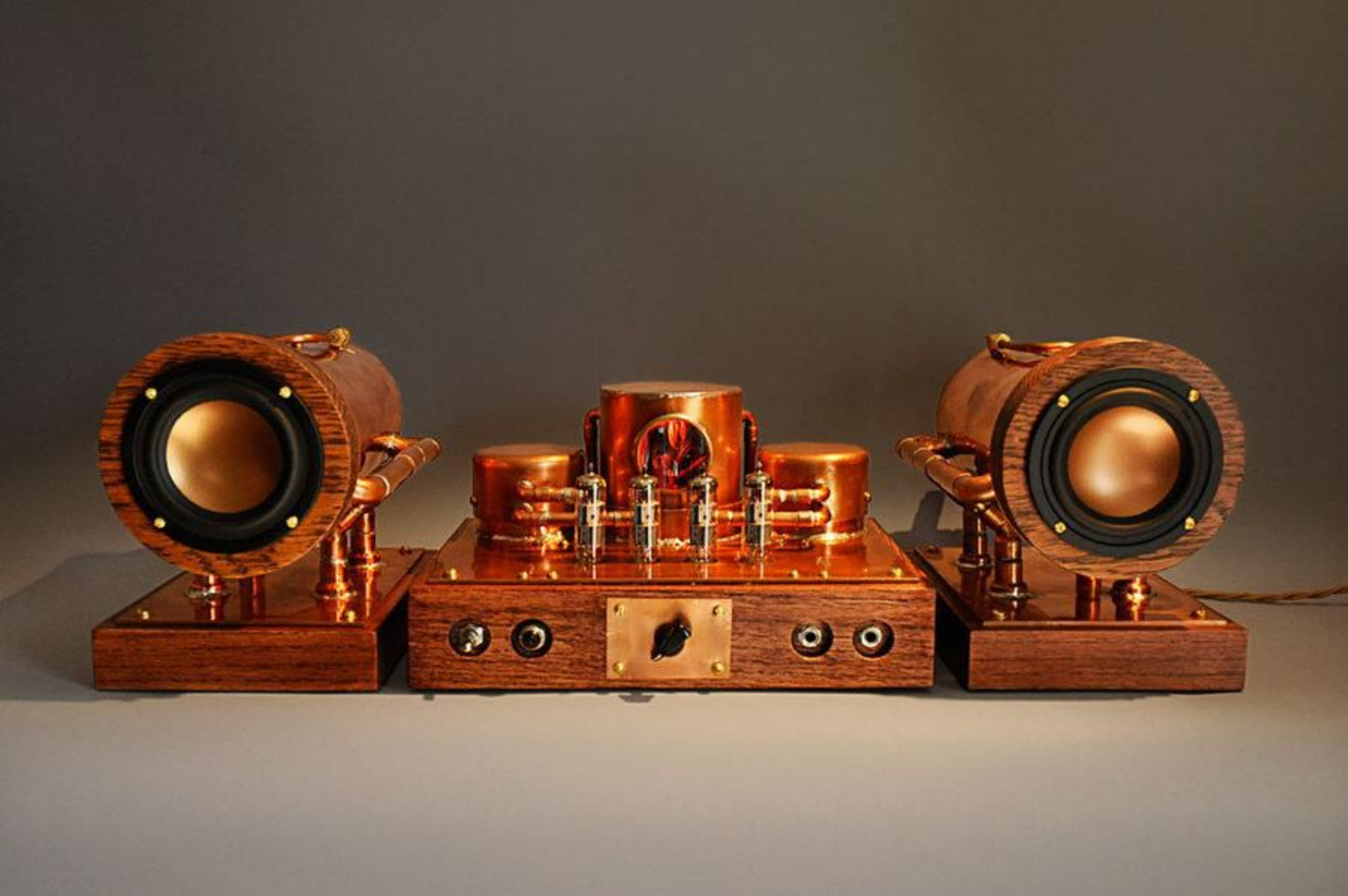} &
\includegraphics[width=0.20\linewidth]{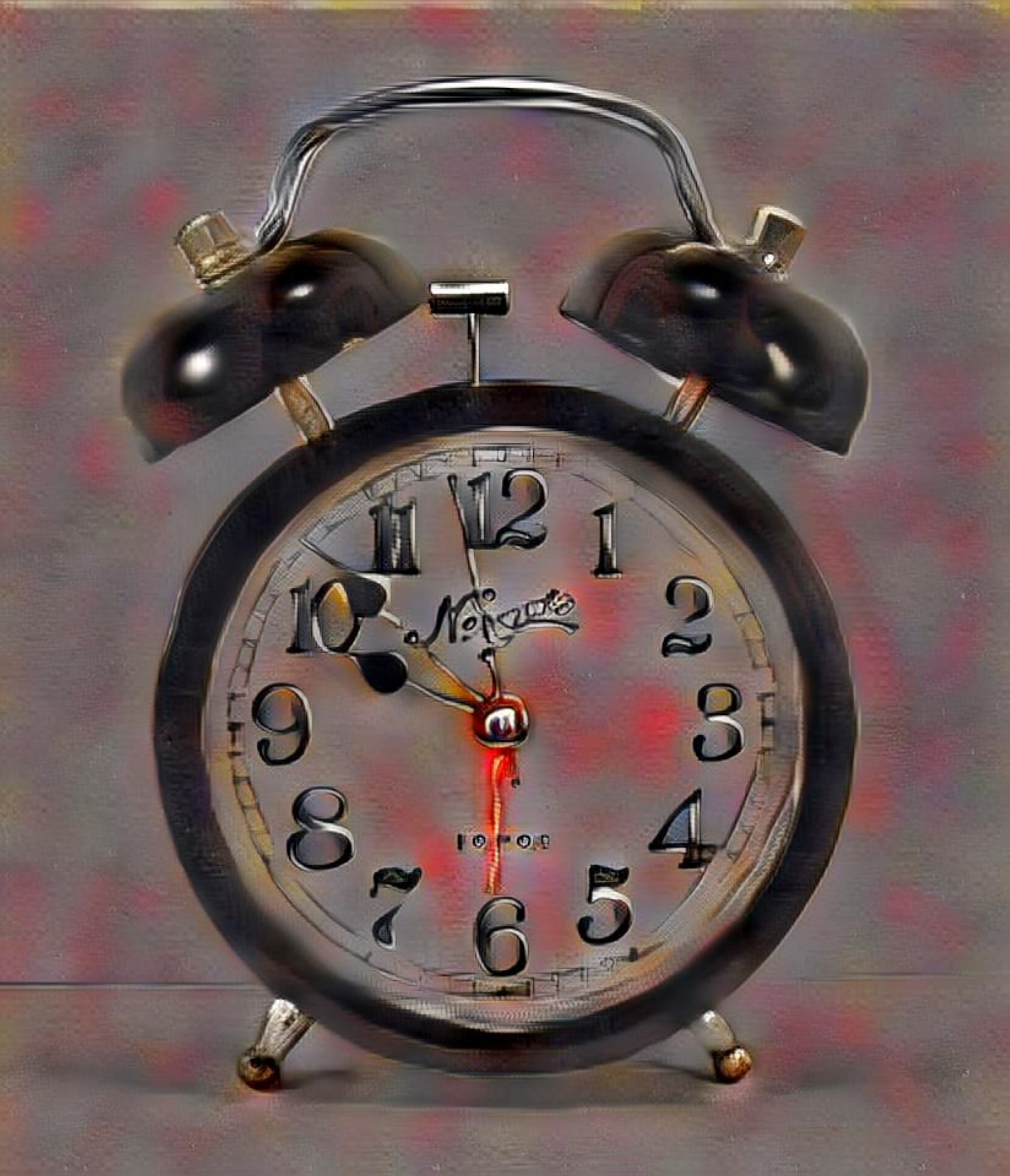} &
\includegraphics[width=0.20\linewidth]{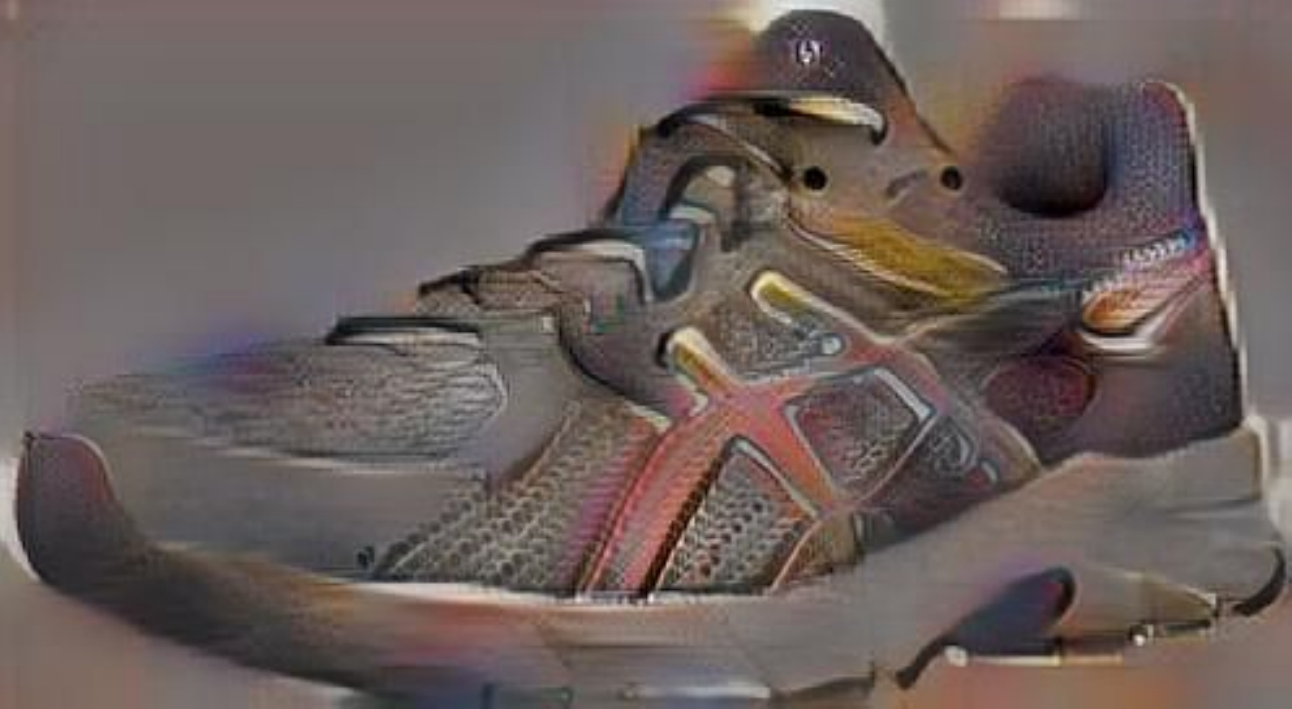} \\
\hline
\includegraphics[width=0.20\linewidth]{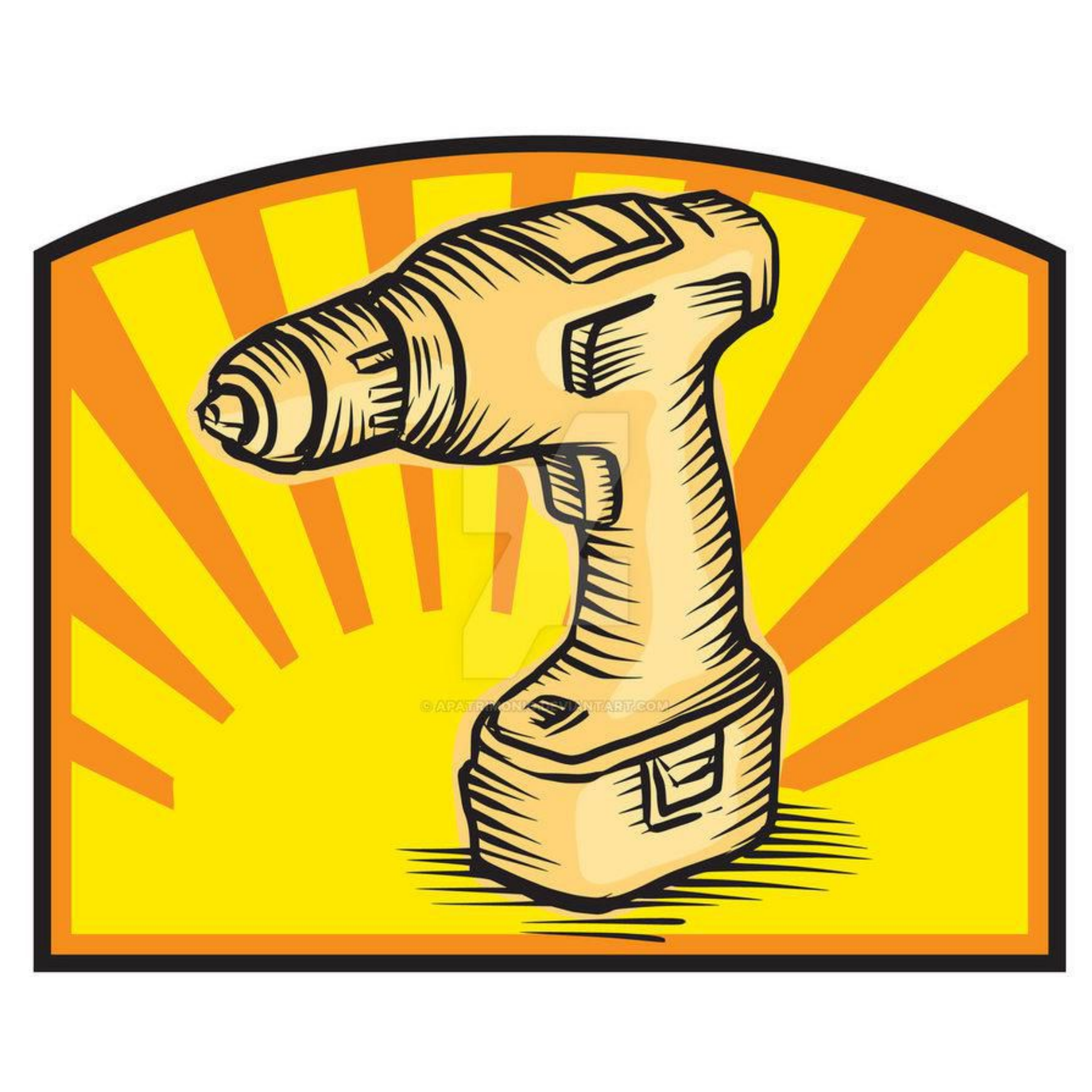} &
\includegraphics[width=0.20\linewidth]{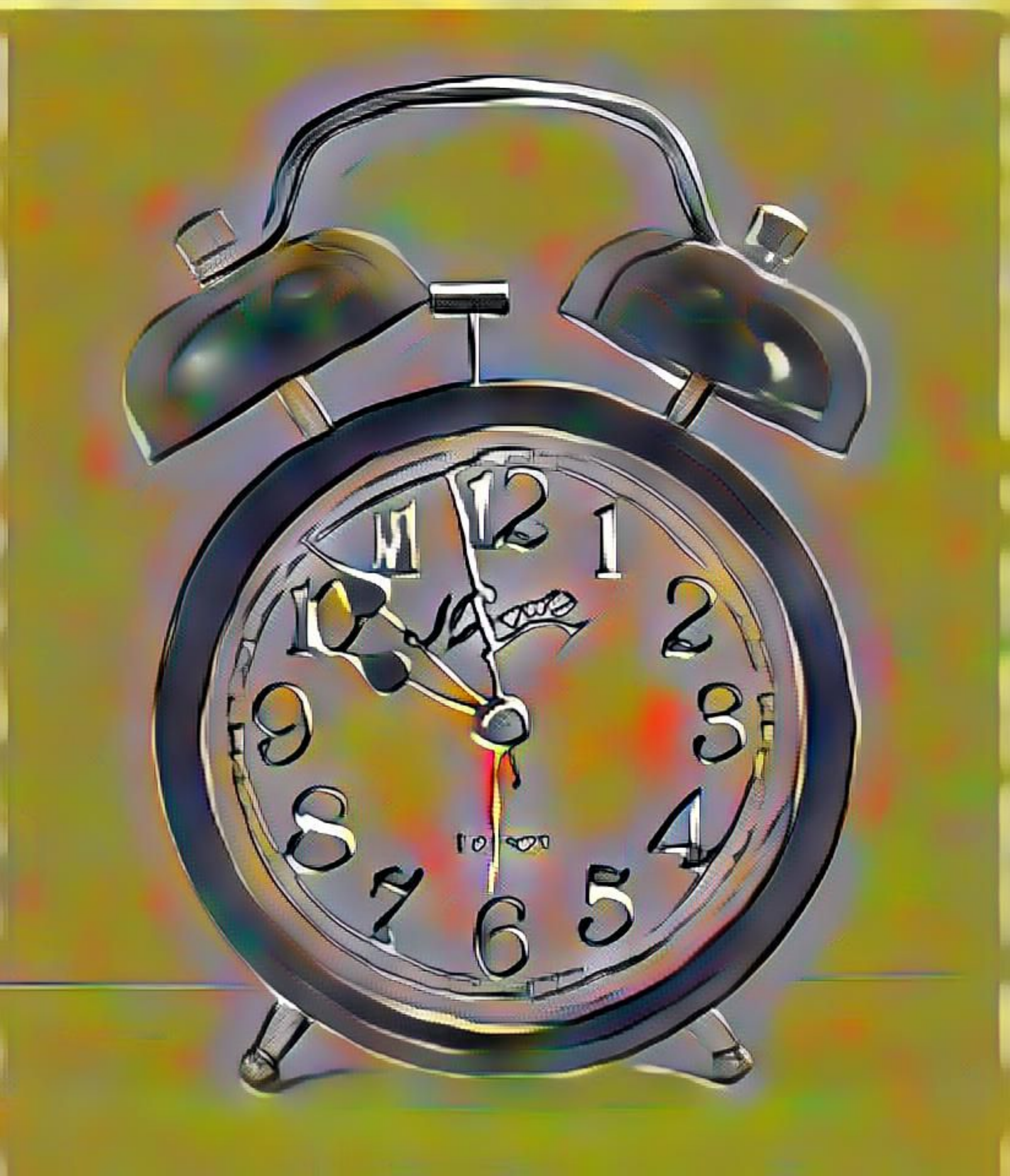} &
\includegraphics[width=0.20\linewidth]{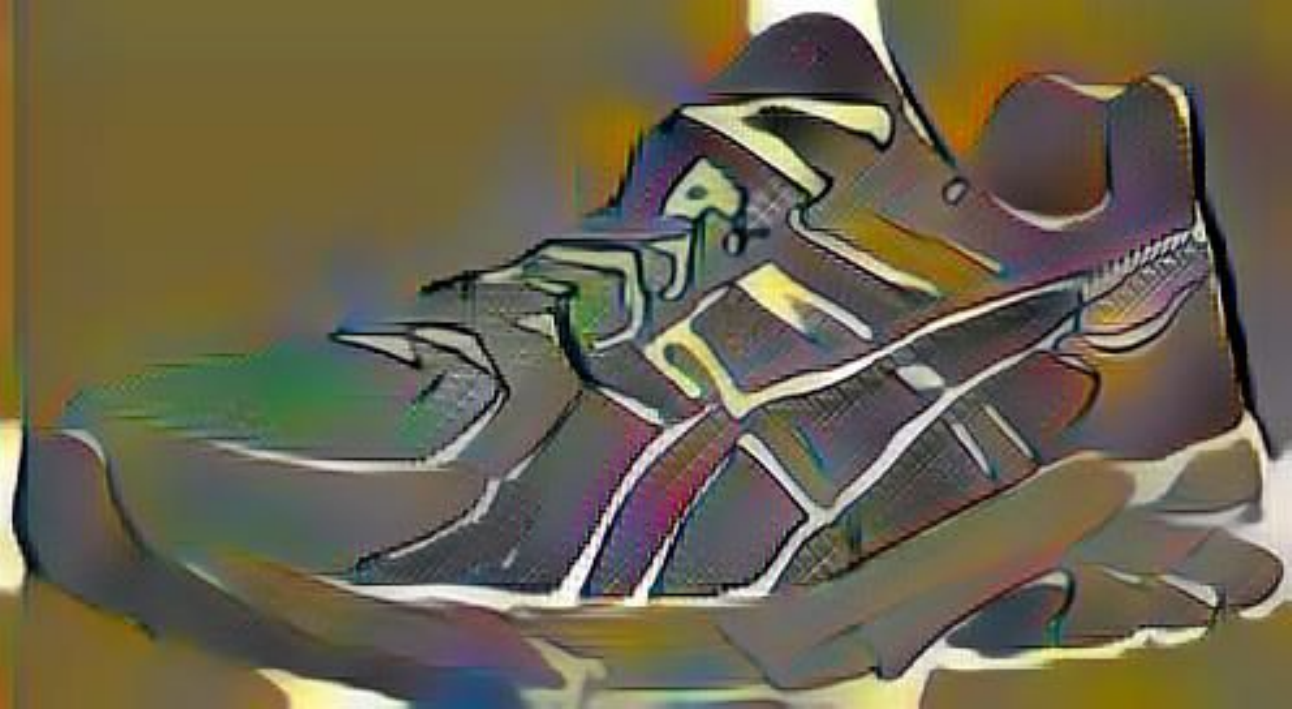} \\
\hline
\includegraphics[width=0.20\linewidth]{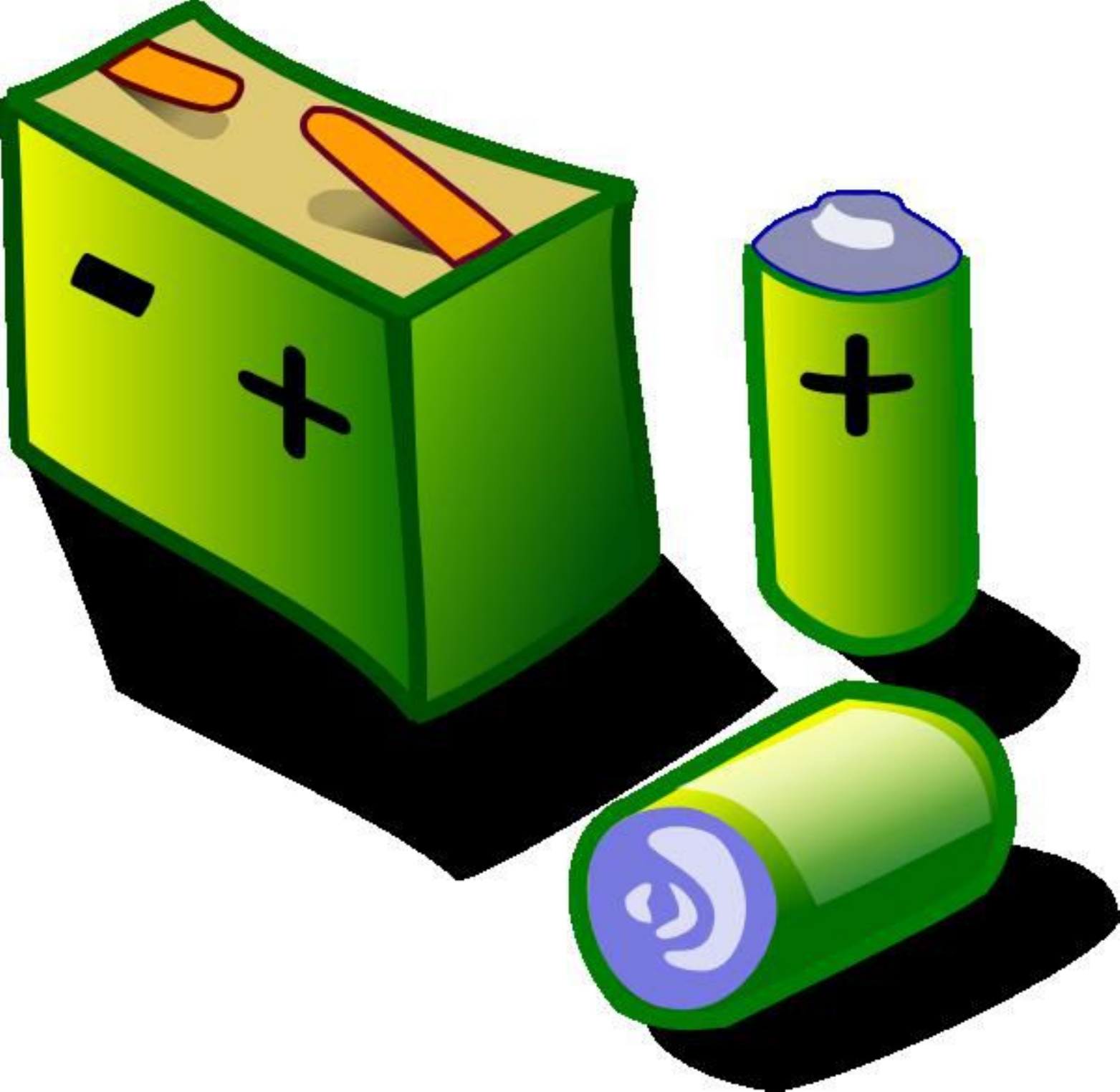} &
\includegraphics[width=0.20\linewidth]{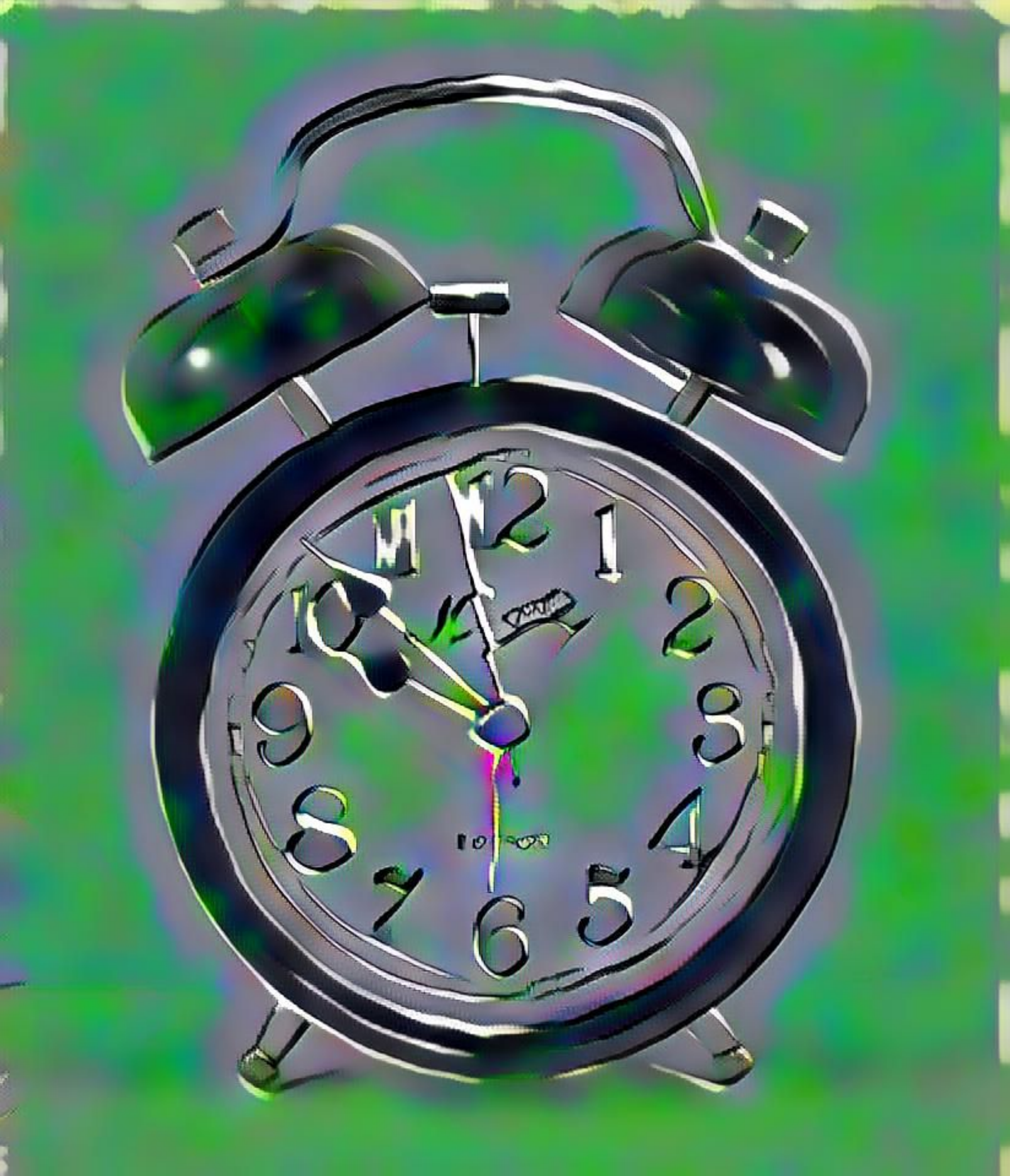} &
\includegraphics[width=0.20\linewidth]{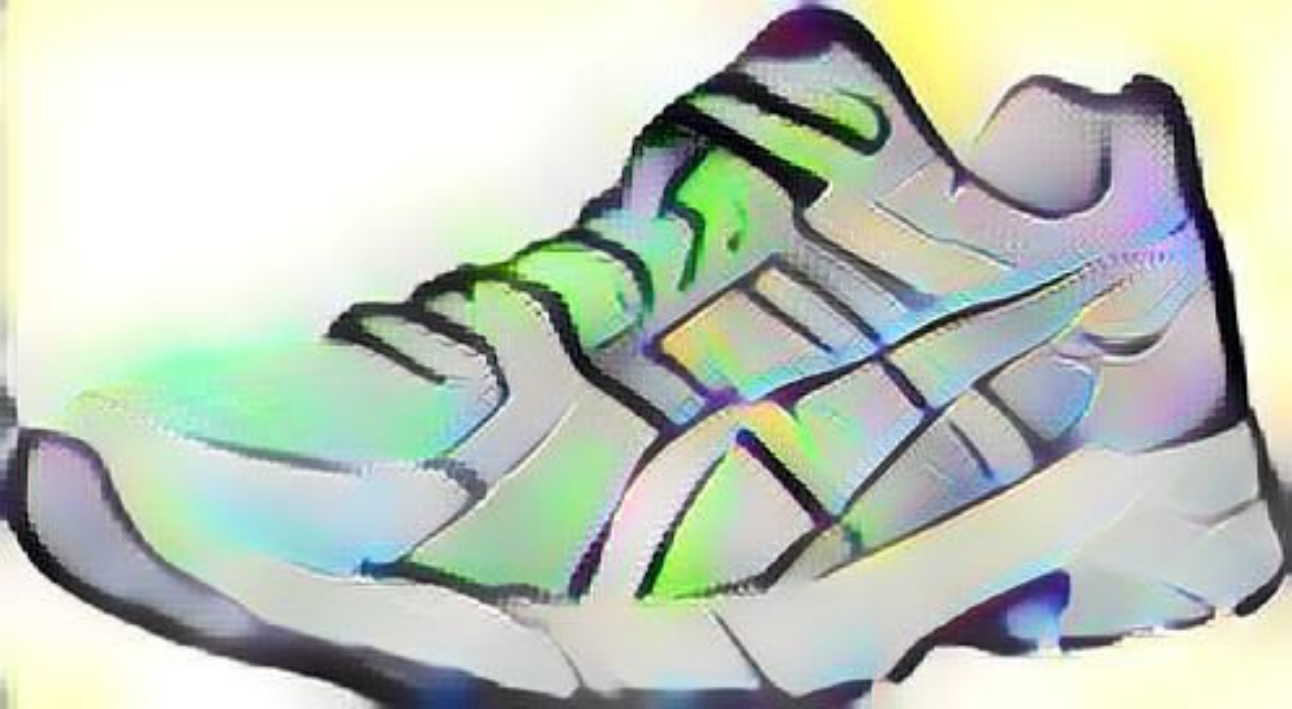} \\
\hline
\end{tabular}

\caption{Source augmentation by style transfer allows to generate a large amount of different variations from each image, borrowing the style from any other image and by keeping the original semantic content. 
All the images are taken from OfficeHome dataset and the style transfer is performed using AdaIN. 
}
\label{rethinking:fig:examples}
\end{figure}

\noindent$\bullet$\hspace{2mm} \textbf{A simple and effective style transfer data augmentation approach for domain generalization}. We show how the method AdaIN \cite{Huang_2017_ICCV_adain}, that is able to perform style transfer in real time, can be re-purposed for data augmentation, combining semantic and texture information of the available source data (see Figure \ref{rethinking:fig:examples}). The extended training set allows to get top target results, outperforming existing state of the art approaches.\\

\noindent$\bullet$\hspace{2mm} \textbf{We designed tailored strategies to integrate for the first time style transfer data augmentation with the current state of the art approaches}. The obtained results indicate that the original advantage of those methods almost always disappears when compared with the data augmented baseline. \\

The scenario described by this analysis clearly suggests the need of rethinking domain generalization baselines. On one side simple data augmentation strategies should be envisaged to increase source data variability compatible with orthogonal feature and model generalization approaches. On the other, new cross-source adaptive strategies should be designed to build over images generated by style transfer approaches.

\subsection{Source augmentation by style transfer}

We focus on the multi-source domain generalization setting where $S =\{S_1,\ldots,S_n\}$ denotes the $n$ available data sources with the respective $\{x^s_i,y^s_i\}_{i=1}^{N_s}\in S$ samples, where $y_i$ specifies the object classification label of its $x_i$ image. The main goal is to generalize to an unknown target database  $\{x^t_i,y^t_i\}_{i=1}^{N_t}\in T$, where $T$ shares with $S$ the same set of categories, while each source and the target are drawn from different marginal distributions. 
As indicated by the chosen notation, we disregard the specific domain annotation of each source sample, meaning that we do not need to know the exact source origin of each image. 

We indicate with $C(x^s, \theta_c)$ a basic deep learning classifier parametrized by $\theta_c$  and trained on the source data by minimizing the standard cross-entropy loss $\mathcal{L}(C(x^s, \theta_c), y^s)$. To increase data variability we study how to augment each sample $x^s$ by keeping its semantic content and changing the image style, borrowing it from the other available source data. The 
stylized sample $\tilde{x}^s$ 
obtained from $x^s$ inherits its label $y^s$ and enriches the training set, possibly making the model learned by optimizing $\mathcal{L}(C(\tilde{x}^s, \theta_c), y^s)$ more robust to domain shifts.
Thus, our analysis will consider a two step process, where a deep model $A$ parametrized by $\theta_a$ is first learned on the source data to perform style transfer $x^s \rightarrow \tilde{x}^s = A(x^s, \theta_a)$, and then it is used to perform data augmentation at runtime while learning to classify the image object content. 

\subsubsection{Training the Style Transfer Model} 
To implement $A$ we use AdaIN \cite{Huang_2017_ICCV_adain}, a simple and effective encoder-decoder-based approach that allows style transfer in real time. The encoder $E$ extracts representative features $f_c, f_s$ respectively from the content and the style image, the first are then re-normalized to have the same channel-wise mean and standard deviation of the second as follows:
\begin{equation}
    f_{cs}= \sigma(f_s)\left(\frac{f_c-\mu(f_c)}{\sigma(f_c)}\right) + \mu(f_s)~.
\end{equation}
Finally, the obtained feature $f_{cs}$ is mapped back to the image space through the decoder $D$ minimizing two losses: 
\begin{equation}
\mathcal{L}_{A} = \mathcal{L}_c + \lambda\mathcal{L}_s~.
\end{equation}
Both the losses measure the distance between the features re-extracted through the encoder $E(D(f_{cs}))$ from the stylized output image, and $f_{cs}$. Specifically $\mathcal{L}_c$ focuses on the content information considering the whole final feature output, while $\mathcal{L}_s$ focuses on the style information, measuring the difference of mean and standard deviation of the Relu output of several encoder layers. 

The method has two main hyperparameters $\theta_a=\{\lambda, \alpha\}$. The first controls the degree of the style transfer during training by adjusting the importance of the style loss and is generally kept fixed at $\lambda=10$. The second allows a content-style trade-off at test time by interpolating between the feature maps that are fed to the decoder with $f_{cs\alpha}=D((1-\alpha)f_c + \alpha f_{cs})$. When $\alpha=0$ the network tries to reconstruct the content image, while when $\alpha=1$ it produces the most stylized image.

\subsubsection{Style Transfer as Data Augmentation} 
As already mentioned above, we disregard the source domain labels, thus when training our object classifier $C$ the data batches contain samples extracted from all the source domains.  
Each sample in a batch has the role of content image and any of the remaining instances in the same batch can be selected randomly to work as style provider. In this scenario stylization can happen both from images of the same source domain (e.g. two photos) or from images of different domains (e.g. a photo and a painting). To regulate this process we use a stochastic approach with the transformed image $\tilde{x}^s$ replacing its original version $x^s$ with probability $p$.

We highlight that the described procedure for the application of AdaIN differs from what appeared in previous works. Indeed, both the original approach \cite{Huang_2017_ICCV_adain} and its use for data augmentation in \cite{zhang2020learning}, exploit the style transfer model trained on MS-COCO \cite{mscoco} as content images and paintings mostly collected from WikiArt \cite{wikiart} as style images. In our study we do not allow extra datasets besides those directly involved in the domain generalization task as source domains.

\subsection{Experiments}

We designed our experimental analysis with the aim of running a thorough evaluation of the impact of style transfer data augmentation on domain generalization. Besides observing how this data augmentation can improve the standard learning baseline model, and how it compares with the most recent state of the art DG methods, we are also interested in the effectiveness of their combination. 
In the following we provide details on the chosen data testbeds and sota models, describing how the data augmentation strategy is integrated in each approach.

\subsubsection{Experimental Setting}

We test our method on PACS, VLCS and OfficeHome benchmarks.

We apply the same experimental protocol of \cite{jigsawCVPR19}: the predefined full training data is randomly partitioned in train and validation sets with a 90-10 ratio. The training is performed on the train splits of the 3 source domains while the validation splits are used for model selection. At the end the model is tested on the predefined test split of the left out domain. This split has been defined randomly selecting 30\% of images of the overall dataset.

All our results are obtained by performing an average over 3 runs. In the case of both OfficeHome and VLCS the random 90-10 train-val split was repeated for each run.

\input{Rethinking/PACS_table}

\subsubsection{Comparison methods}
We want to show that \textit{source augmentation by style transfer} does not only allow to obtain performances higher than previous more complex methods, but more importantly that current state of the art methods, designed and built on a baseline that does not take into account the source augmentation by style transfer, loose their effectiveness when applied on this new stronger baseline. For this reason 
we have chosen a number of current state of the art approaches and integrated the source augmentation style transfer into them. 
We performed our choice considering methods that employ different approaches to deal with the Domain Generalization setting. For each of these methods we carefully designed the integration of the source augmentation by style transfer in order to not undermine their specific strategies
For our study we consider as main Baseline a classification model learned on all the source data and na\"{\i}vely applied on the target. We indicate with Original the standard data augmentation with horizontal flippling and random cropping, while we use Stylized to specify the cases where we add style transfer data augmentation.
The behavior of four among the most recent DG methods is evaluated under both these augmentation settings. We dedicate a particular attention to the integration of the style transfer data augmentation strategy with each of the considered approaches. The goal is getting the most out of them without undermining their nature. In particular, considering that the style transfer leads to domain mixing, it is important to not integrate it in procedures that need a separation among source domains.

\vspace{4mm}\noindent\textbf{DG-MMLD \cite{dg_mmld}}

\vspace{4mm} This approach exploits clustering and domain adversarial feature alignment. Since it does not need the source domain labels, the integration of the proposed style transfer data augmentation is straightforward: styles of random images are applied to each content images (inside a batch) with probability $p$, exactly as done for the Baseline.

\vspace{4mm}\noindent\textbf{Epi-FCR \cite{episodic_hospedales}}

\vspace{4mm} Epi-FCR is a meta-learning method which splits the network in two modules, each one is trained by pairing it with a partner that is badly tuned for the domain considered in the current learning episode. The modules are the feature extractor and the classifier which alternatively cover the two roles of learning part and bad reference. After this phase, a final model is learned by integrating the trained modules together with a random classifier used as regularizer. In the first stage, knowing the source domain labels is crucial to choose and set the two network modules, thus mixing the domains with style transfer augmentation could degrade its performance. In the ending stage instead, all the source data are considered together: we applied here the style data augmentation. 

\vspace{4mm}\noindent\textbf{DDAIG \cite{zhou2020deep}}

\vspace{4mm}This is a data augmentation strategy based on a transformation network which is trained so that every synthesized sample keeps the same label of the original image, but fools a domain classifier. In the learning procedure the transformation module, the label classifier and the domain classifier are iteratively updated. In particular the label classifier is trained on all the source data, both original and synthetic: we further extended this set with style transfer augmented data. 

\vspace{4mm}\noindent\textbf{Rotation \cite{lopez_rotation}}

\vspace{4mm} It has been shown that self-supervised knowledge supports domain generalization when combined with supervised learning in a multi-task model. In particular we focused on rotation recognition, where the orientation angle of each image should be recognized among $\{0^\circ, 90^\circ, 180^\circ, 270^\circ\}$. The model minimizes a combination of the supervised and self-supervised loss with linear weight $\sigma$ generally kept lower than 1 to let the supervised model guide the learning process.
In this case the domain labels are not used during training, so the application of the source augmentation by style transfer is straightforward.

\vspace{4mm}\noindent\textbf{Mixup \cite{zhang2018mixup}}

\vspace{4mm} An approach related to data augmentation, originally defined to improve generalization in standard in-domain learning, is Mixup: it interpolates samples and their labels, regularizing a neural network to favor a simple linear behavior between training examples. Its hyper-parameter $\gamma \in \{0,\infty\}$ controls the strength of interpolation between data pairs, recovering the Baseline for $\gamma=0$. In our study we consider Mixup as further reference, and in particular we tested data mixing both at pixel and at feature level \cite{xu2020adversarial}.

\subsubsection{Training Setup}

Our style transfer model $A$ is trained on source data before training the classification model $C$. As already mentioned, $A$ is implemented by AdaIN \cite{Huang_2017_ICCV_adain} and is therefore based on a VGG backbone. It is trained for 20 epochs with a learning rate equal to 5e-5. The hyperparameters $\alpha$ and $p$ used in each experiment are specified in the caption of the respective result tables and in depth analysis on the sensitivity of the method to them is presented in Section \ref{rethinking:sec:sensitivity}.

For the classification model $C$ we use AlexNet and ResNet18 backbones. Specifically, Baseline, Rotation and Mixup are trained using SGD with $0.9$ momentum for $30k$ iterations. We set the batch size to $32$ images per source domain: since in all the testbed there are three source domains each data batch contains $96$ images. The learning rate and the weigh decay are respectively fixed to $0.001$ and $0.0001$. Regarding the hyperparameters of the individual algorithms, we empirically set the Rotation auxiliary weight to $\sigma = 0.5$ and for Mixup $\gamma= 0.4$.

We implement Rotation by adding a rotation recognition branch to our Baseline.
For DG-MMLD, Epi-FCR and DDAIG, we use the code provided by the authors integrating different datasets/backbones where needed. The training setup for these experiments is the one defined in their papers for both the Original and Stylized version. 
We report the previously published results whenever possible. In the following we will indicate with a star ($^*$) the results we obtained by running the authors' code.

\subsubsection{Results analysis}

\input{Rethinking/OfficeHome_table}

\input{Rethinking/VLCS_table}

Table \ref{rethinking:tab:pacs} shows results on PACS benchmark with both AlexNet and ResNet18 backbones. 
We get two main outcomes. (1) There is an evident improvement of more than 5 percentage points in the Baseline performance when using the stylized augmented source data with respect to the original case.
Looking at the results for the different domains we can see that improvement is higher for Art Painting, Cartoon and Sketch, than in Photo. 
(2) All the considered state of the art DG methods benefit from the source augmentation. Indeed in absolute terms their performance grows, but at the same time they loose in effectiveness as they cannot outperform the Baseline any more. 

Table \ref{rethinking:tab:officehome} shows results on OfficeHome dataset with ResNet18 backbone. Even if in this case the improvement produced by the source augmentation by style transfer is more limited, the results confirm what we have already observed for PACS. The Stylized Baseline obtains the best accuracy outperforming the competitor state of the art methods, even when those are improved using the same source augmentation.

Table \ref{rethinking:tab:vlcs} reports results on VLCS benchmark with AlexNet backbone. This dataset is particularly challenging and shows a fundamental limit of tackling DG through style transfer data augmentation. Since the domain shift is not originally due to style differences in this testbed, source augmentation by style transfer does not support generalization.

As a final remark, we focus on Mixup. The results over all the considered datasets show that it is not able to generalize across domains and it might perform even worse that the Original Baseline. Between the two considered pixel and feature variants, only the second shows some advantage on PACS, so we focused on it in the other tests. Still, its results remain lower than those obtained by the DG methods both with and without style based data augmentation.

\subsubsection{Analysis of AdaIN hyperparameters}
\label{rethinking:sec:sensitivity}
\begin{figure}
    \centering
    \includegraphics[width=\linewidth]{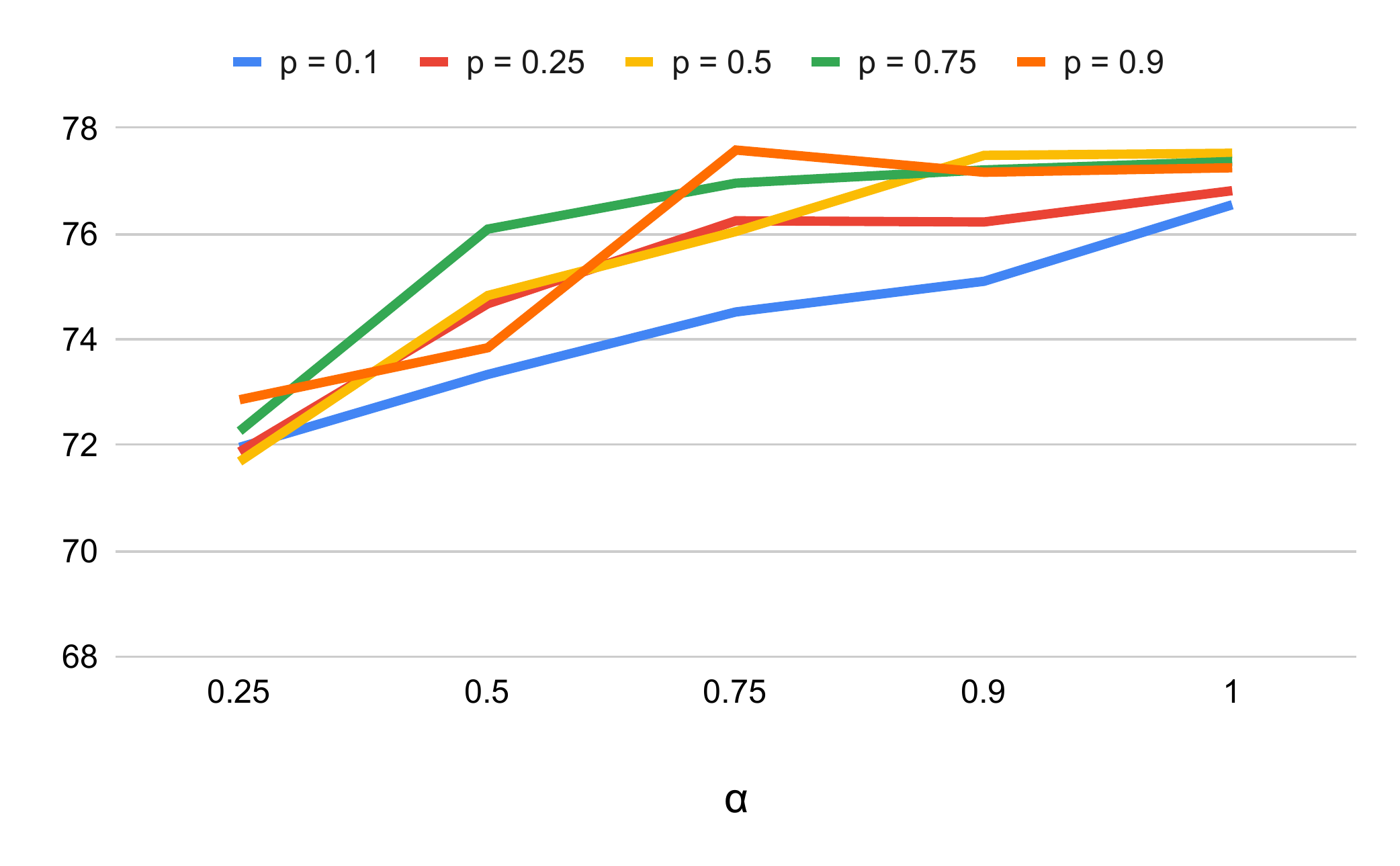}
    \caption{Average accuracy on PACS AlexNet with different values of $p$ when varying $\alpha$.} 
    \label{rethinking:fig:varying_alpha}
\end{figure}

\begin{figure}
    \centering
    \includegraphics[width=\linewidth]{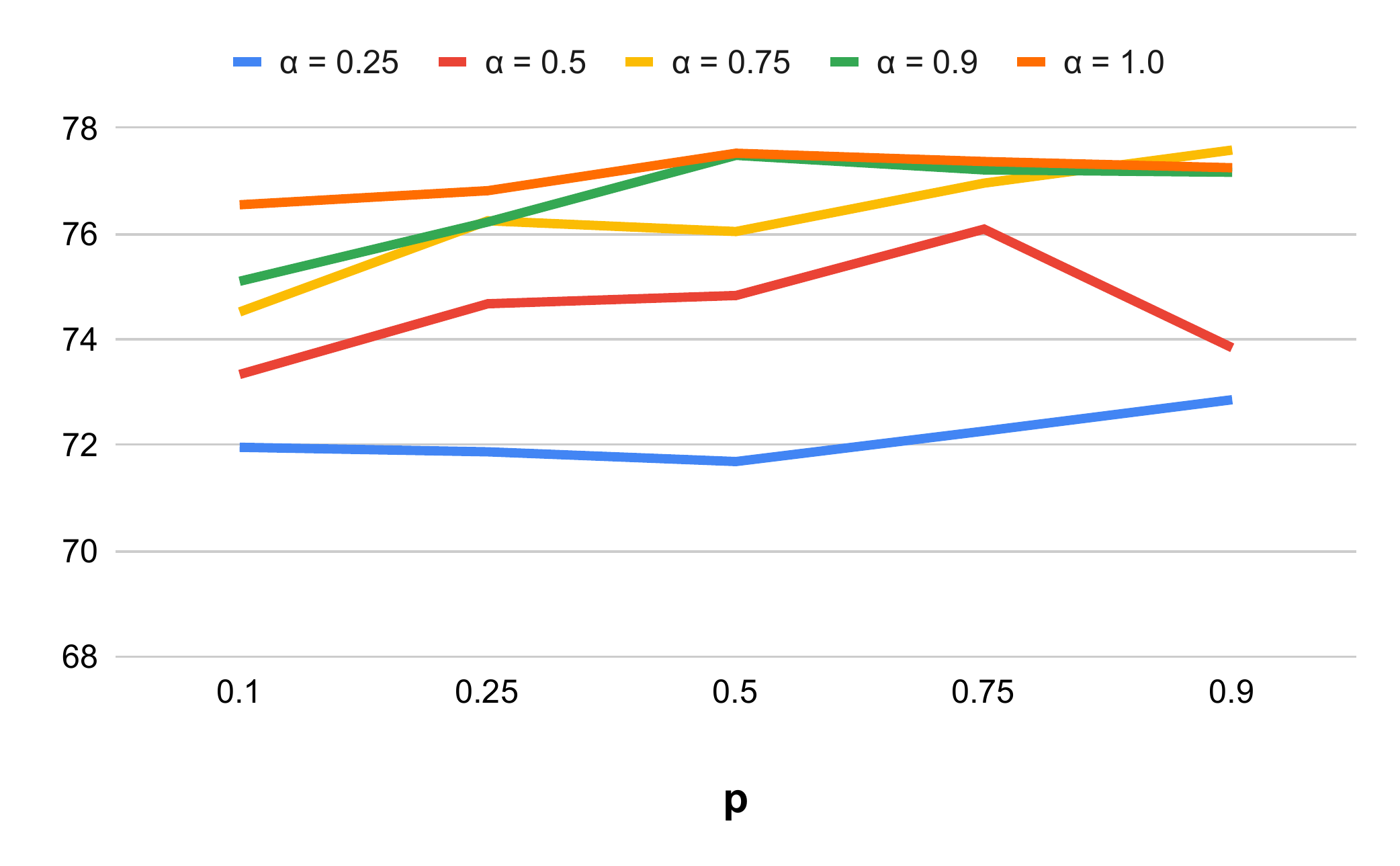}
    \caption{Average accuracy on PACS AlexNet with different values of $\alpha$ when varying $p$.}
    \label{rethinking:fig:varying_p}
\end{figure}

In Figures \ref{rethinking:fig:varying_alpha} and \ref{rethinking:fig:varying_p} we see how the PACS AlexNet results change when varying either $\alpha$ or $p$ by keeping the other fixed. With a low value of $\alpha$ the style transfer is too weak to produce an effective appearance change of the source sample and introduce extra variability. In general the best results are obtained using $\alpha = 1$ regardless of the specific value of $p$. 

For what concerns the value of $p$ we can see that, if $\alpha$ is high enough, even a small $p$ allows to obtain good performance with the best results obtained with $p=0.5$ or $p=0.75$.


\subsubsection{Style transfer from external data vs source data} 
The described procedure for the application of AdaIN differs from what appeared in previous works. Indeed, both the original approach \cite{Huang_2017_ICCV_adain} and its use for data augmentation in \cite{zhang2020learning}, exploit the style transfer model trained on MS-COCO \cite{mscoco} as content images, and paintings mostly collected from WikiArt \cite{wikiart} as style images. In our study we did not allow extra datasets besides those directly involved in the domain generalization task as source domains. The reason is twofold: first, we want to keep the method as simple as possible, without the need of relying on external data; second, to perform a fair benchmark with the competitors DG methods all of them should have access to the same source information.

\begin{figure}
\centering
\begin{tabular}{ccc}

\includegraphics[width=0.27\linewidth]{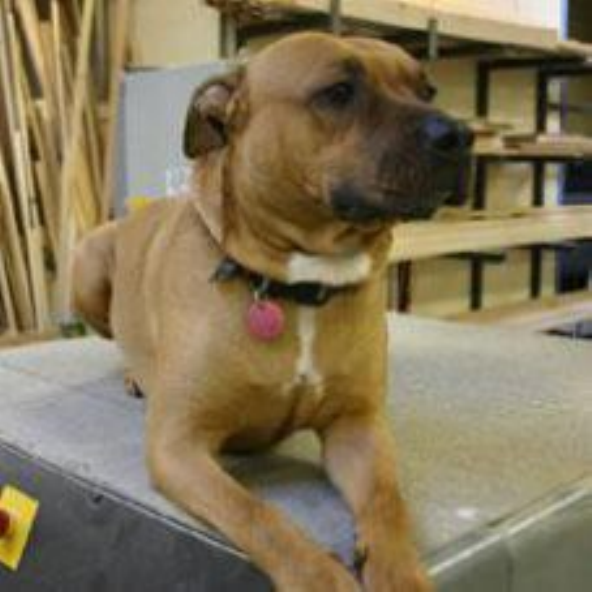} &
\includegraphics[width=0.27\linewidth]{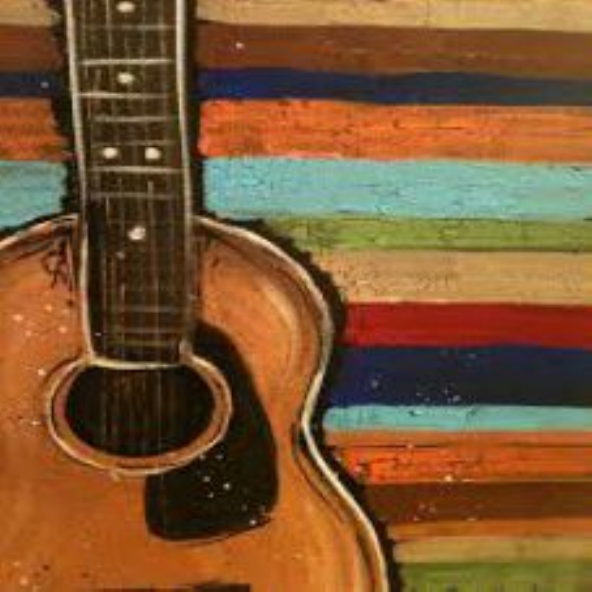} &
\includegraphics[width=0.27\linewidth]{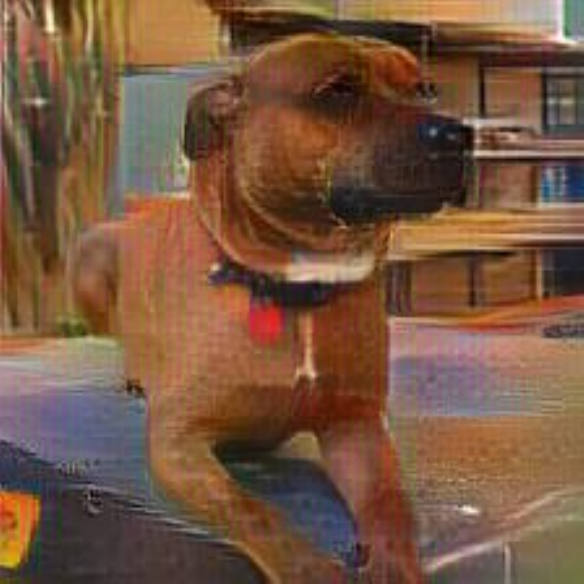}
\end{tabular}\\
\begin{tabular}{cccc}
\includegraphics[width=0.22\linewidth]{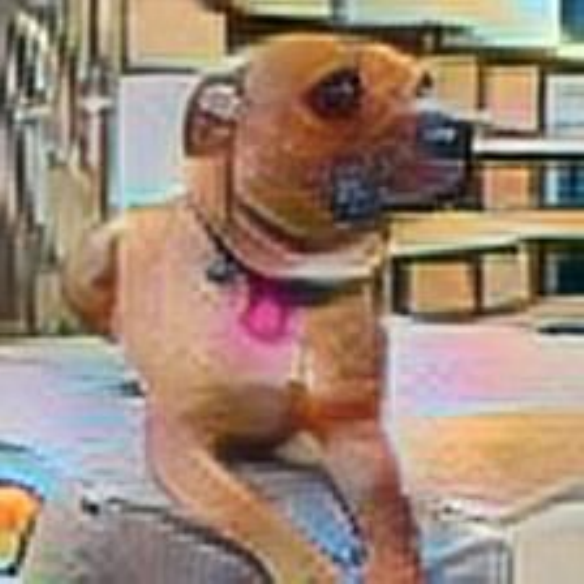} &
\includegraphics[width=0.22\linewidth]{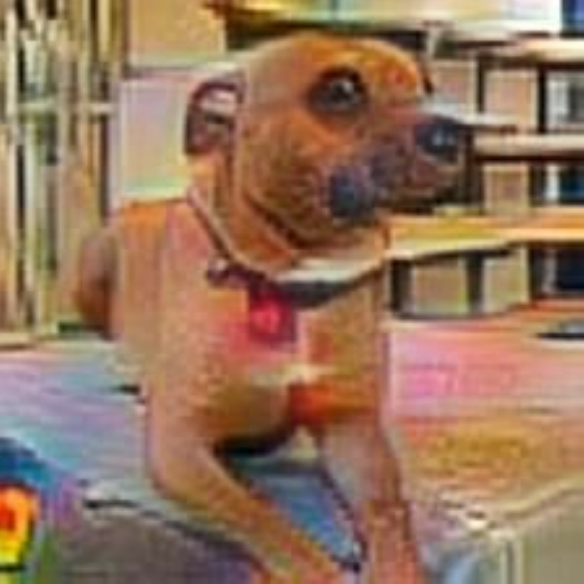} &
\includegraphics[width=0.22\linewidth]{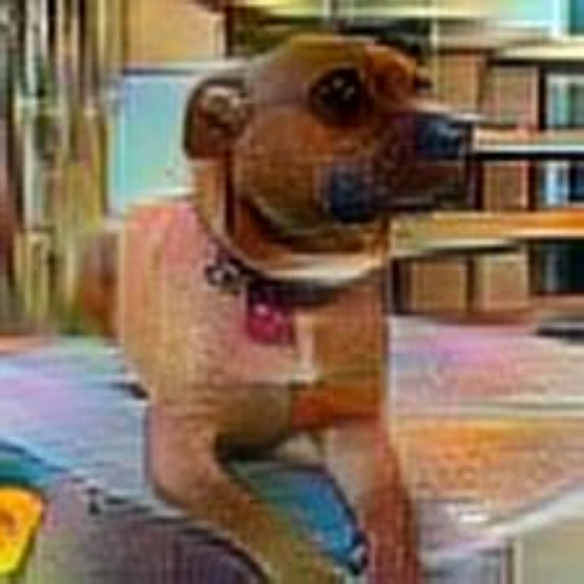} &
\includegraphics[width=0.22\linewidth]{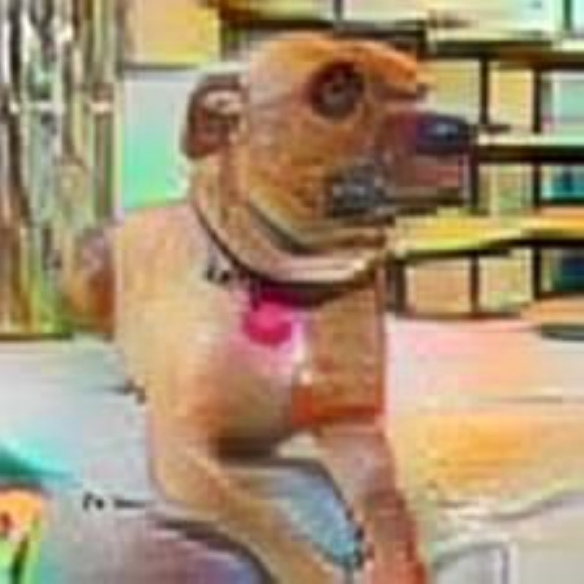}
\end{tabular}

\caption{Example of application of style transfer using AdaIN. The first image comes from the PACS Photo domain and is used as content while the second comes from PACS Art Painting domain and is used as style image. On top right there is the translation performed using AdaIN trained on MS-COCO and WikiArt images. In the second row we see the translations performed using our AdaIN models trained on source data only, respectively when the Art Paintings, Cartoon, Sketch and Photo domains are used as target.}
\label{rethinking:fig:transfer_comparison}
\end{figure}

Still, the interested reader may wonder what would be the effect of using the original AdaIN model trained on MSCOCO and WikiArt. Figure 
\ref{rethinking:fig:transfer_comparison} shows one example obtained in this way. Specifically we consider a dog image drawn from the PACS Photo domain and we analyse the images obtained by borrowing the style form the Art Painting guitar image. 
We compare the stylized sample produced with the MSCOCO-WikiArt AdaIN model against the outcomes of the four AdaIN variants trained on the source with every one of the four domains used as target.

As can be observed, the obtained results in terms of image quality are not so different and, as already indicated by the results discussed above, they are good enough to introduce variability in the source. 
We also run a quantitative analysis: in Table \ref{rethinking:tab:adain_training} we compare the performance of the our Stylized Baseline on PACS AlexNet with the analogous Baseline trained using the augmented data produced with the AdaIN MSCOCO-WikiArt pretrained model. The last one shows a slightly better accuracy which is though not significant if we consider the related standard deviation.

\begin{table}[tb]
    \centering
    \caption{Comparison of AdaIN training strategies}
    \resizebox{\textwidth}{!}{
    \begin{tabular}{c@{~~}|c@{~~}c@{~~}c@{~~}c@{~~}|c}
    \hline
    & Art Painting & Cartoon & Sketch & Photo & Average \\
    \hline
    Stylized Baseline & $71.96$ & $72.47$ & $76.47$ & $88.34$ & $77.31 \pm 1.1$ \\
    MSCOCO-WikiArt  Baseline & $73.00$ & $73.78$ & $76.37$ & $89.04$ & $\textbf{78.05} \pm 0.9$ \\    
    \hline
    \end{tabular}
    }
    \label{rethinking:tab:adain_training}
\end{table}

\subsection{Conclusions}

Among the current state of the art domain generalization methods some are based on data augmentation and use complex generative approaches, while other propose source feature adaptation and meta-learning strategies. Despite being orthogonal among each other, no previous work tried to integrate them. We investigated here a simple and effective style transfer data augmentation strategy for domain generalization and we showed how it overcomes its competitors. Moreover we designed proper combination of this approach with the most relevant existing DG approaches. Our experimental analysis indicates that the performance of the considered methods improves over the respective versions not including the style data augmentation, but surprisingly the methods lose their original effectiveness, not showing any improvement over the new data augmented baseline. 

As other concurrent technical reports \cite{gulrajani2020search}, our work suggests the need of a shading new light on domain generalization problems and calls for novel strategies able to take advantage of the data variability introduced by cross-domain style transfer. 

%% file: Rethinking/PACS_table.tex
\begin{table}
    \centering
    \caption{PACS classification accuracy (\%). We used AdaIN with $\alpha=1.0$ and $p=0.75$ for AlexNet-based experiments and AdaIN with $\alpha=1.0$ and $p=0.90$ for those based on ResNet18.} 
    \resizebox{\textwidth}{!}{
    \begin{tabular}{c@{~~}c@{~~}|c@{~~}c@{~~}c@{~~}c@{~~}|c}
    \hline
    \multicolumn{7}{c}{ AlexNet } \\
    \hline
    & & Painting & Cartoon & Sketch & Photo & Average \\
    \hline
    \multirow{5}{*}{ Original }& Baseline & 66.83 & 70.85 & 59.75 & 89.78 & 71.80 \\
    & Rotation & 65.66 & 71.89 & 62.15 & 89.88 & 72.39 \\
    & DG-MMLD & 69.27 & 72.83 & 66.44 & 88.98 & 74.38 \\
    & Epi-FCR & 64.70 & 72.30 & 65.00 & 86.10 & 72.03 \\
    & DDAIG* & 62.77 & 67.06 & 58.90 & 86.82 & 68.89 \\
    \hline
    \multirow{5}{*}{ Stylized } & Baseline & 71.96 & 72.47 & 76.47 & 88.34 & \textbf{77.31} \\
    
    & Rotation & 71.74 & 73.39 & 75.98 & 89.22 & 77.59 \\
    & DG-MMLD & 70.50 & 70.84 & 75.39 & 88.43 & 76.29 \\
    & Epi-FCR & 65.19 & 69.54 & 71.97 & 83.43 & 72.53 \\
    & DDAIG & 69.35 & 71.10 & 70.99 & 87.70 & 74.79 \\
    \hline 
    \multirow{2}{*} {Mixup} 
    & pixel-level & 66.03 & 68.00 & 51.18 & 88.90 & 68.53 \\
    & feature-level & 67.04 & 69.10 & 55.40 & 88.88 & 70.11 \\
    \hline
    \multicolumn{7}{c}{ ResNet18 } \\
    \hline
    \multirow{5}{*}{ Original } & Baseline & 77.28 & 73.89 & 67.01 & 95.83 & 78.50 \\
    & Rotation & 78.16 & 76.64 & 72.20 & 95.57 & 80.64 \\
    & DG-MMLD & 81.28 & 77.16 & 72.29 & 96.06 & 81.83 \\
    & Epi-FCR & 82.10 & 77.00 & 73.00 & 93.90 & 81.50 \\
    & DDAIG* & 79.41 & 74.81 & 69.29 & 95.22 & 79.68 \\
    \hline
    \multirow{6}{*}{ Stylized } & Baseline & 82.73 & 77.97 & 81.61 & 94.95 & \textbf{84.32} \\
    & Rotation & 79.51 & 79.93 & 82.01 & 93.55 & 83.75 \\
    & DG-MMLD & 80.85 & 77.10 & 77.69 & 95.11 & 82.69 \\
    & Epi-FCR & 80.68 & 78.87 & 76.57 & 92.50 & 82.15 \\
    & DDAIG & 81.02 & 78.75 & 79.67 & 95.07 & 83.63 \\
    \hline 
    \multirow{2}{*} {Mixup} 
    & pixel-level & 78.09 & 71.08 & 66.58 & 93.85 & 77.40 \\
    & feature-level  & 81.20 & 76.41 & 69.67 & 96.31 & 80.90 \\
    \hline
    \end{tabular} 
    }
    \label{rethinking:tab:pacs}
\end{table}

%% file: Rethinking/OfficeHome_table.tex
\begin{table}
    \centering
    \caption{OfficeHome classification accuracy (\%). We used AdaIN with parameters $\alpha=1.0$ and $p=0.1$.} 
    \resizebox{\textwidth}{!}{
    \begin{tabular}{c@{~~}c@{~~}|c@{~~}c@{~~}c@{~~}c@{~~}|c}
    \hline
    \multicolumn{7}{c}{ ResNet18 } \\
    \hline
    & & Art & Clipart & Product & Real World & Average \\
    \hline
    \multirow{5}{*}{ Original } & Baseline & 57.14 & 46.96 & 73.50 & 75.72 & 63.33 \\
    & Rotation & 55.94 & 47.26 & 72.38 & 74.84 & 62.61 \\
    & DG-MMLD* & 58.08 & 49.32 & 72.91 & 74.69 & 63.75 \\
    & Epi-FCR* & 53.34 & 49.66 & 68.56 & 70.14 & 60.43 \\
    				
    & DDAIG* & 57.79 & 48.32 & 73.28 & 74.99 & 63.59 \\
    \hline
    \multirow{5}{*}{ Stylized } & Baseline & 58.71 & 52.33 & 72.95 & 75.00 & \textbf{64.75} \\
    & Rotation & 57.24 & 52.15 & 72.33 & 73.66 & 63.85 \\
    & DG-MMLD & 59.24 & 49.30 & 73.56 & 75.85 & 64.49 \\
    & Epi-FCR & 52.97 & 50.14 & 67.03 & 70.66 & 60.20 \\
    & DDAIG & 58.21 & 50.26 & 73.81 & 74.99 & 64.32 \\
    \hline
    Mixup & feature-level & 58.33 & 39.76 & 70.96 & 72.07 & 60.28 \\
    \hline
    \end{tabular} 
    }
    \label{rethinking:tab:officehome} 
\end{table}

%% file: Rethinking/VLCS_table.tex
\begin{table}
    \centering
    \caption{VLCS classification accuracy (\%). We used AdaIN with parameters are $\alpha=1.0$ and $p=0.75$.}
    \resizebox{\textwidth}{!}{
    \begin{tabular}{c@{~~}c@{~~}|c@{~~}c@{~~}c@{~~}c@{~~}|c}
    \hline
    \multicolumn{7}{c}{ AlexNet } \\
    \hline
    & & CALTECH & LABELME & PASCAL & SUN & Average \\
    \hline
    \multirow{5}{*}{ Original } & Baseline & 94.89 & 59.14 & 71.31 & 64.64 & 72.49 \\
    & Rotation & 94.50 & 61.27 & 68.94 & 63.28 & 72.00 \\
    & DG-MMLD* & 96.94 & 59.10 & 68.48 & 62.06 & 71.64 \\
    & Epi-FCR* & 91.43 & 61.36 & 63.44 & 60.07 & 69.07 \\
    & DDAIG* & 95.75 & 60.18 & 65.48 & 60.78 & 70.55 \\
    \hline
    \multirow{5}{*}{ Stylized } & Baseline & 96.86 & 60.77 & 68.18 & 63.42 & 72.31 \\
    & Rotation & 96.86 & 60.77 & 68.18 & 63.42 & 72.31 \\
    & DG-MMLD & 97.49 & 61.02 & 64.23 & 62.37 & 71.28 \\
    & Epi-FCR & 92.69 & 58.18 & 62.59 & 57.87 & 67.83 \\
    & DDAIG & 97.48 & 60.48 & 65.19 & 62.57 & 71.43 \\
    \hline
    Mixup & feature-level & 94.73 & 62.15 & 69.82 & 62.98 & 72.42 \\
    \hline
    \end{tabular}
    }
    \label{rethinking:tab:vlcs}
\end{table}

%% file: Jigen.tex
\section{Domain Generalization by solving jigsaw puzzles}
\newcommand*{\our}{JiGen\@\xspace}
\newcommand{\argmin}{\mathop{\mathrm{argmin}}}

\textit{Human adaptability relies crucially on the ability to learn and merge knowledge both from supervised and unsupervised learning: the parents point out few important concepts, but then the children fill in the gaps on their own. This is particularly effective, because supervised learning can never be exhaustive and thus learning autonomously allows to discover invariances and regularities that help to generalize.
In this paper we propose to apply a similar approach to the task of object recognition across domains: our model learns the semantic labels in a supervised fashion, and broadens its understanding of the data by learning from self-supervised signals how to solve a jigsaw puzzle on the same images. This secondary task helps the network to learn the concepts of spatial correlation while acting as a regularizer for the classification task. Multiple experiments on the PACS, VLCS, Office-Home and digits datasets confirm our intuition and show that this simple method  outperforms previous domain generalization and adaptation solutions. An ablation study further illustrates the inner workings of our approach.}

\vspace{4mm}In the current gold rush towards artificial intelligent systems it is becoming more and more evident 
that there is little intelligence without the ability to transfer knowledge and generalize across tasks, 
domains and categories \cite{csurka_book}. 
A large portion of computer vision research is dedicated to supervised methods that show remarkable results with 
convolutional neural networks in well defined settings, but still struggle when attempting these types of generalizations.  
Focusing on the ability to generalize across domains, the community has attacked this issue so far 
mainly by regularizing supervised learning processes with techniques that search for intermediate semantic spaces able to capture basic data knowledge regardless of the specific appearance of input images.
Proposed methods range from decoupling image style from the shared object content \cite{Bousmalis:DSN:NIPS16}, 
to pulling data of different domains together and imposing adversarial conditions \cite{Li_2018_CVPR,Li_2018_ECCV}, 
up to generating new samples to better cover the space spanned by any future target \cite{DG_ICLR18,Volpi_2018_NIPS}.

\begin{figure}
    \centering
\includegraphics[width=\textwidth]{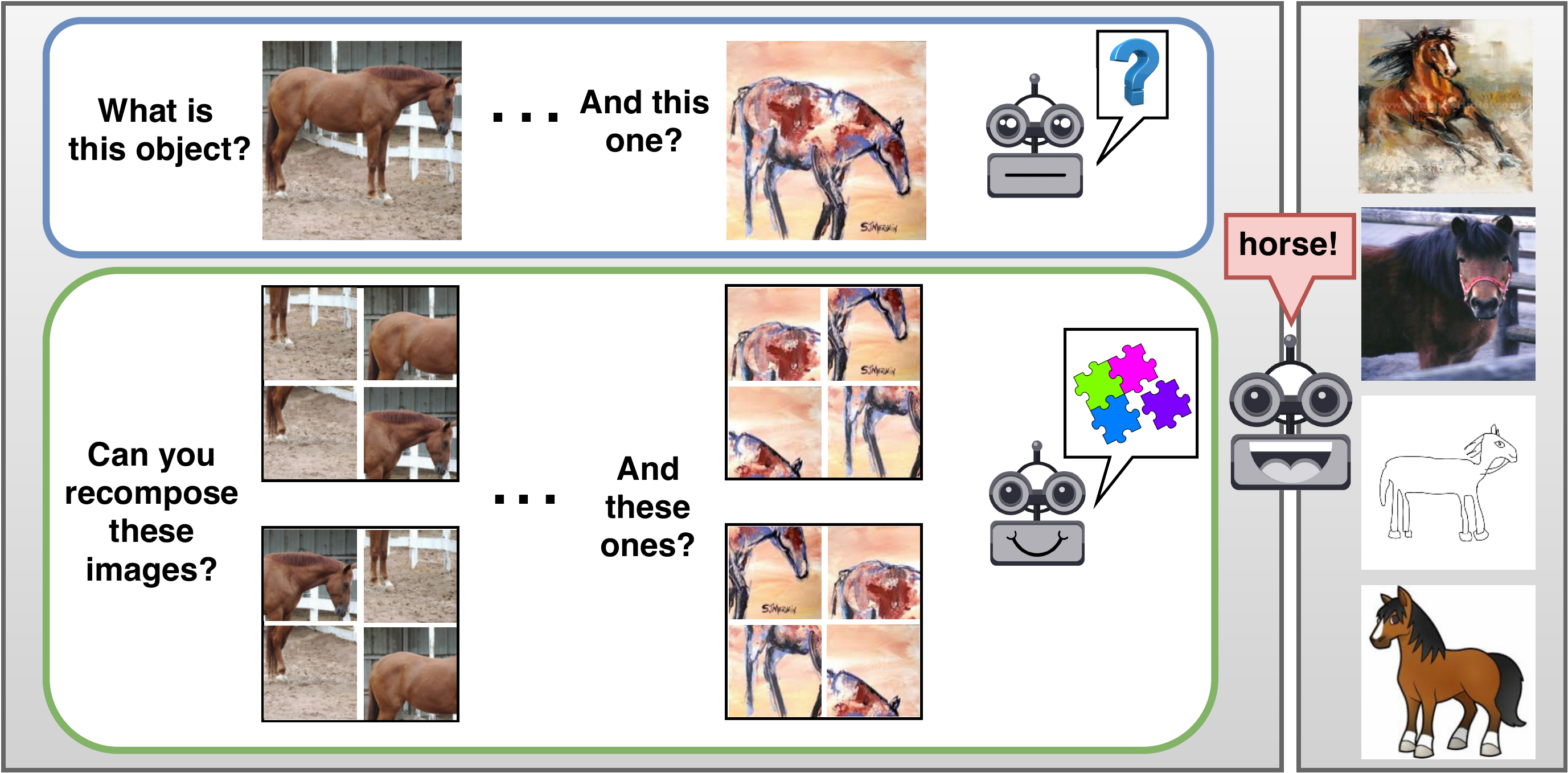}   
    \caption{Recognizing objects across visual domains is a challenging task that requires high generalization abilities.
    Other tasks, based on intrinsic self-supervisory image signals, allow to capture natural invariances and regularities
    that can help to bridge across large style gaps. With \our we learn jointly to classify objects and solve 
    jigsaw puzzles, showing that this supports generalization to new domains.}
    \label{jigen:fig:copertina}
\end{figure}

With the analogous aim of getting general purpose feature embeddings, an alternative research direction has
been recently pursued in the area of unsupervised learning. The main techniques are based on the 
definition of tasks useful to learn visual invariances and regularities by exploiting self-supervisory signals as the 
spatial co-location of patches \cite{NorooziF16,Cruz2017,Noroozi_2018_CVPR}, counting primitives \cite{learningtocount}, 
image coloring \cite{zhang2016colorful}, and video frame ordering \cite{misra2016unsupervised,videosiccv15}. 
The large availability of variable unsupervised data, as well as their very nature 
which makes them free from the labeling bias issue \cite{TorralbaEfros_bias}, 
suggests that the unsupervised tasks may also capture knowledge independent from specific domain style.
Despite their large potential, the existing unsupervised approaches often come with tailored architectures
that need dedicated finetuning strategies 
to re-engineer the acquired knowledge and make it usable as input for a standard supervised training process \cite{Noroozi_2018_CVPR}.
Moreover, this knowledge is generally applied on real-world photos and has not been challenged before across
large domain gaps with images of other nature like paintings or sketches.

This clear separation between learning intrinsic regularities from images and robust classification across domains 
is in contrast with the visual learning strategies of biological systems, and in particular of the human visual system. 
Indeed, numerous studies highlight that infants and toddlers learn both to categorize objects and about regularities at 
the same time \cite{children_learning}. For instance, popular toys for infants teach to recognize different categories by fitting them 
into shape sorters; jigsaw puzzles of animals or vehicles to encourage learning of object parts' spatial relations 
are equally widespread among 12-18 months old. 
This type of joint learning is certainly a key ingredient in the ability of humans to reach sophisticated 
visual generalization abilities at an early age \cite{PLOS}. 

\begin{figure*}
    \centering
\includegraphics[width=\textwidth]{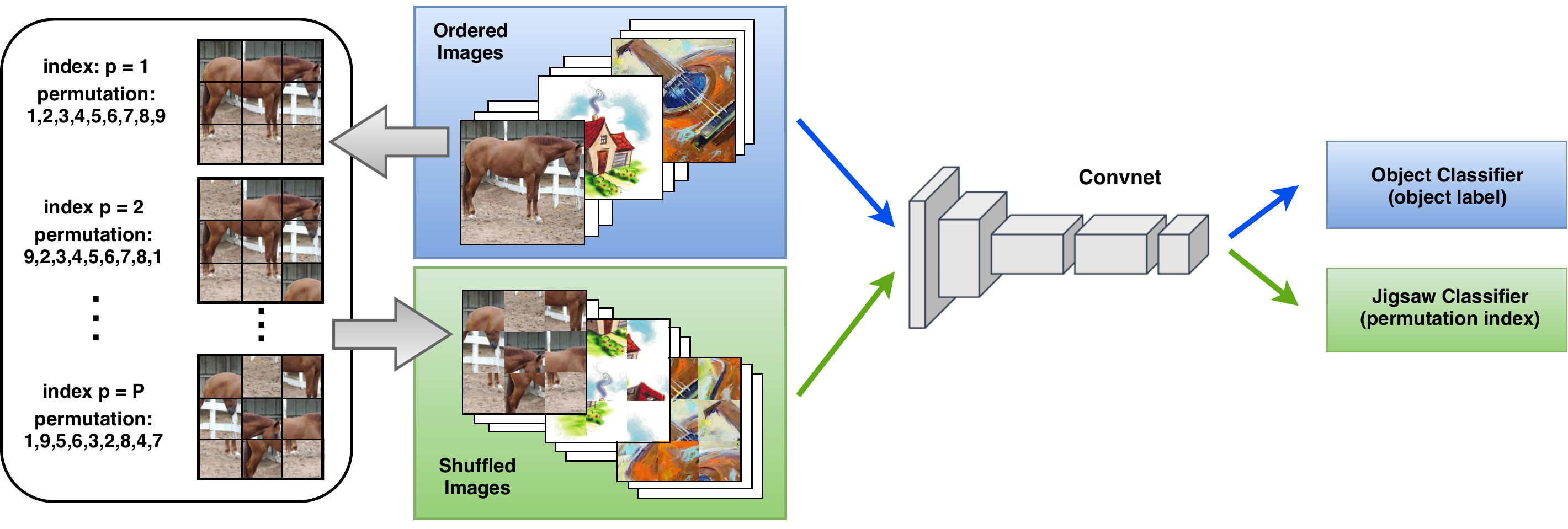}   
    \caption{Illustration of the proposed method \our. 
    We start from images of multiple domains and use a $3\times 3$ grid to decompose them in 9 patches
    which are then randomly shuffled and used to form images of the same dimension of the original ones.
    By using the maximal Hamming distance algorithm in \cite{NorooziF16} we define a set of $P$ patch permutations
    and assign an index to each of them. Both the original ordered and the shuffled images are fed to 
    a convolutional network that is optimized to satisfy two objectives: object classification on the 
    ordered images and jigsaw classification, meaning permutation index recognition, on the shuffled images.}
    \label{jigen:fig:jigen}
\end{figure*}

Inspired by this, we propose the first end-to-end architecture that learns simultaneously how to generalize across 
domains and about spatial co-location of image parts (Figure \ref{jigen:fig:copertina}, \ref{jigen:fig:jigen}). In this work we focus on the unsupervised task of 
recovering an original image from its shuffled parts, also known as solving jigsaw puzzles. We show how this 
popular game can be re-purposed as a side objective to be optimized jointly with object classification over different 
source domains and improve generalization with a simple multi-task process.
We name our Jigsaw puzzle based Generalization method \our.
Differently from previous approaches that deal with separate image patches and recombine their features towards the 
end of the learning process \cite{NorooziF16,Cruz2017,Noroozi_2018_CVPR,Cruz2017}, we move the patch re-assembly at the 
image level and we formalize the jigsaw task as a classification problem over recomposed images with the same dimension 
of the original one. In this way object recognition and patch reordering can share the same network backbone and we
can seamlessly leverage over any convolutional learning structure as well as several pretrained models without the 
need of specific architectural changes.

We demonstrate that \our allows to better capture the shared knowledge among multiple sources
and acts as a regularization tool for a single source. In the case unlabeled samples of the target
data are available at training time, running the unsupervised jigsaw task on them 
contributes to the feature adaptation process and shows competing results with respect
to state of the art unsupervised domain adaptation methods.

\subsection{The JiGen approach}
\label{jigen:sec:approach}

Starting from the samples of multiple source domains, we wish to learn a model that can perform well on any new 
target data population covering the same set of categories. 
Let us assume to observe $S$ domains, with the $i$-th domain containing  $N_i$ labeled instances 
$\{(x_j^{i},y_j^{i})\}_{j=1}^{N_i}$, where $x_j^{i}$ indicates the $j$-th image and $y_j^{i} \in \{1,\ldots,C\}$ 
is its class label. 
The first basic objective of \our is to 
minimize the loss  $\mathcal{L}_c(h(x| \theta_f, \theta_c),y)$ that measures the error between 
the true label $y$ and the label predicted by the deep model function $h$, parametrized by $\theta_f$ and $\theta_c$. 
These parameters define the feature embedding space and the final classifier, respectively for the convolutional
and fully connected parts of the network.
Together with this objective, we ask the network to satisfy a second condition related to solving jigsaw puzzles.
We start by decomposing the source images using a regular $n\times n$ grid of patches, which are then shuffled
and re-assigned to one of the $n^2$ grid positions. Out of the $n^2!$ possible permutations we select a set 
of $P$ elements by following the Hamming distance based algorithm in \cite{NorooziF16}, and we assign an index to each entry. 
In this way we define a second classification task on $K_i$ labeled instances $\{(z_k^{i},p_k^{i})\}_{k=1}^{K_i}$,
where $z_k^{i}$ indicates the recomposed samples and $p_k^i \in \{1,\ldots,P\}$ the related permutation index,
for which we need to minimize the jigsaw loss  $\mathcal{L}_p(h(z| \theta_f, \theta_p),p)$.
Here the deep model function $h$ has the same structure used for object classification and shares with that
the parameters  $\theta_f$. The final fully connected layer dedicated to permutation recognition
is parametrized by $\theta_p$. 
Overall we train the network to obtain the optimal model through 
\begin{align}
    \argmin_{\theta_f, \theta_c, \theta_p} \sum_{i=1}^S  & \sum_{j=1}^{N_i}  \mathcal{L}_c(h(x^i_j | \theta_f, \theta_c),y_j^i) +  \nonumber \\ 
     & \sum_{k=1}^{K_i}\alpha\mathcal{L}_p(h(z^i_k | \theta_f, \theta_p),p_k^i) 
\end{align}
where both  $\mathcal{L}_c$  and $\mathcal{L}_p$ are standard cross-entropy losses. We underline that the
jigsaw loss is also calculated on the ordered images. Indeed, the correct patch sorting corresponds to one of 
the possible permutations and we always include it in the considered subset $P$. On the other way round, the 
classification loss is not influenced by the shuffled images, since this would make object recognition tougher. 
At test time we use only the object classifier to predict on the new target images.

\subsubsection{Extension to Unsupervised Domain Adaptation}
Thanks to the unsupervised nature of the jigsaw puzzle task, we can always extend \our to the unlabeled samples
of target domain when available at training time. This allows us to exploit the jigsaw task for unsupervised domain adaptation. 
In this setting, for the target ordered images we minimize the classifier prediction uncertainty through the empirical entropy loss $\mathcal{L}_E (x^t)= \sum_{y\in \mathcal{Y}} h(x^t|\theta_f, \theta_c)log\{h(x^t|\theta_f, \theta_c)\}$, while for the shuffled target images we 
keep optimizing the jigsaw loss $\mathcal{L}_p(h(z^t | \theta_f, \theta_p),p^t)$.

\subsubsection{Implementation Details}
Overall \our\footnote{Code available at https://github.com/fmcarlucci/JigenDG} has two parameters related to 
how we define the jigsaw task, and three related to the 
learning process. The first two are respectively the grid size $n \times n$ used to define the image patches 
and the cardinality of the patch permutation subset $P$. As we will detail in the following section, \our is 
robust to these values and for all our experiments we kept them fixed, using $3 \times 3$ patch grids and $P=30$.
The remaining parameters are the weights $\alpha$ of the jigsaw loss, and $\gamma$ assigned to the entropy
loss when included in the optimization process for unsupervised domain adaptation. The final third 
parameter regulates the data input process: the shuffled images enter the network together with the original 
ordered ones, hence each image batch contains both of them.  We define a data bias parameter $\beta$ to 
specify their relative ratio. For instance $\beta=0.6$ means that for each batch, $60\%$ of the
images are ordered, while the remaining $40\%$ are shuffled. These last three parameters were chosen by cross
validation on a $10\%$ subset of the source images for each experimental setting.

We designed the \our network making it able to leverage over many possible convolutional deep architectures.
Indeed it is sufficient to remove the existing last fully connected layer of a network and substitute
it with the new object and jigsaw classification layers.
\our is trained with SGD solver, $30$ epochs, batch size $128$,  learning 
rate set to $0.001$ and stepped down to $0.0001$ after $80\%$ of the training epochs. 
We used a simple data augmentation protocol by randomly cropping the images to retain between $80-100\%$ and 
randomly applied horizontal flipping. Following \cite{Noroozi_2018_CVPR} we randomly ($10\%$ probability) 
convert an image tile to grayscale.

\subsection{Experiments}

\begin{table}[!t]
\begin{center} \small
\resizebox{\textwidth}{!}{
\begin{tabular}{@{}c@{~~~}c@{~~~}c@{~~~}c@{~~~}c@{~~~}c|@{~~~}c}
\hline
\multicolumn{2}{c}{\textbf{PACS}}  & \textbf{art\_paint.} & \textbf{cartoon} &  \textbf{sketches} & \textbf{photo} &   \textbf{Avg.}\\ \hline
\multicolumn{7}{@{}c@{}}{\textbf{CFN - Alexnet} }\\
\hline
\multicolumn{2}{@{}c@{}}{\footnotesize{J-CFN-Finetune}} & 47.23  & \textbf{62.18}  & \textbf{58.03}  & 70.18  & 59.41\\
\multicolumn{2}{@{}c@{}}{\footnotesize{J-CFN-Finetune++}} &  51.14  &  58.83  &  54.85  &  73.44  &  59.57\\
\multicolumn{2}{@{}c@{}}{\footnotesize{C-CFN-Deep All}} &  59.69  &  59.88  &  45.66  &  \textbf{85.42}  &  62.66\\
\multicolumn{2}{@{}c@{}}{\footnotesize{C-CFN-\our}} &  \textbf{60.68}  &  60.55  &  55.66  &  82.68  &  \textbf{64.89}\\
\hline
\multicolumn{7}{c}{\textbf{Alexnet}}\\
\hline
\multirow{2}{*}{\cite{hospedalesPACS}}  & Deep All & 63.30 & 63.13 & 54.07 & 87.70 & 67.05\\
& TF & 62.86 & 66.97 & 57.51  & \textbf{89.50} & 69.21\\
\hline
\multirow{3}{*}{\cite{Li_2018_ECCV}} & Deep All & 57.55  & 67.04  & 58.52  & 77.98  & 65.27 \\
& DeepC &  62.30 & 69.58  & 64.45  &  80.72 &  69.26\\
& CIDDG &  62.70 & 69.73 & 64.45  &  78.65 &  68.88\\
\hline
\multirow{2}{*}{\cite{MLDG_AAA18}} & Deep All & 64.91 & 64.28 & 53.08 & 86.67 & 67.24\\
 & MLDG & 66.23 & 66.88 & 58.96 &  88.00& 70.01\\
\hline
\multirow{2}{*}{\cite{Antonio_GCPR18}} & Deep All  & 64.44 & 72.07 & 58.07 &  87.50 & 70.52 \\
& D-SAM & 63.87 & 70.70 & 64.66 & 85.55 & 71.20\\
\hline
& Deep All & 66.68 & 69.41 & 60.02 & \underline{89.98}  & 71.52\\
& \textbf{\our} & \textbf{67.63} & \textbf{71.71} & \textbf{65.18} & 89.00 & \textbf{73.38}\\
\hline
\multicolumn{7}{c}{\textbf{Resnet-18}}\\
\hline
 \multirow{2}{*}{\cite{Antonio_GCPR18}} & Deep All & 77.87 & \underline{75.89} & 69.27 &  95.19 & 79.55\\
 & D-SAM & 77.33 & 72.43 & \textbf{77.83} & 95.30 & \textbf{80.72}\\
\hline
 & Deep All & 77.85  & 74.86  & 67.74  & 95.73  & 79.05 \\
 & \textbf{\our}    & \textbf{79.42}  & \textbf{75.25}  & 71.35  &  \textbf{96.03} &  80.51\\
\hline
\end{tabular}
}
\caption{Domain Generalization results on PACS. The results of \our are average over three repetitions of each run. The top part of the table is dedicated to a comparison with previous methods that use the jigsaw task as a pretext to learn transferable features using a context-free siamese-ennead network (CFN). The central and bottom part of the table show the comparison of \our with several domain generalization methods when using respectively Alexnet and Resnet-18 architectures. 
Each column title indicates the name of the domain used as target. We use the bold font to highlight the best results of the generalization methods, while we underline a result when it is higher than all the others despite produced by the na\"ive Deep All baseline.}
\label{jigen:table:resultsDG_PACS}
\end{center}
\end{table}

\subsubsection{Patch-Based Convolutional Models for Jigsaw Puzzles}
We start our experimental analysis by evaluating the application of existing jigsaw related patch-based convolutional 
architectures and models to the domain generalization task.
We considered two recent works that proposed a jigsaw puzzle solver for 9 shuffled patches from images decomposed by a regular
$3\times3$ grid. Both \cite{NorooziF16} and \cite{Noroozi_2018_CVPR} use a Context-Free Network (CFN) with 9 siamese branches 
that extract features separately from each image patch and then recompose them before entering the final classification layer.
Specifically, each CFN branch is an Alexnet up to the first fully connected layer ($fc6$) and all the branches 
share their weights. Finally, the branches' outputs are concatenated and given as input to the following 
fully connected layer ($fc7$). The jigsaw puzzle task is formalized as a classification problem on a subset
of patch permutations and, once the network is trained on a shuffled version of Imagenet \cite{imagenet}, the
learned weights can be used to initialize the $conv$ layers of a standard Alexnet while the rest of the network is trained 
from scratch for a new target task.  Indeed, according to the original works, the learned representation is able to capture
semantically relevant content from the images regardless of the object labels. 
We followed the instructions in \cite{NorooziF16} and started from the pretrained Jigsaw CFN (J-CFN) model provided by the authors
to run finetuning on the PACS dataset with all the source domain samples aggregated together. 
In the top part of Table \ref{jigen:table:resultsDG_PACS} we indicate with J-CFN-Finetune the results of this experiment
using the jigsaw model proposed in \cite{NorooziF16}, while with J-CFN-Finetune++ the results from the 
advanced model proposed in \cite{Noroozi_2018_CVPR}. 
In both cases the average classification accuracy on the domains is lower than what can be obtained with a standard Alexnet 
model pre-trained for object classification on Imagenet and finetuned on all the source data aggregated together.
We indicate this baseline approach with Deep All and we can use as reference the corresponding values
in the following central part of Table \ref{jigen:table:resultsDG_PACS}. 
We can conclude that, despite its power as an unsupervised pretext task, completely disregarding 
the object labels when solving jigsaw puzzles induces a loss of semantic information that may still be 
crucial for generalization across multiple domains.

To demonstrate the potentialities of the CFN architecture, the authors of \cite{NorooziF16} used it also to train 
a supervised object Classification model on Imagenet (C-CFN) and demonstrated that it can produce results analogous to the
standard Alexnet. With the aim of further testing this network to understand if and how much its peculiar 
siamese-ennead structure can be useful to distill shared knowledge across domains, we considered it as the main 
convolutional backbone for \our.
Starting from the C-CFN model provided by the authors, we run the obtained C-CFN-\our on 
PACS data, as well as its plain object classification version with the jigsaw loss disabled ($\alpha=0$) that we 
indicate as C-CFN-Deep All. From the obtained recognition accuracy we can state that combining the jigsaw puzzle with 
the classification task provides an average improvement in performance, which is the first result to confirm our intuition. 
However, C-CFN-Deep All is still lower than the reference results obtained with standard Alexnet.

For all the following experiments we consider the convolutional architecture of \our built with the same main 
structure of Alexnet or Resnet, using always the image as a whole (ordered or shuffled) instead of relying on 
separate patch-based network branches.

\subsubsection{Multi-Source Domain Generalization}
We compare the performance of \our against several recent domain generalization methods. 
TF is the low-rank parametrized network that was presented together with the dataset PACS in \cite{hospedalesPACS}. 
CIDDG is the conditional invariant deep domain generalization method presented in \cite{Li_2018_ECCV} that 
trains for image classification with two adversarial constraints: one that maximizes the overall domain 
confusion following \cite{Ganin:DANN:JMLR16} and a second one that does the same per-class. 
In the DeepC variant, only this second condition is enabled. 
MLDG \cite{MLDG_AAA18} is a meta-learning approach  that simulates 
train/test domain shift during training and exploit them to optimize the learning model. 
CCSA \cite{doretto2017}  learns an embedding subspace where mapped visual domains 
are semantically aligned and yet maximally separated. 
MMD-AAE \cite{Li_2018_CVPR} is a deep method based on adversarial autoencoders that learns an invariant feature representation 
by aligning the data distributions to an arbitrary prior through the Maximum  Mean Discrepancy (MMD).
SLRC \cite{Ding2017DeepDG} is based on a single domain invariant network and multiple domain specific ones and
it applies a low rank constraint among them. 
D-SAM \cite{Antonio_GCPR18} is a method based on the use of domain-specific aggregation modules combined to improve
model generalization: it provides the current sota results on PACS and Office-Home.
For each of these methods, the Deep All baseline indicates the performance of the corresponding network when all the introduced
domain adaptive conditions are disabled.

\begin{table}[!t]
\begin{center} \small
\resizebox{\textwidth}{!}{
\begin{tabular}{@{}c@{~~~}c@{~~~}c@{~~~}c@{~~~}c@{~~~}c|c}
\hline
\multicolumn{2}{c}{\textbf{VLCS}}  & \textbf{Caltech} & \textbf{Labelme} &  \textbf{Pascal} & \textbf{Sun} &   \textbf{Avg.}\\ \hline
\multicolumn{7}{c}{\textbf{Alexnet}}\\
\hline
\multirow{3}{*}{\cite{Li_2018_ECCV}} & Deep All & 85.73	& 61.28	& 62.71	& 59.33	& 67.26  \\
& DeepC & 87.47	& 62.60	& 63.97	& 61.51	& 68.89 \\
& CIDDG & 88.83	& 63.06	& 64.38	& 62.10	& 69.59 \\
\hline
\multirow{2}{*}{\cite{doretto2017}} & Deep All & 86.10 & 55.60 & 59.10 & 54.60	& 63.85 \\
& CCSA & 92.30	& 62.10	& 67.10	& 59.10	& 70.15 \\
\hline
\multirow{2}{*}{\cite{Ding2017DeepDG}} & Deep All &  86.67 & 58.20 & 59.10 & 57.86 & 65.46 \\
& SLRC &  92.76	& 62.34	& 65.25	& 63.54	& 70.97 \\
\hline
\multirow{2}{*}{\cite{hospedalesPACS}} & Deep All & 93.40 & 62.11 & 68.41 & 64.16 & 72.02\\
 & TF & 93.63 & \textbf{63.49} & 69.99 & 61.32 & 72.11\\
\hline
 \cite{Li_2018_CVPR} & MMD-AAE & 94.40&	62.60&	67.70&	\textbf{64.40} &	72.28\\ 
 \hline
\multirow{2}{*}{\cite{Antonio_GCPR18}} & Deep All  & 94.95  & 57.45  & 66.06  & 65.87  &  71.08 \\
 & D-SAM & 91.75  & 56.95  & 58.59  & 60.84  & 67.03\\
\hline
 & Deep All & \underline{96.93} & 59.18 &  \underline{71.96} & 62.57 & 72.66\\
 & \textbf{\our} & \textbf{96.93} & 60.90 & \textbf{70.62} & 64.30 & \textbf{73.19}\\
\hline
\end{tabular}
}
\caption{Domain Generalization results on VLCS. The results of \our are average over three repetitions of each run. 
Each column title indicates the name of the domain used as target. We use the bold font for the best generalization result, while we underline the highest result when produced by the na\"ive Deep All baseline.}
\label{jigen:table:resultsDG_VLCS}
\end{center}
\end{table}

\begin{table}[tbp]
\begin{center}\small
\resizebox{\textwidth}{!}{
\begin{tabular}{@{}c@{~~~}c@{~~~}c@{~~~}c@{~~~}c@{~~~}c|c}
\hline
 \multicolumn{2}{c}{\textbf{Office-Home}}  & \textbf{Art} & \textbf{Clipart} &  \textbf{Product} & \textbf{Real-World} &  \textbf{Avg.}\\ \hline
\multicolumn{7}{c}{\textbf{Resnet-18}}\\
\hline
 \multirow{2}{*}{\cite{Antonio_GCPR18}} & Deep All  & 55.59 & 42.42 & 70.34 & 70.86 & 59.81 \\
 & D-SAM & \textbf{58.03} & 44.37 & 69.22 & 71.45 & 60.77\\
\hline
 & Deep All & 52.15 & 45.86 &  70.86 & \textbf{73.15} & 60.51\\
 & \textbf{\our} & 53.04 & \textbf{47.51} & \textbf{71.47} & 72.79 & \textbf{61.20}\\
\hline
\end{tabular}
}
\caption{Domain Generalization results on Office-Home. The results of \our are average over three repetitions of each run. 
Each column title indicates the name of the domain used as target.}
\label{jigen:table:resultsDG_officehome}
\end{center}
\end{table}

\begin{figure*}
    \hspace{-5mm}
    \centering
    \begin{tabular}{c@{~~~}c@{~~~}}
\includegraphics[height=5cm]{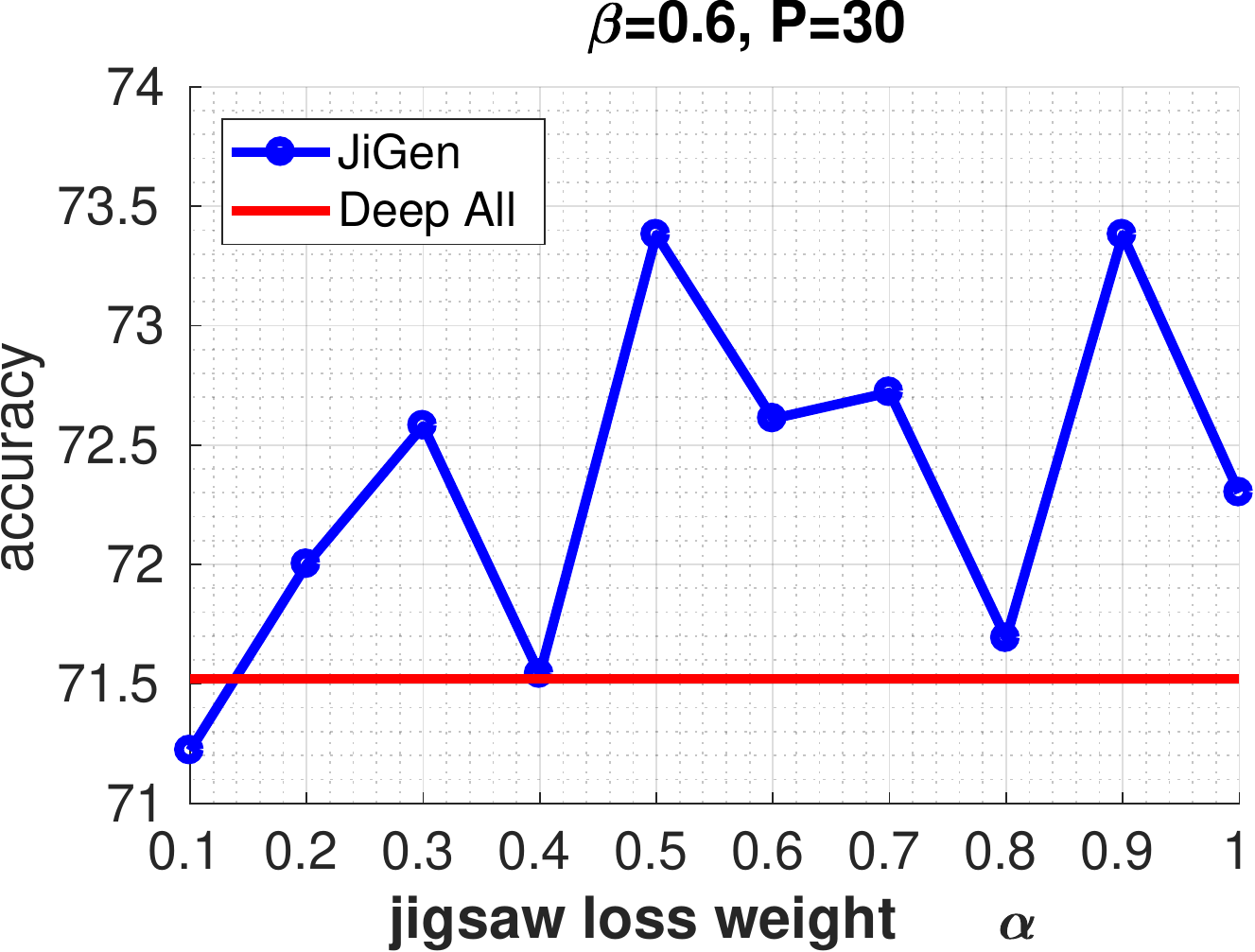}   &  \includegraphics[height=5cm]{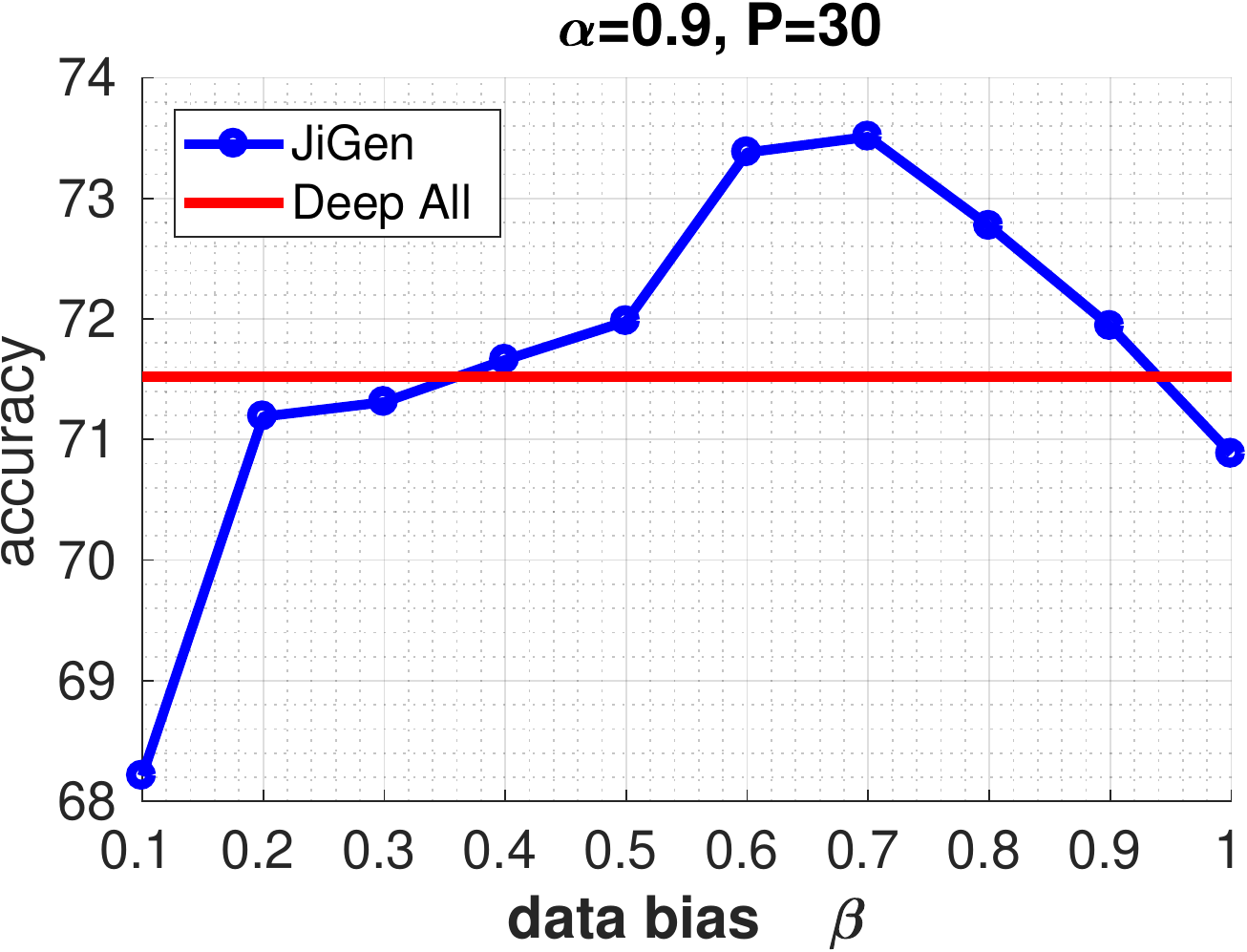} \\
\includegraphics[height=5cm]{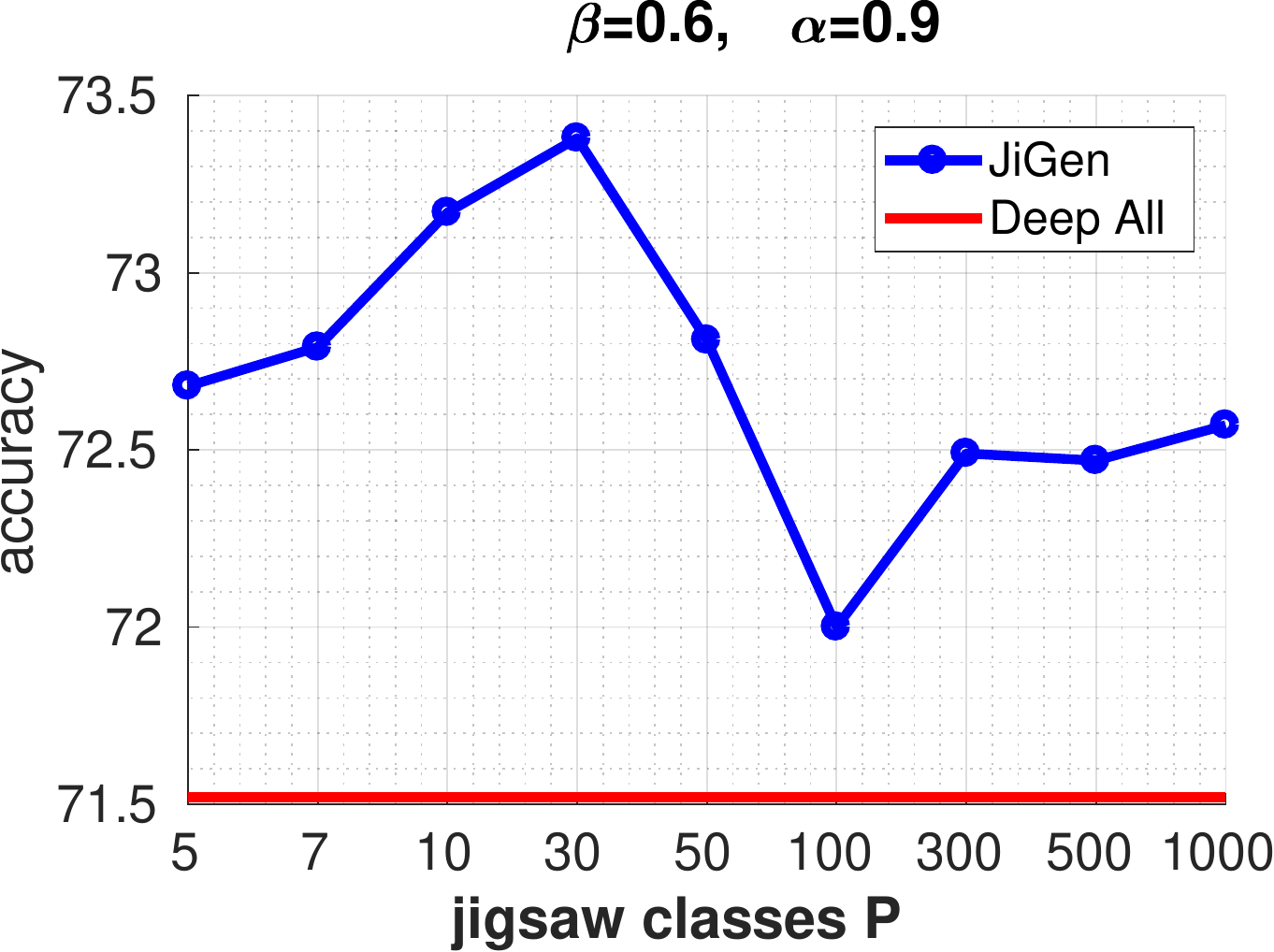}  &  \includegraphics[height=5cm]{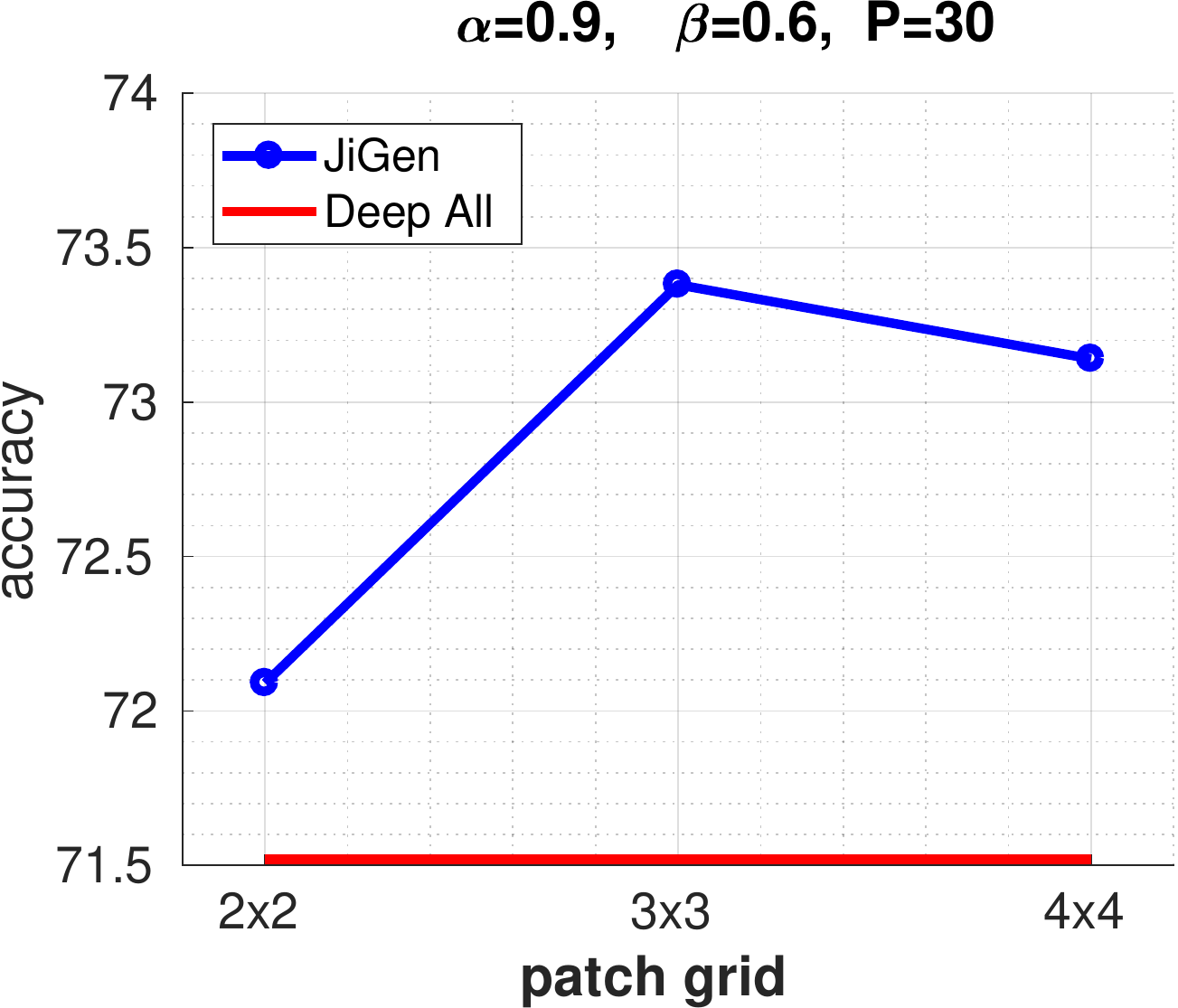}\\
    \end{tabular}
    \caption{Ablation results on the Alexnet-PACS DG setting. The reported accuracy is the global average over all the target domains of the setting, obtained by averaging over three repetitions of each run. The red line represents our Deep All average from Table \ref{jigen:table:resultsDG_PACS}.}
    \label{jigen:fig:ablation}
\end{figure*}

The central and bottom parts of Table \ref{jigen:table:resultsDG_PACS} show the results of \our on the dataset PACS when using 
as backbone architecture Alexnet and Resnet-18\footnote{With Resnet18, to put \our on equal
footing with D-SAM we follow the same data augmentation protocol in \cite{Antonio_GCPR18} and enabled color jittering.}. 
On average \our produces the best result when using Alexnet and it is just slighlty worse than the D-SAM reference for Resnet-18. 
Note however, that in this last case, \our outperforms D-SAM in three out of four target cases and the average advantage of 
D-SAM originate only from its result on sketches. 
On average, \our outperforms also the competing methods on the VLCS and on the Office-Home datasets 
(see respectively Table \ref{jigen:table:resultsDG_VLCS} and \ref{jigen:table:resultsDG_officehome}). 
In particular we remark that VLCS is a tough setting where the most recent works have only presented small gain 
in accuracy with respect to the corresponding Deep All baseline (e.g. TF). Since  \cite{Antonio_GCPR18} did not 
present the results of D-SAM on the VLCS dataset, we used the code provided by the authors\footnote{https://github.com/VeloDC/D-SAM\_public} 
to run these experiments. 
The obtained results show that, although generally able to close large domain gaps across images of different 
styles as in PACS and Office-Home, when dealing with domains all coming from real-world images, the use of 
aggregative modules tend to overfit, not supporting generalization. 

\begin{figure}[!t]\hspace{-5mm}
\centering
    \begin{tabular}{c@{~}c}
\includegraphics[height=4.6cm]{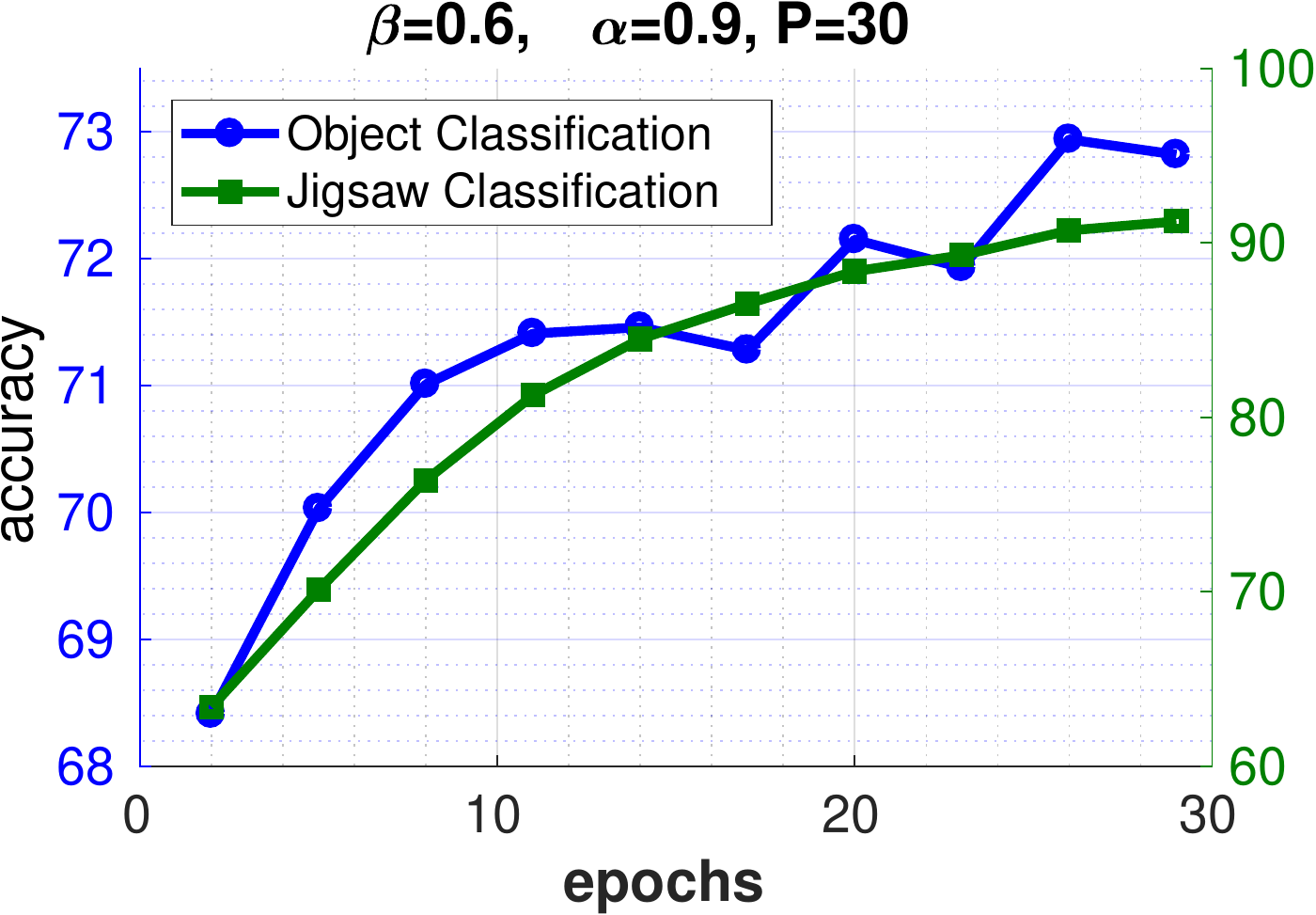}  &  \includegraphics[height=4.6cm]{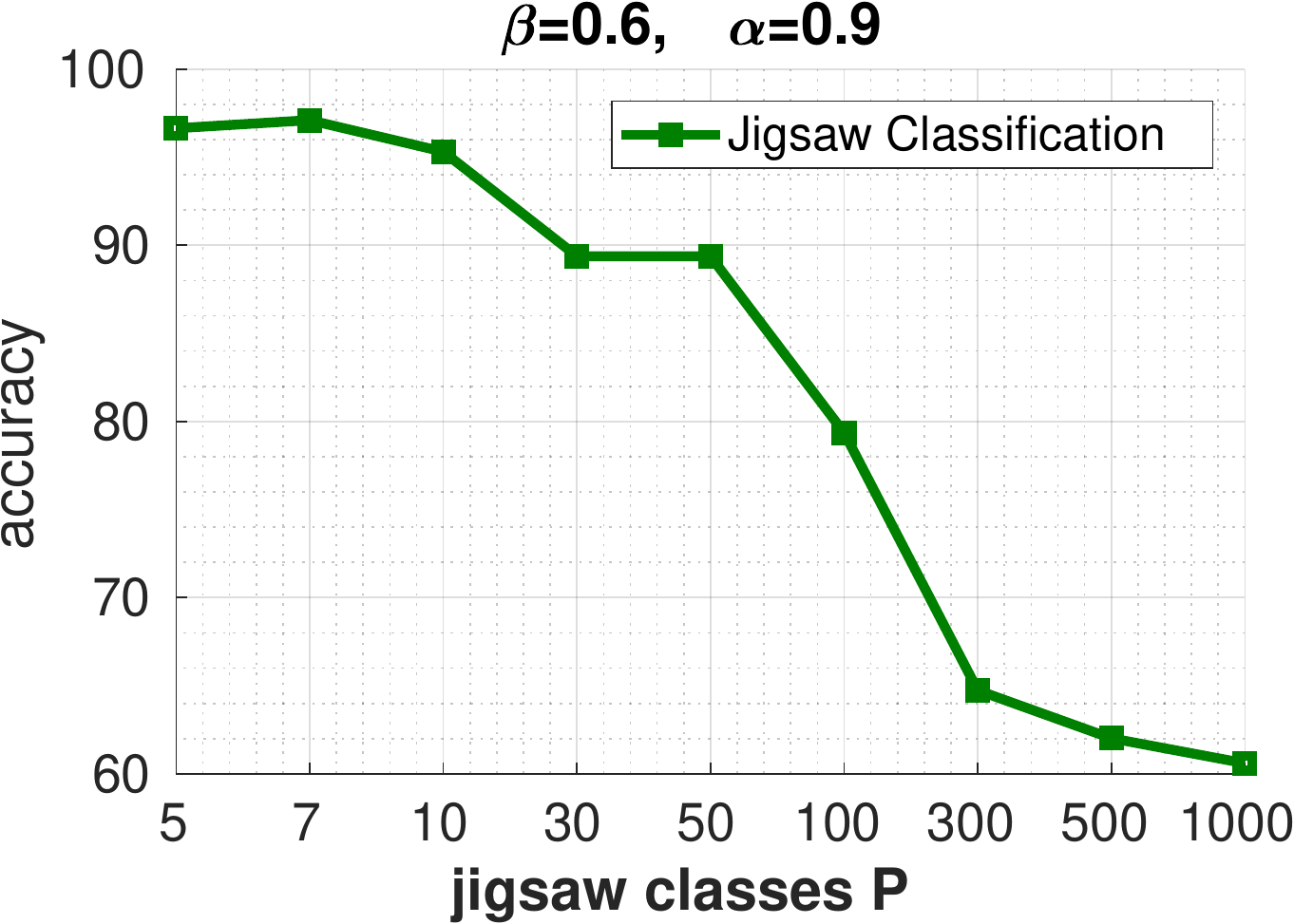}\\
    \end{tabular}
    \caption{Analysis of the behaviour of the jigsaw classifier on the Alexnet-PACS DG setting. For the plot on the left the color of the
    axes refer to each of the matching curve in the graph.}
    \label{jigen:fig:ablation_jigsaw}
\end{figure}

\subsubsection{Ablation}
Here we evaluate the robustness of \our showing its behaviour when the hyperparameter values are systematically changed on the Alexnet-PACS domain generalization setting.
First of all, we focus on the case with fixed number of jigsaw classes $P=30$, ordered/shuffled data bias $\beta=0.6$, as well as patch grid $3 \times 3$ and varying the jigsaw loss weight $\alpha$ in $\{0.1,1\}$. We show the obtained global average
classification accuracy on the first plot on the left of Figure \ref{jigen:fig:ablation}. We notice that only for the very low value $\alpha=0.1$ the obtained accuracy is just slightly lower than the Deep All baseline, while overall we observe an advantage regardless of the specific chosen $\alpha$.
From the second plot in Figure \ref{jigen:fig:ablation}, it appears that the data bias $\beta$ has a more significant impact on the result than $\alpha$. Indeed, $\beta<0.5$ implies that the amount of shuffled images fed to the network is higher than the respective amount of ordered ones. In those cases object classification assumes almost a secondary role with respect to the jigsaw reordering task, which implies a classification accuracy equal or lower than the Deep All baseline. On the other way round, for $\beta\geq0.5$ the performance increases and gets to its maximum to then decrease again. This behaviour is logically related to the fact that $\beta=1$ implies that we are considering only ordered images: the jigsaw loss encourages the network to correctly recognize always the same permutation class, which may increase the risk of overfitting.
The third plot in Figure \ref{jigen:fig:ablation} shows the change in performance when the number of jigsaw
classes $P$ varies between 5 and 1000. We started from a low number, with the same order of 
magnitude of the number of object classes in PACS, and we grew till 1000 which is the number used
for the experiments in \cite{NorooziF16}. We observe an overall variation of 1.5 percentage point
in the accuracy, but it always remains higher than the Deep All baseline. 
Finally, although for all our experiments we used images decomposed in 9 patches from a $3\times3$ grid, we also ran a further test to check the accuracy when changing the grid size and consequently the patch number. Even in this case, the range of variation is limited when passing from a $2\times2$ to a $4\times4$ grid, confirming the conclusions of robustness already obtained for this parameter in \cite{NorooziF16} and \cite{Cruz2017}. Moreover all the results are better than the Deep All reference.

In all our experiments we are using the jigsaw puzzle as a side task to help generalization, but it is also 
interesting to check if the jigsaw classifier is producing meaningful results. We show its recognition accuracy when 
testing it on the same images used to evaluate the object classifier but with shuffled patches. In Figure
\ref{jigen:fig:ablation_jigsaw}, the first plot on the left shows the accuracy over the learning epochs
for the object and jigsaw classifiers indicating that both grows simultaneously (on different scales).
The plot on the right of the same figure shows the jigsaw recognition accuracy when changing the number of
jigsaw classes $P$: of course the performance decreases when the task becomes more difficult, but overall
the obtained results indicate that the jigsaw model is always effective in reordering the shuffled image
patches.

\begin{figure}\hspace{-0.4cm}
\centering
\begin{tabular}{@{}c@{~~}c@{}}
\includegraphics[height=4cm]{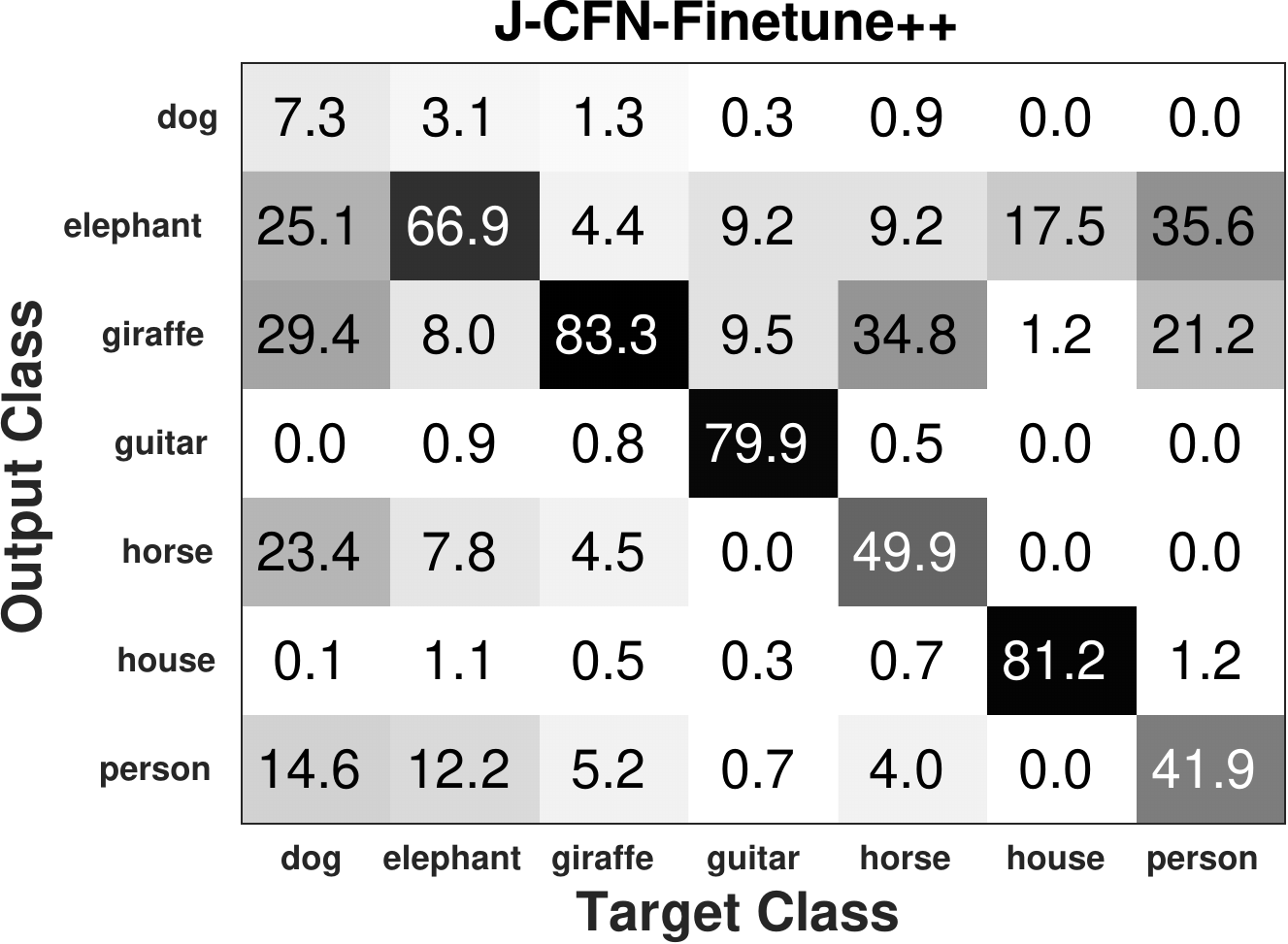} & 
\includegraphics[height=4cm]{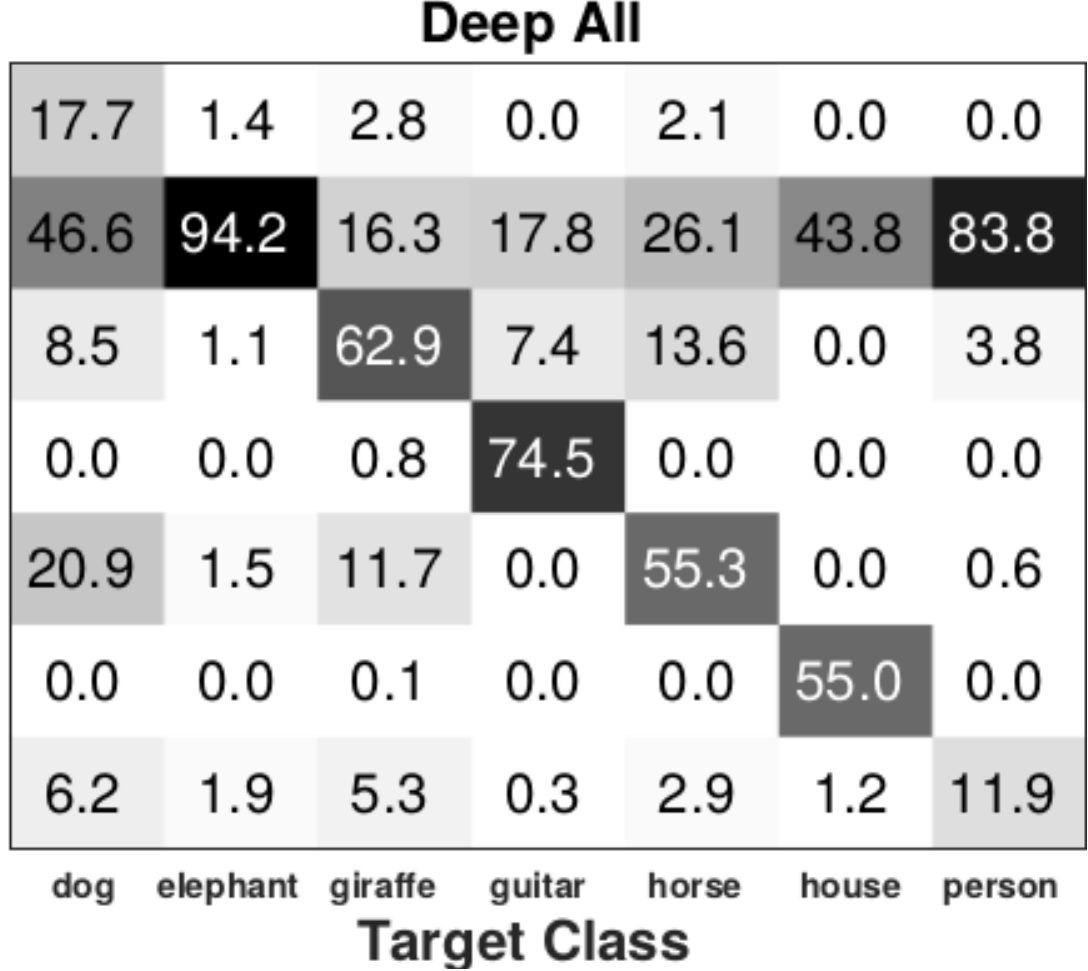} \\
\multicolumn{2}{c}{\includegraphics[height=4cm]{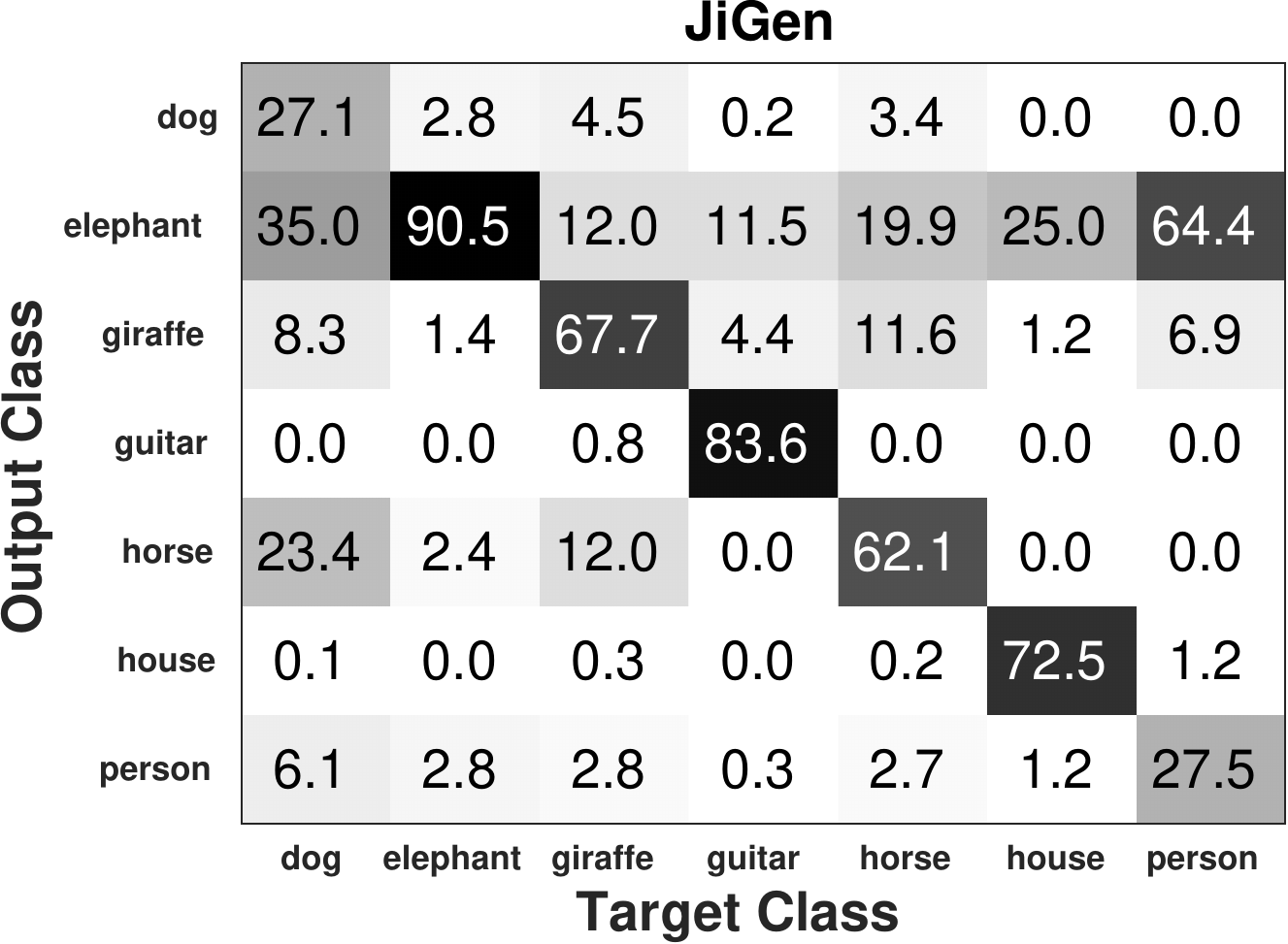}}
\end{tabular}
\caption{Confusion matrices on Alexnet-PACS DG setting, when sketches is used as target domain.}
\label{jigen:table:confmat}
\end{figure}

To further evaluate how \our is using the jigsaw puzzle task to learn spatial correlations, we consider 
the results per class produced when using the challenging sketches domain of PACS as target and comparing 
the obtained accuracy against J-CFN-Finetune++ and Deep All. The confusion matrices in Figure \ref{jigen:table:confmat},
indicate that for four out of seven categories, J-CFN-Finetune++ that leverages over the jigsaw model trained to 
relocate patch features, is actually doing a good job, better than Deep All. With \our we improve over Deep All 
for the same categories by exploiting the knowledge from solving jigsaw puzzles at image level and we keep the 
advantage of Deep All for the remaining categories.

\begin{figure*}
\centering
\includegraphics[width=0.4\textwidth]{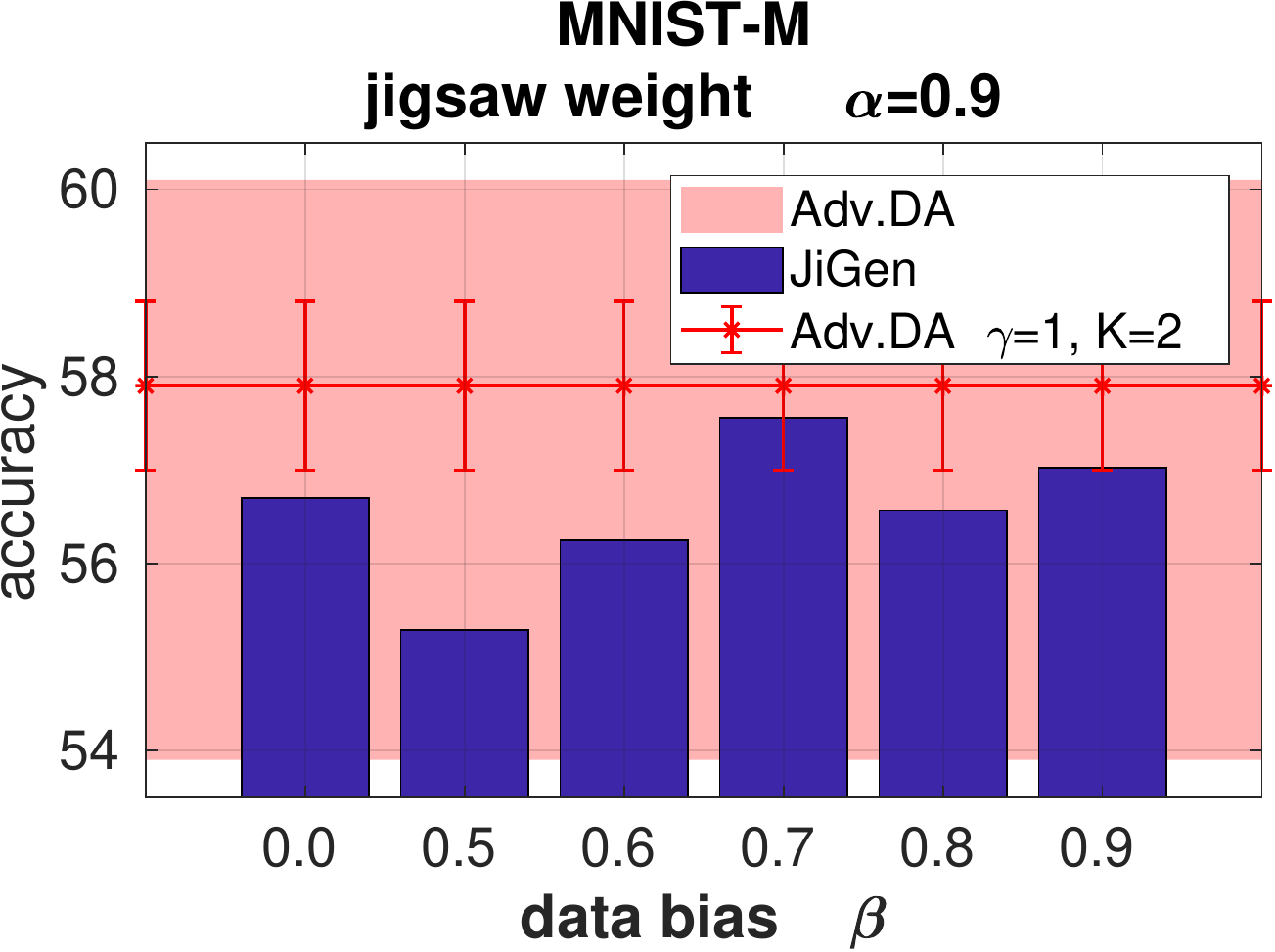}
\includegraphics[width=0.4\textwidth]{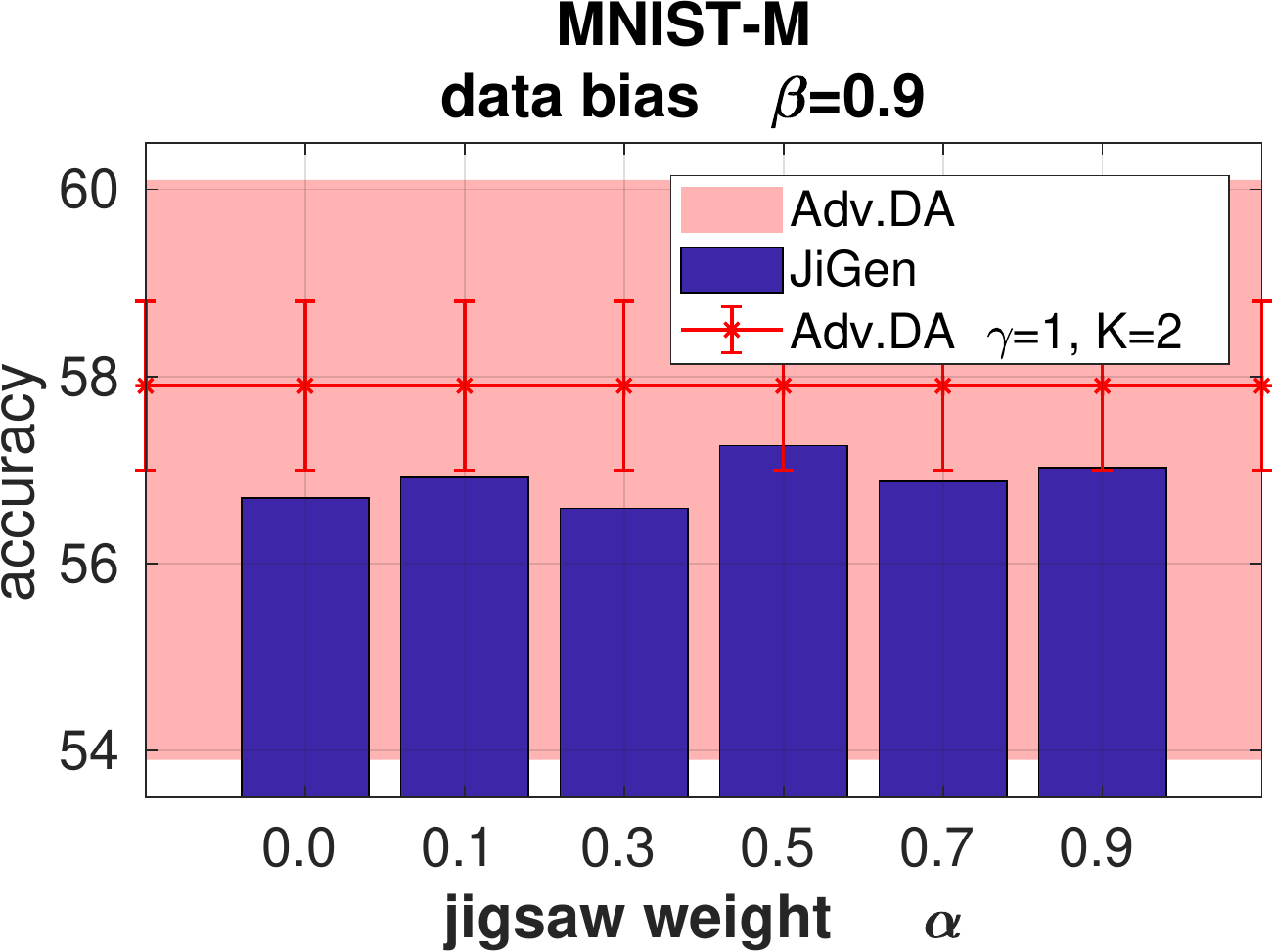}
\includegraphics[width=0.4\textwidth]{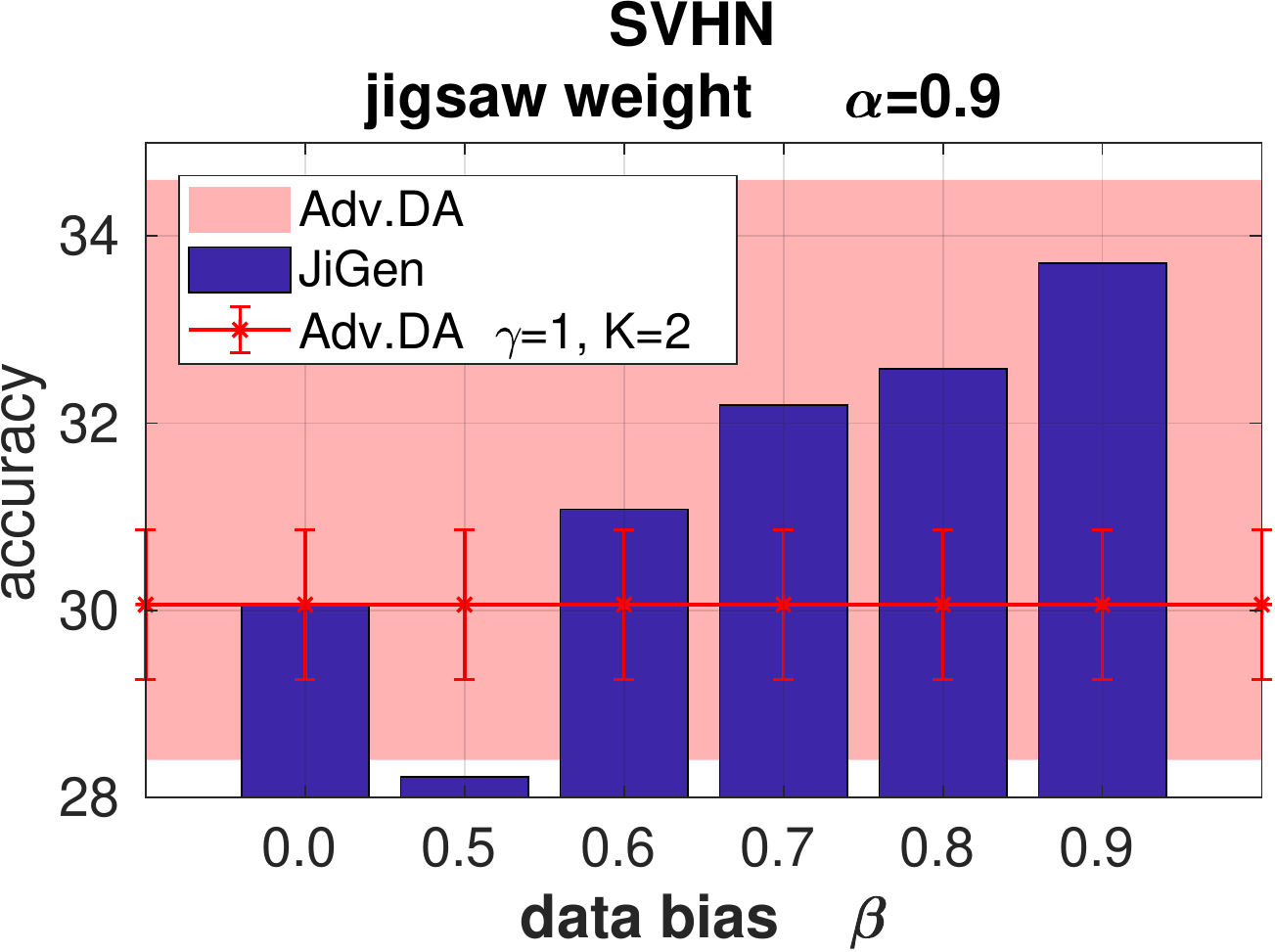}  
\includegraphics[width=0.4\textwidth]{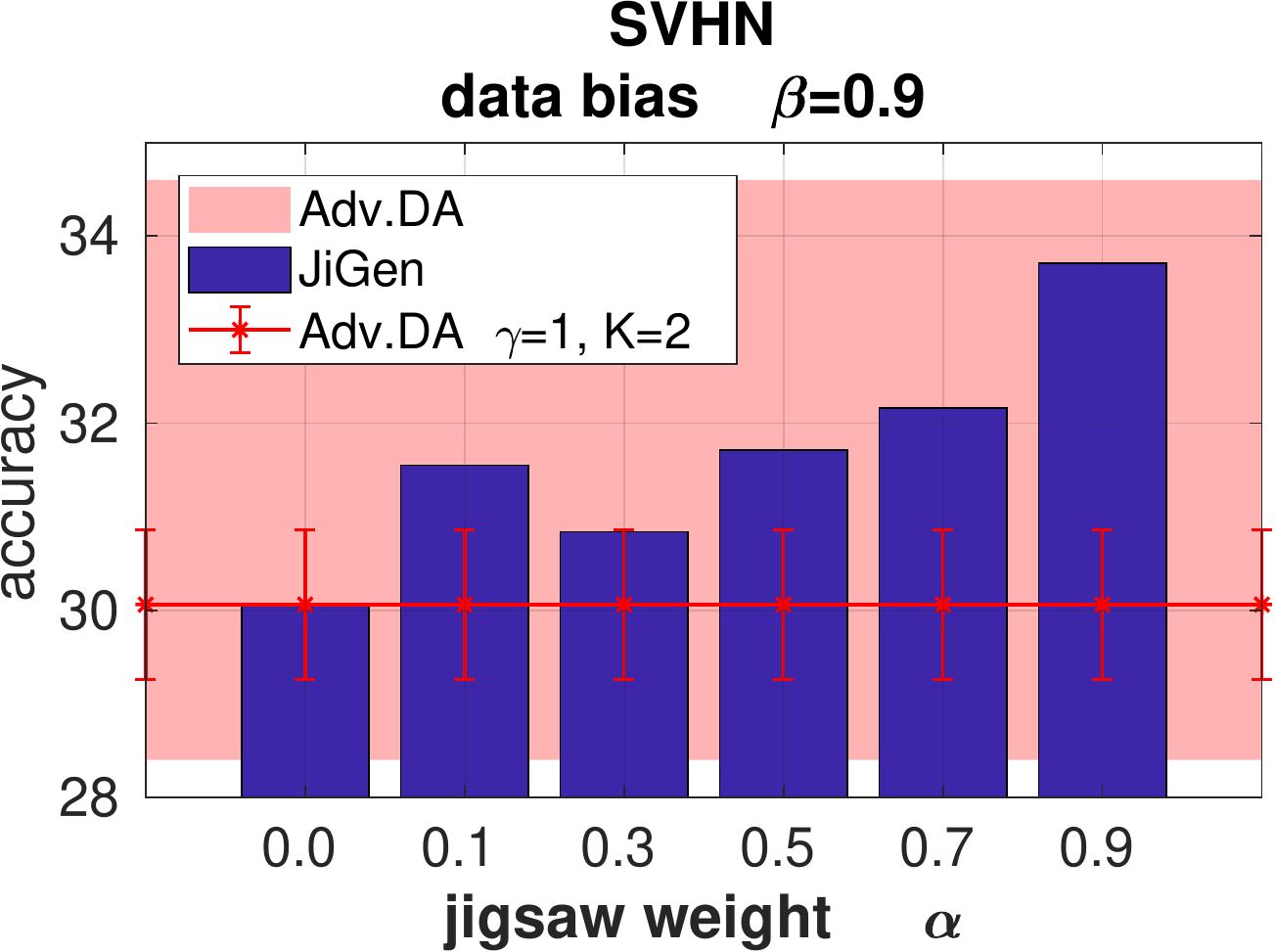}
    \caption{Single Source Domain Generalization experiments. We analyze the performance
    of \our in comparison with the method Adv.DA \cite{Volpi_2018_NIPS}. The
    shaded background area covers the overall range of results of Adv.DA obtained when changing
    the hyper-parameters of the method. The reference result of Adv.DA ($\gamma=1$, $K=2$) together 
    with its standard deviation is indicated here by the horizontal red line.
    The blue histogram bars show the performance of \our when changing the jigsaw weight $\alpha$
    and data bias $\beta$.}
    \label{jigen:fig:singesource}
\end{figure*}

\subsubsection{Single Source Domain Generalization}
The generalization ability of a model is intrinsically related to the way in which it was trained so it
depends both on the used data and on its own learning structure. To understand the potentialities of 
\our we decided to limit the variability in the training data by considering experiments 
with a single source domain, so that we can better focus on the regularization effect provided by 
the  jigsaw task in the learning process.

For these experiments we compare against the generalization method based on adversarial data augmentation (Adv.DA) recently presented in  \cite{Volpi_2018_NIPS}. This work proposes an iterative procedure
that perturbs the data to make them hard to recognize under the current model and then combine them with the
original samples while solving the classification task.
We reproduced the experimental setting used  in \cite{Volpi_2018_NIPS} and adopt a similar 
result display style with bar plots for experiments on the MNIST-M and SVHN target datasets when training on MNIST.
In Figure \ref{jigen:fig:singesource} we show the performance of \our when varying the data bias $\beta$ 
and the jigsaw weight $\alpha$. With the red background shadow we indicate the overall range 
covered by Adv.DA results when changing its parameters\footnote{The whole set of results is provided as supplementary material of \cite{Volpi_2018_NIPS}.}, while the horizontal line is the reference Adv.DA results
around which the authors of \cite{Volpi_2018_NIPS} ran their parameter ablation analysis. The figure indicates that,
although Adv.DA can reach high peak values, it is also very sensitive to the chosen hyperparameters. 
On the other hand, \our is much more stable and it is always better than the lower accuracy value of Adv.DA with 
a single exception for SVHN and data bias 0.5, but we know from the ablation analysis, that this
corresponds to a limit case for the proper combination of object and jigsaw classification.
Moreover, \our gets close to Adv.DA reference results for MNIST-M and significantly outperform it for SVHN. 

\begin{table}[tb]
\begin{center} \small
\resizebox{\textwidth}{!}{
\begin{tabular}{@{}c@{}c@{}c@{~~}c@{~~}c@{~~}c|@{~~}c}
\hline
\multicolumn{2}{@{}c@{}}{\textbf{PACS-DA}}  & \textbf{art\_paint.} & \textbf{cartoon} &  \textbf{sketches} & \textbf{photo} &   \textbf{Avg.}\\ \hline
\multicolumn{7}{@{}c@{}}{\textbf{Resnet-18}}\\
\hline
\multirow{3}{*}{\cite{mancini2018boosting}}& Deep All & 74.70 & 72.40 & 60.10 & 92.90 & 75.03\\
& Dial & 87.30 & 85.50 & 66.80 & 97.00 &  84.15 \\
& DDiscovery  &  \textbf{87.70}	& \textbf{86.90}	& 69.60 & 97.00		& 85.30\\
\hline
& Deep All & 77.85  & 74.86  & 67.74  & 95.73  & 79.05 \\
& \hspace{-3mm}\textbf{\our$_{\alpha^s=\alpha^t=0.7}$} &  84.88 &	81.07 &	\textbf{79.05} &	\textbf{97.96} &	\textbf{85.74} \\
\hline
& \our$_{\alpha^t=0.1}$ &  85.58 &	82.18 &	78.61  &	98.26 &	86.15 \\
& \our$_{\alpha^t=0.3}$ &  85.08 &	81.28 &	81.50 &	97.96 &	86.46 \\
& \our$_{\alpha^t=0.5}$ &  85.73 &	82.58 &	78.34 &	98.10 &	86.19 \\
& \our$_{\alpha^t=0.9}$ &  85.32 &	80.56 &	79.93 &	97.63 &	85.86 \\
\hline
\end{tabular}
}
\caption{Multi-source Domain Adaptation results on PACS obtained as average over three repetitions for each run. Besides considering the same jigsaw loss weight for source and target samples $\alpha^s=\alpha^t$, we also tuned just the target jigsaw loss weight while keeping $\alpha^s=0.7$ showing that we can get even higher results.}
\label{jigen:table:resultsDA_PACS}
\end{center}
\end{table}

\subsubsection{Unsupervised Domain Adaptation}
When unlabeled target samples are available at training time we can let the jigsaw puzzle task involve these data.
Indeed patch reordering does not need image labels and running the jigsaw optimization process on both source 
and target data may positively influence the source classification model for adaptation. 
To verify this intuition we considered again the PACS dataset and used it in the same unsupervised domain 
adaptation setting of \cite{mancini2018boosting}. This previous work proposed a method to first discover 
the existence of multiple latent domains in the source data and then differently adapt their knowledge to the
target depending on their respective similarity. It has been shown that this domain discovery 
(DDiscovery) technique outperforms other powerful adaptive approaches as Dial
\cite{CarlucciICIAP17} when the source actually includes multiple domains. Both these methods exploit 
the minimization of the entropy loss as an extra domain alignment condition: in this way
the source model when predicting on the target samples is encouraged to assign maximum prediction 
probability to a single label rather than distributing it over multiple class options. 
For a fair comparison we also turned on the entropy loss for \our with weight $\gamma=0.1$.
Moreover, we considered two cases for the jigsaw loss: either keeping the weight $\alpha$ already
used for the PACS-Resnet-18 DG experiments for both the source and target data ($\alpha=\alpha^s=\alpha^t=0.7$),
or treating the domain separately with a dedicated lower weight for the jigsaw target loss
($\alpha^s=0.7$, $\alpha^t=[0.1,0.3,0.5,0.7]$). 
The results for this setting are summarized in Table \ref{jigen:table:resultsDA_PACS}. The obtained accuracy 
indicates that \our outperforms the competing methods on average and in particular the difficult task of 
recognizing images of sketches is the one that better shows its advantage. Furthermore, the advantage remains
true regardless of the specific choice of the target jigsaw loss weight.

\subsection{Conclusions}

In this paper we showed for the first time that generalization across domains can be achieved effectively by learning to classify and learning intrinsic invariances in images at the same time. We focused on learning spatial co-location of image parts, and proposed a simple yet powerful framework that can accommodate a wide spectrum of pre-trained convolutional architectures. Our method \our can be 
seamlessly used for domain adaptation and generalization scenarios, always with great effectiveness as shown by the experimental results.

We see this paper as opening the door to a new research thread in domain adaptation and generalization. While here we focused on a specific type of invariance, several other regularities could be learned possibly leading to an even stronger benefit. Also, the simplicity of our approach calls for testing its effectiveness in applications different from object categorization, like semantic segmentation and person re-identification, where the domain shift effect strongly impact the deployment of methods in the wild. 

%% file: iciap.tex
\section{Tackling PDA with self-supervision}
\renewcommand{\our}{JiGen\xspace}

\textit{Domain adaptation approaches have shown promising results in reducing the marginal distribution difference
among visual domains. They allow to train reliable models that work over datasets of different nature (photos, paintings etc.), but they still struggle when the domains do not share an identical label space. 
In the partial domain adaptation setting, where the target covers only a subset of the source classes, it is 
challenging to reduce the domain gap without incurring in negative transfer. Many solutions just keep the standard 
domain adaptation techniques by adding heuristic sample weighting strategies. In this work we show how the 
self-supervisory signal obtained from the spatial co-location of patches can be used to define a side task 
that supports adaptation regardless of the exact label sharing condition across domains. 
We build over a recent work that introduced a jigsaw puzzle task for domain generalization: we
describe how to reformulate this approach for partial domain adaptation and we show how it boosts
existing adaptive solutions when combined with them. The obtained experimental results on 
three datasets supports the effectiveness of our approach.}

\vspace{4mm}Today the most popular synonym of Artificial Intelligence is Deep Learning: 
new convolutional neural network architectures constantly hit the headlines by improving the state of the 
art for a wide variety of machine learning problems and applications with impressive results.
The large availability of annotated data, as well as the assumption of training and testing
on the same domain and label set, are important ingredients of this success. However
this closed set condition is not realistic and the learned models cannot be said fully intelligent. 
Indeed, when trying to summarize several definitions of intelligence from dictionaries, psychologists 
and computer scientists of the last fifty years, it turns out that all of them highlight as fundamental 
the  ability to adapt and achieve goals in a wide range of environments and conditions \cite{AI}.
Domain Adaptation (DA) and Domain Generalization (DG) methods are trying to go over this 
issue and allow the application of deep learning models in the wild. Many DA and DG approaches have been 
developed for the object classification task to reduce the domain gap across samples obtained from different
acquisition systems, different illumination conditions and visual styles, but most of them keep a strong 
control on the class set, supposing that the trained model will be deployed exactly on the same categories 
observed during training. When part of the source classes are missing at test time, those models show 
a drop in performance which indicates the effect of negative transfer in this 
Partial Domain Adaptation (PDA) setting. 
The culprit must be searched in the need of solving two challenging tasks at the same time: one that exploits
all the available source labeled data to train a reliable classification model in the source domain 
and another that estimates and minimizes the marginal distribution difference between source and target, 
but disregards the potential presence of a conditional distribution shift. 
Very recently it has been shown that this second task may be substituted with self-supervised objectives 
which are agnostic with respect to the domain identity of each sample. 
In particular, \cite{jigsawCVPR19} exploits image patch shuffling 
and reordering as a side task over multiple sources: it leverages the intrinsic regularity 
of the spatial co-location of patches and generalizes to new domains. This information appears 
also independent from the specific class label of each image, which makes it an interesting 
reference knowledge also when the class set of source and target are only partially overlapping. 
We dedicate this work to investigate how 
the jigsaw puzzle task of \cite{jigsawCVPR19} performs in the PDA setting and how it can be
reformulated to reduce the number of needed learning parameters. 
The results on three different datasets indicate that our approach outperforms several competitors 
whose adaptive solutions include specific strategies to 
down-weight the samples belonging to classes supposedly absent from the target. We also discuss 
how such a re-scaling  process can be combined with the jigsaw puzzle obtaining further 
gains in performance.

\subsection{Solving jigsaw puzzles for Partial Domain Adaptation}

\begin{figure}[t]
  \centering
\includegraphics[width=\textwidth]{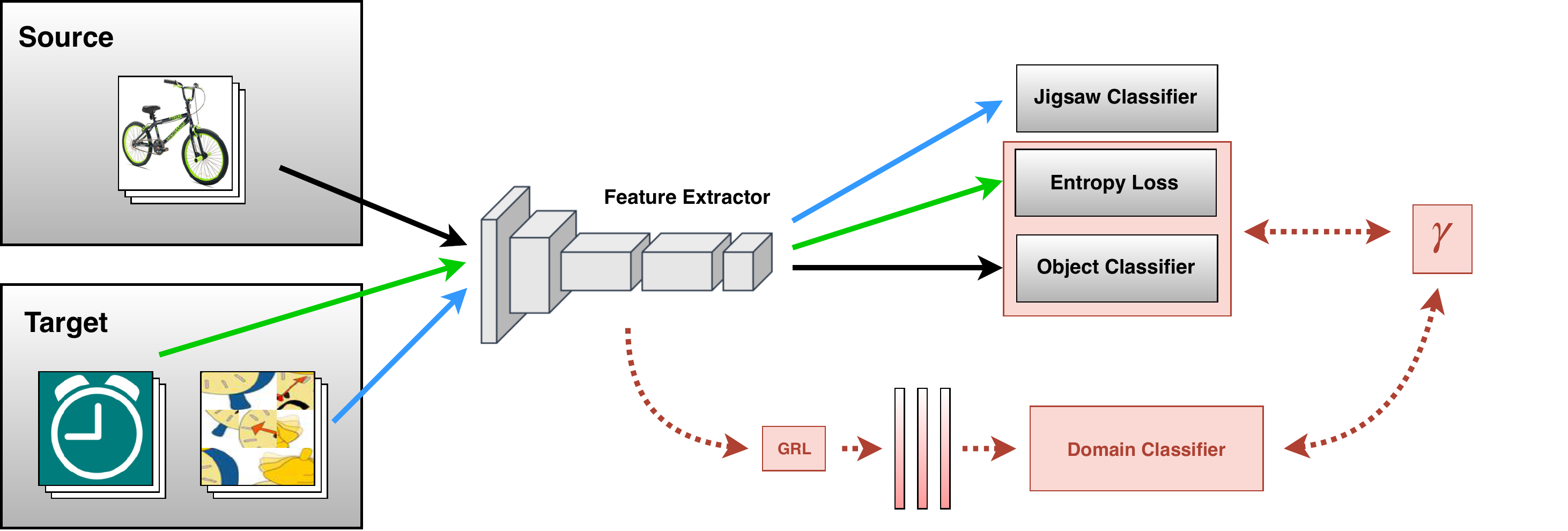}
\caption{Schematic representation of our SSPDA approach. All the parts in gray describe the main blocks 
of the network with the solid line arrows indicating the contribution of each group of training samples 
to the corresponding final tasks and related optimization objectives according to the assigned blue/green/black
colors. 
The blocks in red illustrate the domain adversarial 
classifier with the gradient reversal layer (GRL) and source sample weighting procedure (weight $\gamma$) that can be added to SSPDA}
\label{iciap:fig:scheme}
\end{figure}

\subsubsection{Problem Setting}
Let us introduce the technical terminology for the PDA scenario. We have $n^s$ annotated samples from a source domain $\mathcal{D}_s = \{(\mathbf{x}_i^s,\mathbf{y}_i^s)\}_{i=1}^{n^s}$, drawn from the distribution $S$, and $n^t$ unlabeled examples of the target domain $\mathcal{D}_t = \{\mathbf{x}_j^t\}_{j=1}^{n^t}$ drawn from a different distribution $T$. The label space of the target domain is contained in that of the source domain $\mathcal{Y}_t \subseteq \mathcal{Y}_s$. Thus, besides dealing with the marginal shift $S\neq T$ as in standard unsupervised domain adaptation, it is necessary to take care of the difference in the label space which makes the problem even more challenging. If this information is neglected and the matching between the whole source and target data is forced, any adaptive method may incur in a degenerate case producing worse performance than its plain non-adaptive version. Still the objective remains that of learning both class discriminative and domain invariant feature models 
which can be formulated as a multi-task learning problem \cite{Caruana:1997}. Instead of just focusing on the explicit reduction of the feature domain discrepancy, one could consider some inherent characteristics shared by any visual domain regardless of the assigned label and derive a learning problem to solve together with the main classification task. By leveraging the inductive bias of related objectives, multi-task learning regularizes the overall model and improves generalization having as an
implicit consequence the reduction of the domain bias. This reasoning is at the basis of the recent work \cite{jigsawCVPR19}, which proposed to use jigsaw puzzle as a side task for closed set domain adaptation and generalization: the model named JiGen is described in details in the next subsection.

\subsubsection{Jigsaw Puzzle Closed Set Adaptation}
Starting from the $n^s$ labeled and $n^t$ unlabeled images, the method in \cite{jigsawCVPR19} decomposes them according to an $3\times 3$ grid obtaining $9$ squared patches from every sample, which are then moved from their original location and re-positioned randomly to form a shuffled version $\mathbf{z}$ of the original image $\mathbf{x}$. Out of all the $9!$ possibilities, a set of $p=1,\ldots, P$ permutations are chosen on the basis of their maximal reciprocal Hamming distance \cite{NorooziF16} and used to define a jigsaw puzzle classification task which consists in recognizing the index $p$ of the permutation used to scramble a certain sample.
All the original $\{(\mathbf{x}_i^s,\mathbf{y}_i^s)\}_{i=1}^{n^s}$, $\{\mathbf{x}_j^t\}_{j=1}^{n^t}$ as well as the shuffled versions of the images $\{(\mathbf{z}_k^s,\mathbf{p}_k^s)\}_{k=1}^{K^s}$ , $\{(\mathbf{z}_k^t,\mathbf{p}_k^t)\}_{k=1}^{K^t}$ 
are given as input to a multi-task
deep network where the convolutional feature extraction backbone is indicated by $G_f$ and is parametrized by $\theta_f$, 
while the classifier $G_c$ of the object labels and $G_p$ of the permutation indices, are parametrized respectively by $\theta_c$ and $\theta_p$. The source samples are involved both in the object classification and in the jigsaw puzzle classification task, while the unlabeled target samples deal only with the puzzle task. To further exploit the available target data, the uncertainty of the estimated prediction $\hat{\mathbf{y}}^t=G_c(G_f(\mathbf{x}^t))$ is evaluated through
the entropy $H  
=-\sum_{l=1}^{|\mathcal{Y}_s|}\hat{y}^t_{l} \log \hat{y}^t_{l}$
and minimized to enforce the decision boundary to pass through low-density areas.

Overall the end-to-end JiGen multi-task network is trained by optimizing the following objective 
\begin{align} \small
    \arg \min_{\theta_f, \theta_c, \theta_p} ~~ & 
    \frac{1}{n^s} \sum_{i=1}^{n^s}   \mathcal{L}_c(G_c(G_f(\mathbf{x}^s_i),y_i^s)) +  \alpha_s \frac{1}{K^s}\sum_{k=1}^{K^s}\mathcal{L}_p(G_p(G_f(\mathbf{z}^s_k),p_k^s)) +  \nonumber \\ 
      &
    \eta~\frac{1}{n^t} \sum_{j=1}^{n^t}  H(G_c(G_f(\mathbf{x}^t_j))) +  \alpha_t \frac{1}{K^t}
    \sum_{k=1}^{K^t}\mathcal{L}_p(G_p(G_f(\mathbf{z}^t_k),p_k^t)) ~,
    \label{iciap:equationJIGEN}
\end{align}
where $\mathcal{L}_c$ and $\mathcal{L}_p$ are cross entropy losses for both the object and puzzle classifiers. In the closed set scenario, the experimental evaluation of \cite{jigsawCVPR19} 
showed that tuning two different hyperparameters $\alpha_s$ and $\alpha_t$ respectively for the source and target puzzle classification loss is beneficial with respect to just using a single value $\alpha=\alpha_s=\alpha_t$, while it is enough to assign a small value to $\eta$ ($\sim 10^{-1}$). 

\subsubsection{Jigsaw Puzzle for Partial Domain Adaptation}
We investigate here if the approach in \cite{jigsawCVPR19} can be extended to the PDA setting and how to improve it to deal with the specific characteristics of the considered scenario.
The two $\mathcal{L}_p$ terms in (\ref{iciap:equationJIGEN}) provide a domain shift reduction effect on the learned feature representation, however their co-presence seem redundant: indeed the features are already chosen to minimize the 
source classification loss and the self-supervised jigsaw puzzle task on the target back-propagates its effect 
directly on the learned features inducing a cross-domain adjustment. By following this logic, we decided to drop the source jigsaw puzzle term, which corresponds to setting $\alpha_s=0$. This choice has a double positive effect: on one side it allows to reduce the number of hyper-parameters in the learning process leaving space for the introduction of other complementary learning conditions, on the other we let the self-supervised module focus only on the samples from the target without involving the extra classes of the source. In the following we indicate this approach as SSPDA: Self-Supervised Partial Domain Adaptation. A schematic illustration of the method is presented in Figure \ref{iciap:fig:scheme}.

\subsection{Combining Self-Supervision with other PDA Strategies}
\label{iciap:otherpda}
To further enforce the focus on the shared classes, SSPDA can be extended to integrate a weighting mechanism analogous to that presented in \cite{PADA_eccv18}. The source classification output on the target data are accumulated as follow 
$\boldsymbol{\gamma} = \frac{1}{n^t}\sum_{j=1}^{n^t}\hat{\mathbf{y}}_j^t$
and normalized $\boldsymbol{\gamma} \leftarrow \boldsymbol{\gamma}/ \max(\boldsymbol{\gamma})$, obtaining a $|\mathcal{Y}_t|$-dimensional vector that quantifies the contribution of each source class. 
Moreover, we can easily integrate a domain discriminator $G_d$ with a gradient reversal layers as in \cite{Ganin:DANN:JMLR16}, and adversarially maximize the related binary cross-entropy to increase the domain confusion, taking also into consideration the defined class weighting procedure for the source samples. In more formal terms, the final objective of our multi-task problem is
\begin{align} \small
    \arg \min_{\theta_f, \theta_c, \theta_p} \max_{\theta_d} ~~ 
    \frac{1}{n^s} \sum_{i=1}^{n^s}  \gamma_{y} \Big( \mathcal{L}_c(G_c(G_f(\mathbf{x}^s_i),y_i^s)) +  \lambda \log (G_d(G_f(\mathbf{x}^s_i))) \Big)+  \nonumber \\ 
    \frac{1}{n^t} \sum_{j=1}^{n^t} \gamma_{y} \Big( \eta  H(G_c(G_f(\mathbf{x}^t_j))) +  \lambda \log (1 - G_d(G_f(\mathbf{x}^t_j)))\Big)+ \nonumber \\
    \alpha_t \frac{1}{K^t}
    \sum_{k=1}^{K^t}\mathcal{L}_p(G_p(G_f(\mathbf{z}^t_k),p_k^t))~,
    \label{iciap:equationOUR}
\end{align}

\noindent where 
$\lambda$ is a hyper-parameter that adjusts the importance of the introduced domain discriminator. We adopted the same scheduling of \cite{Ganin:DANN:JMLR16} to update the value of $\lambda$, so that the importance of the domain discriminator increases with the training epochs,
from 0 at the first epoch to $\lambda$-max at the last epoch, 
avoiding the noisy signal at the early stages of the learning procedure. 
When $\lambda=0$ and $\gamma_{y} = 1/|\mathcal{Y}_s| $ we fall back to SSPDA. 
Note that here the entropy term $H$ is also weighted with a per-class score $\gamma$ based on the predicted target class.

\subsection{Experiments}

\subsubsection{Implementation Details}
We implemented all our deep methods in PyTorch. Specifically the main backbone of our SSPDA network is a ResNet-50 pre-trained on ImageNet and corresponds to the feature extractor defined as $G_f$, while the specific object and puzzle classifiers $G_c,G_p$ are implemented each by an ending fully connected layer. The domain classifier $G_d$ is introduced by adding three fully connected layers after the last pooling layer of the main backbone, and using a sigmoid function for the last activation as in \cite{Ganin:DANN:JMLR16}. By training the network end-to-end we fine-tune all the feature layers, while $G_c, G_p$ and $G_d$ are learned from scratch. We train the model with backpropagation using SGD with momentum set at $0.9$, weight decay  $0.0005$ and initial learning rate $0.0005$. We use a batch size of 64 (32 source samples + 32 target samples) and, following \cite{jigsawCVPR19}, we shuffle the tiles of each input image with probability $1 - \beta$, with $\beta = 0.7$. Shuffled samples are only used for the auxiliary jigsaw task, therefore only unshuffled (original) samples are passed to $G_d$ and $G_c$ for domain and label predictions. The entropy weight $\eta$ and jigsaw task weight $\alpha_t$ are set respectively to 0.2 and 1. Our data augmentation protocol is the same of \cite{jigsawCVPR19}. 

\vspace{4mm}\noindent\textbf{Model Selection}

\vspace{4mm}As standard practice, we used $10\%$ of the source training domain to define a validation set on which the model is evaluated after each epoch $e$. The obtained accuracy $A_e$ is dynamically averaged with the value obtained at the previous epoch with $A_e\leftarrow w A_{e-1} + (1-w) A_{e}$. The final model to apply on the target is chosen as
the one producing the top accuracy over all the epochs $e=1,\ldots, E$. We noticed that this procedure leads to a more reliable selection of the best trained model, preventing to choose one that might have overfitted on the validation set. For all our experiments we kept $w=0.6$.
We underline that this smoothing procedure was applied uniformly on all our experiments. Moreover the 
hyper-parameters of our model are the same for all the domain pairs within 
each dataset and also across all the datasets. In other words we did not select a tailored set of parameters for
each sub-task of a certain dataset which could lead to further performance gains, a procedure used 
in previous works \cite{PADA_eccv18,SAN}.

\subsubsection{Results of SSPDA}
Here we present and discuss the obtained classification accuracy results on the three considered datasets: Office-31 in Table \ref{iciap:tab:office31}, Office-Home in Table \ref{iciap:tab:office-home} and VisDA in Table \ref{iciap:tab:visda2017}. Each table is
organized in three horizontal blocks: the first one shows the results obtained with standard DA methods, the second block illustrates the performance with algorithms designed to deal with PDA and the third one includes the scores of \our and SSPDA. Only Table \ref{iciap:tab:office31} has an extra fourth block that we will discuss in details in the following section.

Both \our and SSPDA exceed all plain DA methods and present accuracy value comparable to those of the PDA methods. In particular
SSPDA is always better than PADA \cite{PADA_eccv18} on average, and for both Office-Home and VisDA it also outperforms all the other competing PDA methods with the only exception of IAFN \cite{featurenorm_PDA}. We highlight that this approach uses a competitive version of ResNet-50 as backbone, with extra bottleneck fully connected layers which add about 2 million 
parameters to the standard version of ResNet-50 that we adopted.

\begin{table}[t]
\centering
\caption{Classification accuracy in the PDA setting defined on the Office-31 dataset with all the 31 classes used for each source domain, and a fixed set of 10 classes used for each target domain. The results are obtained using 10 random crop predictions on each target image and are averaged over three repetitions of each run.}
\resizebox{\textwidth}{!}{
\begin{tabular}{l@{~~~}C{1.3cm}C{1.3cm}C{1.3cm}C{1.3cm}C{1.3cm}C{1.3cm}|C{1.3cm}}
\hline
       &       \multicolumn{6}{c|}{\textbf{Office-31}} &       \\
       & \textbf{A}$\rightarrow$\textbf{W} & \textbf{D}$\rightarrow$\textbf{W} & \textbf{W}$\rightarrow$\textbf{D} & \textbf{A} $\rightarrow$\textbf{D} & \textbf{D}$\rightarrow$\textbf{A} & \textbf{W}$\rightarrow$\textbf{A} & \textbf{Avg.} \\
\hline
Resnet-50           & $75.37$ & $94.13$ & $98.84$ & $79.19$ & $81.28$ & $85.49$ & $85.73$ \\ 
\hline
DAN\cite{Long:2015}               & $59.32$ & $73.90$ & $90.45$ & $61.78$ & $74.95$ & $67.64$ & $71.34$ \\ 
DANN\cite{Ganin:DANN:JMLR16}      & $75.56$ & $96.27$ & $98.73$ & $81.53$ & $82.78$ & $86.12$ & $86.50$ \\ 
ADDA\cite{Hoffman:Adda:CVPR17}    & $75.67$ & $95.38$ & $99.85$ & $83.41$ & $83.62$ & $84.25$ & $87.03$ \\ 
RTN\cite{long2016unsupervised}    & $78.98$ & $93.22$ & $85.35$ & $77.07$ & $89.25$ & $89.46$ & $85.56$ \\ 
\hline
IWAN \cite{IWAN}                & $89.15$ & $99.32$ & $99.36$ & $90.45$ & $\textbf{95.62}$ & $94.26$ & $94.69$ \\ 
SAN \cite{SAN}                  & $93.90$ & $99.32$ & $99.36$ & $94.27$ & $94.15$ & $88.73$ & $94.96$ \\ 
PADA\cite{PADA_eccv18}            & $86.54$ & $\mathbf{99.32}$ & $\mathbf{100}$ & $82.17$ & $92.69$ & $\mathbf{95.41}$ & $92.69$ \\ 
TWIN \cite{TWIN_PDA}    & $86.00$ & $99.30$ & $\mathbf{100}$ & $86.80$ & $94.70$ & $94.50$ & $93.60$ \\

\hline\hline
\our \cite{jigsawCVPR19} & $ 92.88 $ & $92.43$   & $98.94$  & $89.6$  & $84.06$  & $92.94$    &   $91.81$  \\
SSPDA & $91.52$ & $92.88$  & $98.94$  & $90.87$  & $90.61$  & $94.36$   &  $93.20$  \\
\hline
SSPDA-$\gamma$ & $99.32$ & $94.69$  & $99.36$  & $96.39$  & $86.36$  & $94.22$   & $95.06$    \\
SSPDA-PADA & $\textbf{99.66}$ & $94.46$  & $99.57$  &   $\textbf{97.67}$ & $87.33$  & $94.26$   &  $\textbf{95.49}$   \\
\hline
\end{tabular}
}
    \label{iciap:tab:office31} 
\end{table}

\begin{table}[t]
\caption{Classification accuracy in the PDA setting defined on the Office-Home dataset with all the 65 classes used for each source domain, and a fixed set of 25 classes used for each target domain.
The results are obtained by averaging over three repetitions of each run.}
\centering
\resizebox{\textwidth}{!}{
\begin{tabular}{l@{~~~}C{1.4cm}C{1.4cm}C{1.4cm}C{1.4cm}C{1.4cm}C{1.4cm}C{1.4cm}C{1.4cm}C{1.4cm}C{1.4cm}C{1.4cm}C{1.4cm}|C{1.4cm}}
\hline
       &       \multicolumn{12}{c|}{\textbf{Office-Home}} &       \\
       & \textbf{Ar}$\rightarrow$\textbf{Cl} & \textbf{Ar}$\rightarrow$\textbf{Pr} & \textbf{Ar}$\rightarrow$\textbf{Rw} & \textbf{Cl} $\rightarrow$\textbf{Ar} & \textbf{Cl}$\rightarrow$\textbf{Pr} & \textbf{Cl}$\rightarrow$\textbf{Rw} & \textbf{Pr}$\rightarrow$\textbf{Ar} & \textbf{Pr}$\rightarrow$\textbf{Cl}  & \textbf{Pr}$\rightarrow$\textbf{Rw}  & \textbf{Rw}$\rightarrow$\textbf{Ar}  & \textbf{Rw}$\rightarrow$\textbf{Cl}  & \textbf{Rw}$\rightarrow$\textbf{Pr} &  \textbf{Avg.} \\
\hline
Resnet-50          & $38.57$ & $60.78$ & $75.21$ & $39.94$ & $48.12$ & $52.90$ & $49.68$ & $30.91$ & $70.79$ & $65.38$ & $41.79$ & $70.42$ & $53.71$ \\ 
\hline
DAN\cite{Long:2015}               & $44.36$ & $61.79$ & $74.49$ & $41.78$ & $45.21$ & $54.11$ & $46.92$ & $38.14$ & $68.42$ & $64.37$ & $45.37$ & $68.85$ & $54.48$ \\
DANN\cite{Ganin:DANN:JMLR16}      & $44.89$ & $54.06$ & $68.97$ & $36.27$ & $34.34$ & $45.22$ & $44.08$ & $38.03$ & $68.69$ & $52.98$ & $34.68$ & $46.50$ & $47.39$ \\
RTN\cite{long2016unsupervised}   & $49.37$ & $64.33$ & $76.19$ & $47.56$ & $51.74$ & $57.67$ & $50.38$ & $41.45$ & $75.53$ & $70.17$ & $51.82$ & $74.78$ & $59.25$ \\
\hline
IWAN \cite{IWAN}                 & $53.94$ & $54.45$ & $78.12$ & $61.31$ & $47.95$ & $63.32$ & $54.17$ & $52.02$ & $81.28$ & $\textbf{76.46}$ & $56.75$ & $\textbf{82.90}$ & $63.56$\\ 
SAN \cite{SAN}                   & $44.42$ & $68.68$ & $74.60$ & $67.49$ & $64.99$ & $\textbf{77.80}$ & $59.78$ & $44.72$ & $80.07$ & $72.18$ & $50.21$ & $78.66$ & $65.30$\\ 
PADA\cite{PADA_eccv18}           & $51.95$ & $67.00$ & $78.74$ & $52.16$ & $53.78$ & $59.03$ & $52.61$ & $43.22$ & $78.79$ & $73.73$ & $56.60$ & $77.09$ & $62.06$\\
HAFN\cite{featurenorm_PDA}           & $53.35$ & $72.66$ & $80.84$ & $64.16$ & $65.34$ & $71.07$ & $66.08$ & $51.64$ & $78.26$ & $72.45$ & $55.28$ & $79.02$ & $67.51$\\
IAFN\cite{featurenorm_PDA}           & $\textbf{58.93}$ & $\textbf{76.25}$ & $\textbf{81.42}$ & $\textbf{70.43}$ & $\textbf{72.97}$ & $77.78$ & $\textbf{72.36}$ & $\textbf{55.34}$ & $80.40$ & $75.81$ & $60.42$ & $79.92$ & $\mathbf{71.83}$\\
\hline\hline
\our \cite{jigsawCVPR19} & $53.19$ & $65.45$ & $81.30$ & $68.84$ & $58.95$ & $74.34$ & $69.94$ & $50.95$ & $\textbf{85.38}$ & $75.60$ & $60.02$ & $81.96$ & $68.83$    \\
SSPDA & $52.02$ & $63.64$ & $77.95$ & $65.66$ & $59.31$ & $73.48$ & $70.49$ & $51.54$ & $84.89$ & $76.25$ & $\textbf{60.74}$ & $80.86$ & $68.07$ \\
\hline
\end{tabular}
}
    \label{iciap:tab:office-home} 
\end{table}

\begin{table}[t]
    \caption{Classification accuracy in the PDA setting defined on VisDA2017 dataset with all the 12 classes used for each source domain, and a fixed set of 6 classes used for each target domain.
    The results are obtained using 10 random crop predictions on each target image and are averaged over three repetitions of each run.}
\centering
\resizebox{0.4\textwidth}{!}{
\begin{tabular}{l@{~~~}C{3cm}} 
\hline
     \multicolumn{2}{c}{\textbf{VisDA2017}}        \\
       & \textbf{Syn.}$\rightarrow$\textbf{Real} \\
\hline
Resnet-50          & $45.26$ \\ 
\hline
DAN\cite{Long:2015}               & $47.60$ \\
DANN\cite{Ganin:DANN:JMLR16}      & $51.01$  \\
RTN\cite{long2016unsupervised}   & $50.04$ \\
\hline
PADA\cite{PADA_eccv18}           & $53.53$ \\
HAFN\cite{featurenorm_PDA} & $65.06$ \\
IAFN\cite{featurenorm_PDA} & $67.65$\\
\hline\hline
\our \cite{jigsawCVPR19} & $68.33$ \\
SSPDA & $\textbf{68.89}$ \\
\hline
\end{tabular}
}
\label{iciap:tab:visda2017} 
\end{table}

\begin{figure}[ht!]
\centering
\begin{tabular}{c@{~~~}c@{~~~}c}
     \includegraphics[width=0.3\textwidth]{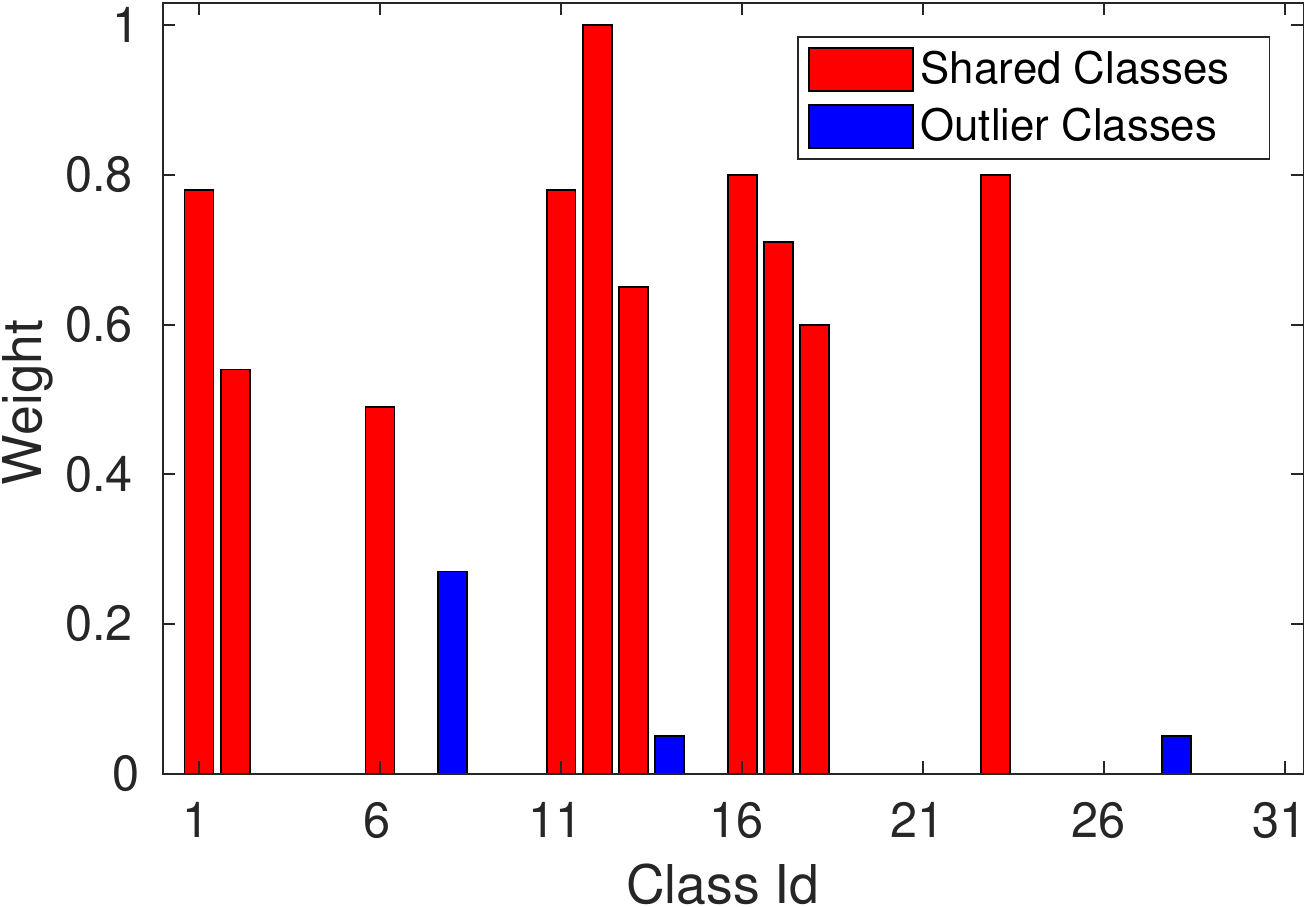} &  \includegraphics[width=0.3\textwidth]{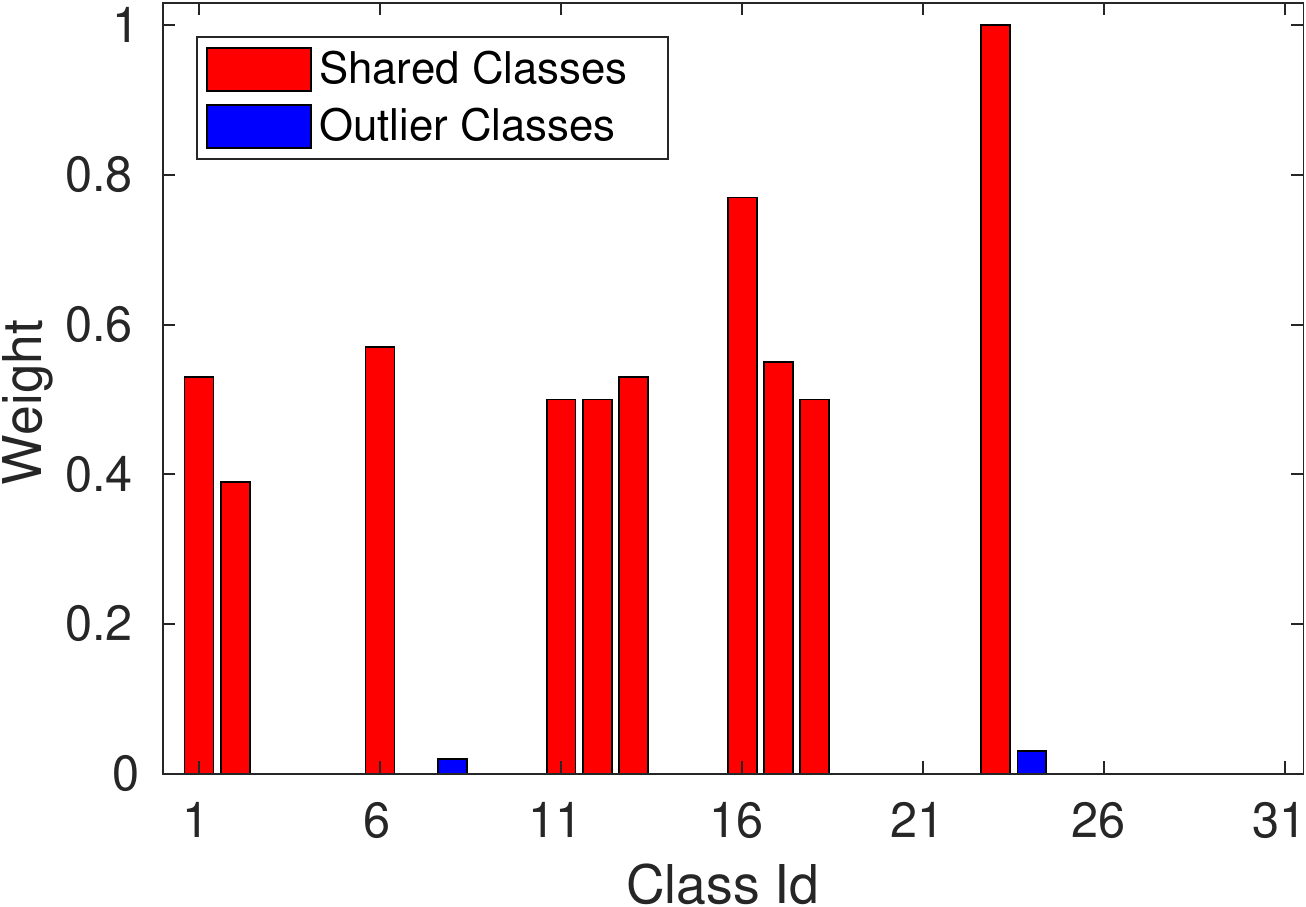} &
     \includegraphics[width=0.3\textwidth]{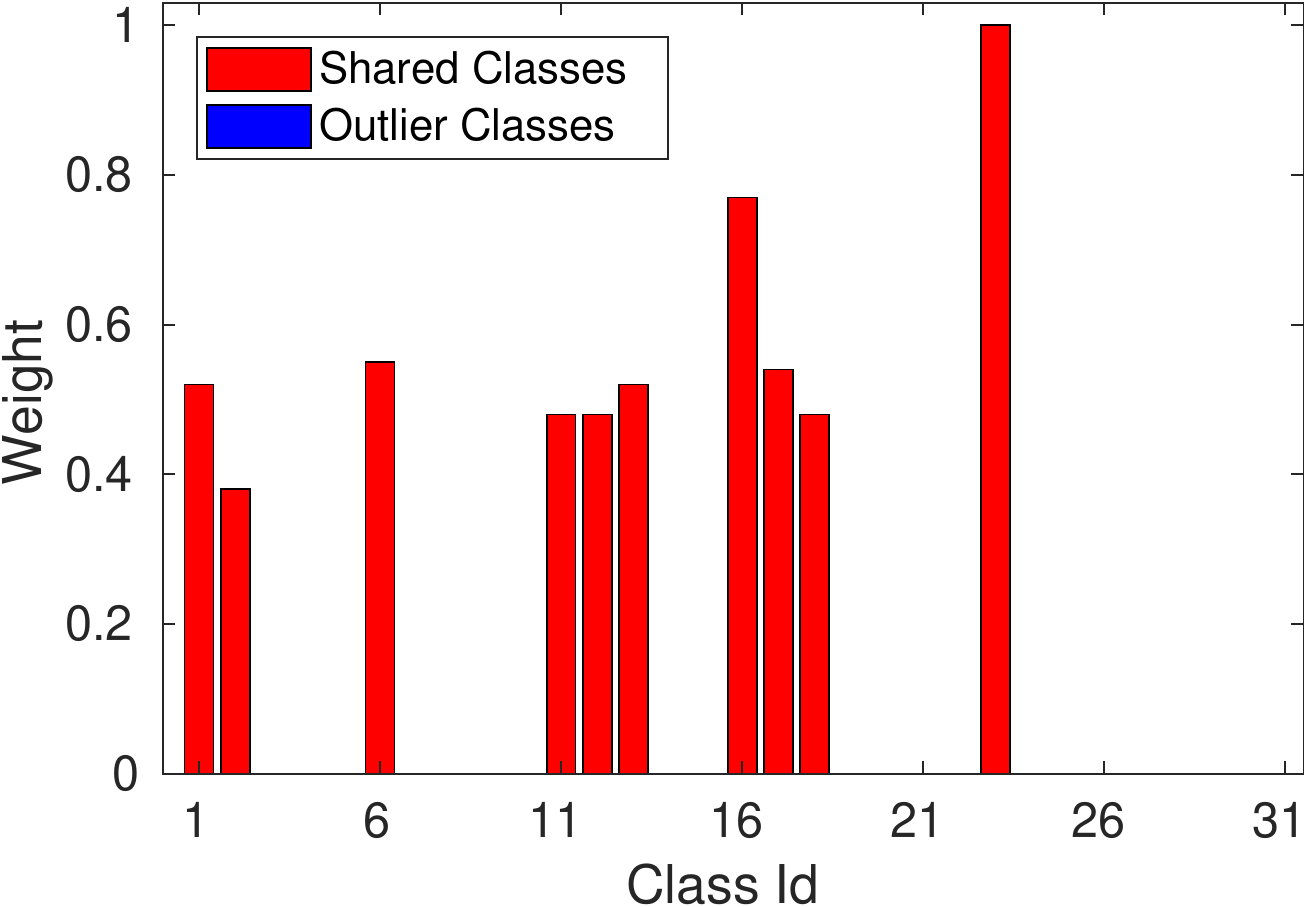}\\
     ~~~PADA  & ~~~SSPDA-$\gamma$ & ~~~SSPDA-PADA\\
\end{tabular}
\caption{Histogram showing the elements of the $\gamma$ vector, corresponding to the class weight learned by PADA, SSPDA-$\gamma$ and SSPDA-PADA for the A$\rightarrow$W experiment.}
\label{iciap:fig:bars}
\end{figure}

\subsubsection{Results of SSPDA combined with  other PDA strategies}
To analyze the combination of SSPDA with the standard PDA source re-weighting technique and the adversarial domain classifier, we extended the experiments on the Office-31 dataset. The bottom part of Table \ref{iciap:tab:office31} reports
the obtained results when we add the estimate of the target class statistics through the weight $\gamma$ (SSPDA-$\gamma$) and when also the domain classifier is included in the network as in \cite{PADA_eccv18} (SSPDA-PADA).
In the first case, estimating the target statistics helps the network to focus only on the shared categories, with an average accuracy improvement of two percentage points over the plain SSPDA. Moreover, since the technique to evaluate $\gamma$ is the same used in \cite{PADA_eccv18}, we can state that the advantage comes from a better alignment of the domain features, thus from the introduction of the self-supervised jigsaw task. Indeed, by comparing the $\gamma$ values on the A$\rightarrow$W domain shift we observe that SSPDA-$\gamma$ is more precise in identifying the missing classes of the target (see Figure \ref{iciap:fig:bars}). 
In the second case, since the produced features are already well aligned across domains, we fixed $\lambda$-max to $0.1$ and observed a further small average improvement, with the largest advantage when the A domain is used as source. From the last bar plot on the right of Figure \ref{iciap:fig:bars} we also observe a further improvement in the identification of the missing target classes.

\subsection{Conclusions}

In this paper we discussed how the self-supervised jigsaw puzzle task can be used for domain adaptation in the challenging partial setting with some of the source classes missing in the target. Since the high-level knowledge captured by the spatial co-location of patches is unsupervised with respect to the image object content, this task can be applied on the unlabeled target samples and help to close the domain gap without suffering from negative transfer. Moreover we showed that the proposed solution can be seamlessly integrated with other existing partial domain adaptation methods and it contributes to a reliable identification of the categories absent in the target with a consequent further improvement in the recognition results.  
In the future we plan to further explore the jigsaw puzzle task also in the open-set scenario where the target contains new unknown classes with respect to the source. 

%% file: oshot.tex
\section{One-Shot unsupervised cross-domain detection}
\renewcommand*{\our}{OSHOT\@\xspace}

\textit{Despite impressive progress in object detection over the last years, it is still an open challenge to reliably detect objects across visual domains. 
All current approaches access a sizable amount of target data at training time. This is a heavy assumption, as often it is not possible to anticipate the domain where a detector will be used, nor to access it in advance for data acquisition. Consider for instance the task of monitoring image feeds from social media: as every image is uploaded by a different user it belongs to a different target domain that is impossible to foresee during training. 
Our work addresses this setting, presenting an object detection algorithm able to perform unsupervised adaptation across domains by using only one target sample, seen at test time. We introduce a multi-task architecture that one-shot adapts to any incoming sample by iteratively solving a self-supervised task on it. We further enhance this auxiliary adaptation with cross-task pseudo-labeling. A thorough benchmark analysis against the most recent cross-domain detection methods and a detailed ablation study show the advantage of our approach.}

\vspace{4mm}Social media feed us every day with an unprecedented amount of visual data. 
Images are uploaded by various actors, from corporations to political parties, institutions, entrepreneurs and private citizens, with roughly $10^2M$ unique images shared everyday on Twitter,  Facebook and  Instagram. For the sake of freedom of expression, control over their content is limited, and their vast majority is uploaded without any textual description of their content. Their sheer magnitude makes it imperative to use algorithms to monitor and make sense of them, finding the right balance between protecting the privacy of citizens and their right of expression, and tracking fake news (often associated with malicious intentions) while fighting illegal and hate content. This in most cases boils down to the ability to automatically associate as many tags as possible to images, which in turns means determining which objects are present in a scene.  

\begin{figure}[tb]
    \centering
    \includegraphics[width=\textwidth]{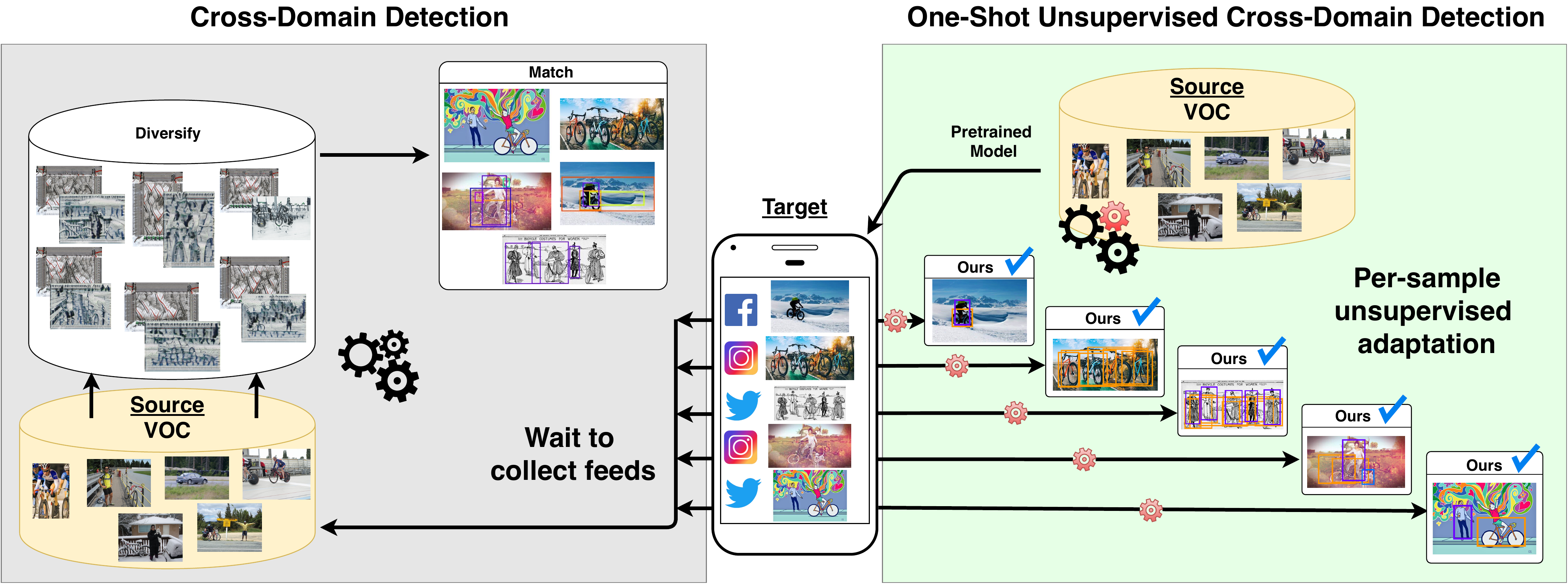}
    \caption{Each social media image comes from a different domain. Existing Cross-Domain Detection algorithms (e.g. \cite{diversify_match_Kim_2019_CVPR} in the left gray box) struggle to adapt in this setting. OSHOT (right) is able to adapt across domains from one single target image, thanks to the combined use of self-supervision and pseudo-labeling
    } 
    \label{oshot:fig:scheme}
\end{figure}

Object detection has been largely investigated since the infancy of computer vision  \cite{viola2001rapid,dalal2005histograms} and continues to attract a large attention in the current deep learning era \cite{girshick2014rich,dai2016r,zhang2018single,liu2018receptive}.
Most of the algorithms assume that training and test data come from the same visual domain \cite{girshick2014rich,girshick2015fast,ren2015faster}. Recently, 
some authors have started to investigate the more challenging yet realistic scenario where the detector is trained on data from a visual source domain, and deployed at test time in a different target domain \cite{Long:2015,LongZ0J17,dcoral,Hoffman:Adda:CVPR17}. This setting is usually referred to as cross-domain detection and heavily relies on concepts and results from the domain adaptation literature \cite{Long:2015,ganin2014unsupervised,Goodfellow:GAN:NIPS2014}. Specifically, it inherits the standard tansductive logic, according to which unsupervised target data is available at training time together with annotated source data, and can be used to adapt across domains. 
This approach is not suitable, neither effective, for monitoring social media feeds.
Consider for instance the scenario depicted in Figure \ref{oshot:fig:scheme}, where there is an incoming stream of images from various social media and the detector is asked to look for instances of the class bicycle. The images come continuously, but they are produced by different users that share them on different social platforms. Hence, even though they might contain the same object, each of them has been acquired by a different person, in a different context, under different viewpoints and illuminations. In other words, each image comes from a different visual domain, distinct from the visual domain where the detector has been trained. This poses two key challenges to current cross-domain detectors: (1) to adapt to the target data, these algorithms need first to gather feeds, and only after  enough target data has been collected they can learn to adapt and start performing on the incoming images; (2) even if the algorithms have learned to adapt on target images from the feed up to time $t$, there is no guarantee that the images that will arrive from time $t+1$ will come from the same target domain.

This is the scenario we address.
We focus on cross-domain detection when only one target sample is available for adaptation, without any form of supervision.
We propose an object detection method able to adapt from one target image, hence suitable for the social media scenario described above. Specifically, we build a multi-task deep architecture that adapts across domains by leveraging over a pretext task.
This auxiliary knowledge is further guided by a cross-task pseudo-labeling that injects the locality specific of object detection into self-supervised learning. The result is an  architecture able to perform unsupervised adaptive object detection from a single image.
Extensive experiments show the power of our method compared to previous state-of-the-art approaches.
\noindent To summarize, the contributions of our paper are as follows:\\
(1) we introduce the One-Shot Unsupervised Cross-Domain Detection setting, a cross-domain detection scenario where the target domain changes from sample to sample, hence adaptation can be learned only from one image. This scenario is especially relevant for monitoring social media image feeds. We are not aware of previous works addressing it.\\
(2) We propose OSHOT, the first cross-domain object detector able to perform one-shot unsupervised adaptation.  Our approach leverages over self-supervised one-shot learning guided by a cross-task pseudo-labeling procedure, embedded into a multi-task architecture. A thorough ablation study showcases the importance of each component.\\
(3) We present a new experimental setup for studying one-shot unsupervised cross-domain adaptation, designed on three existing databases plus a new test set collected from social media feed. We compare against recent algorithms in cross-domain adaptive detection \cite{Saito_2019_CVPR,diversify_match_Kim_2019_CVPR} and one-shot unsupervised learning \cite{Cohen_2019_ICCV}, achieving the state-of-the-art. \\

We make the code of our project available at \url{https://github.com/VeloDC/oshot_detection}.

\subsection{Method}
\label{oshot:sec:method}

\vspace{4mm}\noindent\textbf{Problem Setting}

\vspace{4mm}We introduce the one-shot unsupervised cross-domain detection scenario where our goal is to predict on a single image $x^t$, with $t$ being any target domain not available at training time, starting from $N$ annotated samples of the source domain $S=\{x^s_{i},y^s_{i}\}_{i=1}^N$. Here the structured labels $y^s=(c,b)$ describe class identity $c$ and bounding box location $b$ in each image $x^s$, and we aim to obtain $y^t$ that precisely detects objects in $x^t$ despite the domain shift.

\vspace{4mm}\noindent\textbf{\our strategy} 

\begin{figure*}[tb]
    \centering
\includegraphics[width=\textwidth]{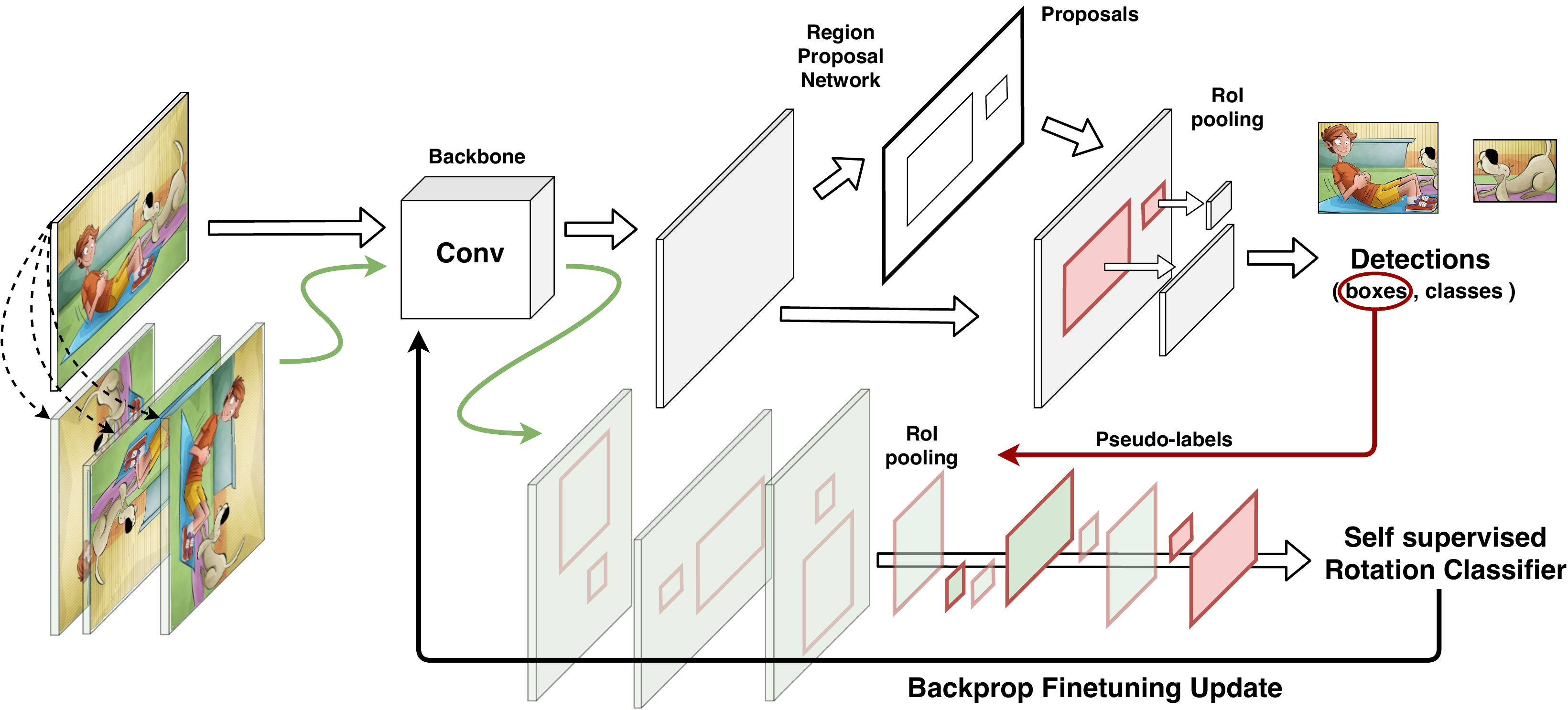}   
\caption{Visualization of the adaptive phase of \our with cross-task pseudo-labeling. 
The target image passes through the network and produces detections. While the class information is not used, the identified boxes are exploited to select object regions from the feature maps of the rotated image. The obtained region-specific feature vectors are finally sent to the rotation classifier. A number of subsequent finetuning iterations allows to adapt the convolutional backbone to the domain represented by the test image}
    \label{oshot:fig:fasterRCNN}
\end{figure*}

\vspace{4mm}To pursue the described goal, our strategy is to train the parameters of a detection learning model such that it can be ready to get the maximal performance on a single unsupervised sample from a new domain after few gradient update steps on it. Since we have no ground truth on the target sample, we implement this strategy by learning a representation that exploits inherent data information as that captured by a self-supervised task, and then finetune it on the target sample (see Figure \ref{oshot:fig:fasterRCNN}). 
Thus, we design our \our to include (1) an initial pretraining phase where we extend a standard deep detection model adding an image rotation classifier, and (2) a following adaptation stage where the network features are updated on the single target sample by further optimization of the rotation objective. 
Moreover, we exploit pseudo-labeling to focus the auxiliary task on the local object context. A clear advantage of this solution is that we decouple source training from target testing, with no need to access the source data while adapting on the target sample.\\

\vspace{4mm}\noindent\textbf{Preliminaries} 

\vspace{4mm}We leverage on Faster R-CNN \cite{ren2015faster} as our base detection model. It is a two-stage detector with three main components: an initial block of convolutional layers, a region proposal network (RPN) and a region-of-interest (ROI) based classifier.  The bottom layers transform any input image $x$ into its convolutional feature map $G_{f}(x|\theta_{f})$ where $\theta_{f}$ is used to parametrize the feature extraction model. The feature map is then used by RPN to generate candidate object proposals. Finally the ROI-wise classifier predicts the category label from the feature vector obtained using ROI-pooling. 
The training objective combines the loss of both RPN and ROI, each of them composed by two terms:

\begin{equation}
\begin{aligned}
\mathcal{L}_{d}(G_{d}(G_{f}(x|\theta_{f})|\theta_{d}), y)= & \big(\mathcal{L}_{class}(c^*) + \mathcal{L}_{regr}(b) \big)_{RPN} + \\
& \big( \mathcal{L}_{class}(c) + \mathcal{L}_{regr}(b) \big)_{ROI}~.
\end{aligned}
\end{equation}

Here $\mathcal{L}_{class}$ is a classification loss to evaluate the object recognition accuracy, while $\mathcal{L}_{regr}$ is a regression loss on the box coordinates for better localization. 
To maintain a simple notation we summarize the role of ROI and RPN with the function $G_{d}(G_{f}(x|\theta_{f})|\theta_{d})$ parametrized by $\theta_{d}$. Moreover, we use $c^*$ to highlight that RPN deals with a binary classification task to separate foreground and background objects, while ROI deals with the multi-class objective  needed to discriminate among $c$ foreground object categories. As mentioned above, ROI and RPN are applied in sequence: they both elaborate on the feature maps produced by the convolutional block, and then influence each other in the final optimization of the multi-task (classification, regression) objective function.

\vspace{4mm}\noindent\textbf{\our pretraining} 

\vspace{4mm}As a first step, we extend Faster R-CNN to include image rotation recognition. Formally, to each source training image $x^s$ we apply four geometric transformations $R(x,\alpha)$ where $\alpha= q\times90^{\circ}$ indicates rotations with $q \in \{1,\ldots,4\}$.
In this way we obtain a new set of samples $\{R(x)_j, q_j\}_{j=1}^M$ where we dropped the $\alpha$ without loss of generality.
We indicate the auxiliary rotation classifier and its parameters respectively with $G_{r}$ and $\theta_{r}$
and we train our network to optimize the following multi-task objective\\

\begin{equation}
\begin{aligned}
    \argmin_{\theta_{f}, \theta_{d}, \theta_{r}} \ & \sum_{i=1}^N\mathcal{L}_{d}(G_{d}(G_{f}(x^s_i|\theta_{f})|\theta_{d}),y^s_i) +  \lambda \sum_{j=1}^M\mathcal{L}_{r}(G_{r}(G_{f}(R(x^s)_j|\theta_{f})|\theta_{r}), q^s_j)~,
\end{aligned}
\end{equation}

where $\mathcal{L}_{r}$ is the cross-entropy loss.
When solving this problem, we can design $G_{r}$ in two different ways. Indeed it can either be a Fully Connected layer that na\"{\i}vely takes as input the feature map produced by the whole (rotated) image $G_{r}(\cdot |\theta_r) = \mbox{FC}_{\theta_r}(\cdot)$, or it can exploit the ground truth location of each object with a subselection of the features only from its bounding box in the original map  
$G_{r}(\cdot |\theta_r) = \mbox{FC}_{\theta_r}(boxcrop(\cdot))$. The $boxcrop$ operation includes pooling to rescale the feature dimension before entering the final FC layer. 
In this last case the network is encouraged to focus only on the object orientation without introducing noisy information from the background and provides better results with respect to the whole image option as we will discuss in Section \ref{oshot:sec:ablation}.
In practical terms, both in the case of image and box rotations, we randomly pick one rotation angle per instance, rather than considering all four of them: this avoids any troublesome unbalance between rotated and non-rotated data when solving the multi-task optimization problem.

\vspace{4mm}\noindent\textbf{\our adaptation} 

\vspace{4mm}Given the single target image $x^t$, we finetune the backbone's parameters $\theta_{f}$ by iteratively solving a self-supervised task on it. 
This allows to adapt the original feature representation both to the content and to the style of the new sample. Specifically, we start from the rotated versions $R(x^t)$  of the provided sample and optimize the rotation classifier through 

\begin{equation}
    \argmin_{\theta_{f}, \theta_{r}}  \mathcal{L}_{r}(G_{r}(G_{f}(R(x^t)|\theta_{f})|\theta_{r}),q^t)~.
    \label{oshot:eq:finetuning}
\end{equation}

This process involves only $G_{f}$ and $G_{r}$, while the RPN and ROI detection components described by $G_{d}$ remain unchanged. In the following we use $\gamma$ to indicate the number of gradient steps (i.e. iterations), with $\gamma=0$ corresponding to the \our{} pretraining phase. At the end of the finetuning process, the inner feature model is  described by $\theta^*_f$ and the detection prediction on $x^t$ is obtained by $y^{t*} = G_{d}(G_{f}(x^t|\theta^*_{f})|\theta_{d})$.\\

\vspace{4mm}\noindent\textbf{Cross-task pseudo-labeling}

\vspace{4mm}As in the pretraining phase, also in the adaptation stage we have two possible choices to design $G_{r}$: either considering the whole feature map $G_{r}(\cdot |\theta_r) = \mbox{FC}_{\theta_r}(\cdot)$, or focusing on the object locations $G_{r}(\cdot |\theta_r) = \mbox{FC}_{\theta_r}(pseudoboxcrop(\cdot))$. For both variants we include dropout to prevent overfitting on the single target sample. With $pseudoboxcrop$ we mean a localized feature extraction operation analogous to that discussed for pretraining, but obtained through a particular form of cross-task self-training. 
Specifically, we follow the self-training strategy used in  \cite{kim2019selftraining,inoue2018cross} with a cross-task variant: instead of reusing the pseudo-labels produced by the source model on the target to update the detector, we exploit them for the self-supervised rotation classifier.
In this way we keep the advantage of the self-training initialization, largely reducing the risks of error propagation due to wrong class pseudo-labels.

More practically, we start from the $(\theta_{f},\theta_{d})$ model parameters of the pretraining stage and we get the feature maps from all the rotated versions of the target sample $G_{f}(\{R(x^t),q\}|\theta_{f})$, $q={1,\ldots,4}$. 
Only the feature map produced by the original image (i.e. $q=4$) is provided as input to the RPN and ROI network components to get the predicted detection $y^{t}=(c,b)=G_{d}(G_{f}(x^t|\theta_{f})|\theta_{d})$. This pseudo-label is composed by the class label $c$ and the bounding box location $b$. We discard the first and consider only the second to localize the region containing an object in all the four feature maps, also recalibrating the position to compensate for the orientation of each map. Once passed through this pseudoboxcrop operation, the obtained features are used to finetune the rotation classifier, updating the bottom convolutional network block.

\subsection{Experiments}

\subsubsection{Datasets}

\begin{figure}[tb]
    \centering
    \includegraphics[width=\textwidth]{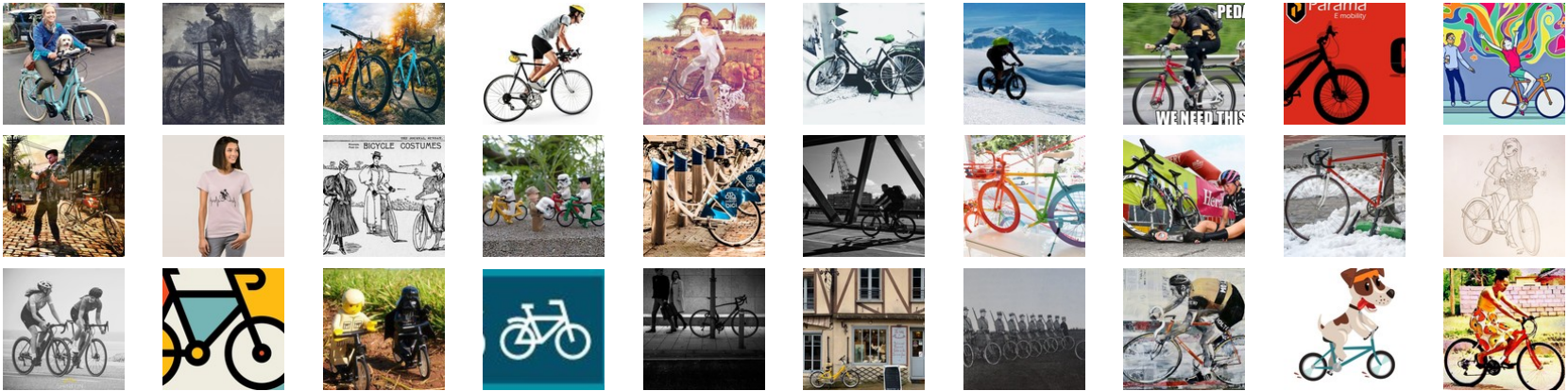}
    \caption{The Social Bikes concept-dataset. %
    A random data acquisition from multiple users/feeds leads to a target distribution with several, uneven domain shifts}
    \label{oshot:fig:socialbikes}
\end{figure}

\vspace{4mm}\noindent\textbf{Social Bikes} 

\vspace{4mm}Social Bikes is our new concept-dataset containing 30 images of scenes with persons/bicycles collected from {Twitter}, {Instagram} and {Facebook} by searching for \#bike tags. Square crops of the full dataset are presented in Figure \ref{oshot:fig:socialbikes}: images acquired randomly from social feeds show diverse style properties and cannot be grouped under a single shared domain.

\subsubsection{Performance analysis}
\label{oshot:sec:setup}

\vspace{4mm}\noindent\textbf{Experimental Setup} 

\vspace{4mm}We evaluate \our on several testbeds using the described datasets. In the following we will use an arrow $Source \rightarrow Target$ to indicate the experimental setting.
Our base detector is Faster-RCNN \cite{massa2018mrcnn} with a ResNet-50 \cite{he2016deep} backbone pre-trained on ImageNet, RPN with 300 top proposals after non-maximum-supression, anchors at three scales (128, 256, 512) and three aspect ratios (1:1, 1:2, 2:1).
For all our experiments we set the IoU threshold at 0.5 for the mAP results, and report the average of three independent runs. 

\vspace{4mm}\noindent\emph{\our pretraining.} We always resize the image's shorter size to 600 pixels and apply random horizontal flipping. Unless differently specified, we train the base network for 70k iterations using SGD with momentum set at $0.9$, the initial learning rate is $0.001$ and decays after 50k iterations. We use a batch size of 1, keep batch normalization layers fixed for both pretraining and adaptation phases and freeze the first 2 blocks of ResNet50. The weight of the auxiliary task is set to $\lambda=0.05$.

\vspace{4mm}\noindent\emph{\our adaptation.} We increase the weight of the auxiliary task to $\lambda=0.2$ to speed up adaptation and keep all other training hyperparameters fixed. For each test instance, we finetune the initial model on the auxiliary task for 30 iterations before testing. 

\vspace{4mm}\noindent\emph{Benchmark methods.} We compare \our with the following algorithms. FRCNN: baseline Faster-RCNN with ResNet50 backbone, trained on the source domain and deployed on the target without further adaptation. DivMatch \cite{diversify_match_Kim_2019_CVPR}:
cross-domain detection algorithm that, by exploiting target data, creates multiple randomized domains via CycleGAN and aligns their representations using an adversarial loss. SW \cite{Saito_2019_CVPR}: adaptive detection algorithm that aligns source and target features based on global context similarity. For both DivMatch and SW, we use a ResNet-50 backbone pretrained on ImageNet for fair comparison. Since all cross-domain algorithms need target data in advance and are not designed to work in our one-shot unsupervised setting, we provide them with the advantage of 10 target images accessible during training and randomly selected at each run. We collect average precision statistics during inference under the favorable assumption that the target domain will not shift after deployment.

\input{oshot/social}

\vspace{4mm}\noindent\textbf{Adapting to social feeds} 

\vspace{4mm}When data is collected from multiple sources, the  assumption that all target images originate from the same underlying distribution does not hold and standard cross-domain detection methods are penalized regardless of the number of seen target samples. We pretrain the source detector on Pascal VOC, and deploy it on Social Bikes. We consider only the bicycle and person annotations for this target, since all other instances of VOC classes are scarce.

\vspace{4mm}\noindent\emph{Results.} We report results in Table \ref{oshot:table:social}. \our outperforms all considered competitors, with a mAP score of $64.4$. Despite granting them access to the full target, adaptive algorithms incur in negative transfer due to data scarcity and large variety of target styles.

\vspace{4mm}\noindent\textbf{Large distribution shifts} 

\vspace{4mm}Artistic images are difficult benchmarks for cross-domain methods. Unpredictable perturbations in shape and color are challenging to detectors trained only on realistic images. 
We investigate this setting by training the source detector on Pascal VOC an deploying it on Clipart, Comic and Watercolor datasets.

\vspace{4mm}\noindent\emph{Results.} Table \ref{oshot:table:VOC2AMD} summarizes results on the three adaptation splits. We can see how \our with 30 finetuning iterations outperforms all competitors, with mAP gains ranging from $7.5$ points on Clipart to $9.2$ points on Watercolor. Cross-detection methods perform poorly in this setting, despite using 9 more samples in the adaptation phase compared to \our that only uses the test sample. These results confirm that they are not designed to tackle data scarcity conditions and exhibit negligible improvements compared to the baseline.

\input{oshot/voc2amd}
\input{oshot/city2foggy}

\vspace{4mm}\noindent\textbf{Adverse weather} 

\vspace{4mm}Some peculiar environmental conditions, such as fog, may be disregarded in source data acquisition,
yet adaptation to these circumstances is crucial for real world applications.
We assess the performance of \our on Cityscapes $\rightarrow$ FoggyCityscapes. We train our base detector on Cityscapes for 30k iterations without stepdown, as in \cite{cai2019exploring}. We select the best performing model on the Cityscapes validation split and deploy it to FoggyCityscapes.

\vspace{4mm}\noindent\emph{Results.} Experimental evaluation in Table \ref{oshot:table:C2F} shows that \our outperforms all compared approaches. Without finetuning iterations, performance using the auxiliary rotation task increases compared to the baseline. Subsequent finetuning iterations on the target sample improve these results, and 30 iterations yield models able to outperform the second-best method by $5$ mAP. Cross-domain algorithms used in this setting struggle to surpass the baseline (DivMatch) or suffer negative transfer (SW).

\vspace{4mm}\noindent\textbf{Cross-camera transfer} 

\vspace{4mm}Dataset bias between training and testing is unavoidable in practical applications, as for urban scene scenarios collected in different cities and with different cameras.
Subtle changes in illumination conditions and camera resolution might preclude a model trained on one realistic domain to optimally perform in another realistic but different domain. 
We test adaptation between KITTI and Cityscapes in both directions. For cross-domain evaluation we consider only the label car as standard practice.

\vspace{4mm}\noindent\emph{Results.} In Table \ref{oshot:table:KITTI}, \our improves by $7$ mAP points on KITTI $\rightarrow$ Cityscapes compared to the FRCNN baseline. DivMatch and SW both show a gain in this split, with SW obtaining the highest mAP of $39.2$ in the ten-shot setting. We argue that this is not surprising considering that, as shown in the visualization of Table \ref{oshot:table:KITTI}, the Cityscapes images share all a uniform visual style. As a consequence, 10 target images may be enough for standard cross-domain detection methods. Despite visual style homogeneity, the diversity among car instances in Cityscapes is high enough for learning a good car detection model. This is highlighted by the results in Cityscapes $\rightarrow$ KITTI task, for which adaptation performance for all methods is similar, and \our with $\gamma=0$ obtains the highest mAP of $75.4$. The FRCNN baseline on KITTI scores a high mAP of $75.1$: in this favorable condition detection doesn't benefit from adaptation.

\input{oshot/KITTI}

\subsubsection{Comparison with One-Shot Style Transfer}
\label{oshot:sec:oneshotstyle}

Although not specifically designed for cross-domain detection, in principle it is possible to apply one-shot style transfer methods as an alternative solution for our setting. We use BiOST \cite{Cohen_2019_ICCV}, the current state-of-the-art method for one-shot transfer, to modify the style of the target sample towards that of the source domain before performing inference. Due to the time-heavy requirements to perform BiOST on each test sample\footnote{To get the style update, BiOST trains of a double-variational autoencoder using the entire source besides the single target sample. As advised by the authors through personal communications, we trained the model for 5 epochs.}, we test it on Social Bikes and on a random subset of 100 Clipart images that we name Clipart100. We compare performance and time requirements of \our and BiOST on these two targets. Speed has been computed on an RTX2080Ti with full precision settings.

\input{oshot/biost}

\vspace{4mm}\noindent\emph{Results.} Table \ref{oshot:table:biost} shows summary mAP results using BiOST and \our. On Clipart100, the baseline FRCNN detector obtains $27.9$ mAP. We can see how BiOST is effective in the adaptation from one-sample, gaining $1.9$ points over the baseline, however it is outperformed by \our, which obtains $30.7$ mAP. On Social Bikes, while \our still outperforms the baseline, BiOST incurs in negative transfer, indicating that it was not able to effectively modify the source's style on the images we collected. Furthermore, BiOST is affected by two strong issues: (1) as already mentioned, it has an extremely high time complexity, with more than 6 hours needed to modify the style of a single source instance;  (2) it works under the strict assumption of accessing at the same time the entire source training set and the target sample. Due to these weaknesses, and the fact that \our still outperforms BiOST, we argue that existing one-shot translation methods are not suitable for one shot unsupervised cross-domain adaptation.

\subsubsection{Ablation Study}
\label{oshot:sec:ablation}

\vspace{4mm}\noindent\textbf{Detection error analysis} 

\begin{figure}[tb]
    \centering
    \includegraphics[width=\textwidth]{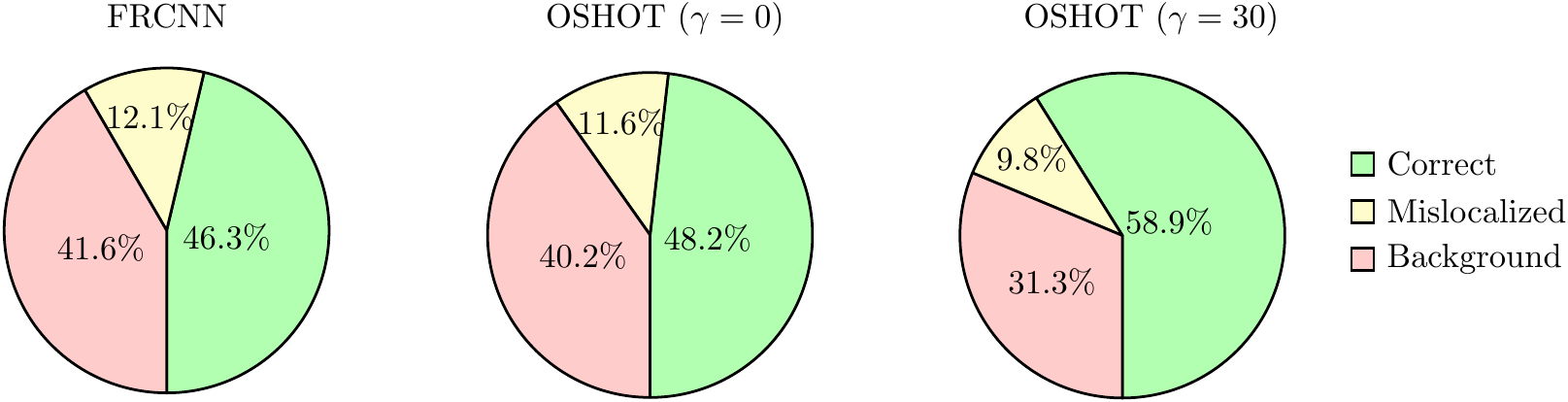}
    \caption{Detection error analysis on the most confident detections on Clipart}
    \label{oshot:fig:error}
\end{figure}

\vspace{4mm}Following \cite{hoiem2012diagnosing}, we provide detection error analysis for VOC $\rightarrow$ Clipart setting in Figure \ref{oshot:fig:error}. We select the 1000 most confident detections, and assign error classes based on IoU with ground truth (IoUgt). Errors are categorized as: correct (IoUgt $\geqslant$ 0.5), mislocalized (0.3 $\leqslant$ IoUgt $<$ 0.5) and background (IoUgt $<$ 0.3). Results show that, compared to the baseline FRCNN model, the regularization effect of adding a self-supervised task at training time ($\gamma = 0$) marginally increases the quality of detections. Instead subsequent finetuning iterations on the test sample substantially improve the number of correct detections, while also decreasing both false positives and mislocalization errors.

\vspace{4mm}\noindent\textbf{Cross-task pseudo-labeling ablation} 

\vspace{4mm}As explained in Section \ref{oshot:sec:method} we have two options in the \our adaptation phase: either considering the whole image, or focusing on pseudo-labeled bounding boxes
obtained from the detector after the first \our pretraining stage. For all the experiments presented above we focused on the second case. Indeed by
solving the auxiliary task only on objects, we limit the use of background features which may mislead the network towards solutions of the rotation task not based on relevant semantic information 
(e.g.: finding fixed patterns in images, exploiting watermarks). We validate our choice by comparing it against using the rotation task on the entire image in both training and adaptation phases. Table $\ref{oshot:table:SS-Patch Ablation}$ shows results for VOC $\rightarrow$ AMD and Cityscapes $\rightarrow$ Foggy Cityscapes using \our. We observe that the choice of rotated regions is critical for the effectiveness of the algorithm. Solving the rotation task on objects using pseudo-annotations results in mAP improvements that range from $2.9$ to $5.9$ points, indicating that we learn better features for the main task.

\input{oshot/regions}

\vspace{4mm}\noindent\textbf{Self-supervised iterations} 

\begin{figure}[tb]
    \centering
    \includegraphics[width=\textwidth]{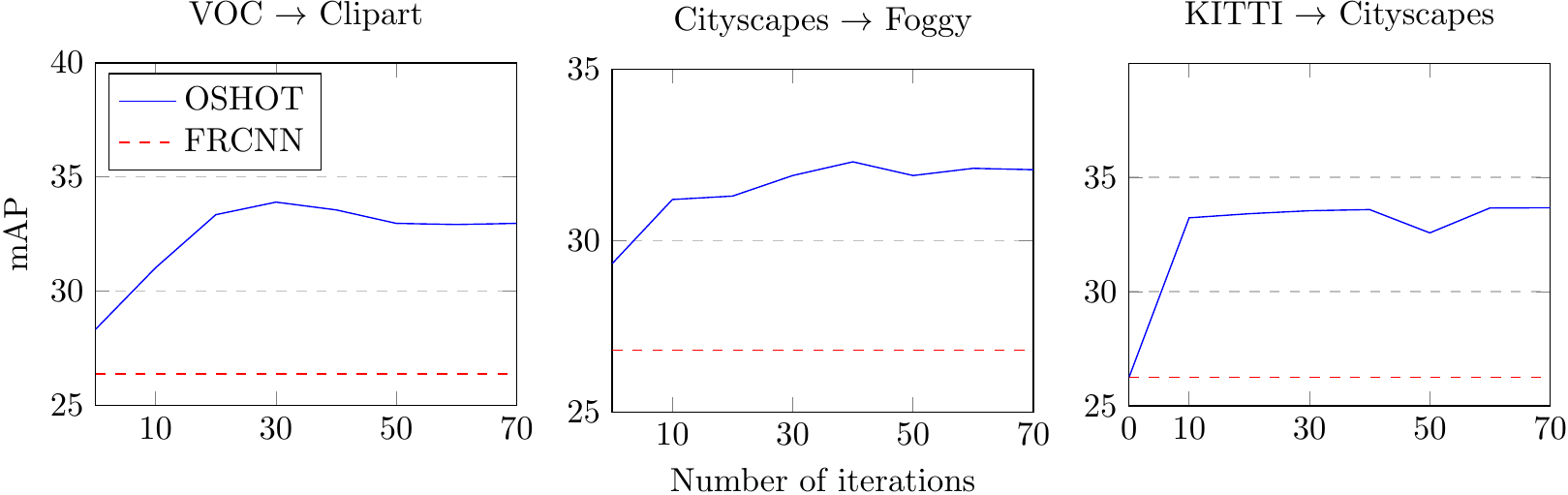}
    \caption{Performance of \our at different self-supervised iterations}
    \label{oshot:fig:Adaptive-Step}
\end{figure}

\vspace{4mm}We study the effects of adaptating with up to $\gamma = 70$ iterations on VOC $\rightarrow$ Clipart, Cityscapes $\rightarrow$ FoggyCityscapes and KITTI $\rightarrow$ Cityscapes. Results are shown in Figure \ref{oshot:fig:Adaptive-Step}. We observe a positive correlation between number of finetuning iterations and final mAP of the model in the earliest steps. This correlation is strong for the first 10 iterations and gets to a plateau after about 30 iterations: increasing $\gamma$ beyond this point doesn't affect the final results.

{}

\subsection{Conclusions}
This paper introduced the one-shot unsupervised cross-domain detection scenario, which is extremely relevant for monitoring image feeds on social media, where algorithms are called to adapt to a new visual domain from one single image. We showed that existing cross-domain detection methods suffer in this setting, as they are all explicitly designed to adapt from far larger quantities of target data. We presented OSHOT, the first deep architecture able to reduce the domain gap between source and target distribution by leveraging over one single target image. Our approach is based on a multi-task structure that exploits self-supervision and cross-task self-labeling. Extensive quantitative experiments and a qualitative analysis clearly demonstrate its effectiveness.

%% file: oshot/social.tex
\begin{table}[t]
\centering
\resizebox{0.9\textwidth}{!}{
\begin{tabular}{c}
\begin{tabular}{c@{~~}c@{~~}c@{~~}|c@{~~}}
\hline
\multicolumn{4}{c}{\textsl{\textbf{One-Shot} Target}}\\
\hline
Method & person & bicycle & mAP\\ \hline
FRCNN &  67.7 & 56.6 & 62.1 \\ \hline
\textbf{\textit{\our}} ($\gamma = 0$) & 72.1 & 52.8 & 62.4 \\
\textbf{\textit{\our}} ($\gamma = 30$) & 69.4 & 59.4 & \textbf{64.4}\\ \hline
\multicolumn{4}{c}{\textsl{\textbf{Full} Target}}\\
\hline
DivMatch  \cite{diversify_match_Kim_2019_CVPR} & 63.7 & 51.7 & 57.7 \\
SW \cite{Saito_2019_CVPR}  & 63.2 & 44.3 & 53.7 \\\hline 
\end{tabular}
\\
\qquad
\centering
\includegraphics[width=0.8\textwidth,valign=m]{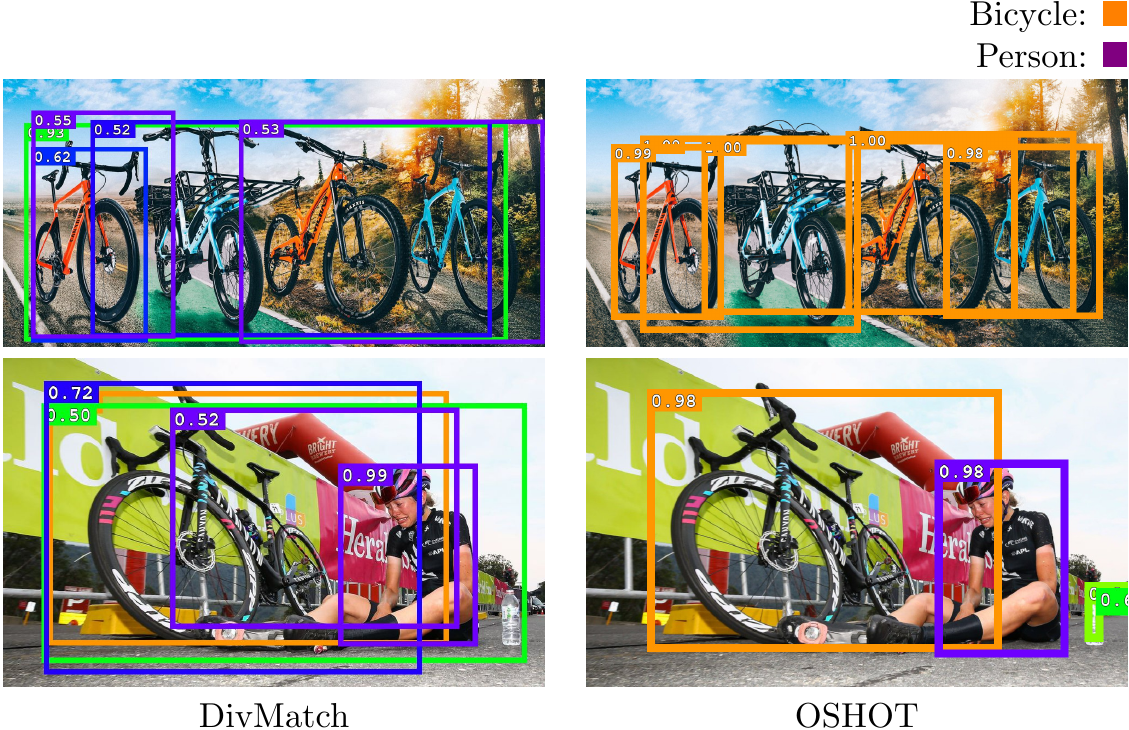}
\end{tabular}
}
\caption{(top) VOC $\rightarrow$ Social Bikes mAP results;  (bottom) visualization of DivMatch and OSHOT  detections. The number associated with each bounding box indicates the model's confidence in localization. Examples show how OSHOT detection is accurate, while most DivMatch boxes are false positives}
\label{oshot:table:social}
\end{table}

%% file: oshot/voc2amd.tex
\begin{table*}[tb]
\centering
\begin{adjustbox}{width=1\textwidth}
\subfloat[VOC $\rightarrow$ Clipart]{{}\begin{tabular}{c@{~~}c@{~~}c@{~~}c@{~~}c@{~~}c@{~~}c@{~~}c@{~~}c@{~~}c@{~~}c@{~~}c@{~~}c@{~~}c@{~~}c@{~~}c@{~~}c@{~~}c@{~~}c@{~~}c@{~~}c@{~~}|c@{~~}}
\hline
\multicolumn{22}{c}{\textsl{\textbf{One-Shot} Target}}\\
\hline
Method & aero & bike & bird & boat & bottle & bus & car & cat & chair & cow & table & dog & horse & mbike & person & plant & sheep & sofa & train & tv & mAP\\ \hline
FRCNN & 18.5 & 43.3 & 20.4 & 13.3 & 21.0 & 47.8 & 29.0 & 16.9 & 28.8 & 12.5 & 19.5 & 17.1 & 23.8 & 40.6 & 34.9 & 34.7 & 9.1 & 18.3 & 40.2 & 38.0 & 26.4 \\ \hline
\textbf{\textit{\our}} ($\gamma = 0$) & 23.1 & 55.3 & 22.7 & 21.4 & 26.8 & 53.3 & 28.9 & 4.6 & 31.4 & 9.2 & 27.8 & 9.6 & 30.9 & 47.0 & 38.2 & 35.2 & 11.1 & 20.4 & 36.0 & 33.6 & 28.3 \\
\textbf{\textit{\our}} ($\gamma = 10$) & 25.4 & 61.6 & 23.8 & 21.1 & 31.3 & 55.1 & 31.6 & 5.3 & 34.0 & 10.1 & 28.8 & 7.3 & 33.1 & 59.9 & 44.2 & 38.8 & 15.9 & 19.1 & 39.5 & 33.9 & 31.0\\
\textbf{\textit{\our}} ($\gamma = 30$) & 25.4 & 56.0 & 24.7 & 25.3 & 36.7 & 58.0 & 34.4 & 5.9 & 34.9 & 10.3 & 29.2 & 11.8 & 46.9 & 70.9 & 52.9 & 41.5 & 21.1 & 21.0 & 38.5 & 31.8 & \textbf{33.9}\\ \hline\hline
\multicolumn{22}{c}{\textsl{\textbf{Ten-Shot} Target}}\\
\hline
DivMatch  \cite{diversify_match_Kim_2019_CVPR} & 19.5 & 57.2 & 17.0 & 23.8 & 14.4 & 25.4 & 29.4 & 2.7 & 35.0 & 8.4 & 22.9 & 14.2 & 30.0 & 55.6 & 50.8 & 30.2 & 1.9 & 12.3 & 37.8 & 37.2 & 26.3 \\ 
SW \cite{Saito_2019_CVPR} & 21.5 & 39.9 & 21.7 & 20.5 & 32.7 & 34.1 & 25.1 & 8.5 & 33.2 & 10.9 & 15.2 & 3.4 & 32.2 & 56.9 & 46.5 & 35.4 & 14.7 & 15.2 & 29.2 & 32.0 & 26.4 \\ \hline
\end{tabular}}
\end{adjustbox}
\begin{adjustbox}{width=1\textwidth}
\subfloat[VOC $\rightarrow$ Comic]{\begin{tabular}{c@{~~}c@{~~}c@{~~}c@{~~}c@{~~}c@{~~}c@{~~}|c@{~~}}
\hline
\multicolumn{8}{c}{\textsl{\textbf{One-Shot} Target}}\\
\hline
Method & bike & bird & car & cat & dog & person &  mAP\\ \hline
FRCNN & 25.2 & 10.0 & 21.1 & 14.1 & 11.0 & 27.1 & 18.1 \\ \hline
\textbf{\textit{\our}} ($\gamma = 0$) & 26.9 & 11.6 & 22.7 & 9.1 & 14.2 & 28.3 & 18.8 \\
\textbf{\textit{\our}} ($\gamma = 10$) & 35.5 & 11.7 & 25.1 & 9.1 & 15.8 & 34.5 & 22.0 \\
\textbf{\textit{\our}} ($\gamma = 30$) & 35.2 & 14.4 & 30.0 & 14.8 & 20.0 & 46.7 & \textbf{26.9} \\  \hline\hline
\multicolumn{8}{c}{\textsl{\textbf{Ten-Shot} Target}}\\
\hline
DivMatch  \cite{diversify_match_Kim_2019_CVPR} & 27.1 & 12.3 & 26.2 & 11.5 & 13.8 & 34.0 & 20.8 \\ 
SW \cite{Saito_2019_CVPR} & 21.2 & 14.8 & 18.7 & 12.4 & 14.9 & 43.9 & 21.0 \\ \hline
\end{tabular}} \hspace{0.5cm}
\subfloat[VOC $\rightarrow$ Watercolor]{\begin{tabular}{c@{~~}c@{~~}c@{~~}c@{~~}c@{~~}c@{~~}c@{~~}|c@{~~}}
\hline
\multicolumn{8}{c}{\textsl{\textbf{One-Shot} Target}}\\
\hline
Method & bike & bird & car & cat & dog & person &  mAP\\ \hline
FRCNN & 62.5 & 39.7 & 43.4 & 31.9 & 26.7 & 52.4 & 42.8 \\ \hline
\textbf{\textit{\our}} ($\gamma = 0$) & 70.2 & 46.7 & 45.5 & 31.2 & 27.2 & 55.7 & 46.1\\
\textbf{\textit{\our}} ($\gamma = 10$) & 70.2 & 46.7 & 48.1 & 30.9 & 32.3 & 59.9 & 48.0\\
\textbf{\textit{\our}} ($\gamma = 30$) & 77.1 & 44.7 & 52.4 & 37.3 & 37.0 & 63.3 & \textbf{52.0}\\ \hline\hline
\multicolumn{8}{c}{\textsl{\textbf{Ten-Shot} Target}}\\
\hline
DivMatch  \cite{diversify_match_Kim_2019_CVPR} & 64.6 & 44.1 & 44.6 & 34.1 & 24.9 & 60.0 & 45.4 \\
SW \cite{Saito_2019_CVPR}  & 66.3 & 41.1 & 41.1 & 30.5 & 20.5 & 52.3 & 42.0 \\
\hline
\end{tabular}}
\end{adjustbox}
\caption{mAP results for VOC $\rightarrow$ AMD}
\label{oshot:table:VOC2AMD}
\end{table*}

%% file: oshot/city2foggy.tex
\begin{table*}[tb]
\centering
\begin{adjustbox}{width=\textwidth}
\centering
\begin{tabular}{c@{~~}c@{~~}c@{~~}c@{~~}c@{~~}c@{~~}c@{~~}c@{~~}c@{~~}|c@{~~}}
\hline
\multicolumn{10}{c}{\textsl{\textbf{One-Shot} Target}}\\
\hline
Method & person & rider & car & truck & bus & train & mcycle & bicycle & mAP\\ \hline
FRCNN & 30.4 & 36.3 & 41.4 & 18.5  & 32.8 & 9.1 & 20.3 & 25.9 & 26.8 \\ \hline
\textbf{\textit{\our}} ($\gamma = 0$) & 31.8 & 42.0 & 42.6 & 20.1 & 31.6 & 10.6 & 24.8 & 30.7 & 29.3 \\
\textbf{\textit{\our}} ($\gamma = 10$) & 31.9 & 41.9 & 43.0 & 19.7 & 38.0 & 10.4 & 25.5 & 30.2 & 30.1\\
\textbf{\textit{\our}} ($\gamma = 30$) & 32.1 & 46.1 & 43.1 & 20.4 & 39.8 & 15.9 & 27.1 & 32.4 & \textbf{31.9}\\ \hline\hline
\multicolumn{10}{c}{\textsl{\textbf{Ten-Shot} Target}}\\
\hline
DivMatch  \cite{diversify_match_Kim_2019_CVPR} & 27.6 & 38.1 & 42.9 & 17.1 & 27.6 & 14.3 & 14.6 & 32.8 & 26.9 \\
SW \cite{Saito_2019_CVPR}  & 25.5 & 30.8 & 40.4 & 21.1 & 26.1 & 34.5 & 6.1 & 13.4 & 24.7 \\\hline \hline
\multicolumn{10}{c}{\textsl{\textbf{Full} Target}}\\
\hline
DivMatch  \cite{diversify_match_Kim_2019_CVPR} & 32.3 & 43.5 & 47.6 & 23.9 & 38.0 & 23.1 & 27.6 & 37.2 & 34.2 \\
SW \cite{Saito_2019_CVPR}  & 31.3 & 32.1 & 47.4 & 19.6 & 28.8 & 41.0 & 9.8 & 20.1 & 28.8 \\ \hline
\end{tabular}
\end{adjustbox}
\caption{mAP results for Cityscapes $\rightarrow$ FoggyCityscapes}
\label{oshot:table:C2F}
\end{table*}

%% file: oshot/KITTI.tex
\begin{table}[tb]
\centering
\resizebox{0.9\textwidth}{!}{
\begin{tabular}{c}
\begin{tabular}{c@{~~}c@{~~}c@{~~}c@{~~}c@{~~}c@{~~}c@{~~}|c@{~~}}
\hline
\multicolumn{3}{c}{\textsl{\textbf{One-Shot} Target}}\\
\hline
Method & KITTI $\rightarrow$ Cityscapes & Cityscapes $\rightarrow$ KITTI\\ \hline
FRCNN & 26.5 & 75.1 \\ \hline
\textbf{\textit{\our}} $\gamma = 0$ & 26.2 & \textbf{75.4} \\
\textbf{\textit{\our}} $\gamma = 10$ & 33.2 & 75.3 \\
\textbf{\textit{\our}} $\gamma = 30$ & \textbf{33.5} & 75.0 \\ \hline\hline
\multicolumn{3}{c}{\textsl{\textbf{Ten-Shot} Target}}\\
\hline
DivMatch  \cite{diversify_match_Kim_2019_CVPR} & 37.9 & 74.1 \\ 
SW \cite{Saito_2019_CVPR}  & 39.2 & 74.6 \\ \hline 
\end{tabular}
\\ \\
\includegraphics[width=0.8\textwidth,valign=m]{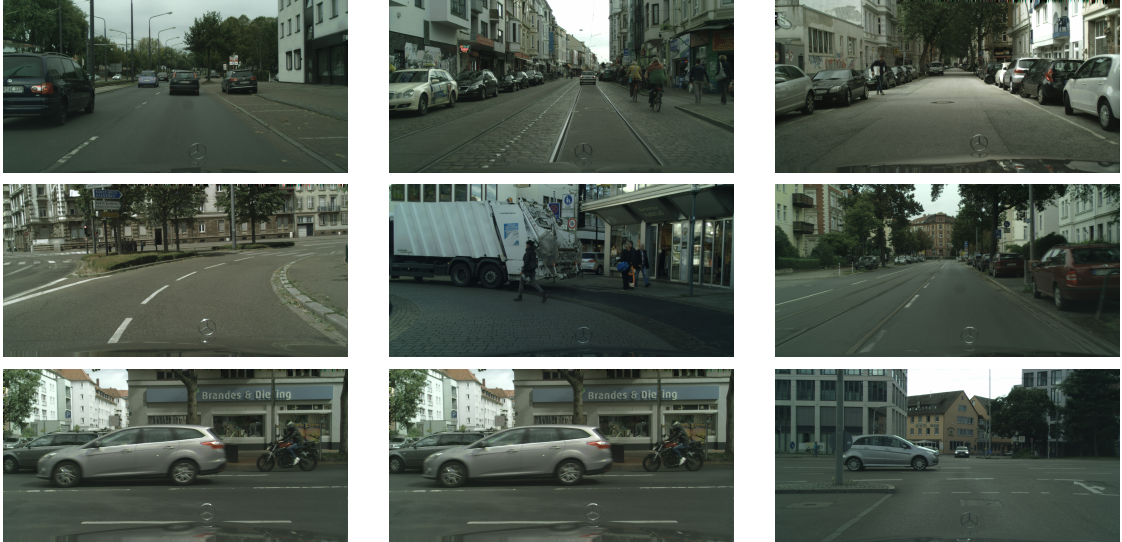}
\end{tabular}
}
\caption{mAP of car class in KITTI/Cityscapes detection experiments (top). Random samples from the Cityscapes dataset, showing low intra-domain variance (bottom)}
\label{oshot:table:KITTI}
\end{table}

%% file: oshot/biost.tex
\begin{table*}[tb]
\centering
\begin{adjustbox}{width=\textwidth}
\begin{tabular}{c@{~~}c@{~~}c@{~~}c@{~~}}
\hline
 & FRCNN & BiOST \cite{Cohen_2019_ICCV} & OSHOT ($\gamma=30$)\\ \hline
mAP on Clipart100 & 27.9 & 29.8 & \textbf{30.7} \\ \hline
mAP on Social Bikes & 62.1 & 51.1 & \textbf{64.4} \\ \hline
Adaptation time (seconds per sample) & - & $\sim 2.4*10^{4}$ & 7.8 \\
\hline
\end{tabular}
\end{adjustbox}
\caption{Comparison between baseline, one-shot syle transfer and OSHOT in the one-shot unsupervised cross-domain detection setting}
\label{oshot:table:biost}
\end{table*}

%% file: oshot/regions.tex
\begin{table}[tb]
\centering
\resizebox{0.9\textwidth}{!}{
\begin{tabular}{c@{~~}c@{~~}c@{~~}}
\hline
& $G_r(image)$ & $G_r(pseudoboxcrop)$ \\ \hline
VOC $\rightarrow$ Clipart  & 31.0 & \textbf{33.9} \\
VOC $\rightarrow$ Comic  & 21.0 & \textbf{26.9} \\
VOC $\rightarrow$ Watercolor  & 48.2 & \textbf{52.0} \\
Cityscapes $\rightarrow$ Foggy Cityscapes  & 27.7 & \textbf{31.9}\\
\hline
\end{tabular}
}
\caption{Rotating image vs rotating objects via pseudo-labeling on \our}
\label{oshot:table:SS-Patch Ablation}
\end{table}

%% file: conclusions.tex
\section{Conclusions}

Dataset bias is an ever-present problem in Computer Vision, as data acquisition and subsequent annotation is often performed in controlled settings, making the training distribution unlikely to correspond to the testing one. Hence, research in Domain Generalization and Domain Adaptation has flourished, with methods at \emph{feature}, \emph{model} and \emph{data} levels.

We contributed to these research areas by presenting novel approaches for Domain Generalization and Domain Adaptation that are based on data aggregations and transformations.

In DSAM \cite{Antonio_GCPR18}, we proposed a model-based algorithm for DG that achieves generalization by explicitly separating agnostic and specific components within the architecture. The approach has shown promising results, although it requires source-awareness during training in the form of domain labels for each image.

With self-supervised solutions, we overcome the need of source domain labels by aligning features with a proxy task. The auxiliary loss enhances model regularization for out-of-distribution generalization \cite{jigsawCVPR19}, while the pretext labels allow unsupervised learning on the unlabeled target set for Domain Adaptation \cite{jigsawCVPR19} and Partial Domain Adaptation \cite{tackling_iciap19}. Furthermore, by having the source pre-text task shared with the test sample, we've designed an object detection model that can adapt on individual images on the fly by performing the auxiliary task on them \cite{oshot}. These approaches, despite being relatively simple, have been shown to be competitive with more elaborate methods for generalization and adaptation \cite{bucci2021self}. They might, however, introduce many additional hyperparameters to balance auxiliary losses \cite{tackling_iciap19}, a weakness that is shared with most Domain Adaptation literature. Another limitation lies in the choice of the best self-supervised task, specifically, the multi-task inductive bias is effective as much as the auxiliary task is semantically related with the primary task. This was most evident in OSHOT, when we used the entire image for self-supervised pre-training and adaptation. There, we could see how the model started to exploit visual cues unrelated to the detection task, such as the position of the ground. At the moment, the best thing we can do is manual searching of the better correlated task with expensive trainings.

We've also shown how non-nai\"ve data-augmentation via style-transfer is a simple and inexpensive strategy to improve generalization on new distributions \cite{borlino2021rethinking}. Established Domain Generalization methods that  leverage the multi-source nature of data fail to take advantage of the augmented sources and their original benefit disappears when compared with the baseline, highlighting the need for new algorithms that can properly incorporate data-augmentation.

\section{Future Work}

Future work could expand more on the use of transformations for Domain Adaptation. Specifically, the success of self-supervision on the unlabeled target depends on how semantically related the auxiliary task is with respect to the main task. This makes the choice of auxiliary task crucial, however, a brute-force search to find the best fitting task is expensive. One solution could do by leveraging a linear combination of tasks (task-ensemble) to approximate the semantics of the main problem. This method could be effective at improving performances on the main objective, but would still raise the computational cost for model training. A more refined approach could instead use meta-learning to align different losses for main and auxiliary objectives to improve Domain Adaptation and One-Shot Adaptation regardless of the chosen task. 

Self-supervision could also be used to enhance model validation in Domain Adaptation. Hyperparameter optimization is an open problem in DA, as it is not possible to validate on the unlabeled target set. Despite this limitation, most methods in this literature tend to introduce multiple additional tasks/losses that need to be tuned. A self-supervised task that is semantically close to the main task could potentially overcome this limitation by acting as a proxy objective for validating any algorithm's hyperparameters.

Future work will also focus on expanding the OSHOT method. One possible direction is to explore different tasks other than object detection, as the training and adaptation phases use a general multi-task paradigm that can be easily extended. In particular, tasks for which semantics of  target images are abundant, such as semantic segmentation or aerial image understanding, are good candidates. Furthermore, meta-learning is a natural extension of OSHOT, as it allows to optimize a model to perform well on a given task when a different objective has been applied beforehand. In our case, we could train the detection task to better detect boxes on the target image when the model has been fine-tuned on the auxiliary task.

Finally, given the effectiveness and convenience of style-transfer based data-augmentation in Domain Generalization, future research should focus on integrating these transformations in generalization pipelines. Most DG approaches that expect well separated input distribution could be relaxed to allow for samples with mixed source styles.

%% file: thesis.bbl
\begin{thebibliography}{145}
\providecommand{\natexlab}[1]{#1}
\providecommand{\url}[1]{\texttt{#1}}
\expandafter\ifx\csname urlstyle\endcsname\relax
  \providecommand{\doi}[1]{doi: #1}\else
  \providecommand{\doi}{doi: \begingroup \urlstyle{rm}\Url}\fi

\bibitem[Angeletti et~al.(2018)Angeletti, Caputo, and Tommasi]{LOAD_ICRA}
G.~Angeletti, B.~Caputo, and T.~Tommasi.
\newblock Adaptive deep learning through visual domain localization.
\newblock In \emph{International Conference on Robotic Automation (ICRA)},
  2018.

\bibitem[Asano et~al.(2020)Asano, Rupprecht, and Vedaldi]{asano20a-critical}
Y.~M. Asano, C.~Rupprecht, and A.~Vedaldi.
\newblock A critical analysis of self-supervision, or what we can learn from a
  single image.
\newblock In \emph{ICLR}, 2020.

\bibitem[Balaji et~al.(2018)Balaji, Sankaranarayanan, and
  Chellappa]{NIPS2018_metareg}
Y.~Balaji, S.~Sankaranarayanan, and R.~Chellappa.
\newblock Metareg: Towards domain generalization using meta-regularization.
\newblock In \emph{NeurIPS}, 2018.

\bibitem[Baxter(1997)]{Baxter1997}
J.~Baxter.
\newblock A bayesian/information theoretic model of learning to learn via
  multiple task sampling.
\newblock \emph{Machine Learning}, 28\penalty0 (1):\penalty0 7--39, Jul 1997.
\newblock ISSN 1573-0565.
\newblock \doi{10.1023/A:1007327622663}.
\newblock URL \url{https://doi.org/10.1023/A:1007327622663}.

\bibitem[Ben-David et~al.(2010)Ben-David, Blitzer, Crammer, Kulesza, Pereira,
  and Vaughan]{hdivergence}
S.~Ben-David, J.~Blitzer, K.~Crammer, A.~Kulesza, F.~Pereira, and J.~Vaughan.
\newblock A theory of learning from different domains.
\newblock \emph{Machine Learning}, 79:\penalty0 151--175, 2010.

\bibitem[Benaim and Wolf(2018)]{oneshotNIPS2018}
S.~Benaim and L.~Wolf.
\newblock One-shot unsupervised cross domain translation.
\newblock In \emph{NeurIPS}. 2018.

\bibitem[Bisanz et~al.(1983)Bisanz, Bisanz, and Kail]{children_learning}
J.~Bisanz, G.~L. Bisanz, and R.~Kail, editors.
\newblock \emph{Learning in Children: Progress in Cognitive Development
  Research}.
\newblock Springer-Verlag, 1983.

\bibitem[Blanchard et~al.(2011)Blanchard, Lee, and Scott]{NIPS2011_blanchard}
G.~Blanchard, G.~Lee, and C.~Scott.
\newblock Generalizing from several related classification tasks to a new
  unlabeled sample.
\newblock In \emph{NIPS}. 2011.

\bibitem[Borlino et~al.(2021)Borlino, D'Innocente, and
  Tommasi]{borlino2021rethinking}
F.~C. Borlino, A.~D'Innocente, and T.~Tommasi.
\newblock Rethinking domain generalization baselines.
\newblock \emph{arXiv preprint arXiv:2101.09060}, 2021.

\bibitem[Bousmalis et~al.(2016)Bousmalis, Trigeorgis, Silberman, Krishnan, and
  Erhan]{Bousmalis:DSN:NIPS16}
K.~Bousmalis, G.~Trigeorgis, N.~Silberman, D.~Krishnan, and D.~Erhan.
\newblock {Domain Separation Networks}.
\newblock In \emph{NeurIPS}, 2016.

\bibitem[Bucci et~al.(2019)Bucci, D'Innocente, and Tommasi]{tackling_iciap19}
S.~Bucci, A.~D'Innocente, and T.~Tommasi.
\newblock Tackling partial domain adaptation with self-supervision.
\newblock In \emph{ICIAP}, 2019.

\bibitem[Bucci et~al.(2021)Bucci, D'Innocente, Liao, Carlucci, Caputo, and
  Tommasi]{bucci2021self}
S.~Bucci, A.~D'Innocente, Y.~Liao, F.~M. Carlucci, B.~Caputo, and T.~Tommasi.
\newblock Self-supervised learning across domains.
\newblock \emph{IEEE Transactions on Pattern Analysis and Machine
  Intelligence}, 2021.

\bibitem[Cai et~al.(2019)Cai, Pan, Ngo, Tian, Duan, and Yao]{cai2019exploring}
Q.~Cai, Y.~Pan, C.-W. Ngo, X.~Tian, L.~Duan, and T.~Yao.
\newblock Exploring object relation in mean teacher for cross-domain detection.
\newblock In \emph{CVPR}, 2019.

\bibitem[Cao et~al.(2018{\natexlab{a}})Cao, Long, Wang, and Jordan]{SAN}
Z.~Cao, M.~Long, J.~Wang, and M.~I. Jordan.
\newblock Partial transfer learning with selective adversarial networks.
\newblock In \emph{CVPR}, 2018{\natexlab{a}}.

\bibitem[Cao et~al.(2018{\natexlab{b}})Cao, Ma, Long, and Wang]{PADA_eccv18}
Z.~Cao, L.~Ma, M.~Long, and J.~Wang.
\newblock Partial adversarial domain adaptation.
\newblock In \emph{ECCV}, 2018{\natexlab{b}}.

\bibitem[Carlucci et~al.(2017{\natexlab{a}})Carlucci, Porzi, Caputo, Ricci, and
  Bul{\`{o}}]{CarlucciICIAP17}
F.~M. Carlucci, L.~Porzi, B.~Caputo, E.~Ricci, and S.~R. Bul{\`{o}}.
\newblock Just {DIAL:} domain alignment layers for unsupervised domain
  adaptation.
\newblock In \emph{ICIAP}, 2017{\natexlab{a}}.

\bibitem[Carlucci et~al.(2017{\natexlab{b}})Carlucci, Porzi, Caputo, Ricci, and
  Rota~Bul{\`o}]{carlucci2017auto}
F.~M. Carlucci, L.~Porzi, B.~Caputo, E.~Ricci, and S.~Rota~Bul{\`o}.
\newblock Autodial: Automatic domain alignment layers.
\newblock In \emph{ICCV}, 2017{\natexlab{b}}.

\bibitem[Carlucci et~al.(2019{\natexlab{a}})Carlucci, D'Innocente, Bucci,
  Caputo, and Tommasi]{jigsawCVPR19}
F.~M. Carlucci, A.~D'Innocente, S.~Bucci, B.~Caputo, and T.~Tommasi.
\newblock Domain generalization by solving jigsaw puzzles.
\newblock In \emph{Proceedings of the IEEE Conference on Computer Vision and
  Pattern Recognition}, pages 2229--2238, 2019{\natexlab{a}}.

\bibitem[Carlucci et~al.(2019{\natexlab{b}})Carlucci, Russo, Tommasi, and
  Caputo]{ADAGE}
F.~M. Carlucci, P.~Russo, T.~Tommasi, and B.~Caputo.
\newblock Hallucinating agnostic images to generalize across domains.
\newblock In \emph{ICCV-Workshop}, 2019{\natexlab{b}}.

\bibitem[Caron et~al.(2018)Caron, Bojanowski, Joulin, and Douze]{caron2018deep}
M.~Caron, P.~Bojanowski, A.~Joulin, and M.~Douze.
\newblock Deep clustering for unsupervised learning of visual features.
\newblock In \emph{ECCV}, 2018.

\bibitem[Caruana(1993)]{Caruana93multitasklearning:}
R.~Caruana.
\newblock Multitask learning: A knowledge-based source of inductive bias.
\newblock In \emph{Proceedings of the Tenth International Conference on Machine
  Learning}, pages 41--48. Morgan Kaufmann, 1993.

\bibitem[Caruana(1997)]{Caruana:1997}
R.~Caruana.
\newblock Multitask learning.
\newblock \emph{Machine Learning}, 28\penalty0 (1):\penalty0 41--75, 1997.

\bibitem[Chen et~al.(2018)Chen, Li, Sakaridis, Dai, and
  Van~Gool]{Chen_2018_CVPR}
Y.~Chen, W.~Li, C.~Sakaridis, D.~Dai, and L.~Van~Gool.
\newblock Domain adaptive faster r-cnn for object detection in the wild.
\newblock In \emph{CVPR}, 2018.

\bibitem[Cohen and Wolf(2019)]{Cohen_2019_ICCV}
T.~Cohen and L.~Wolf.
\newblock Bidirectional one-shot unsupervised domain mapping.
\newblock In \emph{ICCV}, 2019.

\bibitem[Cordts et~al.(2016)Cordts, Omran, Ramos, Rehfeld, Enzweiler, Benenson,
  Franke, Roth, and Schiele]{cordts2016cityscapes}
M.~Cordts, M.~Omran, S.~Ramos, T.~Rehfeld, M.~Enzweiler, R.~Benenson,
  U.~Franke, S.~Roth, and B.~Schiele.
\newblock The cityscapes dataset for semantic urban scene understanding.
\newblock In \emph{Proc. of the IEEE Conference on Computer Vision and Pattern
  Recognition (CVPR)}, 2016.

\bibitem[Cruz et~al.(2017)Cruz, Fernando, Cherian, and Gould]{Cruz2017}
R.~S. Cruz, B.~Fernando, A.~Cherian, and S.~Gould.
\newblock Visual permutation learning.
\newblock In \emph{CVPR}, 2017.

\bibitem[Csurka(2017)]{csurka_book}
G.~Csurka, editor.
\newblock \emph{Domain Adaptation in Computer Vision Applications}.
\newblock Advances in Computer Vision and Pattern Recognition. Springer, 2017.

\bibitem[Dai et~al.(2016)Dai, Li, He, and Sun]{dai2016r}
J.~Dai, Y.~Li, K.~He, and J.~Sun.
\newblock R-fcn: Object detection via region-based fully convolutional
  networks.
\newblock In \emph{NIPS}, 2016.

\bibitem[Dalal and Triggs(2005)]{dalal2005histograms}
N.~Dalal and B.~Triggs.
\newblock Histograms of oriented gradients for human detection.
\newblock In \emph{CVPR}, 2005.

\bibitem[Deng et~al.(2009)Deng, Dong, Socher, jia Li, Li, and
  Fei-Fei]{imagenet}
J.~Deng, W.~Dong, R.~Socher, L.~jia Li, K.~Li, and L.~Fei-Fei.
\newblock Imagenet: A large-scale hierarchical image database.
\newblock In \emph{CVPR}, 2009.

\bibitem[Ding and Fu(2017)]{Ding2017DeepDG}
Z.~Ding and Y.~Fu.
\newblock Deep domain generalization with structured low-rank constraint.
\newblock \emph{IEEE TIP}, 27:\penalty0 304--313, 2017.

\bibitem[D'Innocente and Caputo(2018)]{Antonio_GCPR18}
A.~D'Innocente and B.~Caputo.
\newblock Domain generalization with domain-specific aggregation modules.
\newblock In \emph{GCPR}, 2018.

\bibitem[D'Innocente et~al.(2020)D'Innocente, Borlino, Bucci, Caputo, and
  Tommasi]{oshot}
A.~D'Innocente, F.~C. Borlino, S.~Bucci, B.~Caputo, and T.~Tommasi.
\newblock One-shot unsupervised cross-domain detection.
\newblock \emph{arXiv preprint arXiv:2005.11610}, 2020.

\bibitem[Doersch and Zisserman(2017{\natexlab{a}})]{doersch2017multi}
C.~Doersch and A.~Zisserman.
\newblock Multi-task self-supervised visual learning.
\newblock In \emph{The IEEE International Conference on Computer Vision
  (ICCV)}, 2017{\natexlab{a}}.

\bibitem[Doersch and Zisserman(2017{\natexlab{b}})]{multitaskSSL}
C.~Doersch and A.~Zisserman.
\newblock Multi-task self-supervised visual learning.
\newblock In \emph{ICCV}, 2017{\natexlab{b}}.

\bibitem[Dosovitskiy et~al.(2014)Dosovitskiy, Springenberg, Riedmiller, and
  Brox]{NIPS2014_geometric}
A.~Dosovitskiy, J.~T. Springenberg, M.~Riedmiller, and T.~Brox.
\newblock Discriminative unsupervised feature learning with convolutional
  neural networks.
\newblock In \emph{NeurIPS}, 2014.

\bibitem[Duong et~al.(2015)Duong, Cohn, Bird, and Cook]{duong2015low}
L.~Duong, T.~Cohn, S.~Bird, and P.~Cook.
\newblock Low resource dependency parsing: Cross-lingual parameter sharing in a
  neural network parser.
\newblock In \emph{Proceedings of the 53rd Annual Meeting of the Association
  for Computational Linguistics and the 7th International Joint Conference on
  Natural Language Processing (Volume 2: Short Papers)}, volume~2, pages
  845--850, 2015.

\bibitem[Everingham et~al.(2010)Everingham, Van~Gool, Williams, Winn, and
  Zisserman]{everingham2010pascal}
M.~Everingham, L.~Van~Gool, C.~K. Williams, J.~Winn, and A.~Zisserman.
\newblock The pascal visual object classes (voc) challenge.
\newblock \emph{IJCV}, 88\penalty0 (2):\penalty0 303--338, 2010.

\bibitem[Ferguson et~al.(2018)Ferguson, Franconeri, and Waxman]{PLOS}
B.~Ferguson, S.~L. Franconeri, and S.~R. Waxman.
\newblock Very young infants learn abstract rules in the visual modality.
\newblock \emph{PLOS ONE}, 13\penalty0 (1):\penalty0 1--14, 01 2018.
\newblock \doi{10.1371/journal.pone.0190185}.
\newblock URL \url{https://doi.org/10.1371/journal.pone.0190185}.

\bibitem[Finn et~al.(2017)Finn, Abbeel, and Levine]{MAML}
C.~Finn, P.~Abbeel, and S.~Levine.
\newblock Model-agnostic meta-learning for fast adaptation of deep networks.
\newblock \emph{arXiv preprint arXiv:1703.03400}, 2017.

\bibitem[Ganin and Lempitsky(2015)]{ganin2014unsupervised}
Y.~Ganin and V.~Lempitsky.
\newblock Unsupervised domain adaptation by backpropagation.
\newblock In \emph{ICML}, 2015.

\bibitem[Ganin et~al.(2016)Ganin, Ustinova, Ajakan, Germain, Larochelle,
  Laviolette, Marchand, and Lempitsky]{Ganin:DANN:JMLR16}
Y.~Ganin, E.~Ustinova, H.~Ajakan, P.~Germain, H.~Larochelle, F.~Laviolette,
  M.~Marchand, and V.~Lempitsky.
\newblock Domain-adversarial training of neural networks.
\newblock \emph{J. Mach. Learn. Res.}, 17\penalty0 (1):\penalty0 2096--2030,
  2016.

\bibitem[Geiger et~al.(2013)Geiger, Lenz, Stiller, and Urtasun]{kitti}
A.~Geiger, P.~Lenz, C.~Stiller, and R.~Urtasun.
\newblock Vision meets robotics: The kitti dataset.
\newblock \emph{The International Journal of Robotics Research}, 32\penalty0
  (11):\penalty0 1231--1237, 2013.

\bibitem[Ghifary et~al.(2015)Ghifary, Kleijn, Zhang, and
  Balduzzi]{DGautoencoders}
M.~Ghifary, W.~B. Kleijn, M.~Zhang, and D.~Balduzzi.
\newblock Domain generalization for object recognition with multi-task
  autoencoders.
\newblock In \emph{ICCV}, 2015.

\bibitem[Gidaris et~al.(2018)Gidaris, Singh, and
  Komodakis]{gidaris2018unsupervised}
S.~Gidaris, P.~Singh, and N.~Komodakis.
\newblock Unsupervised representation learning by predicting image rotations.
\newblock In \emph{ICLR}, 2018.

\bibitem[Gidaris et~al.(2020)Gidaris, Bursuc, Komodakis, Pérez, and
  Cord]{gidaris2020learning}
S.~Gidaris, A.~Bursuc, N.~Komodakis, P.~Pérez, and M.~Cord.
\newblock Learning representations by predicting bags of visual words.
\newblock In \emph{CVPR}, 2020.

\bibitem[Girshick(2015)]{girshick2015fast}
R.~Girshick.
\newblock Fast r-cnn.
\newblock In \emph{ICCV}, 2015.

\bibitem[Girshick et~al.(2014)Girshick, Donahue, Darrell, and
  Malik]{girshick2014rich}
R.~Girshick, J.~Donahue, T.~Darrell, and J.~Malik.
\newblock Rich feature hierarchies for accurate object detection and semantic
  segmentation.
\newblock In \emph{CVPR}, 2014.

\bibitem[Goodfellow et~al.(2014)Goodfellow, Pouget-Abadie, Mirza, Xu,
  Warde-Farley, Ozair, Courville, and Bengio]{Goodfellow:GAN:NIPS2014}
I.~Goodfellow, J.~Pouget-Abadie, M.~Mirza, B.~Xu, D.~Warde-Farley, S.~Ozair,
  A.~Courville, and Y.~Bengio.
\newblock Generative adversarial nets.
\newblock In \emph{{NeurIPS}}, 2014.

\bibitem[Gulrajani and Lopez-Paz(2020)]{gulrajani2020search}
I.~Gulrajani and D.~Lopez-Paz.
\newblock In search of lost domain generalization.
\newblock \emph{arXiv preprint arXiv:2007.01434}, 2020.

\bibitem[Guo et~al.(2018)Guo, Haque, Huang, Yeung, and Fei-Fei]{guo2018dynamic}
M.~Guo, A.~Haque, D.-A. Huang, S.~Yeung, and L.~Fei-Fei.
\newblock Dynamic task prioritization for multitask learning.
\newblock In \emph{European Conference on Computer Vision}, pages 282--299.
  Springer, 2018.

\bibitem[He et~al.(2016)He, Zhang, Ren, and Sun]{he2016deep}
K.~He, X.~Zhang, S.~Ren, and J.~Sun.
\newblock Deep residual learning for image recognition.
\newblock In \emph{Conference on Computer Vision and Pattern Recognition
  (CVPR)}, 2016.

\bibitem[Hoffman et~al.(2012)Hoffman, Kulis, Darrell, and
  Saenko]{hoffman_eccv12}
J.~Hoffman, B.~Kulis, T.~Darrell, and K.~Saenko.
\newblock Discovering latent domains for multisource domain adaptation.
\newblock In \emph{ECCV}, 2012.

\bibitem[Hoffman et~al.(2014)Hoffman, Darrell, and Saenko]{Hoffman_CVPR2014}
J.~Hoffman, T.~Darrell, and K.~Saenko.
\newblock Continuous manifold based adaptation for evolving visual domains.
\newblock In \emph{CVPR}, 2014.

\bibitem[Hoffman et~al.(2018)Hoffman, Tzeng, Park, Zhu, Isola, Saenko, Efros,
  and Darrell]{cycada}
J.~Hoffman, E.~Tzeng, T.~Park, J.-Y. Zhu, P.~Isola, K.~Saenko, A.~Efros, and
  T.~Darrell.
\newblock {C}y{CADA}: Cycle-consistent adversarial domain adaptation.
\newblock In \emph{ICML}, 2018.

\bibitem[Hoiem et~al.(2012)Hoiem, Chodpathumwan, and Dai]{hoiem2012diagnosing}
D.~Hoiem, Y.~Chodpathumwan, and Q.~Dai.
\newblock Diagnosing error in object detectors.
\newblock In \emph{ECCV}, 2012.

\bibitem[Huang and Belongie(2017)]{Huang_2017_ICCV_adain}
X.~Huang and S.~Belongie.
\newblock Arbitrary style transfer in real-time with adaptive instance
  normalization.
\newblock In \emph{ICCV}, 2017.

\bibitem[Huang et~al.(2020)Huang, Wang, Xing, and
  Huang]{huang2020selfchallenging}
Z.~Huang, H.~Wang, E.~P. Xing, and D.~Huang.
\newblock Self-challenging improves cross-domain generalization.
\newblock In \emph{ECCV}, 2020.

\bibitem[Inoue et~al.(2018)Inoue, Furuta, Yamasaki, and Aizawa]{inoue2018cross}
N.~Inoue, R.~Furuta, T.~Yamasaki, and K.~Aizawa.
\newblock Cross-domain weakly-supervised object detection through progressive
  domain adaptation.
\newblock In \emph{CVPR}, 2018.

\bibitem[Jang et~al.(2018)Jang, Devin, Vanhoucke, and Levine]{grasp2vec}
E.~Jang, C.~Devin, V.~Vanhoucke, and S.~Levine.
\newblock Grasp2vec: Learning object representations from self-supervised
  grasping.
\newblock In \emph{CoRL}, 2018.

\bibitem[Jenni et~al.(2020)Jenni, Jin, and Favaro]{jenni2020steering}
S.~Jenni, H.~Jin, and P.~Favaro.
\newblock Steering self-supervised feature learning beyond local pixel
  statistics.
\newblock In \emph{CVPR}, 2020.

\bibitem[Kendall et~al.(2018)Kendall, Gal, and Cipolla]{kendall2017multi}
A.~Kendall, Y.~Gal, and R.~Cipolla.
\newblock Multi-task learning using uncertainty to weigh losses for scene
  geometry and semantics.
\newblock In \emph{Conference on Computer Vision and Pattern Recognition
  (CVPR)}, 2018.

\bibitem[Khodabandeh et~al.(2019)Khodabandeh, Vahdat, Ranjbar, and
  Macready]{robust_Khodabandeh_2019_ICCV}
M.~Khodabandeh, A.~Vahdat, M.~Ranjbar, and W.~G. Macready.
\newblock A robust learning approach to domain adaptive object detection.
\newblock In \emph{ICCV}, 2019.

\bibitem[Khosla et~al.(2012)Khosla, Zhou, Malisiewicz, Efros, and
  Torralba]{ECCV12_Khosla}
A.~Khosla, T.~Zhou, T.~Malisiewicz, A.~Efros, and A.~Torralba.
\newblock Undoing the damage of dataset bias.
\newblock In \emph{ECCV}, 2012.

\bibitem[Kim et~al.(2019{\natexlab{a}})Kim, Choi, Kim, and
  Kim]{kim2019selftraining}
S.~Kim, J.~Choi, T.~Kim, and C.~Kim.
\newblock Self-training and adversarial background regularization for
  unsupervised domain adaptive one-stage object detection.
\newblock In \emph{ICCV}, 2019{\natexlab{a}}.

\bibitem[Kim et~al.(2019{\natexlab{b}})Kim, Jeong, Kim, Choi, and
  Kim]{diversify_match_Kim_2019_CVPR}
T.~Kim, M.~Jeong, S.~Kim, S.~Choi, and C.~Kim.
\newblock Diversify and match: A domain adaptive representation learning
  paradigm for object detection.
\newblock In \emph{CVPR}, 2019{\natexlab{b}}.

\bibitem[Kokkinos(2017)]{ubernet}
I.~Kokkinos.
\newblock Ubernet: Training a `universal' convolutional neural network for
  low-, mid-, and high-level vision using diverse datasets and limited memory.
\newblock In \emph{Computer Vision and Pattern Recognition (CVPR)}, 2017.

\bibitem[Krizhevsky et~al.(2012)Krizhevsky, Sutskever, and
  Hinton]{NIPS2012alexnet}
A.~Krizhevsky, I.~Sutskever, and G.~E. Hinton.
\newblock Imagenet classification with deep convolutional neural networks.
\newblock In \emph{NeurIPS}, 2012.

\bibitem[LeCun et~al.(1998)LeCun, Bottou, Bengio, and
  Haffner]{lecun1998gradient}
Y.~LeCun, L.~Bottou, Y.~Bengio, and P.~Haffner.
\newblock Gradient-based learning applied to document recognition.
\newblock \emph{IEEE}, 86\penalty0 (11):\penalty0 2278--2324, 1998.

\bibitem[Lee et~al.(2019)Lee, Zhu, Srinivasan, Shah, Savarese, Fei-Fei, Garg,
  and Bohg]{visiontouch}
M.~A. Lee, Y.~Zhu, K.~Srinivasan, P.~Shah, S.~Savarese, L.~Fei-Fei, A.~Garg,
  and J.~Bohg.
\newblock Making sense of vision and touch: Self-supervised learning of
  multimodal representations for contact-rich tasks.
\newblock In \emph{ICRA}, 2019.

\bibitem[Legg and Hutter(2007)]{AI}
S.~Legg and M.~Hutter.
\newblock A collection of definitions of intelligence.
\newblock \emph{Frontiers in Artificial Intelligence and Applications (FAIA)},
  157:\penalty0 17, 2007.

\bibitem[Li et~al.(2017)Li, Yang, Song, and Hospedales]{hospedalesPACS}
D.~Li, Y.~Yang, Y.-Z. Song, and T.~M. Hospedales.
\newblock Deeper, broader and artier domain generalization.
\newblock In \emph{ICCV}, 2017.

\bibitem[Li et~al.(2018{\natexlab{a}})Li, Yang, Song, and
  Hospedales]{MLDG_AAA18}
D.~Li, Y.~Yang, Y.~Song, and T.~M. Hospedales.
\newblock Learning to generalize: Meta-learning for domain generalization.
\newblock In \emph{AAAI}, 2018{\natexlab{a}}.

\bibitem[Li et~al.(2019)Li, Zhang, Yang, Liu, Song, and
  Hospedales]{episodic_hospedales}
D.~Li, J.~Zhang, Y.~Yang, C.~Liu, Y.-Z. Song, and T.~M. Hospedales.
\newblock Episodic training for domain generalization.
\newblock In \emph{ICCV}, 2019.

\bibitem[Li et~al.(2018{\natexlab{b}})Li, Jialin~Pan, Wang, and
  Kot]{Li_2018_CVPR}
H.~Li, S.~Jialin~Pan, S.~Wang, and A.~C. Kot.
\newblock Domain generalization with adversarial feature learning.
\newblock In \emph{CVPR}, 2018{\natexlab{b}}.

\bibitem[Li et~al.(2018{\natexlab{c}})Li, Tian, Gong, Liu, Liu, Zhang, and
  Tao]{Li_2018_ECCV}
Y.~Li, X.~Tian, M.~Gong, Y.~Liu, T.~Liu, K.~Zhang, and D.~Tao.
\newblock Deep domain generalization via conditional invariant adversarial
  networks.
\newblock In \emph{ECCV}, 2018{\natexlab{c}}.

\bibitem[Lin et~al.(2014)Lin, Maire, Belongie, Hays, Perona, Ramanan,
  Doll{\'a}r, and Zitnick]{mscoco}
T.-Y. Lin, M.~Maire, S.~Belongie, J.~Hays, P.~Perona, D.~Ramanan,
  P.~Doll{\'a}r, and C.~L. Zitnick.
\newblock Microsoft coco: Common objects in context.
\newblock In \emph{ECCV}, 2014.

\bibitem[Liu et~al.(2018)Liu, Huang, et~al.]{liu2018receptive}
S.~Liu, D.~Huang, et~al.
\newblock Receptive field block net for accurate and fast object detection.
\newblock In \emph{Proceedings of the European Conference on Computer Vision
  (ECCV)}, pages 385--400, 2018.

\bibitem[Long et~al.(2015)Long, Cao, Wang, and Jordan]{Long:2015}
M.~Long, Y.~Cao, J.~Wang, and M.~I. Jordan.
\newblock Learning transferable features with deep adaptation networks.
\newblock In \emph{ICML}, 2015.

\bibitem[Long et~al.(2016)Long, Zhu, Wang, and Jordan]{long2016unsupervised}
M.~Long, H.~Zhu, J.~Wang, and M.~I. Jordan.
\newblock Unsupervised domain adaptation with residual transfer networks.
\newblock In \emph{NeurIPS}, 2016.

\bibitem[Long et~al.(2017)Long, Zhu, Wang, and Jordan]{LongZ0J17}
M.~Long, H.~Zhu, J.~Wang, and M.~I. Jordan.
\newblock Deep transfer learning with joint adaptation networks.
\newblock In \emph{ICML}, 2017.

\bibitem[Mancini et~al.(2018{\natexlab{a}})Mancini, Bulo, Caputo, and
  Ricci]{MassiRAL}
M.~Mancini, S.~R. Bulo, B.~Caputo, and E.~Ricci.
\newblock Robust place categorization with deep domain generalization.
\newblock \emph{IEEE Robotics and Automation Letters (RA-L)},
  2018{\natexlab{a}}.

\bibitem[Mancini et~al.(2018{\natexlab{b}})Mancini, Bul{\`o}, Caputo, and
  Ricci]{mancini2018best}
M.~Mancini, S.~R. Bul{\`o}, B.~Caputo, and E.~Ricci.
\newblock Best sources forward: domain generalization through source-specific
  nets.
\newblock \emph{arXiv preprint arXiv:1806.05810}, 2018{\natexlab{b}}.

\bibitem[Mancini et~al.(2018{\natexlab{c}})Mancini, Karaoguz, Ricci, Jensfelt,
  and Caputo]{mancini2018kitting}
M.~Mancini, H.~Karaoguz, E.~Ricci, P.~Jensfelt, and B.~Caputo.
\newblock Kitting in the wild through online domain adaptation.
\newblock In \emph{IROS}, 2018{\natexlab{c}}.

\bibitem[Mancini et~al.(2018{\natexlab{d}})Mancini, Porzi, Rota~Bul\`o, Caputo,
  and Ricci]{mancini2018boosting}
M.~Mancini, L.~Porzi, S.~Rota~Bul\`o, B.~Caputo, and E.~Ricci.
\newblock Boosting domain adaptation by discovering latent domains.
\newblock In \emph{CVPR}, 2018{\natexlab{d}}.

\bibitem[Mancini et~al.(2019)Mancini, Rota~Bulò, Caputo, and Ricci]{adagraph}
M.~Mancini, S.~Rota~Bulò, B.~Caputo, and E.~Ricci.
\newblock Adagraph: Unifying predictive and continuous domain adaptation
  through graphs.
\newblock In \emph{CVPR}, 2019.

\bibitem[Massa and Girshick(2018)]{massa2018mrcnn}
F.~Massa and R.~Girshick.
\newblock {maskrcnn-benchmark: Fast, modular reference implementation of
  Instance Segmentation and Object Detection algorithms in PyTorch}.
\newblock \url{https://github.com/facebookresearch/maskrcnn-benchmark}, 2018.
\newblock Accessed: 22/08/2019.

\bibitem[Matsuura and Harada(2020)]{dg_mmld}
T.~Matsuura and T.~Harada.
\newblock Domain generalization using a mixture of multiple latent domains.
\newblock In \emph{AAAI}, 2020.

\bibitem[Matsuura et~al.(2018)Matsuura, Saito, and Harada]{TWIN_PDA}
T.~Matsuura, K.~Saito, and T.~Harada.
\newblock Twins: Two weighted inconsistency-reduced networks for partial domain
  adaptation.
\newblock \emph{arXiv:1812.07405}, 2018.

\bibitem[Misra et~al.(2016{\natexlab{a}})Misra, Shrivastava, Gupta, and
  Hebert]{misra2016cross}
I.~Misra, A.~Shrivastava, A.~Gupta, and M.~Hebert.
\newblock Cross-stitch networks for multi-task learning.
\newblock In \emph{Proceedings of the IEEE Conference on Computer Vision and
  Pattern Recognition}, pages 3994--4003, 2016{\natexlab{a}}.

\bibitem[Misra et~al.(2016{\natexlab{b}})Misra, Zitnick, and
  Hebert]{misra2016unsupervised}
I.~Misra, C.~L. Zitnick, and M.~Hebert.
\newblock {Shuffle and Learn: Unsupervised Learning using Temporal Order
  Verification}.
\newblock In \emph{ECCV}, 2016{\natexlab{b}}.

\bibitem[Mordan et~al.()Mordan, Thome, Henaff, and Cord]{mordanrevisiting}
T.~Mordan, N.~Thome, G.~Henaff, and M.~Cord.
\newblock Revisiting multi-task learning with rock: a deep residual auxiliary
  block for visual detection supplementary material.

\bibitem[Motiian et~al.(2017{\natexlab{a}})Motiian, Jones, Iranmanesh, and
  Doretto]{fewshotNIPS17}
S.~Motiian, Q.~Jones, S.~Iranmanesh, and G.~Doretto.
\newblock Few-shot adversarial domain adaptation.
\newblock In \emph{NeurIPS}, 2017{\natexlab{a}}.

\bibitem[Motiian et~al.(2017{\natexlab{b}})Motiian, Piccirilli, Adjeroh, and
  Doretto]{doretto2017}
S.~Motiian, M.~Piccirilli, D.~A. Adjeroh, and G.~Doretto.
\newblock Unified deep supervised domain adaptation and generalization.
\newblock In \emph{ICCV}, 2017{\natexlab{b}}.

\bibitem[Muandet et~al.(2013)Muandet, Balduzzi, and Sch\"{o}lkopf]{shallowDG}
K.~Muandet, D.~Balduzzi, and B.~Sch\"{o}lkopf.
\newblock Domain generalization via invariant feature representation.
\newblock In \emph{ICML}, 2013.

\bibitem[Netzer et~al.(2011)Netzer, Wang, Coates, Bissacco, Wu, and
  Ng]{netzer2011reading}
Y.~Netzer, T.~Wang, A.~Coates, A.~Bissacco, B.~Wu, and A.~Y. Ng.
\newblock Reading digits in natural images with unsupervised feature learning.
\newblock In \emph{NIPS-Workshop}, 2011.

\bibitem[Nichol(2016)]{wikiart}
K.~Nichol.
\newblock Painter by numbers, {WikiArt}, 2016.
\newblock URL \url{https://www.kaggle.com/c/painter-by-numbers}.

\bibitem[Noroozi and Favaro(2016{\natexlab{a}})]{NorooziF16}
M.~Noroozi and P.~Favaro.
\newblock Unsupervised learning of visual representations by solving jigsaw
  puzzles.
\newblock In \emph{ECCV}, 2016{\natexlab{a}}.

\bibitem[Noroozi and Favaro(2016{\natexlab{b}})]{noroozi2016unsupervised}
M.~Noroozi and P.~Favaro.
\newblock Unsupervised learning of visual representations by solving jigsaw
  puzzles.
\newblock In \emph{European Conference on Computer Vision}, pages 69--84.
  Springer, 2016{\natexlab{b}}.

\bibitem[Noroozi et~al.(2017)Noroozi, Pirsiavash, and Favaro]{learningtocount}
M.~Noroozi, H.~Pirsiavash, and P.~Favaro.
\newblock Representation learning by learning to count.
\newblock In \emph{ICCV}, 2017.

\bibitem[Noroozi et~al.(2018)Noroozi, Vinjimoor, Favaro, and
  Pirsiavash]{Noroozi_2018_CVPR}
M.~Noroozi, A.~Vinjimoor, P.~Favaro, and H.~Pirsiavash.
\newblock Boosting self-supervised learning via knowledge transfer.
\newblock In \emph{CVPR}, 2018.

\bibitem[Owens and Efros(2018)]{audiovisual}
A.~Owens and A.~A. Efros.
\newblock Audio-visual scene analysis with self-supervised multisensory
  features.
\newblock In \emph{ECCV}, 2018.

\bibitem[Pathak et~al.(2016)Pathak, Kr\"ahenb\"uhl, Donahue, Darrell, and
  Efros]{pathakCVPR16context}
D.~Pathak, P.~Kr\"ahenb\"uhl, J.~Donahue, T.~Darrell, and A.~Efros.
\newblock Context encoders: Feature learning by inpainting.
\newblock In \emph{CVPR}, 2016.

\bibitem[Peng et~al.(2019)Peng, Bai, Xia, Huang, Saenko, and Wang]{domainnet}
X.~Peng, Q.~Bai, X.~Xia, Z.~Huang, K.~Saenko, and B.~Wang.
\newblock Moment matching for multi-source domain adaptation.
\newblock In \emph{Proceedings of the IEEE International Conference on Computer
  Vision}, pages 1406--1415, 2019.

\bibitem[Ponce et~al.(2007)Ponce, Hebert, Schmid, and
  Zisserman]{ponce2007toward}
J.~Ponce, M.~Hebert, C.~Schmid, and A.~Zisserman.
\newblock \emph{Toward category-level object recognition}, volume 4170.
\newblock Springer, 2007.

\bibitem[Ren et~al.(2015)Ren, He, Girshick, and Sun]{ren2015faster}
S.~Ren, K.~He, R.~Girshick, and J.~Sun.
\newblock Faster r-cnn: Towards real-time object detection with region proposal
  networks.
\newblock In \emph{NIPS}, 2015.

\bibitem[Ren and Lee(2018)]{ren-cvpr2018}
Z.~Ren and Y.~J. Lee.
\newblock Cross-domain self-supervised multi-task feature learning using
  synthetic imagery.
\newblock In \emph{CVPR}, 2018.

\bibitem[Rosenbaum et~al.(2017)Rosenbaum, Klinger, and
  Riemer]{rosenbaum2017routing}
C.~Rosenbaum, T.~Klinger, and M.~Riemer.
\newblock Routing networks: Adaptive selection of non-linear functions for
  multi-task learning.
\newblock \emph{arXiv preprint arXiv:1711.01239}, 2017.

\bibitem[Ruder12 et~al.(2017)Ruder12, Bingel, Augenstein, and
  S{\o}gaard]{ruder122017sluice}
S.~Ruder12, J.~Bingel, I.~Augenstein, and A.~S{\o}gaard.
\newblock Sluice networks: Learning what to share between loosely related
  tasks.
\newblock \emph{stat}, 1050:\penalty0 23, 2017.

\bibitem[Russo et~al.(2018)Russo, Carlucci, Tommasi, and
  Caputo]{russo17sbadagan}
P.~Russo, F.~M. Carlucci, T.~Tommasi, and B.~Caputo.
\newblock From source to target and back: symmetric bi-directional adaptive
  gan.
\newblock In \emph{CVPR}, 2018.

\bibitem[Saenko et~al.(2010)Saenko, Kulis, Fritz, and Darrell]{Saenko:2010}
K.~Saenko, B.~Kulis, M.~Fritz, and T.~Darrell.
\newblock Adapting visual category models to new domains.
\newblock In \emph{ECCV}, 2010.

\bibitem[Saito et~al.(2018)Saito, Watanabe, Ushiku, and
  Harada]{saito2017maximum}
K.~Saito, K.~Watanabe, Y.~Ushiku, and T.~Harada.
\newblock Maximum classifier discrepancy for unsupervised domain adaptation.
\newblock \emph{CVPR}, 2018.

\bibitem[Saito et~al.(2019)Saito, Ushiku, Harada, and Saenko]{Saito_2019_CVPR}
K.~Saito, Y.~Ushiku, T.~Harada, and K.~Saenko.
\newblock Strong-weak distribution alignment for adaptive object detection.
\newblock In \emph{CVPR}, 2019.

\bibitem[Sakaridis et~al.(2018)Sakaridis, Dai, and
  Van~Gool]{sakaridis2018semantic}
C.~Sakaridis, D.~Dai, and L.~Van~Gool.
\newblock Semantic foggy scene understanding with synthetic data.
\newblock \emph{IJCV}, 126\penalty0 (9):\penalty0 973--992, 2018.

\bibitem[Sankaranarayanan et~al.(2018)Sankaranarayanan, Balaji, Castillo, and
  Chellappa]{sankaranarayanan2017generate}
S.~Sankaranarayanan, Y.~Balaji, C.~D. Castillo, and R.~Chellappa.
\newblock Generate to adapt: Aligning domains using generative adversarial
  networks.
\newblock In \emph{CVPR}, 2018.

\bibitem[Sermanet et~al.(2018)Sermanet, Lynch, Chebotar, Hsu, Jang, Schaal, and
  Levine]{SSLvideo}
P.~Sermanet, C.~Lynch, Y.~Chebotar, J.~Hsu, E.~Jang, S.~Schaal, and S.~Levine.
\newblock Time-contrastive networks: Self-supervised learning from video.
\newblock In \emph{ICRA}, 2018.

\bibitem[Shankar et~al.(2018{\natexlab{a}})Shankar, Piratla, Chakrabarti,
  Chaudhuri, Jyothi, and Sarawagi]{CrossGrad}
S.~Shankar, V.~Piratla, S.~Chakrabarti, S.~Chaudhuri, P.~Jyothi, and
  S.~Sarawagi.
\newblock Generalizing across domains via cross-gradient training.
\newblock \emph{arXiv preprint arXiv:1804.10745}, 2018{\natexlab{a}}.

\bibitem[Shankar et~al.(2018{\natexlab{b}})Shankar, Piratla, Chakrabarti,
  Chaudhuri, Jyothi, and Sarawagi]{DG_ICLR18}
S.~Shankar, V.~Piratla, S.~Chakrabarti, S.~Chaudhuri, P.~Jyothi, and
  S.~Sarawagi.
\newblock Generalizing across domains via cross-gradient training.
\newblock In \emph{ICLR}, 2018{\natexlab{b}}.

\bibitem[Sun and Saenko(2016)]{dcoral}
B.~Sun and K.~Saenko.
\newblock Deep coral: Correlation alignment for deep domain adaptation.
\newblock In \emph{Workshop of the European Conference on Computer Vision
  (ECCV-Workshop)}, 2016.

\bibitem[Tobin et~al.(2017)Tobin, Fong, Ray, Schneider, Zaremba, and
  Abbeel]{Tobin2017DomainRF}
J.~Tobin, R.~Fong, A.~Ray, J.~Schneider, W.~Zaremba, and P.~Abbeel.
\newblock Domain randomization for transferring deep neural networks from
  simulation to the real world.
\newblock In \emph{International Conference on Intelligent Robots and Systems
  (IROS)}, 2017.

\bibitem[Torralba and Efros(2011)]{TorralbaEfros_bias}
A.~Torralba and A.~A. Efros.
\newblock Unbiased look at dataset bias.
\newblock In \emph{CVPR}, 2011.

\bibitem[Tzeng et~al.(2017)Tzeng, Hoffman, Darrell, and
  Saenko]{Hoffman:Adda:CVPR17}
E.~Tzeng, J.~Hoffman, T.~Darrell, and K.~Saenko.
\newblock Adversarial discriminative domain adaptation.
\newblock In \emph{CVPR}, 2017.

\bibitem[Venkateswara et~al.(2017)Venkateswara, Eusebio, Chakraborty, and
  Panchanathan]{venkateswara2017Deep}
H.~Venkateswara, J.~Eusebio, S.~Chakraborty, and S.~Panchanathan.
\newblock Deep hashing network for unsupervised domain adaptation.
\newblock In \emph{CVPR}, 2017.

\bibitem[Viola and Jones(2001)]{viola2001rapid}
P.~Viola and M.~Jones.
\newblock Rapid object detection using a boosted cascade of simple features.
\newblock In \emph{CVPR}, 2001.

\bibitem[Volpi et~al.(2018)Volpi, Namkoong, Sener, Duchi, Murino, and
  Savarese]{Volpi_2018_NIPS}
R.~Volpi, H.~Namkoong, O.~Sener, J.~Duchi, V.~Murino, and S.~Savarese.
\newblock Generalizing to unseen domains via adversarial data augmentation.
\newblock In \emph{NeurIPS}, 2018.

\bibitem[Wang et~al.(2019)Wang, Ge, Lipton, and Xing]{wang2019learning}
H.~Wang, S.~Ge, Z.~Lipton, and E.~P. Xing.
\newblock Learning robust global representations by penalizing local predictive
  power.
\newblock In \emph{NeurIPS}, 2019.

\bibitem[Wang and Gupta(2015)]{videosiccv15}
X.~Wang and A.~Gupta.
\newblock Unsupervised learning of visual representations using videos.
\newblock In \emph{ICCV}, 2015.

\bibitem[Wulfmeier et~al.(2018)Wulfmeier, Bewley, and
  Posner]{Wulfmeier2017IncrementalAD}
M.~Wulfmeier, A.~Bewley, and I.~Posner.
\newblock Incremental adversarial domain adaptation for continually changing
  environments.
\newblock 2018.

\bibitem[Xie et~al.(2019)Xie, Yu, Wang, Wang, and
  Zhang]{Xie_2019_ICCV_Workshops}
R.~Xie, F.~Yu, J.~Wang, Y.~Wang, and L.~Zhang.
\newblock Multi-level domain adaptive learning for cross-domain detection.
\newblock In \emph{ICCV Workshops}, 2019.

\bibitem[{Xu} et~al.(2019){Xu}, {Xiao}, and {López}]{lopez_rotation}
J.~{Xu}, L.~{Xiao}, and A.~M. {López}.
\newblock Self-supervised domain adaptation for computer vision tasks.
\newblock \emph{IEEE Access}, 7:\penalty0 156694--156706, 2019.

\bibitem[Xu et~al.(2020)Xu, Zhang, Ni, Li, Wang, Tian, and
  Zhang]{xu2020adversarial}
M.~Xu, J.~Zhang, B.~Ni, T.~Li, C.~Wang, Q.~Tian, and W.~Zhang.
\newblock Adversarial domain adaptation with domain mixup.
\newblock In \emph{AAAI}, 2020.

\bibitem[Xu et~al.(2018)Xu, Chen, Zuo, Yan, and Lin]{cocktail_CVPR18}
R.~Xu, Z.~Chen, W.~Zuo, J.~Yan, and L.~Lin.
\newblock Deep cocktail network: Multi-source unsupervised domain adaptation
  with category shift.
\newblock In \emph{CVPR}, 2018.

\bibitem[Xu et~al.(2019)Xu, Li, Yang, and Lin]{featurenorm_PDA}
R.~Xu, G.~Li, J.~Yang, and L.~Lin.
\newblock Larger norm more transferable: An adaptive feature norm approach for
  unsupervised domain adaptation.
\newblock In \emph{ICCV}, 2019.

\bibitem[Yang et~al.(2015)Yang, Luo, Change~Loy, and Tang]{Yang_2015_CVPR}
L.~Yang, P.~Luo, C.~Change~Loy, and X.~Tang.
\newblock A large-scale car dataset for fine-grained categorization and
  verification.
\newblock In \emph{CVPR}, 2015.

\bibitem[Yang and Hospedales(2016)]{yang2016trace}
Y.~Yang and T.~M. Hospedales.
\newblock Trace norm regularised deep multi-task learning.
\newblock \emph{arXiv preprint arXiv:1606.04038}, 2016.

\bibitem[Yu et~al.(2018)Yu, Wang, Shelhamer, and Darrell]{shelhamerdeep}
F.~Yu, D.~Wang, E.~Shelhamer, and T.~Darrell.
\newblock Deep layer aggregation.
\newblock In \emph{CVPR}, 2018.

\bibitem[Zhai et~al.(2019)Zhai, Oliver, Kolesnikov, and Beyer]{S4L_iccv19}
X.~Zhai, A.~Oliver, A.~Kolesnikov, and L.~Beyer.
\newblock S4l: Self-supervised semi-supervised learning.
\newblock In \emph{ICCV}, 2019.

\bibitem[Zhang et~al.(2018{\natexlab{a}})Zhang, Cisse, Dauphin, and
  Lopez-Paz]{zhang2018mixup}
H.~Zhang, M.~Cisse, Y.~N. Dauphin, and D.~Lopez-Paz.
\newblock mixup: Beyond empirical risk minimization.
\newblock In \emph{ICLR}, 2018{\natexlab{a}}.

\bibitem[Zhang et~al.(2018{\natexlab{b}})Zhang, Ding, Li, and Ogunbona]{IWAN}
J.~Zhang, Z.~Ding, W.~Li, and P.~Ogunbona.
\newblock Importance weighted adversarial nets for partial domain adaptation.
\newblock In \emph{CVPR}, 2018{\natexlab{b}}.

\bibitem[Zhang et~al.(2016)Zhang, Isola, and Efros]{zhang2016colorful}
R.~Zhang, P.~Isola, and A.~A. Efros.
\newblock Colorful image colorization.
\newblock In \emph{European Conference on Computer Vision}, pages 649--666.
  Springer, 2016.

\bibitem[Zhang et~al.(2018{\natexlab{c}})Zhang, Wen, Bian, Lei, and
  Li]{zhang2018single}
S.~Zhang, L.~Wen, X.~Bian, Z.~Lei, and S.~Z. Li.
\newblock Single-shot refinement neural network for object detection.
\newblock In \emph{Proceedings of the IEEE conference on computer vision and
  pattern recognition}, pages 4203--4212, 2018{\natexlab{c}}.

\bibitem[Zhang and Yang(2017)]{DBLP:journals/corr/ZhangY17aa}
Y.~Zhang and Q.~Yang.
\newblock A survey on multi-task learning.
\newblock \emph{CoRR}, abs/1707.08114, 2017.
\newblock URL \url{http://arxiv.org/abs/1707.08114}.

\bibitem[{Zhang} et~al.(2020){Zhang}, {Zhang}, {Xu}, and
  {Zhang}]{zhang2020learning}
Y.~{Zhang}, Y.~{Zhang}, Q.~{Xu}, and R.~{Zhang}.
\newblock Learning robust shape-based features for domain generalization.
\newblock \emph{IEEE Access}, 8:\penalty0 63748--63756, 2020.

\bibitem[Zhou et~al.(2020)Zhou, Yang, Hospedales, and Xiang]{zhou2020deep}
K.~Zhou, Y.~Yang, T.~Hospedales, and T.~Xiang.
\newblock Deep domain-adversarial image generation for domain generalisation.
\newblock \emph{AAAI}, 2020.

\bibitem[Zhu et~al.(2017)Zhu, Park, Isola, and Efros]{CycleGAN2017}
J.-Y. Zhu, T.~Park, P.~Isola, and A.~A. Efros.
\newblock Unpaired image-to-image translation using cycle-consistent
  adversarial networks.
\newblock In \emph{International Conference on Computer Vision (ICCV)}, 2017.

\end{thebibliography}
